\documentclass{article}
\usepackage[utf8]{inputenc}

\usepackage[final]{neurips_2025}
\usepackage{amsmath}
\usepackage{graphicx}
\usepackage{subcaption}
\usepackage{caption}
\usepackage{amssymb}
\usepackage{comment}

\newcommand{\E}{\mathbb{E}}

\newtheorem{theorem}{Theorem}

\newtheorem{proposition}[theorem]{Proposition}
\newtheorem{lemma}[theorem]{Lemma}

\newtheorem{remark}[theorem]{Remark}
\newtheorem{example}[theorem]{Example}

\numberwithin{equation}{section}

\newcommand{\RR}{\ensuremath{\mathbb R}}

\newcommand{\R}{\mathbb{R}}

\usepackage{algorithm}
\usepackage{float}
\usepackage{bm}            
\usepackage{dsfont}        
\usepackage{mathtools}     
\usepackage{booktabs}      
\usepackage{geometry}
\usepackage{bbm}
\usepackage{soul}
\usepackage{algorithm}
\usepackage{algpseudocode}
\usepackage{enumitem}   
\usepackage{xurl}      
\usepackage{hyperref}   
\usepackage{braket}
\usepackage{stfloats}
\usepackage[capitalize,nameinlink]{cleveref}
\crefname{lemma}{lemma}{lemmas}
\crefname{example}{example}{Example}
\crefname{proposition}{proposition}{propositions}

\usepackage{subcaption}
\def\sint{\begingroup\textstyle\int\endgroup} 

\usepackage[utf8]{inputenc} 
\usepackage[T1]{fontenc}    
\usepackage{hyperref}       
\usepackage{url}            
\usepackage{booktabs}       
\usepackage{amsfonts}       
\usepackage{nicefrac}       
\usepackage{microtype}      
\usepackage{tabularx}

\usepackage{colortbl} 
\usepackage[table,xcdraw]{xcolor} 
\usepackage{algpseudocode}

\definecolor{darkblue}{rgb}{0.20,0.20,0.80}
\definecolor{darkred}{rgb}{0.68,0.05,0.0}
\definecolor{darkgreen}{rgb}{0.0,0.29,0.29}
\definecolor{darkpurple}{rgb}{0.47,0.09,0.29}

\hypersetup{
   colorlinks = true,
   citecolor  = darkblue,
   linkcolor  = darkred,
   filecolor  = darkblue,
   urlcolor   = darkpurple,
}

\newcounter{customsubsub}[subsubsection]

\renewcommand{\thecustomsubsub}{\thesubsubsection.\arabic{customsubsub}}

\renewcommand{\subsubsubsection}[1]{%
  \par\noindent
  \refstepcounter{customsubsub}%
  \vspace{1.5ex}%
  \textbf{\thecustomsubsub\quad #1}%
\par\nobreak

}

\Crefname{customsubsub}{Section}{Sections}

\title{Cross-fluctuation phase transitions reveal sampling dynamics in diffusion models}

\author{%
  {\bfseries Sai Niranjan Ramachandran}\thanks{School of Computation, Information and Technology, Technical University of Munich, Germany}\, \thanks{Munich Center for Machine Learning (MCML) }%
  \and
  {\bfseries Manish Krishan Lal}\footnotemark[1]\, \footnotemark[2]%
  \and
  {\bfseries Suvrit Sra}\footnotemark[1]\, \footnotemark[2]%
}

\begin{document}

\maketitle

\vskip -0.2in

\begin{abstract}

We analyse how the sampling dynamics of distributions evolve in score-based diffusion models using \emph{cross-fluctuations}, a centered-moment statistic from statistical physics. Specifically, we show that starting from an unbiased isotropic normal distribution, samples undergo sharp, discrete transitions, eventually forming distinct events of a desired distribution while progressively revealing finer structure. As this process is reversible, these transitions also occur in reverse, where intermediate states progressively merge, tracing a path back to the initial distribution.  We demonstrate that these transitions can be detected as discontinuities in $n^{\text{th}}$-order cross-fluctuations. For variance-preserving SDEs, we derive a closed-form for these cross-fluctuations that is efficiently computable for the reverse trajectory. 
We find that detecting these transitions directly boosts sampling efficiency, accelerates class-conditional and rare-class generation, and improves two zero-shot tasks--image classification and style transfer--without expensive grid search or retraining. We also show that this viewpoint unifies classical coupling and mixing from finite Markov chains with continuous dynamics while extending to stochastic SDEs and non Markovian samplers. Our framework therefore bridges discrete Markov chain theory, phase analysis, and modern generative modeling.
\end{abstract}

\section{Introduction}Diffusion models are now a cornerstone of generative systems. They learn to reverse a process that gradually erases structure from data, transforming it into featureless (isotropic) noise. During inference, the model starts with this noise and progressively refines it, adding structure step by step until realistic data samples emerge.~\citep{ho2020denoisingdiffusionprobabilisticmodels,sohl2015deep}. These models can generate lifelike images, 3D scenes, audio, and even molecular structures ~\citep{rombach2022highresolutionimagesynthesislatent,blattmann2023stable,chen2022analog,corso2022diffdock,corso2023particle,karnewar2023holodiffusion,kynkäänniemi2024applyingguidancelimitedinterval}. They also excel in complex tasks like protein design ~\citep{jumper2021highly} and zero-shot vision tasks ~\citep{li2023diffusionmodelsecretlyzeroshot,meng2021sdedit,rombach2022highresolutionimagesynthesislatent}.

Yet, the sampling process remains a black box. Each step blends thousands of values in ways that are difficult to anticipate. Like other generative paradigms ~\citep{hutchinson201950,hendrycks2023overview,rudin2019stop,sharkey2025open}, small tweaks in code or hyperparameters can turn a perfect sample into a failure or introduce unintended biases. This unpredictability underscores the need for a clear understanding of how successful and failed outcomes diverge as the sampler operates.

We provide a clear framework by defining a user-specified goal (e.g., a class label, an aesthetic style, or a semantic attribute) as a \emph{desirable event}, which is a set of outcomes that align with the user's objective. As the sampler evolves \emph{backward} in time, it iteratively generates a distribution over possible outcomes at each step. Initially, paths corresponding to different outcomes,  whether desirable or undesirable, may appear indistinguishable. However, we demonstrate that at certain critical intermediate steps, these paths undergo \emph{discrete phase transitions}, 
where they become distinguishable as they separate into distinct regions of the outcome space. These transitions occur when the sampling process reveals structural differences between outcomes that align with the user's goal (desirable) and those that do not (undesirable), marked by a breakdown of path separability.

\subsection*{Main contributions}
The main contributions of this paper are the following:
\begin{enumerate}[leftmargin=2em]
\setlength{\itemsep}{1pt}
\item \textbf{Framework and theory.} We present a rigorous framework using fluctuation theory from statistical physics \citep{chaikin1995principles,kivelson2024statistical,landau_statistical_1980,pathria2011statistical} to identify and quantify \emph{discrete phase transitions in the diffusion process} (\Cref{sec:fluc-primer}, \Cref{sec:primer-fluc-th,sec:phase-transit}). This framework incorporates user-specified goals as desirable events and shows how their dynamics can be tracked \;(\Cref{sec:methodology}). Moreover, we establish a connection between ensemble based techniques in statistical mechanics as well as classical mixing-coupling results for Markov chains \citep{aldous-fill-2002,levin2017markov,saloffcoste1997lectures}, using a generalised notion of phase transitions, thereby providing a unified perspective on probability flows (\Cref{sec:eqv,sec:coupling-mixing,sec:phase-transit}).

\item \textbf{Practical toolkit.} We develop a practical toolkit with clear diagnostic criteria for identifying and leveraging these \emph{discrete transitions} (\Cref{sec:methodology}, \Cref{app:theoretical,sec:experimental}). The proposed \Cref{alg:template} systematically identifies discrete transitions in cross-fluctuations to characterise the sampling dynamics of desired outcomes. This enables precise intervention to enhance the generation and likelihood of desirable samples, thereby improving overall model performance.

\item \textbf{Illustrative applications.} We illustrate the power of the proposed framework and toolkit  through several applications: accelerated sampling (\Cref{sec:warmup}), conditional generation (\Cref{sec:class-cond}), rare-class coverage (\Cref{sec:rare-class}), and improved zero-shot performance in classification (\Cref{sec:zero-shot}) and style transfer (\Cref{sec:zero-shot-style}). Remarkably, a simple fluctuation driven tweak can improve the performance of baseline methods.  
\end{enumerate}

The remainder of this paper is organised as follows: background (\Cref{sec:background}); methodology (\Cref{sec:methodology}); experiments (\Cref{sec:experiments}); theoretical foundations (Appendix~\ref{app:theoretical}); implementation details and additional experiments (Appendix~\ref{sec:experimental}); and limitations and related-work context (Appendix~\ref{sec:limitations}--\ref{sec:related}).

\section{Background}
\label{sec:background}
We first review variance-preserving diffusion models and the fluctuation-theoretic statistics that let us detect emergent phases. For detailed notation used throughout the paper, see \Cref{app:notation}.

\subsection{Diffusion models in continuous and discrete time}
\label{sec:diff-primer}

Let $\mathbf{x}_{0} \sim p_{0}$ be  a sample from a distribution with support $\mathcal{X} \subseteq \mathbb{R}^{d}$.
The \emph{forward} variance-preserving stochastic differential equation (SDE) of \citet{song2021scorebased} is
\begin{equation}
\label{eq:vp-sde}
\mathrm{d}\mathbf{x}_{t}
  = -\tfrac12\beta(t)\,\mathbf{x}_{t}\,\mathrm{d}t
    + \sqrt{\beta(t)}\,\mathrm{d}\mathbf{w}_{t},
  \qquad t\in[0,T],
\end{equation}
where $\beta(t)>0$ is the noise schedule and $\mathbf{w}_{t}$ is a standard Wiener process.  
For linear schedules and sufficiently large $T$ we have
$p_{T}\approx\mathcal{N}(\mathbf{0},\mathbf{I})$. Removing stochasticity yields the \textbf{probability-flow ODE} (PF-ODE) with identical marginals,
\begin{equation}
\label{eq:pf-ode}
\frac{\mathrm{d}\mathbf{x}_{t}}{\mathrm{d}t}
  = -\tfrac12\beta(t)\,\mathbf{x}_{t}
    -\beta(t)\,
      \underbrace{\nabla_{\mathbf{x}}\log p_{t}(\mathbf{x}_{t})}_{\text{score}} .
\end{equation}
A neural network $s_{\theta}(\mathbf{x},t)$ learns an approximation of the score; sampling integrates the \emph{reverse} ODE from $t=T$ back to~$0$. In discrete time, the Denoising Diffusion Probabilistic Model (DDPM) update of \citet{ho2020denoisingdiffusionprobabilisticmodels} is
\begin{equation}
\label{eq:ddpm}
\mathbf{x}_{t}
  = \sqrt{1-\beta_{t}}\;\mathbf{x}_{t-1}
    + \sqrt{\beta_{t}}\;\varepsilon_{t},
  \qquad
  \varepsilon_{t}\sim\mathcal{N}(\mathbf{0},\mathbf{I}),
  \quad t=1,\ldots,T .
\end{equation}
Both the continuous and the discrete processes admit the closed-form marginal
\begin{equation}
\label{eq:marginal}
p_{t}(\mathbf{x}_{t}\mid\mathbf{x}_{0})
  = \mathcal{N}\!\bigl( J(t)\,\mathbf{x}_{0},
                       \bigl(1-J^{2}(t)\bigr)\mathbf{I}\bigr),
  \qquad
  J(t)=\exp \bigl(-\tfrac12\!\sint_{0}^{t}\!\beta(s)\,\mathrm{d}s\bigr),
\end{equation}
which separates signal from noise cleanly, where $J(t)$ is known as the signal-attenuation factor.

\subsection{Measuring statistical distinctiveness}
\label{sec:fluc-primer}

The central insight of our work is the observation that given coarse categories of data e.g, object classes, tracking their \emph{statistical distinctiveness} throughout the diffusion process suffices to pinpoint the \emph{emergence} of structure during generation.

\begin{figure}[h!]
    \centering
    \captionsetup{font=small, labelfont=bf}
    \includegraphics[width=0.3\textwidth]{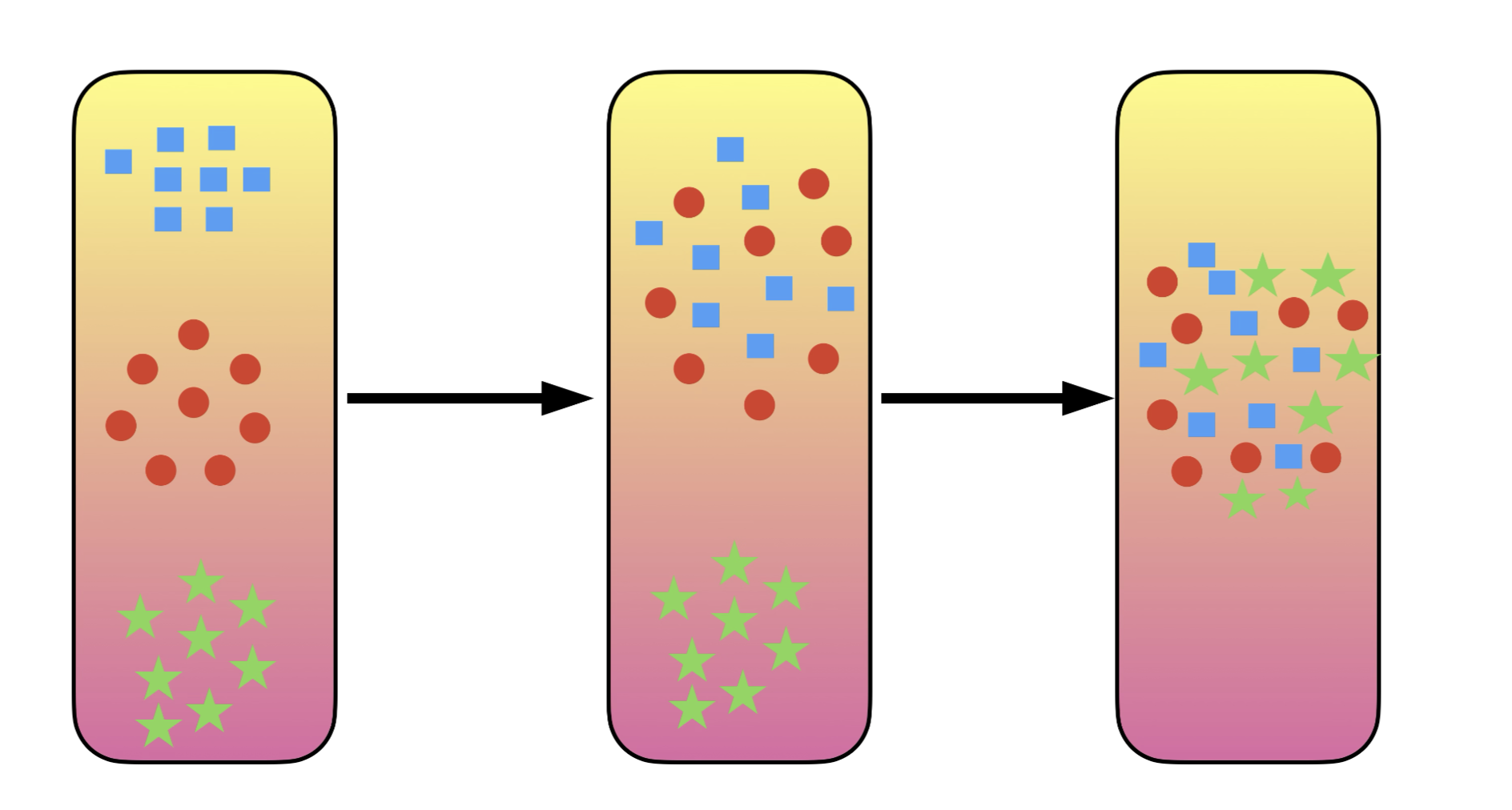}
    \caption{The addition of noise causes distinct categories of data to "merge" through the forward diffusion process as statistical properties progressively converge to those of the standard normal distribution.}
    \label{fig:intro}
\end{figure}
As \Cref{fig:intro} shows, the statistical properties of data show a well-defined convergence on noise addition and conversely show divergence as the sampling process progresses. We formalise this notion through the use of fluctuation theory adapted from physics.

\subsubsection{Basics of fluctuation theory}
Fluctuation theory supplies statistics that reveal when trajectories become (in)distinguishable—our operational signature of a phase transition. We present a formal framework for analyzing these fluctuations in vector-valued state spaces, which generalises the concepts from statistical physics and provides a direct connection to modern tools like Centered Kernel Alignment (CKA) (\cite{kornblith2019similarity}).

\paragraph{Fluctuations.}
Let $(\Omega, \mathcal{F}, P)$ be a probability space. The state of the system is described by a vector-valued random variable $\rho: \Omega \to \mathbb{R}^d$. The fundamental object for our analysis is the \textbf{\(n\)\textsuperscript{th}-order fluctuation}, a random tensor representing the \(n\)\textsuperscript{th} centered moment at a single point $\omega \in \Omega$:
\begin{equation}
\label{eq:fluc-def}
\mathcal{F}^{(n)}_\rho(\omega) := \bigotimes_{k=1}^{n} \bigl(\rho(\omega) - \mathbb{E}[\rho]\bigr).
\end{equation}
This is a tensor of rank $n$. For instance, for $n=2$, this gives $\mathcal{F}^{(2)}_\rho(\omega) = (\rho(\omega) - \mathbb{E}[\rho])(\rho(\omega) - \mathbb{E}[\rho])^\top$, which is a $d \times d$  matrix. Its expectation, $\mathbb{E}[\mathcal{F}^{(2)}_\rho]$, is the familiar \textbf{covariance matrix} of $\rho$. We assume these random fluctuation tensors (and their expectations) reside in suitable Hilbert spaces $\mathcal{H}_n$ (for rank-$n$ tensors) equipped with an inner product $\langle \cdot, \cdot \rangle_{\mathcal{H}_n}$. For matrices ($n=2$), $\mathcal{H}_2$ is typically the space of $d \times d$ matrices with the Frobenius inner product, $\langle A, B \rangle_F = \text{Tr}(A^\top B)$.

\paragraph{Joint dynamics and conditional expectations.}
We consider two disjoint events, $\Omega_1$ and $\Omega_2$. While samples are drawn independently from each, the events themselves are \textbf{coupled}: they originate from the same data manifold and evolve under the \textbf{same stochastic dynamics}. This coupling justifies the comparison of their respective \textbf{conditional expectations of the fluctuation tensor}:
$\mathbb{E}_1[\mathcal{F}^{(n)}_\rho] = \mathbb{E}[\mathcal{F}^{(n)}_\rho | \omega \in \Omega_1]$ and $\mathbb{E}_2[\mathcal{F}^{(n)}_\rho] = \mathbb{E}[\mathcal{F}^{(n)}_\rho | \omega \in \Omega_2]$.
For $n=2$, these are the conditional covariance matrices for each event.

\paragraph{Normalised and unnormalised cross-fluctuations.}
We now redefine our key statistics based on these conditional expectations.
\begin{enumerate}
    \item \textbf{Unnormalised cross-fluctuation ($G$):} This measures the alignment of the conditional expected fluctuation tensors of the two events.
    \begin{equation}
    \label{eq:m-unnorm}
    G^{(n)}_{\rho}(\Omega_{1},\Omega_{2}) := \langle \mathbb{E}_1[\mathcal{F}^{(n)}_\rho], \mathbb{E}_2[\mathcal{F}^{(n)}_\rho] \rangle_{\mathcal{H}_n}.
    \end{equation}

    \item \textbf{Within event fluctuation ($\widehat{F}$):} This measures the magnitude (or self-alignment) of the conditional expected fluctuation tensor for a single event.
    \begin{equation}
    \label{eq:wide-hat-fluc}
    \widehat{F}^{(2n)}_{\rho}(\Omega_{i}) := \langle \mathbb{E}_k[\mathcal{F}^{(n)}_\rho], \mathbb{E}_k[\mathcal{F}^{(n)}_\rho] \rangle_{\mathcal{H}_n} = \|\mathbb{E}_k[\mathcal{F}^{(n)}_\rho]\|_{\mathcal{H}_n}^2.
    \end{equation}
    Note the $2n$ in the symbol $\widehat{F}$ is a convention.

  \item \textbf{Normalised cross-fluctuation ($\mathcal{M}$):} This is the magnitude of the cosine similarity between the conditional expected fluctuation tensors.
    \begin{equation}
    \label{eq:m-normalised}
    \mathcal{M}^{(n)}_{\rho}(\Omega_{1},\Omega_{2}) := \frac{\|G^{(n)}_{\rho}(\Omega_{1},\Omega_{2})\|}{\sqrt{\widehat{F}^{(2n)}_{\rho}(\Omega_{1}) \widehat{F}^{(2n)}_{\rho}(\Omega_{2})}} = \frac{|\langle \mathbb{E}_1[\mathcal{F}^{(n)}_\rho], \mathbb{E}_2[\mathcal{F}^{(n)}_\rho] \rangle_{\mathcal{H}_n}|}{\|\mathbb{E}_1[\mathcal{F}^{(n)}_\rho]\|_{\mathcal{H}_n} \|\mathbb{E}_2[\mathcal{F}^{(n)}_\rho]\|_{\mathcal{H}_n}}.
    \end{equation}
\end{enumerate}
The normalized cross-fluctuation, $\mathcal{M}^{(n)}_{\rho} \in [0,1]$, measures the structural similarity between the events' $n$-th order moments. A value of $\mathcal{M}^{(n)}_{\rho} \approx 1$ indicates that the events' statistical shapes have aligned, a condition we define as a ``merge.'' Conversely, a value significantly less than 1 implies distinguishability. For the important special case of $n=2$, $\mathcal{M}^{(2)}_{\rho}$ quantifies the similarity between the conditional covariance matrices ($\Sigma_k$) of the two events and is connected to CKA.
\section{A practical toolkit for diffusion models}
\label{sec:methodology}
We now present the theoretical construction for our methodology. The main practical outcome is \Cref{alg:return-M-and-istar}, an algorithm whose applications are demonstrated in \Cref{sec:warmup,sec:class-cond,sec:rare-class,sec:zero-shot,sec:zero-shot-style}.

\vspace{-1em}
\paragraph{Embedding in diffusion time.}
Let $p_i$ be the marginal at step $i=0,\ldots,n$ of a forward diffusion
process (For simplicity, we focus on the PF-ODE case\footnote{A detailed treatment for the PF-ODE appears in \Cref{sec:sr-es-disj}, and the SDE case is discussed in \Cref{sec:sde-pullback}.}), with $p_0=p_{\text{desired}}$, where $p_{\text{{desired}}}$ represents a constructed distribution that encodes properties of interest, and
$p_n=\mathcal{N}(\mathbf{0},\mathbf{I})$.
Choose two disjoint events $\Omega_{1,0},\Omega_{2,0}\subseteq\operatorname{supp}(p_0)$
and propagate them to $\Omega_{1,i},\Omega_{2,i}$.
As the forward trajectory converges to white noise, the regions coincide at 
$i=n$ in theory, but due to hyper-contractivity and numerical effects, they often appear to merge much earlier. We thus redefine the operator $\mathcal{M}^{(n)}_{\rho}$ to capture this \emph{discrete transition} (\Cref{sec:phase-transit}),

\begin{equation}
\label{eq:time-dependent-M}
\widetilde{\mathcal{M}}^{(n)}_{\rho}(i)=
\begin{cases}
  \mathcal{M}^{(n)}_{\rho}\!\bigl(\Omega_{1,i},\Omega_{2,i}\bigr),
     & d\bigl(\widehat{F}^{(2n)}_\rho(\Omega_{1,i}), \widehat{F}^{(2n)}_\rho(\Omega_{2,i})\bigr) \! > \varepsilon, \\
  1, & \text{otherwise},
\end{cases}
\end{equation}

where $d(\cdot,\cdot)$ is a distance metric which for our case is the $\ell_1$ distance between the traces of
$\widehat{F}^{(n)}_\rho(\cdot)$ and $\varepsilon>0$ is a fixed threshold.\footnote{Details on the topological equivalence of a metric space on  within-event fluctuations and the  absolute distance of  normalised cross-fluctuation from $1$ can be found at \Cref{sec:cross-fluc-fluc-th}}
The earliest index
\[
  i^{\star}=\min\bigl\{\,i : \widetilde{\mathcal{M}}^{(n)}_{\rho}(i)=1\bigr\}
\]
generalises the coupling time of a Markov chain
\citep{aldous-fill-2002,levin2017markov} (\Cref{sec:coupling-mixing}). The operator undergoes a \emph{discrete} phase transition dictated by $\varepsilon$. Conceptually, \emph{the smooth distance function has been replaced by a "relu" like non-linearity to capture numerical effects}. Further details on this kind of phase transition and how it \emph{differs} from the traditional notion of a phase-transition can be found in (\Cref{sec:phase-transit}).  

We justify our choice of the forward trajectory in \Cref{sec:approx-reverse}, establishing that the detected transitions also occur in the sampling trajectory of a trained diffusion model. Crucially, this approach allows us to rely solely on the computationally efficient empirical forward process. We tailor $p_{\text{desired}}$ and $(\Omega_{1,0},\Omega_{2,0})$ to illuminate
specific phenomena.
\Cref{alg:return-M-and-istar} implements these steps. Note that, when $p_0$ has compact support (e.g., images), the PF-ODE ensures spatial
continuity of $\mathcal{M}^{(n)}_\rho$~\citep[Thm~5.2]{coddington1955theory}; any jump thus indicates a genuine
phase transition (for SDEs see \Cref{sec:sde-pullback}). A single forward Monte-Carlo sweep yields unbiased estimates of all terms in
$\widetilde{\mathcal{M}}^{(n)}_\rho$ see \Cref{sec:mcmc}.
\setlength{\textfloatsep}{0pt}  
\addtolength{\intextsep}{-10pt} 
\begin{algorithm}[H]\small
\caption{Compute $\bigl\{\widetilde{\mathcal{M}}^{(n)}_\rho(i)\bigr\}_{i=0}^{n}$
        and the first merger step $i^{\star}$}
\label{alg:return-M-and-istar}
\begin{algorithmic}[1]
    \Require $p_{\text{data}}$, steps $n$, forward sampler $\mathsf{PF}$, task objective $\mathcal{L}$, fluctuation $F^{(n)}_\rho$,
            metric $d$, threshold $\varepsilon$
    \State $p_{\mathrm{desired}} \gets \textsc{DefineDesired}(p_{\text{data}},\mathcal{L})$
    \State $(\Omega_{1,0},\Omega_{2,0}) \gets \textsc{PickEvents}(p_{\mathrm{desired}},\mathcal{L})$
    \State $\mathbf{M} \gets [\,]$  \Comment{dynamic array for $\mathcal{M}^{(n)}_\rho(i)$}
    \State $i^\star \gets n$        \Comment{merge no later than white noise}
    \For{$i = 0$ \textbf{to} $n$}
        \If{$i > 0$}
            \State $(\Omega_{1,i},\Omega_{2,i}) \gets
                   \textsc{ForwardStep}(\Omega_{1,i-1},\Omega_{2,i-1})$
        \Else
            \State $(\Omega_{1,i},\Omega_{2,i}) \gets (\Omega_{1,0},\Omega_{2,0})$
        \EndIf
        \State $\delta \gets d \bigl(F_\rho^{(n)}(\Omega_{1,i}),
                                     F_\rho^{(n)}(\Omega_{2,i})\bigr)$
        \State $\displaystyle
              \mathcal{M}^{(n)}_\rho(i) \gets
              \begin{cases}
                \mathcal{M}^{(n)}_\rho(\Omega_{1,i},\Omega_{2,i}),
                             & \delta>\varepsilon,\\
                1, & \delta\le\varepsilon
              \end{cases}$
        \State append $\widetilde{\mathcal{M}}^{(n)}_\rho(i)$ to $\mathbf{M}$
        \If{$\widetilde{\mathcal{M}}^{(n)}_\rho(i)=1$ \textbf{and} $i^\star=n$}
            \State $i^\star \gets i$  \Comment{first step where events merge}
        \EndIf
    \EndFor \\
    \Return $\mathbf{M},\; i^\star$
\end{algorithmic}
\label{alg:template}
\end{algorithm}

\section{Demonstrations of our methodology }
\label{sec:experiments}

We next present case studies that show how our machinery can diagnose and subsequently improve the
behaviour of diffusion samplers on practical tasks.\emph{ Our aim is to illustrate simple recipes for working with user-defined events that need not require a deep dive into the underlying theory but at the same time showcase the utility of our framework.}

\subsection{Warm-up: Predicting convergence of the data distribution}
\label{sec:warmup}
In this case study, we abuse notation and let the sequence $\{p_{0},\dots,p_{n}\}$ represent the marginals under the PF-ODE for a data distribution $p_{0}.$ Assume that the support of the data distribution,
$\mathcal{D}_{0}=\operatorname{supp}(p_{0})$,  is \emph{essentially disjoint} from the support of
the Gaussian endpoint,
$\mathcal{D}_{n}=\operatorname{supp}(p_{n})
                 =\operatorname{supp}\bigl(\mathcal{N}(\mathbf{0}_{d},\mathbf{I}_{d})\bigr)$.
In high dimensions, this is realistic because
$\mathcal{D}_{n}\approx\sqrt{d}\,S^{d-1}$ and therefore has asymptotically
negligible volume~\citep{vershynin2018high}%
\footnote{%
Note that, $\mathcal{D}_{n}$ becomes a null set for $d\to\infty$, so overlap
with any finite-volume data manifold is vanishingly unlikely. Concentration inequalities showing rapid decay as a function of increasing $n$ can be found, e.g., in~\citep{vershynin2018high}. Formally, we assume that if $\mathcal{D}^{\ast} = \mathcal{D}_{0} \cup \mathcal{D}_{n}$, then $P_{0}(\mathcal{D}^{\ast}), P_{n}(\mathcal{D}^{\ast}) \to 0$.}

We now show the setup of \Cref{alg:template} in this setting. Let $p_{\mathrm{desired}}$ the desired distribution be defined over the union of all supports $\cup_{i=0}^{n}\mathcal{D}_{n}$ such that,
\begin{equation}
  P_{\mathrm{desired}}(Z)=
    \begin{cases}
      P_{0}(Z), & Z\subseteq\mathcal{D}_{0},\\[4pt]
      0,        & \text{otherwise},
    \end{cases}
\end{equation}
where $P_{\mathrm{desired}},P_{0}$ are the probability masses associated with $p_{\mathrm{desired}},p_{0}$ respectively. It is easy to see that the PF-ODE on $\hat{p}_{0}=p_{\mathrm{desired}}$ leads to the set of marginals that we term as an \textbf{augmented process}, $\text{aug}=\{\hat{p}_{i}\}_{i=0}^{n}$ defined as,
\begin{equation}
  \hat{P}_{i}(Z)=
    \begin{cases}
      P_{i}(Z), & Z\subseteq\mathcal{D}_{i},\\[4pt]
      0,        & \text{otherwise},
    \end{cases}
\end{equation}
where $P_{i}$ (resp.\ $\hat{P}_{i}$) is the probability measure associated
with $p_{i}$ (resp.\ $\hat{p}_{i}$).
Operationally, \text{aug} coincides with the original PF-ODE but with an
explicitly enlarged state space.\footnote{The above construction is designed to measure how probability mass transfers away from any marginal $p_{i}$ towards another marginal $p_{j}$ where $j > i$. In our work we only concern ourselves with the case where $j=n$ so as to probe convergence to stationarity.}

Set $\Omega_{1,0}=\mathcal{D}_{0}$ and $\Omega_{2,0}=\mathcal{D}_{n}$.  Because
$\Omega_{2,0}$ is already the support of the stationary distribution,
$\Omega_{2,i}=\Omega_{2,0}$ for all $i>0$.  Detecting the smallest
index $i$ for which
$M^{(n)}_{\rho,\Omega_{1},\Omega_{2}}(i)\!\approx\!1$
therefore amounts to  \emph{deducing the onset of convergence} to the stationary distribution, this is in practice equivalent to a multivariate normality test on $p_{i}$.
We adopt the D'Agostino-Pearson omnibus test
\citep{4fb818aa-ec5d-3643-a3a8-c3b27442e46c,272b2fa8-f3b3-371d-b34e-a4c6938458a1}, the $p$-value threshold is the conventional $0.05$. We evaluate MNIST~\citep{deng2012mnist}, CIFAR-10~\citep{cifar10}, and a
compressed INT-8 variant of ImageNet~\citep{deng2009imagenet,imagenet_int8},
using the popular DDPM noise schedule~\citep{ho2020denoisingdiffusionprobabilisticmodels}.
Results appear in \Cref{fig:ddpm-norm} in \Cref{sec:conv}.  
\paragraph{Acceleration via early stopping.}
Let $i^{\star}$ denote the smallest index flagged as Gaussian by the test. Starting the \emph{reverse} sampler from $t=i^{\star}$, rather than the
conventional $t=n$, preserves visual quality while saving $n-i^{\star}$ steps,
as summarised in \Cref{tab:ddpm-acc} across datasets. By convention, throughout our work a $(\downarrow)$ sign denotes that lower values are better and vice-versa for an $(\uparrow)$.  For ImageNet we employ the
state-of-the-art DiT-XL/2 model at $512\times512$
resolution~\citep{Peebles2022DiT}; open-source checkpoints are used for the
other datasets. A comparable observation was reported by \citet{raya2024spontaneousiop}, who
derive $i^{\star}$ analytically under stronger assumptions on the data. Intriguingly, an
unrelated theorem on Brownian equilibrium time predicts
$i^{\star}$ surprisingly well for our data in \Cref{sec:vp-mixing}; clarifying this connection is left
to future work, but the theorem already provides a useful heuristic.


\begin{table}[H]
  \centering
  \small
  \begin{tabular}{|l|c|c|c|}
    \hline
    \rowcolor[gray]{0.8}
    \textbf{Model / Dataset} & \textbf{FID}$(\downarrow)$ & \textbf{Steps}$(\downarrow)$ & \textbf{GFLOPs}$(\downarrow)$ \\ \hline
    DiT-XL/2 (ImageNet, full) & $3.42\pm0.21$ & 250 & 4100 \\ \hline
    DiT-XL/2 (ImageNet, ours) & $\mathbf{3.37}\pm0.31$ & \textbf{175} & \textbf{2870} \\ \hline
    DDPM (MNIST, full)        & $2.27\pm0.19$ & 1000 & 2000 \\ \hline
    DDPM (MNIST, ours)        & $2.29\pm0.17$ & \textbf{600} & \textbf{1200} \\ \hline
    DDPM (CIFAR-10, full)     & $3.62\pm0.35$ & 500  & 6000 \\ \hline
    DDPM (CIFAR-10, ours)     & $\mathbf{3.47}\pm0.34$ & \textbf{300} & \textbf{3600} \\ \hline
  \end{tabular}
  \vskip 0.1in 
  \caption{ Acceleration achieved by stopping reverse diffusion at
           $t=i^{\star}$ instead of $t=n$ (DDPM schedule).  FID scores are
           averaged over three runs with $95\%$ confidence intervals. }
  \label{tab:ddpm-acc}
\end{table}
\emph{Thus, starting the reverse sampler from the observed convergence 
time yields the same perceptual quality while reducing wall-clock cost—
often by thousands of gigaflops (GFLOPs)}.

\begin{figure}[h!]
    \centering
    \captionsetup{font=small, labelfont=bf}
    \includegraphics[width=0.6\textwidth]{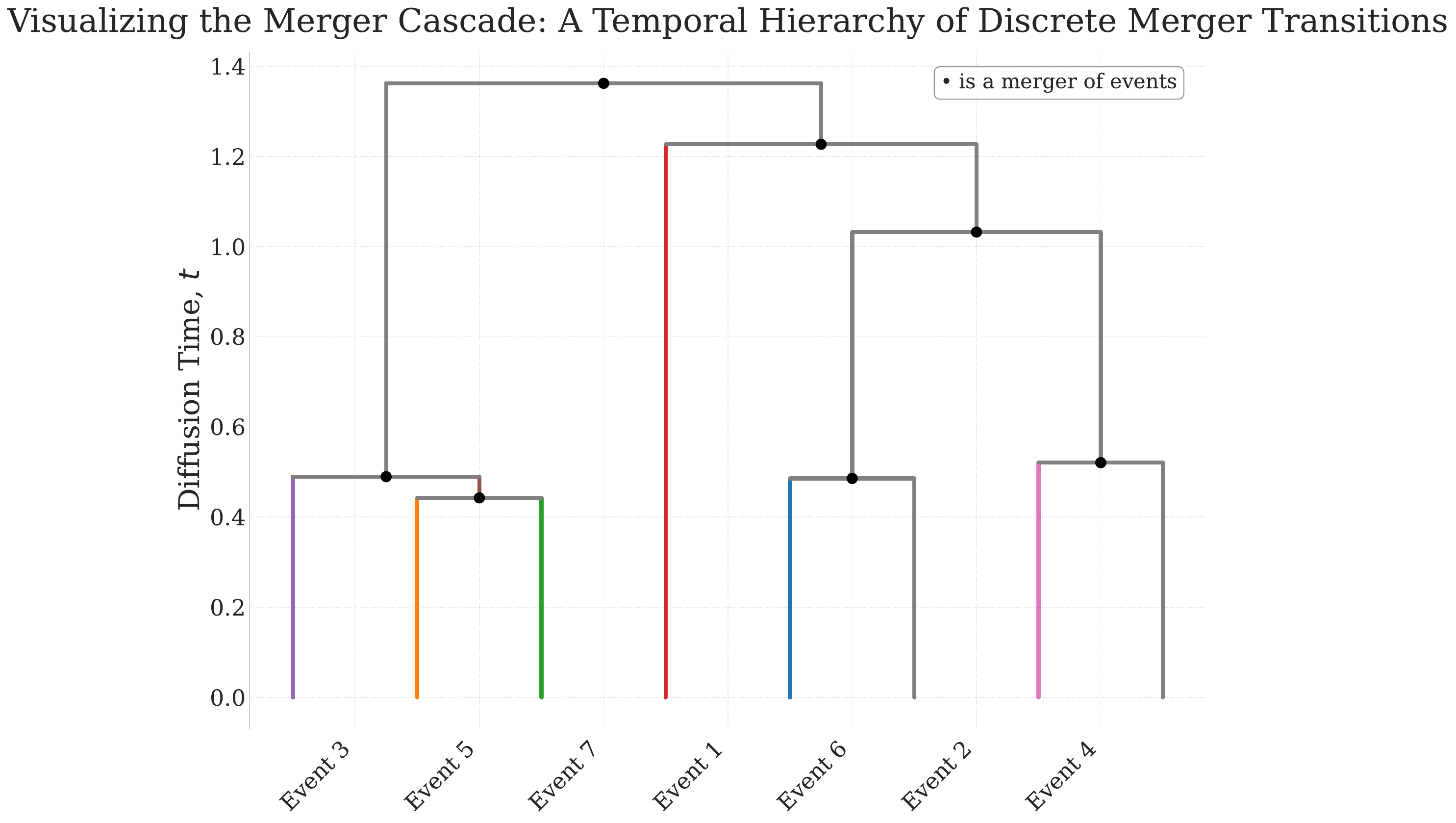}
    \caption{\textbf{Visualizing the merger cascade.} This temporal hierarchy illustrates how distinct, mutually disjoint events merge as the diffusion process evolves. Time $t$ flows upward. At $t=0$, events are distinct leaves. As they diffuse, pairs whose fluctuation tensors become indistinguishable undergo a discrete merger event (black dot), and their branches merge. This cascade continues until all discriminating information has vanished into a single cluster.}
    \label{fig:merger_cascade}
\end{figure}

\paragraph{Visualizing the merger cascade.}
Before detailing our next set of applications, it is instructive to visualise mergers occurring over a disjoint partition of events. \Cref{fig:merger_cascade} presents the ``merger cascade,'' a temporal hierarchy visualizing how disjoint events merge over diffusion time $t$. Events start as distinct leaves at $t=0$. As time progresses upwards, branches join at a black dot---a \emph{discrete merger event} (\Cref{sec:phase-transit})---at the precise moment their fluctuation tensors become indistinguishable. This dendrogram provides an intuitive map of how structural similarities are lost, motivating the targeted interventions that follow.

\subsection{Class-conditional generation}
\label{sec:class-cond}

We now study the behaviour of \Cref{alg:template} when $p_{\text{desired}}$ is just the data split
into $K$ \emph{mutually disjoint classes}
$\{\Omega_{1,0},\dots,\Omega_{K,0}\}$ such that  
$\bigcup_{k=1}^{K}\Omega_{k,0}=\Omega$ where $\Omega$ is the sample space of the data distribution
and $\Omega_{k,0}\cap\Omega_{\ell,0}=\varnothing$ whenever $k\neq\ell$.
Throughout we write $\Omega_{k,t}$ for the image of $\Omega_{k,0}$ under the
forward diffusion map at (discrete) time~$t$. A formal algorithm summarizing this modification is present in \Cref{alg:class}.

\paragraph{A second-order statistic via centred kernel alignment.}
Higher-order fluctuations for vector states become unwieldy, so we fix $n=2$, leveraging covariance matrices to efficiently capture fourth-order moments. This captures non-linear feature information while maintaining second-order structure (\Cref{sec:cka}). For completeness, we also demonstrate an experiment with higher-order fluctuations, simplified using Isserlis'/Wick's theorem \citep{isserlis_1918_1431593,wick1950evaluation} (\Cref{sec:hi-ord-fluc}). Let us for the $n=2$ case define,
\begin{equation}
    \Sigma_{k,t}\;=
              \operatorname{Cov}\bigl(\rho\mid\Omega_{k,t}\bigr)
              \in\mathbb{R}^{d\times d}.
\label{eq:sigma-t}
\end{equation}

The normalised second-order cross-fluctuation $\mathcal{M}^{(2)}_{\rho}(\Omega_{k,t},\Omega_{\ell,t})$ is then the \emph{centred kernel alignment} (CKA)
between $\Sigma_{k,t}$ and $\Sigma_{\ell,t}$
\citep{cortes2012algorithms,cristianini2001kernel,kornblith2019similarity}:
\begin{equation}
\label{eq:cka-simplified}
\text{CKA}_{k\ell}(t)=
\frac{\operatorname{Tr}\bigl(\Sigma_{k,t}^{\!\top}\Sigma_{\ell,t}\bigr)}
     {\sqrt{\operatorname{Tr} \bigl(\Sigma_{k,t}^{\!\top}\Sigma_{k,t}\bigr)
            \operatorname{Tr}\bigl(\Sigma_{\ell,t}^{\!\top}\Sigma_{\ell,t}\bigr)}}.
\end{equation}
where $G^{(2)}_{\rho}(\Omega_{k,t},\Omega_{l,t}) = \operatorname{Tr} \bigl(\Sigma_{k,t}^{\!\top}\Sigma_{\ell,t}\bigr), \widehat{F}^{(4)}_{\rho}(\Omega_{k,t})= \operatorname{Tr} \bigl(\Sigma_{k,t}^{\!\top}\Sigma_{k,t}\bigr), \widehat{F}^{(4)}_{\rho}(\Omega_{l,t})= \operatorname{Tr} \bigl(\Sigma_{l,t}^{\!\top}\Sigma_{l,t}\bigr)$. We convert the above definition \eqref{eq:cka-simplified} into a discrete phase transition indicator \eqref{eq:time-dependent-M} detecting whether the events have “merged / not merged’’ by the rule  
\begin{equation}
\mathcal{M}^{(2)}_{k\ell}(t)=
\begin{cases}
\text{CKA}_{k\ell}(t), & 
  \text{if } d\bigl(\widehat{F}^{(2n)}_\rho(\Omega_{1,i}), \widehat{F}^{(2n)}_\rho(\Omega_{2,i})\bigr) \! > \varepsilon,\\
1, & \text{otherwise},
\end{cases}
\label{eq:mk-l}
\end{equation}
Because the forward SDE is isotropic, $\Sigma_{k,t}$ contracts in the radial
direction only; its spectrum therefore collapses towards a scaled identity.
Hyper-contractivity implies that the merger of two classes is governed by the
largest eigenvalues (See \Cref{sec:cka} for further details)  
\[
\lambda_{k}^{\max}(t)=\lambda_{\max}(\Sigma_{k,t}),
\qquad
\lambda_{\ell}^{\max}(t)=\lambda_{\max}(\Sigma_{\ell,t}),
\]
so we may replace the trace metric by  
$d\bigl(\lambda_{k}^{\max}(t),\lambda_{\ell}^{\max}(t)\bigr)
      <\varepsilon$.
In practice, we fix  
\[
\varepsilon \approx \frac{\max_{k}\lambda_{k}^{\max}(0)}{400},
\]
and use absolute eigenvalue differences for~$d$.
This criterion yields clear merger times across datasets;
trajectories  are plotted in \Cref{sec:class-cond-exp}.

\paragraph{Practical payoff: class-wise \emph{Interval guidance}.}
The state-of-the-art guidance algorithm, Interval guidance (IG) \citep{kynkäänniemi2024applyingguidancelimitedinterval}
argues that class conditioning should be applied only for a sub-interval 
$t\in(t_{\text{start}},t_{\text{end}})$ for optimal results.
The reference implementation runs an expensive grid search as detailed in \Cref{sec:class-cond-exp}. Our fluctuation analysis supplies both bounds for free:
\begin{itemize}
\item $t_{\text{start}}=i^{\star}$, the convergence index from
      \Cref{sec:warmup};
\item $t_{\text{end}}=\max\bigl\{t:
      \mathcal{M}^{(2)}_{k\ell}(t)<1\text{ for some }\ell\neq k\bigr\}$.
\end{itemize}
Using these class-specific windows, we match or improve IG performance on
ImageNet, CIFAR-10, and MNIST datasets. Our method is capable of cutting the
combinatorial search cost by orders of magnitude. We present the results in Table \ref{tab:class-cond}. We utilise the metrics Frechet Inception Distance (FID) \cite{heusel2017gans}, Precision, Recall, and their more suitable variants for generative models such as Density and Coverage \cite{naeem2020reliable}. (We also show results for the StableDiffusion \cite{rombach2022highresolutionimagesynthesislatent} model using Imagenet and the fine-grained classification dataset OxfordIIITPets in \Cref{sec:class-cond-exp} along with further results.)
\vskip 0.1in

\begin{table*}[ht]
  \centering
  \small
  \resizebox{\textwidth}{!}{%
    \begin{tabular}{|l|c|c|c|c|c|}
      \hline
      \rowcolor[gray]{0.8}
      \textbf{Model/Dataset} & \textbf{FID} ($\downarrow$) & \textbf{Precision} ($\uparrow$) &
      \textbf{Recall} ($\uparrow$) & \textbf{Density} ($\uparrow$) & \textbf{Coverage} ($\uparrow$) \\ \hline
      DiT-XL/2 (Imagenet, IG Baseline) & $3.22 \pm 0.16$ & $0.78 \pm 0.01$ &
      $0.23 \pm 0.05$ & $0.83 \pm 0.01$ & $0.35 \pm 0.02$ \\ \hline
      DiT-XL/2 (Imagenet, IG Ours) & $\mathbf{2.86 \pm 0.15}$ & $\mathbf{0.83 \pm 0.02}$ &
      $\mathbf{0.26 \pm 0.04}$ & $\mathbf{0.85 \pm 0.01}$ & $\mathbf{0.39 \pm 0.02}$ \\ \hline
      DDPM (MNIST, IG Baseline) & $2.15 \pm 0.06$ & $0.80 \pm 0.02$ &
      $0.25 \pm 0.01$ & $0.85 \pm 0.01$ & $0.36 \pm 0.02$ \\ \hline
      DDPM (MNIST, IG Ours) & $\mathbf{1.99 \pm 0.11}$ & $\mathbf{0.85 \pm 0.01}$ &
      $\mathbf{0.28 \pm 0.03}$ & $\mathbf{0.89 \pm 0.02}$ & $\mathbf{0.40 \pm 0.01}$ \\ \hline
      DDPM (CIFAR10, IG Baseline) & $3.32 \pm 0.25$ & $0.77 \pm 0.01$ &
      $0.19 \pm 0.14$ & $0.81 \pm 0.03$ & $0.32 \pm 0.02$ \\ \hline
      DDPM (CIFAR10, IG Ours) & $\mathbf{3.01 \pm 0.14}$ & $\mathbf{0.79 \pm 0.02}$ &
      $\mathbf{0.22 \pm 0.01}$ & $\mathbf{0.84 \pm 0.01}$ & $\mathbf{0.35 \pm 0.04}$ \\ \hline
    \end{tabular}
  }
  \caption{Class conditional generation}
  \label{tab:class-cond}
\end{table*}

\emph{Monitoring when class supports merge lets us place guidance exactly where it helps, eliminating costly grid searches
and boosting class-conditional quality with negligible overhead.}

\subsection{Rare-class generation}
\label{sec:rare-class}

We reuse the same modification of \Cref{alg:template} as in \Cref{sec:class-cond} but focus on \emph{tail
classes}-categories under-represented in the training data, and therefore
poorly synthesised by off-the-shelf diffusion models.
Concretely, we study the following datasets,

\begin{itemize}
\item \textbf{CUB-200} (200 bird species), and
\item \textbf{iNaturalist 2019} (fine-grained flora \& fauna).
\end{itemize}

Both have been identified as failure cases for Stable Diffusion
\citep{samuel2024generating}.  
For each species $k$, we denote the source event by
$\Omega_{k,0}$ and its forward image at timestep $t$ by $\Omega_{k,t}$.

\paragraph{Merger-aware guidance window.}
Adopting the notation of \Cref{sec:warmup,sec:class-cond},
let 
\[
t_{\text{conv}} \;=\; i^{\star},
\qquad
t_{\text{merge},k} \;=\;
\min\bigl\{\,t:\,
        \widetilde{\mathcal{M}}^{(2)}_{\rho}(\Omega_{k,t},\Omega_{\ell,t})=1
        \text{ for some } \ell\neq k
      \bigr\}.
\]
The class-specific guidance interval is then 
$(t_{\text{start},k},\,t_{\text{end},k})
\;=\;
\bigl(t_{\text{conv}},\,t_{\text{merge},k}\bigr),
$
that is, from global data convergence up to-but not beyond-the first time
the target class becomes indistinguishable from any other class.
\paragraph{Interval guidance with corrupted exemplars.}
We observe that for this setting as compared to naive interval guidance, a slight tweak using corrupted exemplars can yield better results \footnote{We observed no appreciable gain for the setup in \Cref{sec:class-cond}, hence omit this tactic there. }. Inside this window, we apply
\emph{Interval Guidance} \citep{kynkäänniemi2024applyingguidancelimitedinterval}
and interpolate with a single exemplar\,
$x_{\text{ref}}\!\sim\!\Omega_{k,0}$ that is forward-noised in lock-step with
the individual sample being generated.  The conditioning term is effectively therefore the pair
$\bigl(z_{t},\,x_{\text{ref},t}\bigr)$, making the scheme a
class-specific variant of ILVR \citep{choi2021ilvrconditioningmethoddenoising} with intervals.
No additional hyper-parameters are introduced; we keep the noise schedule and
scaling constants from \Cref{sec:class-cond}. A formal version of the algorithm is presented in \Cref{alg:opt-intp}, refer to \Cref{sec:rare-class-app} for further details.

\Cref{tab:cond-rare} summarises the results using standard metrics, where we compare  three settings:  
\textit{(i)} naïve Interval Guidance over the full horizon,  
\textit{(ii)} merger-aware Interval Guidance as in \Cref{sec:class-cond} and  
\textit{(iii)} merger-aware IG + corrupted exemplar (ours).
Across metrics, the merger-aware schedules consistently outperform the
baselines, with the exemplar variant giving the largest gains. 

We conclude that \emph{as fluctuation-driven merger times generalise from mainstream classes to the tail
classes without re-tuning, leveraging them yields a lightweight yet effective
strategy for rare-class synthesis, mitigating class imbalance at essentially
zero extra computational cost.}


\begin{table*}[t]
  \centering
  \small
  \resizebox{\textwidth}{!}{%
    \begin{tabular}{|l|c|c|c|c|c|}
      \hline
      \rowcolor[gray]{0.8}
      \textbf{Model/Dataset} & \textbf{CLIP Similarity} ($\uparrow$) & \textbf{Precision} ($\uparrow$) &
      \textbf{Recall} ($\uparrow$) & \textbf{Density} ($\uparrow$) & \textbf{Coverage} ($\uparrow$) \\ \hline
      SD (iNaturalist, IG baseline)      & $0.21 \pm 0.03$ & $0.70 \pm 0.01$ & $0.15 \pm 0.01$ & $0.75 \pm 0.01$ & $0.27 \pm 0.02$ \\ \hline
      SD (iNaturalist, IG Ours)          & $0.24 \pm 0.02$ & $0.73 \pm 0.01$ & $0.18 \pm 0.01$ & $0.78 \pm 0.02$ & $0.30 \pm 0.04$ \\ \hline
      SD (iNaturalist, IG-ILVR Ours)     & $\mathbf{0.27 \pm 0.01}$ & $\mathbf{0.76 \pm 0.01}$ & $\mathbf{0.19 \pm 0.02}$ & $\mathbf{0.81 \pm 0.01}$ & $\mathbf{0.31 \pm 0.03}$ \\ \hline
      SD (CUB200, IG baseline)           & $0.24 \pm 0.05$ & $0.75 \pm 0.01$ & $0.14 \pm 0.01$ & $0.79 \pm 0.02$ & $0.30 \pm 0.02$ \\ \hline
      SD (CUB200, IG Ours)               & $0.26 \pm 0.01$ & $0.78 \pm 0.05$ & $0.17 \pm 0.01$ & $0.82 \pm 0.02$ & $\mathbf{0.33 \pm 0.01}$ \\ \hline
      SD (CUB200, IG-ILVR Ours)          & $\mathbf{0.27 \pm 0.02}$ & $\mathbf{0.82 \pm 0.02}$ & $\mathbf{0.18 \pm 0.02}$ & $\mathbf{0.85 \pm 0.01}$ & $0.31 \pm 0.02$ \\ \hline
    \end{tabular}
  }
  \vskip 0.05in
  \caption{Rare class generation using StableDiffusion}
  \label{tab:cond-rare}
\end{table*}


\subsection{Zero-shot classification}
\label{sec:zero-shot}

Once again we continue with a setting of \Cref{alg:template} as is in \Cref{sec:class-cond}:  
where each sample $z\!\sim\!p_{0}$ belongs to exactly one class event
$\Omega_{k,0}$.
Given a trained \emph{class-conditional} diffusion model
$f_{\theta}$ with label index set $\Lambda=\{1,\dots,K\}$,
the goal is to predict  
\[
k^{\star}
\;=\;
\arg\max_{k\in\Lambda}\;p_{k}(z),
\]
where $p_{k}(z)$ estimates the posterior class probability. We build upon the procedure proposed by \cite{li2023diffusionmodelsecretlyzeroshot} for this task. Let $z_{t}(z,\varepsilon)\in\mathbb{R}^{d}$ denote the forward-noised
version of $z$ at step $t$, obtained with seed
$\varepsilon\sim\mathcal{N}(\mathbf{0},\mathbf{I})$.
For that configuration, define the \emph{logit}
\[
s_{k}(t,\varepsilon)
\;=\;
-\bigl\|
      f_{\theta}\bigl(z_{t}(z,\varepsilon),\varepsilon,k\bigr)
      -\varepsilon
  \bigr\|_{2}^{2},
\qquad k\in\Lambda.
\]
With $N$ noise seeds
$\{\varepsilon_{n}\}_{n=1}^{N}$ and weights $w(t)$,
our importance-weighted class score becomes
\begin{equation}
\label{eq:zs-classifier-split}
p_{k}(z)
\;=\;
\sum_{t=t_{\mathrm{start},k}}^{t_{\mathrm{stop},k}}
      w(t)\;\Bigl[\;
      \frac1N\sum\nolimits_{n=1}^{N}
          q_{k}\bigl(t,\varepsilon_{n}\bigr)
      \Bigr],\quad q_{k}(t,\varepsilon)
\;=\;
\frac{\exp\bigl(s_{k}(t,\varepsilon)\bigr)}
     {\sum_{\ell\in\Lambda}\exp\bigl(s_{\ell}(t,\varepsilon)\bigr)}.
\end{equation}
where $ q_{k}(t,\varepsilon)$ is the corresponding softmax probability. The choices \textbf{uniform}, \textbf{inverse-SNR}
and \textbf{truncated inverse-SNR} correspond to three concrete $w$ distributions,
each summing to~1 on its support. For the inverse-SNR strategies, the weights at each time are proportional to the inverse of the Signal to Noise Ratio (SNR) \citep{kingma2021variational} at that time,  while uniform assigns a constant weight. The lower bound is set to $t_{\mathrm{start},k} = 0 \; \forall k$ for the uniform and inverse-SNR strategies. For the truncated case we set $t_{\mathrm{start},k} = 20$ for all $k$; (\Cref{sec:zsc}) the upper bound
\[
t_{\mathrm{stop},k}
\;=\;
\min\bigl\{
  t : \widetilde{\mathcal{M}}^{(2)}_{\rho}(\Omega_{k,t},\Omega_{\ell,t})=1
      \text{ for some } \ell\neq k
\bigr\}
\]
ensures that no timestep beyond the
first class merger contributes to \eqref{eq:zs-classifier-split}. A formal version of the above algorithm is present in \Cref{alg:zs-cl} along with further details in \Cref{sec:zsc}. 
Using a Stable Diffusion backbone
\citep{rombach2022highresolutionimagesynthesislatent},
we obtain the accuracies in \Cref{tab:class-cond}.
Merger-aware weighting improves on the uniform baseline of
\citet{li2023diffusionmodelsecretlyzeroshot}, where $t_{\text{start},k}=0$ and $t_{\mathrm{stop},k}=T$ for all classes $k$, where $T$ is the horizon of the diffusion process; truncated inverse-SNR performs best.
\vskip 0.05in 
\begin{table}[H]
  \centering
  \small
  \begin{tabular}{|l|c|c|c|}
    \hline
    \rowcolor[gray]{0.8}
    \textbf{Method} &
    \textbf{ImageNet} $\uparrow$ &
    \textbf{CIFAR-10} $\uparrow$ &
    \textbf{Oxford-IIIT Pets} $\uparrow$
    \\ \hline
    SD, uniform ($\,$\citeauthor{li2023diffusionmodelsecretlyzeroshot}$)$
      & $54.96\!\pm\!0.67$ & $84.67\!\pm\!1.23$ & $82.87\!\pm\!0.39$ \\ \hline
    SD, uniform (ours)
      & $57.91\!\pm\!0.53$ & $85.17\!\pm\!0.17$ & $86.17\!\pm\!0.26$ \\ \hline
    SD, inverse-SNR
      & $64.17\!\pm\!0.33$ & $87.26\!\pm\!0.67$ & $88.17\!\pm\!0.29$ \\ \hline
    SD, trunc.\ inverse-SNR
      & $\mathbf{65.28\!\pm\!0.46}$ &
        $\mathbf{88.38\!\pm\!0.43}$ &
        $\mathbf{89.15\!\pm\!0.26}$ \\ \hline
    CLIP RN-50
      & $58.41\!\pm\!0.35$ & $75.42\!\pm\!0.26$ & $85.61\!\pm\!0.29$ \\ \hline
    OpenCLIP ViT-H/14
      & $76.91\!\pm\!0.75$ & $96.87\!\pm\!0.59$ & $94.61\!\pm\!0.37$ \\ \hline
  \end{tabular}
  \vskip 0.05in 
  \caption{Zero-shot multi-class accuracy (\%); $\pm$\,95\% CI over five runs.}
  \label{tab:class-acc}
\end{table}
 
That truncated inverse-SNR outperforms uniform weighting suggests the corrupted
marginals remain most discriminative shortly \emph{before} class supports
merge.  \Cref{sec:bin-clf} analyses this effect in detail via linear probes for binary classification between Imagenet classes. 

\emph{Thus, merger times not only optimise guidance but also delimit the timesteps that
matter for zero-shot recognition, yielding tangible gains at negligible cost.}


\subsection{Zero-shot style transfer}
\label{sec:zero-shot-style}

 Let $p_{0}$ be a \emph{source} image distribution on $\mathcal{X}\subset\mathbb{R}^{d}$
and let $p^{\star}=\mathcal{T}_{\!\text{style}}(p_{0})$ be the same semantic
content rendered in a new artistic style.  We assume

\begin{enumerate}
\item \textbf{Bijectivity.}  $\mathcal{T}_{\!\text{style}}:\mathcal{X}\to\mathcal{X}$
      is invertible on the supports, so $p^{\star}$ is a valid density and
      $\operatorname{supp}(p^{\star})=\mathcal{T}_{\!\text{style}}\!\bigl(\operatorname{supp}(p_{0})\bigr)$.

\item \textbf{Structure preservation in the Fourier domain.}  
      For a density $q$ write
      $\widehat{q}(\boldsymbol{\xi})=\int_{\mathcal{X}}q(\mathbf{x})\,e^{-2\pi i\boldsymbol{\xi}^{\top}\mathbf{x}}\mathrm{d}\mathbf{x}$
      and define the frequency-domain norm
      $
        \lVert p-q\rVert_{f}
        =\lVert\widehat{p}-\widehat{q}\rVert_{L^{2}(\mathbb{R}^{d})}.
      $
      We posit the regularity bound
      \begin{equation}
      \label{eq:fourier-regularity}
        \lVert p_{0}-p^{\star}\rVert_{f}\;\le\;\delta,
        \qquad \text{with }0<\delta\ll1.
      \end{equation}
\end{enumerate}

\paragraph{Fluctuation adaptation lemma for style transfer.}
Let $\rho(\mathbf{x})=\mathbf{x}$ be the identity state operator for $p_{0}$
and let $\rho^{\star}=\rho\circ\mathcal{T}_{\!\text{style}}^{-1}$ for
$p^{\star}$.  \Cref{sec:reg-fluc} proves that for every measurable
$\Omega\subseteq\mathcal{X}$ and each moment order $n\in\{1,2,3,4\}$,
\begin{equation}
\bigl|
  \widehat{F}^{(n)}_{\rho}(\Omega)\;-\;
  \widehat{F}^{(n)}_{\rho^{\star}}\!\bigl(\mathcal{T}_{\!\text{style}}(\Omega)\bigr)
\bigr|
\;\le\;C_{n}\,\delta,
\label{eq:st-tf}
\end{equation}
with a constant $C_{n}$ independent of~$\Omega$.
Hence the fluctuation trajectories of the \emph{source} distribution
$p_{0}$ approximate those of the \emph{target-style} distribution $p^{\star}$
to $O(\delta)$ accuracy.

\paragraph{Choosing a start time for reverse denoising.}
For each semantic class $\lambda\in\Lambda$ let
\[
m_{\lambda}
\;=\;
\min\Bigl\{t:
  \widetilde{\mathcal{M}}^{(2)}_{\rho}\bigl(\Omega_{\lambda,t},\Omega_{\ell,t}\bigr)=1
  \text{ for some }\ell\neq\lambda
\Bigr\}
\]
be its earliest merger time under the \emph{forward} diffusion of $p_{0}$ (see
\Cref{sec:class-cond}).  By the lemma above, $m_{\lambda}$ remains a valid
approximation for the target-style chain.  Therefore, when we apply a
target-style model (e.g.\ Stable Diffusion fine-tuned on van-Gogh paintings)
we can \textbf{kick-start} reverse denoising at $t=m_{\lambda}$ rather than at
the full horizon $t=n$, achieving style transfer in fewer steps. Thus $p_{\mathrm{desired}}$ in \Cref{alg:template} here is the \emph{source distribution} while the diffusion model is actually trained on a different \emph{target distribution}. A formal version of the above algorithm is presented in \Cref{alg:zs-st}. We also note that the current procedure naturally extends to inverse problems using diffusion models, following a similar procedure under conditions analogous to \eqref{eq:st-tf}. These conditions align with prior methodologies for inverse problems \cite{song2022solvinginverseproblemsmedical}. Exploring these extensions is left for future work 

We follow the setup of~\citep{meng2021sdedit} wherein an optimal time needs to be estimated till which noising must occur. Our baseline is the grid search technique used in ~\citep{meng2021sdedit} computed over an entire dataset. We present sample results for the Studio Ghibli \cite{miyazaki_ghibli_2014} and Van Gogh \cite{vanrooijen_vangogh_2019} artistic styles in Table \ref{tab:stf} using the PSNR and MSE metrics. See \Cref{sec:zero-shot-app} for experimental details, figures and empirical verification of the fluctuation adaptation lemma, and additional results for the Elden Ring \cite{eldenring_2022} and Arcane \cite{arcane_2021} styles.  
\vskip 0.05in 
\begin{table}[H]
\centering
\small
\begin{tabular}{|l|cc|cc|}
\hline
\rowcolor[gray]{0.8}
Style & \multicolumn{2}{l|}{Ghibli} & \multicolumn{2}{l|}{van-Gogh} \\ \hline
\rowcolor[gray]{0.9}
Models/Metrics & \multicolumn{1}{l|}{PSNR ($\uparrow$)} & MSE ($\downarrow$) & \multicolumn{1}{l|}{PSNR ($\uparrow$)} & MSE ($\downarrow$) \\ \hline
SD Edit (OxfordIIITPets) & \multicolumn{1}{l|}{$25.67 \pm 1.14$} & $0.09 \pm 0.001$ & \multicolumn{1}{l|}{$26.16 \pm 0.72$} & $0.09 \pm 0.003$ \\ \hline
\textbf{Ours} (OxfordIIITPets) & \multicolumn{1}{l|}{$\mathbf{28.71 \pm 0.86}$} & $\mathbf{0.03 \pm 0.005}$ & \multicolumn{1}{l|}{$\mathbf{28.65 \pm 0.49}$} & $\mathbf{0.03 \pm 0.002}$ \\ \hline
SD Edit (AFHQv2) & \multicolumn{1}{l|}{$27.12 \pm 0.59$} & $0.05 \pm 0.006$ & \multicolumn{1}{l|}{$27.49 \pm 0.27$} & $0.04 \pm 0.004$ \\ \hline
\textbf{Ours} (AFHQ v2) & \multicolumn{1}{l|}{$\mathbf{27.65 \pm 0.57}$} & $\mathbf{0.04 \pm 0.004}$ & \multicolumn{1}{l|}{$\mathbf{28.07 \pm 0.32}$} & $\mathbf{0.03 \pm 0.006}$ \\ \hline
\end{tabular}
\vskip 0.1in 
\caption{Style Transfer results for Ghibli and van-Gogh styles }
\label{tab:stf}
\end{table}

\emph{Thus, as under mild Fourier regularity, the merger schedule learned on a source
dataset transfers directly to any purely stylistic variant of that dataset, the result is a true zero-shot style transfer procedure that inherits the
efficiency gains of \Cref{sec:warmup} without additional tuning.}

\section{Conclusion}
\label{sec:conclusion}

We introduced \emph{cross-fluctuation mergers}: a centred-moment statistic
that fires precisely when two regions of a diffusion trajectory become
indistinguishable. The concept links statistical-physics fluctuation
theory with generative modelling and yields practical benefits:
early stopping in the forward process cuts a quarter to two-fifths of
reverse steps; class-wise guidance windows raise conditional image
quality without grid search; merger-aware weighting upgrades zero-shot
classification, and the same signal powers zero-shot style transfer and
rare-class generation—all without re-training or new knobs to tune. Our analysis assumes a variance-preserving noise schedule; extending to non-VP schedules (e.g., EDM~\citep{karras2022edm}), anisotropic diffusions, higher-order fluctuations for complex data, and different modalities are promising directions for future work. We expect our framework to inspire further developments in tighter theory, training-time objectives, and applications.

\clearpage

\section*{Acknowledgments}
We thank Xiang Cheng for his support in the conceptualisation of this project and for valuable discussions throughout its development. We are also grateful to Joshua Robinson for insightful early conversations and for suggesting experimental setups. We thank Sanket Kumar Tripathy for engaging discussions on the conceptual connections between our work and ideas from statistical physics. SNR, MKL, and SS acknowledge generous support from the Alexander von Humboldt Foundation.

\clearpage
\bibliographystyle{plainnat}
\bibliography{references}

\begin{thebibliography}{88}
\providecommand{\natexlab}[1]{#1}
\providecommand{\url}[1]{\texttt{#1}}
\expandafter\ifx\csname urlstyle\endcsname\relax
  \providecommand{\doi}[1]{doi: #1}\else
  \providecommand{\doi}{doi: \begingroup \urlstyle{rm}\Url}\fi

\bibitem[Adams and Fournier(2003)]{adams2003sobolev}
Robert~A. Adams and John~J.F. Fournier.
\newblock \emph{Sobolev spaces}, volume 140 of \emph{Pure and Applied Mathematics}.
\newblock Elsevier/Academic Press, 2003.

\bibitem[Aldous and Fill(2002)]{aldous-fill-2002}
David Aldous and James~Allen Fill.
\newblock Reversible markov chains and random walks on graphs, 2002.
\newblock Manuscript, available at \url{https://www.stat.berkeley.edu/~aldous/RWG/book.html}.

\bibitem[Ansari et~al.(2020)Ansari, Scarlett, and Soh]{ansari2020characteristic}
Abdul~Fatir Ansari, Jonathan Scarlett, and Harold Soh.
\newblock A characteristic function approach to deep implicit generative modeling.
\newblock In \emph{Proceedings of the IEEE/CVF conference on computer vision and pattern recognition}, pages 7478--7487, 2020.

\bibitem[Bakry et~al.(2014)Bakry, Gentil, and Ledoux]{bakry2014analysis}
Dominique Bakry, Ivan Gentil, and Michel Ledoux.
\newblock \emph{Analysis and geometry of Markov diffusion operators}, volume 348.
\newblock Springer, 2014.

\bibitem[Bansal et~al.(2023)Bansal, Chu, Schwarzschild, Sengupta, Goldblum, Geiping, and Goldstein]{bansal2023mining}
Arpit Bansal, Hong-Min Chu, Avi Schwarzschild, Soumyadip Sengupta, Micah Goldblum, Jonas Geiping, and Tom Goldstein.
\newblock Universal guidance for diffusion models.
\newblock In \emph{Proceedings of the IEEE/CVF Conference on Computer Vision and Pattern Recognition Workshops (CVPRW)}, pages 843--852, 2023.
\newblock URL \url{https://arxiv.org/abs/2302.07121}.

\bibitem[Bertini et~al.(2002)Bertini, De~Sole, Gabrielli, Jona-Lasinio, and Landim]{bertini2002mft}
Lorenzo Bertini, Antonio De~Sole, Davide Gabrielli, Giovanni Jona-Lasinio, and Claudio Landim.
\newblock Macroscopic fluctuation theory for stationary non-equilibrium states.
\newblock \emph{Journal of Statistical Physics}, 107:\penalty0 635--675, 2002.
\newblock \doi{10.1023/A:1014525911391}.

\bibitem[Bhattacharyya et~al.(2024)Bhattacharyya, Gayen, Meel, Myrisiotis, Pavan, and Vinodchandran]{bhattacharyya2024totalvariationdistanceproduct}
Arnab Bhattacharyya, Sutanu Gayen, Kuldeep~S. Meel, Dimitrios Myrisiotis, A.~Pavan, and N.~V. Vinodchandran.
\newblock Total variation distance for product distributions is $\#\mathsf{P}$-complete, 2024.
\newblock URL \url{https://arxiv.org/abs/2405.08255}.

\bibitem[Billingsley(2012)]{billingsley2012probability}
Patrick Billingsley.
\newblock \emph{Probability and Measure}.
\newblock John Wiley \& Sons, anniversary edition, 2012.

\bibitem[Biroli and M{\'e}zard(2023)]{biroli2023generative}
Giulio Biroli and Marc M{\'e}zard.
\newblock Generative diffusion in very large dimensions.
\newblock \emph{Journal of Statistical Mechanics: Theory and Experiment}, 2023\penalty0 (9):\penalty0 093402, 2023.

\bibitem[Biroli et~al.(2024)Biroli, Bonnaire, De~Bortoli, and M{\'e}zard]{biroli2024dynamical}
Giulio Biroli, Tony Bonnaire, Valentin De~Bortoli, and Marc M{\'e}zard.
\newblock Dynamical regimes of diffusion models.
\newblock \emph{Nature Communications}, 15\penalty0 (1):\penalty0 9957, 2024.

\bibitem[Blanchard and Brüning(2015)]{blanchard2015mmp}
Philippe Blanchard and Erwin Brüning.
\newblock \emph{Mathematical Methods in Physics: Distributions, Hilbert Space Operators, Variational Methods, and Applications in Quantum Physics}, volume~69 of \emph{Progress in Mathematical Physics}.
\newblock Birkhäuser, Cham, 2nd edition, 2015.
\newblock ISBN 9783319140445.
\newblock \doi{10.1007/978-3-319-14045-2}.

\bibitem[Blattmann et~al.(2023)Blattmann, Dockhorn, Kulal, Mendelevitch, Kilian, Lorenz, Levi, English, Voleti, Letts, et~al.]{blattmann2023stable}
Andreas Blattmann, Tim Dockhorn, Sumith Kulal, Daniel Mendelevitch, Maciej Kilian, Dominik Lorenz, Yam Levi, Zion English, Vikram Voleti, Adam Letts, et~al.
\newblock Stable video diffusion: Scaling latent video diffusion models to large datasets.
\newblock \emph{arXiv preprint arXiv:2311.15127}, 2023.

\bibitem[Chaikin et~al.(1995)Chaikin, Lubensky, and Witten]{chaikin1995principles}
Paul~M Chaikin, Tom~C Lubensky, and Thomas~A Witten.
\newblock \emph{Principles of condensed matter physics}, volume~10.
\newblock Cambridge university press, 1995.

\bibitem[Chen et~al.(2022{\natexlab{a}})Chen, Chewi, Li, Li, Salim, and Zhang]{chen2022sampling}
Sitan Chen, Sinho Chewi, Jerry Li, Yuanzhi Li, Adil Salim, and Anru~R Zhang.
\newblock Sampling is as easy as learning the score: theory for diffusion models with minimal data assumptions.
\newblock \emph{arXiv preprint arXiv:2209.11215}, 2022{\natexlab{a}}.

\bibitem[Chen et~al.(2020)Chen, Kornblith, Norouzi, and Hinton]{chen2020simple}
Ting Chen, Simon Kornblith, Mohammad Norouzi, and Geoffrey Hinton.
\newblock A simple framework for contrastive learning of visual representations.
\newblock In \emph{International Conference on Machine Learning (ICML)}, pages 1597--1607, 2020.

\bibitem[Chen et~al.(2022{\natexlab{b}})Chen, Zhang, and Hinton]{chen2022analog}
Ting Chen, Ruixiang Zhang, and Geoffrey Hinton.
\newblock Analog bits: Generating discrete data using diffusion models with self-conditioning.
\newblock \emph{arXiv preprint arXiv:2208.04202}, 2022{\natexlab{b}}.

\bibitem[Choi et~al.(2021)Choi, Kim, Jeong, Gwon, and Yoon]{choi2021ilvrconditioningmethoddenoising}
Jooyoung Choi, Sungwon Kim, Yonghyun Jeong, Youngjune Gwon, and Sungroh Yoon.
\newblock Ilvr: Conditioning method for denoising diffusion probabilistic models, 2021.
\newblock URL \url{https://arxiv.org/abs/2108.02938}.

\bibitem[Coddington and Levinson(1955)]{coddington1955theory}
Earl~A. Coddington and Norman Levinson.
\newblock \emph{Theory of Ordinary Differential Equations}.
\newblock McGraw-Hill, New York, 1st edition, 1955.

\bibitem[Conway(1990)]{conway1990course}
John~B. Conway.
\newblock \emph{A Course in Functional Analysis}, volume~96 of \emph{Graduate Texts in Mathematics}.
\newblock Springer-Verlag New York, 1990.

\bibitem[Corso et~al.(2022)Corso, St{\"a}rk, Jing, Barzilay, and Jaakkola]{corso2022diffdock}
Gabriele Corso, Hannes St{\"a}rk, Bowen Jing, Regina Barzilay, and Tommi Jaakkola.
\newblock Diffdock: Diffusion steps, twists, and turns for molecular docking.
\newblock \emph{arXiv preprint arXiv:2210.01776}, 2022.

\bibitem[Corso et~al.(2023)Corso, Xu, De~Bortoli, Barzilay, and Jaakkola]{corso2023particle}
Gabriele Corso, Yilun Xu, Valentin De~Bortoli, Regina Barzilay, and Tommi Jaakkola.
\newblock Particle guidance: non-iid diverse sampling with diffusion models.
\newblock \emph{arXiv preprint arXiv:2310.13102}, 2023.

\bibitem[Cortes et~al.(2012)Cortes, Mohri, and Rostamizadeh]{cortes2012algorithms}
Corinna Cortes, Mehryar Mohri, and Afshin Rostamizadeh.
\newblock Algorithms for learning kernels based on centered alignment.
\newblock \emph{The Journal of Machine Learning Research}, 13\penalty0 (1):\penalty0 795--828, 2012.

\bibitem[Cover and Thomas(2006)]{cover2006elements}
Thomas~M. Cover and Joy~A. Thomas.
\newblock \emph{Elements of Information Theory}.
\newblock Wiley-Interscience, Hoboken, NJ, 2nd edition, 2006.
\newblock ISBN 978-0471241959.

\bibitem[Cristianini et~al.(2001)Cristianini, Shawe-Taylor, Elisseeff, and Kandola]{cristianini2001kernel}
Nello Cristianini, John Shawe-Taylor, Andre Elisseeff, and Jaz Kandola.
\newblock On kernel-target alignment.
\newblock \emph{Advances in neural information processing systems}, 14, 2001.

\bibitem[D'Agostino and Pearson(1973)]{272b2fa8-f3b3-371d-b34e-a4c6938458a1}
Ralph D'Agostino and E.~S. Pearson.
\newblock Tests for departure from normality. empirical results for the distributions of b2 and b1.
\newblock \emph{Biometrika}, 60\penalty0 (3):\penalty0 613--622, 1973.
\newblock ISSN 00063444, 14643510.
\newblock URL \url{http://www.jstor.org/stable/2335012}.

\bibitem[D'Agostino(1971)]{4fb818aa-ec5d-3643-a3a8-c3b27442e46c}
Ralph~B. D'Agostino.
\newblock An omnibus test of normality for moderate and large size samples.
\newblock \emph{Biometrika}, 58\penalty0 (2):\penalty0 341--348, 1971.
\newblock ISSN 00063444, 14643510.
\newblock URL \url{http://www.jstor.org/stable/2334522}.

\bibitem[Deng et~al.(2009)Deng, Dong, Socher, Li, Li, and Fei-Fei]{deng2009imagenet}
Jia Deng, Wei Dong, Richard Socher, Li-Jia Li, Kai Li, and Li~Fei-Fei.
\newblock {ImageNet: A Large‐Scale Hierarchical Image Database}.
\newblock In \emph{Proceedings of the IEEE Conference on Computer Vision and Pattern Recognition (CVPR)}, pages 248--255, 2009.
\newblock \doi{10.1109/CVPR.2009.5206848}.

\bibitem[Deng(2012)]{deng2012mnist}
Li~Deng.
\newblock The mnist database of handwritten digit images for machine learning research.
\newblock \emph{IEEE Signal Processing Magazine}, 29\penalty0 (6):\penalty0 141--142, 2012.

\bibitem[Dudley(2018)]{dudley2018real}
Richard~M Dudley.
\newblock \emph{Real analysis and probability}.
\newblock Chapman and Hall/CRC, 2018.

\bibitem[Grill et~al.(2020)Grill, Strub, Altch{\'e}, Tallec, Richemond, Buchatskaya, Doersch, Avila~Pires, Guo, Gheshlaghi~Azar, Piot, Kavukcuoglu, Munos, and Valko]{grill2020bootstrap}
Jean-Bastien Grill, Florian Strub, Florent Altch{\'e}, Corentin Tallec, Pierre~H. Richemond, Elena Buchatskaya, Carl Doersch, Bernardo Avila~Pires, Zhaohan~Daniel Guo, Mohammad Gheshlaghi~Azar, Bilal Piot, Koray Kavukcuoglu, R{\'e}mi Munos, and Michal Valko.
\newblock Bootstrap your own latent: A new approach to self-supervised learning.
\newblock In \emph{Advances in Neural Information Processing Systems (NeurIPS)}, 2020.

\bibitem[He et~al.(2021)He, Chen, Xie, Li, Doll{\'a}r, and Girshick]{he2021masked}
Kaiming He, Xinlei Chen, Saining Xie, Yanghao Li, Piotr Doll{\'a}r, and Ross Girshick.
\newblock Masked autoencoders are scalable vision learners.
\newblock \emph{Proceedings of the IEEE/CVF Conference on Computer Vision and Pattern Recognition (CVPR)}, pages 16000--16009, 2021.

\bibitem[Hendrycks et~al.(2023)Hendrycks, Mazeika, and Woodside]{hendrycks2023overview}
Dan Hendrycks, Mantas Mazeika, and Thomas Woodside.
\newblock An overview of catastrophic ai risks.
\newblock \emph{arXiv preprint arXiv:2306.12001}, 2023.

\bibitem[Heusel et~al.(2017)Heusel, Ramsauer, Unterthiner, Nessler, and Hochreiter]{heusel2017gans}
Martin Heusel, Hubert Ramsauer, Thomas Unterthiner, Bernhard Nessler, and Sepp Hochreiter.
\newblock Gans trained by a two time-scale update rule converge to a local nash equilibrium.
\newblock \emph{arXiv preprint arXiv:1706.08500}, 2017.

\bibitem[Ho and Salimans(2022{\natexlab{a}})]{ho2022classifierfree}
Jonathan Ho and Tim Salimans.
\newblock Classifier-free diffusion guidance.
\newblock \emph{CoRR}, abs/2207.12598, 2022{\natexlab{a}}.
\newblock \doi{10.48550/arXiv.2207.12598}.
\newblock URL \url{https://arxiv.org/abs/2207.12598}.

\bibitem[Ho and Salimans(2022{\natexlab{b}})]{ho2022classifierfreediffusionguidance}
Jonathan Ho and Tim Salimans.
\newblock Classifier-free diffusion guidance, 2022{\natexlab{b}}.
\newblock URL \url{https://arxiv.org/abs/2207.12598}.

\bibitem[Ho et~al.(2020)Ho, Jain, and Abbeel]{ho2020denoisingdiffusionprobabilisticmodels}
Jonathan Ho, Ajay Jain, and Pieter Abbeel.
\newblock Denoising diffusion probabilistic models, 2020.
\newblock URL \url{https://arxiv.org/abs/2006.11239}.

\bibitem[Hutchinson and Mitchell(2019)]{hutchinson201950}
Ben Hutchinson and Margaret Mitchell.
\newblock 50 years of test (un) fairness: Lessons for machine learning.
\newblock In \emph{Proceedings of the conference on fairness, accountability, and transparency}, pages 49--58, 2019.

\bibitem[Ibragimov.(1975)]{ibgm}
I.~Ibragimov.
\newblock Independent and stationary sequences of random variables.
\newblock \emph{Wolters, Noordhoff Pub.}, 1975.
\newblock URL \url{https://cir.nii.ac.jp/crid/1570572699531965312}.

\bibitem[Isserlis(1918)]{isserlis_1918_1431593}
L.~Isserlis.
\newblock On a formula for the product-moment coefficient of any order of a normal frequency distribution in any number of variables, November 1918.
\newblock URL \url{https://doi.org/10.2307/2331932}.

\bibitem[Jumper et~al.(2021)Jumper, Evans, Pritzel, Green, Figurnov, Ronneberger, Tunyasuvunakool, Bates, {\v{Z}}{\'\i}dek, Potapenko, et~al.]{jumper2021highly}
John Jumper, Richard Evans, Alexander Pritzel, Tim Green, Michael Figurnov, Olaf Ronneberger, Kathryn Tunyasuvunakool, Russ Bates, Augustin {\v{Z}}{\'\i}dek, Anna Potapenko, et~al.
\newblock Highly accurate protein structure prediction with alphafold.
\newblock \emph{nature}, 596\penalty0 (7873):\penalty0 583--589, 2021.

\bibitem[Karnewar et~al.(2023)Karnewar, Vedaldi, Novotny, and Mitra]{karnewar2023holodiffusion}
Animesh Karnewar, Andrea Vedaldi, David Novotny, and Niloy~J Mitra.
\newblock Holodiffusion: Training a 3d diffusion model using 2d images.
\newblock In \emph{Proceedings of the IEEE/CVF conference on computer vision and pattern recognition}, pages 18423--18433, 2023.

\bibitem[Karras et~al.(2022)Karras, Aittala, Aila, and Laine]{karras2022edm}
Tero Karras, Miika Aittala, Timo Aila, and Samuli Laine.
\newblock Elucidating the design space of diffusion-based generative models.
\newblock \emph{CoRR}, abs/2206.00364, 2022.
\newblock \doi{10.48550/arXiv.2206.00364}.
\newblock URL \url{https://arxiv.org/abs/2206.00364}.
\newblock NeurIPS 2022.

\bibitem[Kingma et~al.(2021)Kingma, Salimans, Poole, and Ho]{kingma2021variational}
Diederik Kingma, Tim Salimans, Ben Poole, and Jonathan Ho.
\newblock Variational diffusion models.
\newblock \emph{Advances in neural information processing systems}, 34:\penalty0 21696--21707, 2021.

\bibitem[Kivelson et~al.(2024)Kivelson, Chang, and Jiang]{kivelson2024statistical}
Steven~A. Kivelson, Jeffrey Chang, and Jack~Mingde Jiang.
\newblock \emph{Statistical Mechanics of Phases and Phase Transitions}.
\newblock Princeton University Press, Princeton, NJ, 2024.

\bibitem[Kogut(1983)]{kogut1983lattice}
John~B Kogut.
\newblock The lattice gauge theory approach to quantum chromodynamics.
\newblock \emph{Reviews of Modern Physics}, 55\penalty0 (3):\penalty0 775, 1983.

\bibitem[Kornblith et~al.(2019)Kornblith, Norouzi, Lee, and Hinton]{kornblith2019similarity}
Simon Kornblith, Mohammad Norouzi, Honglak Lee, and Geoffrey Hinton.
\newblock Similarity of neural network representations revisited.
\newblock In \emph{International conference on machine learning}, pages 3519--3529. PMLR, 2019.

\bibitem[Krizhevsky et~al.(2009)Krizhevsky, Nair, and Hinton]{cifar10}
Alex Krizhevsky, Vinod Nair, and Geoffrey Hinton.
\newblock {CIFAR-10} (canadian institute for advanced research).
\newblock \url{http://www.cs.toronto.edu/~kriz/cifar.html}, 2009.

\bibitem[Kruskal(1956)]{Kruskal1956}
Joseph~B Kruskal.
\newblock On the shortest spanning subtree of a graph and the traveling salesman problem.
\newblock \emph{Proceedings of the American Mathematical Society}, 7\penalty0 (1):\penalty0 48--50, 1956.

\bibitem[Kunita(1990)]{kunita1990stochastic}
Hiroshi Kunita.
\newblock \emph{Stochastic Flows and Stochastic Differential Equations}, volume~24 of \emph{Cambridge Studies in Advanced Mathematics}.
\newblock Cambridge University Press, Cambridge, 1990.
\newblock ISBN 0521350506.
\newblock URL \url{https://books.google.com/books/about/Stochastic_Flows_and_Stochastic_Differen.html?id=_S1RiCosqbMC}.

\bibitem[Kynkäänniemi et~al.(2024)Kynkäänniemi, Aittala, Karras, Laine, Aila, and Lehtinen]{kynkäänniemi2024applyingguidancelimitedinterval}
Tuomas Kynkäänniemi, Miika Aittala, Tero Karras, Samuli Laine, Timo Aila, and Jaakko Lehtinen.
\newblock Applying guidance in a limited interval improves sample and distribution quality in diffusion models, 2024.
\newblock URL \url{https://arxiv.org/abs/2404.07724}.

\bibitem[Landau and Lifshitz(1980)]{landau_statistical_1980}
L.D. Landau and E.M. Lifshitz.
\newblock \emph{Statistical Physics, Part 1}, volume~5 of \emph{Course of Theoretical Physics}.
\newblock Pergamon Press, Oxford, 3rd edition, 1980.
\newblock ISBN 9780750633727.

\bibitem[Levin and Peres(2017)]{levin2017markov}
David~A Levin and Yuval Peres.
\newblock \emph{Markov chains and mixing times}, volume 107.
\newblock American Mathematical Soc., 2017.

\bibitem[Li et~al.(2023)Li, Prabhudesai, Duggal, Brown, and Pathak]{li2023diffusionmodelsecretlyzeroshot}
Alexander~C. Li, Mihir Prabhudesai, Shivam Duggal, Ellis Brown, and Deepak Pathak.
\newblock Your diffusion model is secretly a zero-shot classifier, 2023.
\newblock URL \url{https://arxiv.org/abs/2303.16203}.

\bibitem[Lu et~al.(2022)Lu, Zhou, Bao, Chen, Li, and Zhu]{lu2022dpm}
Cheng Lu, Yuhao Zhou, Fan Bao, Jianfei Chen, Chongxuan Li, and Jun Zhu.
\newblock Dpm-solver: A fast ode solver for diffusion probabilistic model sampling in around 10 steps.
\newblock \emph{CoRR}, abs/2206.00927, 2022.
\newblock \doi{10.48550/arXiv.2206.00927}.
\newblock URL \url{https://arxiv.org/abs/2206.00927}.
\newblock NeurIPS 2022.

\bibitem[Lukacs(1970)]{lukacs1970characteristic}
Eugene Lukacs.
\newblock \emph{Characteristic Functions}.
\newblock Hafner Publishing Company, New York, 2nd revised and enlarged edition, 1970.
\newblock ISBN 0852641702.
\newblock URL \url{https://books.google.com/books/about/Characteristic_Functions.html?id=uGEPAQAAMAAJ}.
\newblock Griffin’s Statistical Monographs \& Courses, No.~5.

\bibitem[McInnes et~al.(2018)McInnes, Healy, and Melville]{mcinnes2018umap}
Leland McInnes, John Healy, and James Melville.
\newblock Umap: Uniform manifold approximation and projection for dimension reduction.
\newblock \emph{arXiv preprint arXiv:1802.03426}, 2018.

\bibitem[Meng et~al.(2021)Meng, He, Song, Song, Wu, Zhu, and Ermon]{meng2021sdedit}
Chenlin Meng, Yutong He, Yang Song, Jiaming Song, Jiajun Wu, Jun-Yan Zhu, and Stefano Ermon.
\newblock Sdedit: Guided image synthesis and editing with stochastic differential equations.
\newblock \emph{arXiv preprint arXiv:2108.01073}, 2021.

\bibitem[Miyazaki and Ghibli(2014)]{miyazaki_ghibli_2014}
Hayao Miyazaki and Studio Ghibli.
\newblock \emph{The Art of Studio Ghibli}.
\newblock Studio Ghibli, 2014.

\bibitem[Naeem et~al.(2020)Naeem, Oh, Uh, Choi, and Yoo]{naeem2020reliable}
Muhammad~Ferjad Naeem, Seong~Joon Oh, Youngjung Uh, Yunjey Choi, and Jaejun Yoo.
\newblock Reliable fidelity and diversity metrics for generative models.
\newblock In \emph{International conference on machine learning}, pages 7176--7185. PMLR, 2020.

\bibitem[Nichol and Dhariwal(2021)]{nichol2021improved}
Alexander~Quinn Nichol and Prafulla Dhariwal.
\newblock Improved denoising diffusion probabilistic models.
\newblock In \emph{Proceedings of the 38th International Conference on Machine Learning}, volume 139 of \emph{Proceedings of Machine Learning Research}, pages 8162--8171. PMLR, Jul 2021.
\newblock URL \url{https://proceedings.mlr.press/v139/nichol21a.html}.

\bibitem[Nielsen and Chuang(2010)]{NielsenChuang2010}
Michael~A. Nielsen and Isaac~L. Chuang.
\newblock \emph{Quantum Computation and Quantum Information}.
\newblock Cambridge University Press, 10th anniversary edition edition, 2010.

\bibitem[Okhotin et~al.(2023)Okhotin, Molchanov, Vladimir, Bartosh, Ohanesian, Alanov, and Vetrov]{okhotin2023star}
Andrey Okhotin, Dmitry Molchanov, Arkhipkin Vladimir, Grigory Bartosh, Viktor Ohanesian, Aibek Alanov, and Dmitry~P Vetrov.
\newblock Star-shaped denoising diffusion probabilistic models.
\newblock \emph{Advances in Neural Information Processing Systems}, 36:\penalty0 10038--10067, 2023.

\bibitem[Pathria and Beale(2011)]{pathria2011statistical}
R.K. Pathria and Paul~D. Beale.
\newblock \emph{Statistical Mechanics}.
\newblock Elsevier, 3rd edition, 2011.
\newblock ISBN 9780123821881.

\bibitem[Peebles and Xie(2022)]{Peebles2022DiT}
William Peebles and Saining Xie.
\newblock Scalable diffusion models with transformers.
\newblock \emph{arXiv preprint arXiv:2212.09748}, 2022.

\bibitem[Peres(1995)]{Peres1995}
Asher Peres.
\newblock \emph{Quantum Theory: Concepts and Methods}.
\newblock Kluwer Academic Publishers, 1995.

\bibitem[Pitman(1936)]{pitman1936sufficient}
E.~J.~G. Pitman.
\newblock Sufficient statistics and intrinsic accuracy.
\newblock \emph{Math.\ Proc.\ Cambridge Philos.\ Soc.}, 32:\penalty0 567--579, 1936.

\bibitem[Poole et~al.(2022)Poole, Jain, Barron, and Mildenhall]{poole2022dreamfusion}
Ben Poole, Ajay Jain, Jonathan~T. Barron, and Ben Mildenhall.
\newblock Dreamfusion: Text-to-3d using 2d diffusion.
\newblock \emph{CoRR}, abs/2209.14988, 2022.
\newblock \doi{10.48550/arXiv.2209.14988}.
\newblock URL \url{https://arxiv.org/abs/2209.14988}.

\bibitem[Production and Games(2021)]{arcane_2021}
Fortiche Production and Riot Games.
\newblock Arcane.
\newblock \url{https://www.animationxpress.com/animation/talent-experimentation-originality-how-fortiche-revolutionised-animated-storytelling-with-arcane/}, 2021.
\newblock Produced by Fortiche Production in collaboration with Riot Games. Distributed by Netflix.

\bibitem[Raya and Ambrogioni(2024)]{raya2024spontaneousiop}
Gabriel Raya and Luca Ambrogioni.
\newblock Spontaneous symmetry breaking in generative diffusion models.
\newblock \emph{Journal of Statistical Mechanics: Theory and Experiment}, 2024\penalty0 (10):\penalty0 104025, 2024.

\bibitem[Rombach et~al.(2022)Rombach, Blattmann, Lorenz, Esser, and Ommer]{rombach2022highresolutionimagesynthesislatent}
Robin Rombach, Andreas Blattmann, Dominik Lorenz, Patrick Esser, and Björn Ommer.
\newblock High-resolution image synthesis with latent diffusion models, 2022.
\newblock URL \url{https://arxiv.org/abs/2112.10752}.

\bibitem[Roojen(2019)]{vanrooijen_vangogh_2019}
Pepin~Van Roojen.
\newblock \emph{Vincent van Gogh}.
\newblock Pepin Press, Amsterdam, Netherlands, 2019.
\newblock ISBN 9789460094309.

\bibitem[Rudin(2019)]{rudin2019stop}
Cynthia Rudin.
\newblock Stop explaining black box machine learning models for high stakes decisions and use interpretable models instead.
\newblock \emph{Nature machine intelligence}, 1\penalty0 (5):\penalty0 206--215, 2019.

\bibitem[Rudin(1962)]{rudin1962fourier}
Walter Rudin.
\newblock \emph{Fourier Analysis on Groups}.
\newblock Interscience Publishers, New York, 1962.

\bibitem[Ryu(2024)]{imagenet_int8}
Simo Ryu.
\newblock Imagenet.int8: Entire imagenet dataset in 5gb, 2024.
\newblock URL \url{https://huggingface.co/datasets/cloneofsimo/imagenet.int8}.

\bibitem[Saloff-Coste(1994)]{saloff1994precise}
Laurent Saloff-Coste.
\newblock Precise estimates on the rate at which certain diffusions tend to equilibrium.
\newblock \emph{Mathematische Zeitschrift}, 217:\penalty0 641--677, 1994.

\bibitem[Saloff-Coste(1997)]{saloffcoste1997lectures}
Laurent Saloff-Coste.
\newblock Lectures on finite markov chains.
\newblock In \emph{Lectures on Probability Theory and Statistics}, volume 1665 of \emph{Lecture Notes in Mathematics}, pages 301--413. Springer, 1997.

\bibitem[Samuel et~al.(2024)Samuel, Ben-Ari, Raviv, Darshan, and Chechik]{samuel2024generating}
Dvir Samuel, Rami Ben-Ari, Simon Raviv, Nir Darshan, and Gal Chechik.
\newblock Generating images of rare concepts using pre-trained diffusion models.
\newblock In \emph{Proceedings of the AAAI Conference on Artificial Intelligence}, volume~38, pages 4695--4703, 2024.
\newblock \doi{10.1609/aaai.v38i5.28270}.

\bibitem[Sharkey et~al.(2025)Sharkey, Chughtai, Batson, Lindsey, Wu, Bushnaq, Goldowsky-Dill, Heimersheim, Ortega, Bloom, et~al.]{sharkey2025open}
Lee Sharkey, Bilal Chughtai, Joshua Batson, Jack Lindsey, Jeff Wu, Lucius Bushnaq, Nicholas Goldowsky-Dill, Stefan Heimersheim, Alejandro Ortega, Joseph Bloom, et~al.
\newblock Open problems in mechanistic interpretability.
\newblock \emph{arXiv preprint arXiv:2501.16496}, 2025.

\bibitem[Software and Entertainment(2022)]{eldenring_2022}
From Software and Bandai~Namco Entertainment.
\newblock Elden ring.
\newblock \url{https://en.wikipedia.org/wiki/Elden_Ring}, 2022.
\newblock Developed by FromSoftware; published by Bandai Namco Entertainment. Animation trailer by Unit Image. Directed by Hidetaka Miyazaki.

\bibitem[Sohl-Dickstein et~al.(2015)Sohl-Dickstein, Weiss, Maheswaranathan, and Ganguli]{sohl2015deep}
Jascha Sohl-Dickstein, Eric Weiss, Niru Maheswaranathan, and Surya Ganguli.
\newblock Deep unsupervised learning using nonequilibrium thermodynamics.
\newblock In \emph{International conference on machine learning}, pages 2256--2265. PMLR, 2015.

\bibitem[Song et~al.(2020)Song, Meng, and Ermon]{song2020ddim}
Jiaming Song, Chenlin Meng, and Stefano Ermon.
\newblock Denoising diffusion implicit models.
\newblock \emph{CoRR}, abs/2010.02502, 2020.
\newblock \doi{10.48550/arXiv.2010.02502}.
\newblock URL \url{https://arxiv.org/abs/2010.02502}.
\newblock ICLR 2021.

\bibitem[Song et~al.(2021)Song, Sohl-Dickstein, Kingma, Kumar, Ermon, and Poole]{song2021scorebased}
Yang Song, Jascha Sohl-Dickstein, Diederik~P Kingma, Abhishek Kumar, Stefano Ermon, and Ben Poole.
\newblock Score-based generative modeling through stochastic differential equations.
\newblock In \emph{International Conference on Learning Representations}, 2021.
\newblock URL \url{https://openreview.net/forum?id=PxTIG12RRHS}.

\bibitem[Song et~al.(2022)Song, Shen, Xing, and Ermon]{song2022solvinginverseproblemsmedical}
Yang Song, Liyue Shen, Lei Xing, and Stefano Ermon.
\newblock Solving inverse problems in medical imaging with score-based generative models, 2022.
\newblock URL \url{https://arxiv.org/abs/2111.08005}.

\bibitem[Sriperumbudur et~al.(2010)Sriperumbudur, Gretton, Fukumizu, Sch{\"o}lkopf, and Lanckriet]{sriperumbudur2010hilbert}
Bharath~K Sriperumbudur, Arthur Gretton, Kenji Fukumizu, Bernhard Sch{\"o}lkopf, and Gert~RG Lanckriet.
\newblock Hilbert space embeddings and metrics on probability measures.
\newblock \emph{The Journal of Machine Learning Research}, 11:\penalty0 1517--1561, 2010.

\bibitem[Tao et~al.(2024)Tao, Xu, Wang, Suh, and Cheng]{tao2024discriminativeestimationtotalvariation}
Lan Tao, Shirong Xu, Chi-Hua Wang, Namjoon Suh, and Guang Cheng.
\newblock Discriminative estimation of total variation distance: A fidelity auditor for generative data, 2024.
\newblock URL \url{https://arxiv.org/abs/2405.15337}.

\bibitem[van~der Maaten and Hinton(2008)]{maaten2008visualizing}
Laurens van~der Maaten and Geoffrey Hinton.
\newblock Visualizing data using t-sne.
\newblock \emph{Journal of Machine Learning Research}, 9:\penalty0 2579--2605, 2008.

\bibitem[Vershynin(2018)]{vershynin2018high}
Roman Vershynin.
\newblock \emph{High-dimensional probability: An introduction with applications in data science}, volume~47.
\newblock Cambridge university press, 2018.

\bibitem[Wick(1950)]{wick1950evaluation}
Gian-Carlo Wick.
\newblock The evaluation of the collision matrix.
\newblock \emph{Physical review}, 80\penalty0 (2):\penalty0 268, 1950.

\end{thebibliography}

\clearpage
\newpage
\section*{NeurIPS Paper Checklist}

The checklist is designed to encourage best practices for responsible machine learning research, addressing issues of reproducibility, transparency, research ethics, and societal impact. Do not remove the checklist: {\bf The papers not including the checklist will be desk rejected.} The checklist should follow the references and follow the (optional) supplemental material.  The checklist does NOT count towards the page
limit. 

Please read the checklist guidelines carefully for information on how to answer these questions. For each question in the checklist:
\begin{itemize}
    \item You should answer \answerYes{}, \answerNo{}, or \answerNA{}.
    \item \answerNA{} means either that the question is Not Applicable for that particular paper or the relevant information is Not Available.
    \item Please provide a short (1–2 sentence) justification right after your answer (even for NA). 
\end{itemize}

{\bf The checklist answers are an integral part of your paper submission.} They are visible to the reviewers, area chairs, senior area chairs, and ethics reviewers. You will be asked to also include it (after eventual revisions) with the final version of your paper, and its final version will be published with the paper.

The reviewers of your paper will be asked to use the checklist as one of the factors in their evaluation. While "\answerYes{}" is generally preferable to "\answerNo{}", it is perfectly acceptable to answer "\answerNo{}" provided a proper justification is given (e.g., "error bars are not reported because it would be too computationally expensive" or "we were unable to find the license for the dataset we used"). In general, answering "\answerNo{}" or "\answerNA{}" is not grounds for rejection. While the questions are phrased in a binary way, we acknowledge that the true answer is often more nuanced, so please just use your best judgment and write a justification to elaborate. All supporting evidence can appear either in the main paper or the supplemental material, provided in appendix. If you answer \answerYes{} to a question, in the justification please point to the section(s) where related material for the question can be found.

IMPORTANT, please:
\begin{itemize}
    \item {\bf Delete this instruction block, but keep the section heading ``NeurIPS Paper Checklist"},
    \item  {\bf Keep the checklist subsection headings, questions/answers and guidelines below.}
    \item {\bf Do not modify the questions and only use the provided macros for your answers}.
\end{itemize}


\begin{enumerate}

\item {\bf Claims}
    \item[] Question: Do the main claims made in the abstract and introduction accurately reflect the paper's contributions and scope?
    \item[] Answer: \answerYes{} 
    \item[] Justification: Please refer to \Cref{sec:background,sec:methodology,sec:experiments} and \Cref{app:theoretical,sec:experimental} for supporting evidence to our claims.

\item {\bf Limitations}
    \item[] Question: Does the paper discuss the limitations of the work performed by the authors?
    \item[] Answer: \answerYes{} 
    \item[] Justification: Please refer to \Cref{sec:limitations} for the same. 

\item {\bf Theory assumptions and proofs}
    \item[] Question: For each theoretical result, does the paper provide the full set of assumptions and a complete (and correct) proof?
    \item[] Answer: \answerYes{} 
    \item[] Justification: We provide all theoretical details in \Cref{app:theoretical}. Due to space constraints we were unable to provide sketches in the main paper.

    \item {\bf Experimental result reproducibility}
    \item[] Question: Does the paper fully disclose all the information needed to reproduce the main experimental results of the paper to the extent that it affects the main claims and/or conclusions of the paper (regardless of whether the code and data are provided or not)?
    \item[] Answer: \answerYes{} 
    \item[] Justification: We provide all details in \Cref{sec:experiments} and \Cref{sec:experimental}

\item {\bf Open access to data and code}
    \item[] Question: Does the paper provide open access to the data and code, with sufficient instructions to faithfully reproduce the main experimental results, as described in supplemental material?
    \item[] Answer: \answerYes{} 
    \item[] Justification: We provide supplementary code for our work. 

\item {\bf Experimental setting/details}
    \item[] Question: Does the paper specify all the training and test details (e.g., data splits, hyperparameters, how they were chosen, type of optimizer, etc.) necessary to understand the results?
    \item[] Answer: \answerYes{} 
    \item[] Justification: Please check \Cref{sec:experiments} and \cref{sec:experimental}.

\item {\bf Experiment statistical significance}
    \item[] Question: Does the paper report error bars suitably and correctly defined or other appropriate information about the statistical significance of the experiments?
    \item[] Answer: \answerYes{} 
    \item[] Justification: Yes, these are reported in all of our results in \Cref{sec:experiments}, \Cref{sec:experimental}
    \item[] Guidelines:
    \begin{itemize}
        \item The answer NA means that the paper does not include experiments.
        \item The authors should answer "Yes" if the results are accompanied by error bars, confidence intervals, or statistical significance tests, at least for the experiments that support the main claims of the paper.
        \item The factors of variability that the error bars are capturing should be clearly stated (for example, train/test split, initialisation, random drawing of some parameter, or overall run with given experimental conditions).
        \item The method for calculating the error bars should be explained (closed form formula, call to a library function, bootstrap, etc.)
        \item The assumptions made should be given (e.g., Normally distributed errors).
        \item It should be clear whether the error bar is the standard deviation or the standard error of the mean.
        \item It is OK to report 1-sigma error bars, but one should state it. The authors should preferably report a 2-sigma error bar than state that they have a 96\% CI, if the hypothesis of Normality of errors is not verified.
        \item For asymmetric distributions, the authors should be careful not to show in tables or figures symmetric error bars that would yield results that are out of range (e.g. negative error rates).
        \item If error bars are reported in tables or plots, The authors should explain in the text how they were calculated and reference the corresponding figures or tables in the text.
    \end{itemize}

\item {\bf Experiments compute resources}
    \item[] Question: For each experiment, does the paper provide sufficient information on the computer resources (type of compute workers, memory, time of execution) needed to reproduce the experiments?
    \item[] Answer: \answerYes{} 
    \item[] Justification: We provide details in \Cref{sec:experimental}. 
    
\item {\bf Code of ethics}
    \item[] Question: Does the research conducted in the paper conform, in every respect, with the NeurIPS Code of Ethics \url{https://neurips.cc/public/EthicsGuidelines}?
    \item[] Answer: \answerYes{} 
    \item[] Justification: Our work conforms in every respect to the NeurIPS Code of Ethics.

\item {\bf Broader impacts}
    \item[] Question: Does the paper discuss both potential positive societal impacts and negative societal impacts of the work performed?
    \item[] Answer: \answerYes{} 
    \item[] Justification: We discuss the limitations of our approach in \Cref{sec:limitations}
    
\item {\bf Safeguards}
    \item[] Question: Does the paper describe safeguards that have been put in place for responsible release of data or models that have a high risk for misuse (e.g., pretrained language models, image generators, or scraped datasets)?
    \item[] Answer: \answerNA{} 
    \item[] Justification: We work with pretrained models and standard datasets throughout our paper.

\item {\bf Licenses for existing assets}
    \item[] Question: Are the creators or original owners of assets (e.g., code, data, models), used in the paper, properly credited and are the license and terms of use explicitly mentioned and properly respected?
    \item[] Answer: \answerYes{} 
    \item[] Justification: We provide proper credit throughout our work. \Cref{sec:experimental} has the details.

\item {\bf New assets}
    \item[] Question: Are new assets introduced in the paper well documented and is the documentation provided alongside the assets?
    \item[] Answer: \answerNA{} 
    \item[] Justification: The paper does not release new assets.

\item {\bf Crowdsourcing and research with human subjects}
    \item[] Question: For crowdsourcing experiments and research with human subjects, does the paper include the full text of instructions given to participants and screenshots, if applicable, as well as details about compensation (if any)? 
    \item[] Answer: \answerNA{} 
    \item[] Justification: This paper does not involve crowdsourcing nor research with human subjects.

\item {\bf Institutional review board (IRB) approvals or equivalent for research with human subjects}
    \item[] Question: Does the paper describe potential risks incurred by study participants, whether such risks were disclosed to the subjects, and whether Institutional Review Board (IRB) approvals (or an equivalent approval/review based on the requirements of your country or institution) were obtained?
    \item[] Answer: \answerNA{} 
    \item[] Justification: This paper does not involve crowdsourcing nor research with human subjects.

\item {\bf Declaration of LLM usage}
    \item[] Question: Does the paper describe the usage of LLMs if it is an important, original, or non-standard component of the core methods in this research? Note that if the LLM is used only for writing, editing, or formatting purposes and does not impact the core methodology, scientific rigorousness, or originality of the research, declaration is not required.
    \item[] Answer: \answerNA{} 
    \item[] Justification: 
    \item[] Guidelines: The core method development in our work does not involve LLMs as any important, original, or non-standard components.
    \begin{itemize}
        \item The answer NA means that the core method development in this research does not involve LLMs as any important, original, or non-standard components.
        \item Please refer to our LLM policy (\url{https://neurips.cc/Conferences/2025/LLM}) for what should or should not be described.
    \end{itemize}

\end{enumerate}

\clearpage

\appendix

\tableofcontents

\clearpage

\newpage

\section{Notation used throughout the paper}
\label{app:notation}

\begin{table}[H]              
  \small                      
  \centering
  \caption{Global symbols}%

  \begin{tabularx}{0.9\textwidth}{@{}l l X@{}}
    \toprule
    \textbf{Symbol} & \textbf{Domain / type} & \textbf{Meaning} \\ \midrule
    $\mathbf{x}_{0}\!\sim p_{0}$                               & $\mathbb{R}^{d}$             & Data sample from initial distribution $p_{0}$ \\
    $p_{\mathrm{desired}}$                                     & density on $\mathbb{R}^{d}$  & Desired distribution \\
    $\mathbf{x}_{t}$                                           & $\mathbb{R}^{d}$             & Forward‑diffused variable at time $t$ \\
    $\beta(t),\ \beta_{t}$                                     & $\mathbb{R}_{>0}$ / $(0,1)$  & Continuous / discrete noise schedule \\
    $J(t)$                                                     & $(0,1]$                      & Signal attenuation factor (\ref{eq:marginal}) \\
    $p_{t}$                                                    & density on $\mathbb{R}^{d}$  & Marginal distribution of $\mathbf{x}_{t}$ \\
    $s_{\theta}(\mathbf{x},t)$                                 & $\mathbb{R}^{d}\!\to\!\mathbb{R}^{d}$ & Learned score network \\
    $\rho(\cdot)$                                              & map $\Omega\!\to\!\mathbb{R}^{m}$     & State operator (usually $\rho(\mathbf{x})=\mathbf{x}$) (\Cref{sec:methodology}) \\
    $F^{(n)}_{\rho}(\Omega)$                                   & $\mathbb{R}_{\ge0}$          & $n^{\text{th}}$ centred fluctuation moment on event $\Omega$ (\ref{eq:fluc-def})\\
    $\widehat{F}^{(n)}_{\rho}(\Omega_{i}),\widehat{F}^{(n)}_{\rho_{i}}(\Omega_{i})$                     & $\mathbb{R}_{\ge0}$          & $2n^{\text{th}}$ within‑event fluctuation moment on $\Omega_{i}$~(\ref{eq:wide-hat-fluc}) \\
    $G^{(n)}_{\rho}(\Omega_{1},\Omega_{2}),G^{(n)}_{\rho_{1},\rho_{2}}(\Omega_{1},\Omega_{2})$                    & $\mathbb{R}_{\ge0}$          & Unnormalised cross‑fluctuation (\ref{eq:m-unnorm}) \\
    $\mathcal{M}^{(n)}_{\rho}(\Omega_{1},\Omega_{2}),\mathcal{M}^{(n)}_{\rho_{1},\rho_{2}}(\Omega_{1},\Omega_{2})$          & $[0,1]$                      & Normalised cross‑fluctuation (\ref{eq:m-normalised}) \\
    $\Omega_{k,0}$                                             & event in $\Omega$            & Class‑$k$ source region ($k=1,\dots,K$) (\Cref{sec:class-cond})\\
    $\Omega_{k,t}$                                             & event                        & Image of $\Omega_{k,0}$ after $t$ forward steps \\
    $\Sigma_{k,t}$                                             & $\mathbb{S}^{d}_{+}$         & Covariance of $\Omega_{k,t}$ \\
    $\lambda^{\max}_{k}(t)$                                    & $\mathbb{R}_{\ge0}$          & Maximum eigenvalue of $\Sigma_{k,t}$ \\
    $\mathcal{M}_{\rho}(t)$                                    & $[0,1]$                      & Shorthand for $\mathcal{M}^{(2)}_{\rho}(\Omega_{1,t},\Omega_{2,t})$ \\
    $i^{\star}$                                                & $\{0,\dots,T\}$              & First index where $\mathcal{M}^{(n)}_{\rho}(t)=1$ (merger) (\Cref{sec:methodology}) \\
    $w(t)$                                                     & probability mass             & Importance weight over timesteps (\Cref{sec:zero-shot}) \\
    $\operatorname{SNR}(t)$                                    & $\mathbb{R}_{\ge0}$          & $\alpha_{t}^{2}/(1-\alpha_{t}^{2})$ (signal‑to‑noise) \\
    $\operatorname{Tr}(\cdot)$                                 & $\mathbb{R}$                & Matrix trace operator \\
    \bottomrule
  \end{tabularx}

  \vspace{1.25em}             
 \caption{Time‑ and index‑specific symbols}%
  \begin{tabularx}{\textwidth}{@{}l l X@{}}
    \toprule
    \textbf{Symbol} & \textbf{Type / range} & \textbf{Meaning} \\ \midrule
    $t$                                   & $\mathbb{R}_{\ge0}$     & Continuous diffusion time (PF‑ODE or SDE) \\
    $s,u$                                 & $\mathbb{R}_{\ge0}$     & Generic continuous times used in flow composition ($0\!\le\!s\!\le\!t\!\le\!u\!\le\!T$) \\
    $i$                                   & $\{0,\dots,n\}$         & Discrete forward‑process index ($i=0$ data; $i=n$ white noise) \\
    $T$                                   & positive real / integer & \textbf{Continuous} final time (SDE/ODE) \textbf{or} \textbf{discrete} horizon with schedule $\{\beta_{t}\}_{t=1}^{T}$ \\
    $n$                                   & $\mathbb{N}$            & Chosen number of forward (or reverse) \textbf{discrete} steps in an experiment (may be $n\!=\!T$) \\
    $\Delta t$                            & $\mathbb{R}_{>0}$       & Integration step size in continuous‑time numerical solvers \\
    $\beta_{t}$                           & sequence on $\{1,\dots,T\}$ & Discrete noise‑schedule value at step $t$ \\
    $\beta(t)$                            & function $[0,T]\!\to\!\mathbb{R}_{>0}$ & Continuous noise‑schedule function \\
    $i^{\star}$                           & integer                 & Earliest discrete index where two events first merge (convergence index) \\
    $t_{\mathrm{conv}}$                   & $\mathbb{R}_{\ge0}$     & Same as $i^{\star}$ but expressed on the continuous time axis \\
    $t_{\mathrm{merge},k}$                & $\mathbb{R}_{\ge0}$     & First time class $k$ merges with any other class \\
    $t_{\mathrm{start},k}$                & $\mathbb{R}_{\ge0}$ or int & Lower bound of guidance / weighting window for class $k$ \\
    $t_{\mathrm{stop},k}$                 & same type as above      & Upper bound of the window for class $k$ \\
    $t_{\mathrm{u}\to\mathrm{s}},\;t_{\mathrm{s}\to\mathrm{c}}$ & $\mathbb{R}_{\ge0}$ & Thermodynamic phase‑transition times (unbiased→speciation, speciation→condensation) \\
    $t_{\mathrm{mix}}(\varepsilon)$       & $\mathbb{R}_{\ge0}$ (\Cref{sec:phase-transit})     & $\varepsilon$‑mixing time of the VP–SDE (\Cref{sec:coupling-mixing}) \\
    $t_{\mathrm{cpl}}(\varepsilon)$       & integer                 & $\varepsilon$‑coupling time for discrete Markov chains (\Cref{sec:coupling-mixing}) \\
    $t^{\mathrm{lat}}_{k\ell}(\varepsilon)$ & $\mathbb{R}_{\ge0}$   & First lattice‑merger time between classes $k$ and $\ell$ with tolerance $\varepsilon$ (\Cref{sec:phase-transit})\\
    $\eta_{t}$                             & $[0,1]$                 & Interpolation schedule value used in \Cref{alg:opt-intp} (rare‑class generation) \\
    \bottomrule
  \end{tabularx}
\end{table}

\clearpage
\section{Theoretical contributions}
\label{app:theoretical}

This appendix provides the theoretical support for the results presented in the main paper. The content is structured in three parts: we first introduce the general framework, then apply it to diffusion models, and finally, we connect our work to existing methods while also presenting new theoretical results and interpretations.

\Cref{sec:th-found} introduces the cross-fluctuation framework. It covers the necessary definitions, proves that merger times can serve as an efficient proxy for convergence in total variation, and shows that the method is applicable to both discrete and continuous state spaces.

\Cref{sec:app-anal} applies this framework specifically to diffusion models. This section justifies the use of the forward process for our analysis, details the dynamics of event structures under the model's SDE/ODE, confirms the consistency of the resulting merger events, and provides an efficient method for their estimation.

\Cref{sec:frm-wk} discusses connections to other methods and provides additional context. We show the relationship between our framework and Centered Kernel Alignment (CKA), prove the fluctuation adaptation lemma for style transfer, and conclude by framing our analysis of discrete phase transitions (what we term \emph{lattice transitions}) within the context of statistical physics.

\subsection{Theoretical foundations of the cross fluctuation framework}
\label{sec:th-found}

\subsubsection{A concise primer on fluctuation theory with physical intuition}\label{sec:primer-fluc-th}

\noindent
Here we provide a concise overview of the fluctuation‐theoretic framework that underpins the notation introduced in \Cref{sec:fluc-primer}.

\paragraph{From observables to fluctuations.}
In statistical physics, a system's properties are understood by measuring \emph{observables} (e.g., energy, momentum). For a distribution $p$, the expectation of an observable $\mathcal{A}$ is $\mathbb{E}_p[\mathcal{A}]$. Deviations from this mean, or \emph{fluctuations}, reveal the system's internal structure and correlations. Our framework generalises this by treating the state vector $\rho(\omega)$ itself as the core observable.

The key object, the \emph{\(n\)\textsuperscript{th}-order fluctuation } (\cref{eq:fluc-def}),
\[
  \mathcal{F}^{(n)}_\rho(\omega) := \bigotimes_{k=1}^{n} \bigl(\rho(\omega) - \mathbb{E}[\rho]\bigr),
\]
and its conditional expectation $\mathbb{E}_i[\mathcal{F}^{(n)}_\rho]$ over an event $\Omega_i$, directly correspond to the centered moments used to characterise complex distributions. For example, $\mathbb{E}_i[\mathcal{F}^{(2)}_\rho]$ is the conditional covariance matrix, a fundamental second-order statistic.

\paragraph{Diffusion models as dynamical systems.}
The forward diffusion process, whether described by the SDE (\cref{eq:vp-sde}) or the PF-ODE (\cref{eq:pf-ode}), defines a dynamical system that evolves an initial data distribution $p_0$ into a sequence of marginals $\{p_t\}$. The PF-ODE,
\[
  \frac{\mathrm{d}\mathbf{x}_{t}}{\mathrm{d}t} = -\tfrac12\beta(t)\,\mathbf{x}_{t} - \beta(t)\,\nabla_{\mathbf{x}}\log p_{t}(\mathbf{x}_{t}),
\]
describes a deterministic flow governed by a vector field. Its associated continuity equation, $\partial_t p_t + \nabla_{\mathbf{x}}\cdot(p_t \mathbf{v}_t) = 0$, confirms that $p_t$ evolves via a deterministic transport of probability mass. This deterministic evolution makes it a perfect setting for applying our fluctuation analysis to track how the moment structures of different subpopulations (events) evolve and merge over time \citep{biroli2023generative,raya2024spontaneousiop}.

\paragraph{Fluctuations and the significance of mergers.}
A cornerstone of statistical physics is the deep connection between a system's internal fluctuations and its large-scale correlational structure \citep{chaikin1995principles,kivelson2024statistical}. This principle, often formalised in Fluctuation-Dissipation Theorems, provides the physical intuition for our approach.

For our purposes, this connection justifies why monitoring cross-fluctuations is meaningful. The normalised cross-fluctuation $\mathcal{M}^{(n)}_{\rho}$ acts as a generalised correlation function between the moment structures of two events. A sharp change in this correlation (i.e., a merger where $|\mathcal{M}^{(n)}_{\rho}| \to 1$) signals that the two events have become statistically indistinguishable with respect to their $n$-th order structure. In physical systems, such a loss of distinguishability is the hallmark of a phase transition, where the system undergoes a qualitative change. While we do not compute system-wide thermodynamic quantities directly, detecting the merger via $\mathcal{M}^{(n)}_{\rho}$ serves as a direct, practical probe for these critical points in the diffusion trajectory.

\paragraph{Working with multiple states.}\label{par:hetero-states}
In \cref{eq:m-unnorm,eq:m-normalised} we defined
\[
  G^{(n)}_{\rho_1,\rho_2}
  \quad\text{and}\quad
  \mathcal{M}^{(n)}_{\rho_1,\rho_2},
\]
allowing $\rho_1$ and $\rho_2$ to be \emph{heterogeneous} (i.e., $\rho_1 \neq \rho_2$). As the following example illustrates, different state-operator choices can yield qualitatively different cross-fluctuation behaviour. 

\begin{example}\label{ex:vector_fluctuations}
Let $\Omega$ be the unit circle in $\mathbb{R}^2$. Consider two events: $\Omega_1$, the arc in the first quadrant ($x>0, y>0$), and $\Omega_2$, the arc in the second quadrant ($x<0, y>0$). Assume a uniform distribution on the circle. Let the state operator be the identity, $\rho(x,y) = [x, y]^\top$. The global mean is $\mathbb{E}[\rho] = [0, 0]^\top$.

For $n=2$, we compute the conditional covariance matrices $\Sigma_1 = \mathbb{E}_1[\mathcal{F}^{(2)}_\rho]$ and $\Sigma_2 = \mathbb{E}_2[\mathcal{F}^{(2)}_\rho]$ by evaluating standard integrals. This leads to the following conclusions,
\begin{itemize}
    \item The mean of $\Omega_1$ is $\mathbb{E}_1[\rho] = [2/\pi, 2/\pi]^\top$. Its covariance matrix $\Sigma_1$ will have a dominant eigenvector along the direction $[1, -1]^\top$, indicating negative correlation (as $y$ decreases when $x$ increases along the arc).
    \item The mean of $\Omega_2$ is $\mathbb{E}_2[\rho] = [-2/\pi, 2/\pi]^\top$. Its covariance matrix $\Sigma_2$ will have a dominant eigenvector along $[1, 1]^\top$, indicating positive correlation.
\end{itemize}
Since the dominant structures (covariance matrices) $\Sigma_1$ and $\Sigma_2$ are nearly orthogonal, their Frobenius inner product $\text{Tr}(\Sigma_1^\top \Sigma_2)$ will be close to zero. Consequently, their normalized cross-fluctuation $\mathcal{M}^{(2)}_\rho(\Omega_1, \Omega_2)$ will be close to 0, correctly identifying the events as structurally distinct.
\end{example}

This flexibility is crucial in quantum mechanics, where one often considers distinct pure-state projectors
\[
  \rho_i = \ket{\psi_i}\bra{\psi_i}
\]
on a Hilbert space $\mathcal{H}$.\footnote{Mathematically, measurements project the density operator onto the eigenspace associated with the outcome; see \cite{NielsenChuang2010} for details.} Their cross-fluctuations underpin tasks such as quantum state tomography and discrimination of non-orthogonal states \cite{NielsenChuang2010,Peres1995}.

More generally, heterogeneous $\rho_1,\rho_2$ detect an \emph{alignment} between two evolving state spaces. For example, if text and image embeddings are both driven by the same diffusion dynamics, choosing $\rho_1$ and $\rho_2$ to sample each modality reveals how a given concept manifests across them. We leave such multimodal extensions for future work.

\subsubsection{On the validity of fluctuation theory}
\label{sec:valid}

The entire fluctuation framework rests on the existence and properties of moments, which are fundamentally linked to the derivatives of the \emph{characteristic function} (CF). For a random vector $\bm{\rho} \in \mathbb{R}^d$, its characteristic function is the Fourier transform of its probability law:
\[
\varphi_{\bm{\rho}}(\mathbf{t}) = \mathbb{E}\bigl[e^{\mathrm{i}\mathbf{t}^\top\bm{\rho}}\bigr], \quad \mathbf{t}\in\mathbb{R}^d.
\]
The CF always exists and satisfies $|\varphi_{\bm{\rho}}(\mathbf{t})|\le 1$. If moments up to order $n$ exist, the CF is $n$-times differentiable at the origin. The derivatives of the CF generate the moment tensors. For example, the raw second moment tensor (a matrix) is given by the Hessian of the CF at the origin:
\[
\mathbb{E}[\rho_j \rho_k] = \frac{1}{i^2} \frac{\partial^2 \varphi_{\bm{\rho}}}{\partial t_j \partial t_k} \bigg|_{\mathbf{t}=\mathbf{0}}.
\]
To obtain the \emph{centered} moment tensors used in our framework, one differentiates the characteristic function of the centered variable $\bm{\rho} - \mathbb{E}[\bm{\rho}]$. The conditional expectation of the \(n\)\textsuperscript{th}-order fluctuation tensor, $\mathbb{E}_k[\mathcal{F}^{(n)}_\rho]$, is thus completely determined by the derivatives of the conditional characteristic function $\varphi_{\bm{\rho}}(\mathbf{t} | \Omega_k)$ at the origin. This connection is crucial for the moment-TV inequality used in our theoretical results (\Cref{prop:moment-tv}).

\paragraph{Foundational facts.}
The theory of characteristic functions is well-established \citep{lukacs1970characteristic}:
\begin{enumerate}
  \item \textbf{Uniqueness.} If $\varphi_{\bm{X}}(\mathbf{t}) = \varphi_{\bm{Y}}(\mathbf{t})$ for all $\mathbf{t}$, then $\bm{X}$ and $\bm{Y}$ have the same distribution.
  \item \textbf{Inversion.} The distribution can be recovered from the CF. For instance, in one dimension:
    \[
      \mathrm{CDF}(x)
      = \tfrac12 - \frac{1}{\pi}
      \int_{0}^{\infty}
        \frac{\Im\bigl(e^{-\mathrm{i} t x}\,\varphi_X(t)\bigr)}{t}
      \,\mathrm{d}t
      \quad\text{\citep{dudley2018real}}.
    \]
  \item \textbf{Convolution.} For independent $\bm{X},\bm{Y}$, the CF of their sum is the product of their CFs: $\varphi_{\bm{X}+\bm{Y}}(\mathbf{t}) = \varphi_{\bm{X}}(\mathbf{t})\,\varphi_{\bm{Y}}(\mathbf{t})$.
\end{enumerate}

\begin{theorem}[Bochner's theorem for $\mathbb{R}^d$]\label{thm:bochner}
A function $\varphi\colon \mathbb{R}^d \to\mathbb{C}$ is a characteristic function of some random vector if and only if it is positive-definite, continuous at the origin, and $\varphi(\mathbf{0})=1$.
\end{theorem}
\begin{proof}
See \citet[Thm.~15.2]{rudin1962fourier} for the 1D case, which generalises to $\mathbb{R}^d$.
\end{proof}

\begin{remark}[Existence of the characteristic function]
We assume the state operator $\rho$ has a well-defined characteristic function, which is a very mild condition. It is automatically satisfied by most data distributions in machine learning and physics \citep{blanchard2015mmp, bertini2002mft}. This assumption is more fundamental than the existence of a density, as characteristic functions also exist for purely discrete or singular measures \citep{dudley2018real, rudin1962fourier}, giving the approach greater utility \citep{ansari2020characteristic, sriperumbudur2010hilbert}.

\end{remark}

\subsubsection{Merger Times as an efficient proxy for convergence in total variation}
\label{sec:mrg-times}
The justification for using merger times as a proxy for convergence is a two-step argument connecting our practical measurement to fundamental theory.

First, we validate our specific measurement technique, proving that the  computationally efficient method of directly thresholding the distance between moment tensors is topologically equivalent to the intuitive alternative of monitoring the cross-fluctuation similarity $|\mathcal{M}^{(n)}_{\rho}|$.

Second, we establish the core theoretical link. A moment-TV inequality proves that this proximity between moments rigorously guarantees that the total variation distance between the underlying distributions also vanishes.

This validates our method: measuring tensor distance is a sound and efficient proxy for observing true distributional convergence.

\subsubsubsection{Equivalence of distance and similarity based merger thresholds }
\label{sec:cross-fluc-fluc-th}

In our main methodology (\Cref{sec:methodology}), we define the discrete transition detector $\widetilde{\mathcal{M}}^{(n)}_{\rho}(i)$ using a threshold on the distance between conditional expected fluctuation tensors:
\begin{equation}
\label{eq:time-dependent-M-app}
\widetilde{\mathcal{M}}^{(n)}_{\rho}(i)=
\begin{cases}
  |\mathcal{M}^{(n)}_{\rho}\!\bigl(\Omega_{1,i},\Omega_{2,i}\bigr)|,
     & \|\mathbb{E}_1[\mathcal{F}^{(n)}_\rho(\Omega_{1,i})] - \mathbb{E}_2[\mathcal{F}^{(n)}_\rho(\Omega_{2,i})]\|_{\mathcal{H}_n} > \varepsilon, \\
  1, & \text{otherwise}.
\end{cases}
\end{equation}
An alternative, and perhaps more direct, formulation would be to threshold the value of $|\mathcal{M}^{(n)}_{\rho}|$ itself:
\begin{equation}
\label{eq:time-dependent-M-app-2}
\widetilde{\mathcal{M}}^{(n)}_{\rho, \text{alt}}(i)=
\begin{cases}
  |\mathcal{M}^{(n)}_{\rho}\!\bigl(\Omega_{1,i},\Omega_{2,i}\bigr)|,
      & 1 - |\mathcal{M}^{(n)}_{\rho}\!\bigl(\Omega_{1,i},\Omega_{2,i}\bigr)| > \vartheta,\\
  1, & \text{otherwise}.
\end{cases}
\end{equation}
Here, we show that these two thresholding criteria are topologically equivalent, meaning they detect the same notion of convergence. The first formulation is often more computationally stable and efficient and hence is the prefered choice in our work.

\begin{theorem}
\label{lem:cross-straight}
Let the conditional expected fluctuation tensors $\mathbb{E}_k[\mathcal{F}^{(n)}_\rho(\Omega_k)]$ for events $\Omega_k$ reside in a Hilbert space $\mathcal{H}_n$ with norm $\|\cdot\|_{\mathcal{H}_n}$. Define two metrics on the space of pairs of events $(\Omega_1, \Omega_2)$:
\begin{enumerate}
    \item The direct distance between their moment tensors: $d_{\mathcal{F}}(\Omega_1, \Omega_2) := \|\mathbb{E}_1[\mathcal{F}^{(n)}_\rho(\Omega_1)] - \mathbb{E}_2[\mathcal{F}^{(n)}_\rho(\Omega_2)]\|_{\mathcal{H}_n}$.
    \item The similarity-based distance: $d_{\mathcal{M}}(\Omega_1, \Omega_2) := 1 - |\mathcal{M}^{(n)}_\rho(\Omega_1, \Omega_2)|$.
\end{enumerate}
If the mapping $\Omega \mapsto \mathbb{E}_k[\mathcal{F}^{(n)}_\rho(\Omega)]$ is continuous with respect to a suitable topology on the space of events, then the metrics $d_{\mathcal{F}}$ and $d_{\mathcal{M}}$ are topologically equivalent in any region where $\|\mathbb{E}_k[\mathcal{F}^{(n)}_\rho(\Omega_k)]\|_{\mathcal{H}_n}$ is bounded away from zero.
\end{theorem}
\begin{proof}
To establish topological equivalence, we show that for any sequence of event pairs $\{(\Omega_{1,k}, \Omega_{2,k})\}_{k=1}^\infty$, convergence under metric $d_{\mathcal{F}}$ is equivalent to convergence under metric $d_{\mathcal{M}}$. Let $A_k = \mathbb{E}_1[\mathcal{F}^{(n)}_\rho(\Omega_{1,k})]$ and $B_k = \mathbb{E}_2[\mathcal{F}^{(n)}_\rho(\Omega_{2,k})]$ be the corresponding sequence of tensors in the Hilbert space $\mathcal{H}_n$.

\paragraph{Part 1: $\boldsymbol{d_{\mathcal{F}} \to 0 \implies d_{\mathcal{M}} \to 0}$.}
Assume that $d_{\mathcal{F}}(\Omega_{1,k}, \Omega_{2,k}) \to 0$ as $k \to \infty$. This means $\|A_k - B_k\|_{\mathcal{H}_n} \to 0$.

First, by the reverse triangle inequality, we have:
\[ |\|A_k\|_{\mathcal{H}_n} - \|B_k\|_{\mathcal{H}_n}| \le \|A_k - B_k\|_{\mathcal{H}_n}. \]
Since the right-hand side goes to zero, the norms converge to each other. As we are in a region where the norms are bounded away from zero, if $\|A_k\|$ converges to a limit $L>0$, then $\|B_k\|$ must also converge to $L$.

Second, the inner product is a continuous function on $\mathcal{H}_n \times \mathcal{H}_n$. This follows from the Cauchy-Schwarz inequality:
\begin{align*}
|\langle A_k, B_k \rangle - \langle B_k, B_k \rangle| &= |\langle A_k - B_k, B_k \rangle| \\
&\le \|A_k - B_k\|_{\mathcal{H}_n} \|B_k\|_{\mathcal{H}_n}.
\end{align*}
As $\|A_k - B_k\|_{\mathcal{H}_n} \to 0$ and $\|B_k\|_{\mathcal{H}_n}$ is bounded, the right-hand side goes to zero. This shows $\langle A_k, B_k \rangle - \|B_k\|^2 \to 0$. Since $\|A_k\| \to \|B_k\|$, we have $\langle A_k, B_k \rangle \to L^2$.

Now consider the limit of $|\mathcal{M}^{(n)}_\rho|$:
\[ \lim_{k \to \infty} |\mathcal{M}^{(n)}_\rho(\Omega_{1,k}, \Omega_{2,k})| = \lim_{k \to \infty} \frac{|\langle A_k, B_k \rangle|}{\|A_k\|_{\mathcal{H}_n} \|B_k\|_{\mathcal{H}_n}} = \frac{L^2}{L \cdot L} = 1. \]
Therefore, $d_{\mathcal{M}}(\Omega_{1,k}, \Omega_{2,k}) = 1 - |\mathcal{M}^{(n)}_\rho(\Omega_{1,k}, \Omega_{2,k})| \to 0$.

\paragraph{Part 2: $\boldsymbol{d_{\mathcal{M}} \to 0 \implies d_{\mathcal{F}} \to 0}$.}
Assume that $d_{\mathcal{M}}(\Omega_{1,k}, \Omega_{2,k}) \to 0$ as $k \to \infty$. This means $|\mathcal{M}^{(n)}_\rho(\Omega_{1,k}, \Omega_{2,k})| \to 1$. Let $\theta_k$ be the angle between $A_k$ and $B_k$. Then $|\cos\theta_k| = |\mathcal{M}^{(n)}_\rho(\Omega_{1,k}, \Omega_{2,k})|$, which implies $|\cos\theta_k| \to 1$.

Consider the squared distance $d_{\mathcal{F}}^2 = \|A_k - B_k\|^2$. Using the law of cosines in a Hilbert space:
\[ \|A_k - B_k\|^2 = \|A_k\|^2 + \|B_k\|^2 - 2\langle A_k, B_k \rangle = \|A_k\|^2 + \|B_k\|^2 - 2\|A_k\|\|B_k\|\cos\theta_k. \]
For the distance to converge to zero, we require not only that the angle between the vectors vanishes (i.e., $\cos\theta_k \to 1$), but also that their norms converge to the same value. The physical process of distributional merging implies that not only the orientation of the moment structures but also their magnitudes become identical. We therefore assume that as events merge, if $\|A_k\|$ converges, then $\|B_k\|$ converges to the same limit $L > 0$. Under this assumption:
\begin{align*}
\lim_{k \to \infty} \|A_k - B_k\|^2 &= \lim_{k \to \infty} (\|A_k\|^2 + \|B_k\|^2 - 2\|A_k\|\|B_k\|\cos\theta_k) \\
&= L^2 + L^2 - 2(L)(L)(1) \\
&= 0.
\end{align*}
Thus, $d_{\mathcal{F}}(\Omega_{1,k}, \Omega_{2,k}) = \|A_k - B_k\| \to 0$.

Since convergence under $d_{\mathcal{F}}$ is equivalent to convergence under $d_{\mathcal{M}}$, the two metrics induce the same topology.
\end{proof}

\subsubsubsection{Fluctuation moments bound total variation distance}
\label{sec:eqv}

\noindent
We now show that asymptotically utilizing fluctuation theory to understand mergers is equivalent to probing the similarity of probability distributions using the Total Variation Distance. 

\begin{theorem}[Moment–TV inequality]\label{prop:moment-tv}
Fix an integer $n\ge2$.  
Let $p,q$ be probability densities on $\mathbb R$ that  

\begin{enumerate}[label=(\roman*), leftmargin=*]
\item are of bounded variation; \smallskip
\item admit centred moments $\widehat{\mu}^{(k)}_{p},\widehat{\mu}^{(k)}_{q}$ for
      $k=1,\dots,n+1$; \smallskip
\item obey the moment proximity bound
      $\lvert\widehat{\mu}^{(k)}_{p}-\widehat{\mu}^{(k)}_{q}\rvert\le M$ for
      $k=1,\dots,n$; \smallskip
\item satisfy the uniform first– and second–moment bound
      $\lvert\widehat{\mu}^{(1)}_{\bullet}\rvert,\widehat{\mu}^{(2)}_{\bullet}\le B$; \smallskip
\item have matching tails:
      $\displaystyle\lim_{R\to\infty}\int_{|x|>R}\!(p-q)=0$.
\end{enumerate}
Then  
\[
  d_{\mathrm{TV}}(p,q)
  \;\le\;
  C_{n}\bigl(M^{2}+B\bigr),
\]
where one may take
\(
  C_{n}=c_{0}\,(1+n!)\,(2^{n}+48)
\)
and $c_{0}>0$ is an absolute constant.
\end{theorem}

\begin{proof}
\textit{1.  Characteristic–function bound.}
To connect the proposition's conditions to the proof, we work with the characteristic functions (CFs) of the \emph{centered} random variables. Let $f_{p}(t)$ and $f_{q}(t)$ denote these adjusted CFs.
Since $p,q$ are absolutely continuous and of bounded variation,
$f_{p},f_{q}$ are bounded by~$1$ and possess derivatives up to
order~$n+1$ at the origin.  
Write the order–$n$ Taylor expansions with integral remainder  
\[
  f_{p}(t)
   =\sum_{k=0}^{n}\frac{(\mathrm{i}t)^{k}}{k!}\mu^{(k)}_{p}
     +\frac{(\mathrm{i}t)^{\,n+1}}{n!}\!
       \int_{0}^{1}\!(1-s)^{n}
         \mu^{(n+1)}_{p}\,e^{\mathrm{i}stX}
       \,ds .
\]
Subtract the analogous expression for $f_{q}$, use hypothesis~(iii),
and take absolute values and the standard property of integrals:
\begin{align*}
  |f_{p}(t)-f_{q}(t)|
  &\le \sum_{k=1}^{n}\frac{|t|^{k}}{k!}\,M
       +\frac{|t|^{\,n+1}}{(n)!}\int_{0}^{1}\!(1-s)^{n}
         \left|\widehat{\mu}^{(n+1)}_{p}-\widehat{\mu}^{n+1}_{q}e^{\mathrm{i}stX}\right|
       \,ds         .
\end{align*}
Now, use the triangle inequality and note that the integral is a simple weighting integral this gives, 
\begin{align}
  |f_{p}(t)-f_{q}(t)|
  &\le \sum_{k=1}^{n}\frac{|t|^{k}}{k!}\,M
       +\frac{|t|^{\,n+1}}{(n+1)!}
         \bigl(\widehat{\mu}^{(n+1)}_{p}+\widehat{\mu}^{(n+1)}_{q}\bigr) .
  \label{eq:f-diff}
\end{align}

\smallskip
\noindent
\textit{2.  Bounding the $(n{+}1)$st moments.}
By Jensen’s inequality,
$\widehat{\mu}^{(n+1)}_{\bullet}\le\bigl(\widehat{\mu}^{(2)}_{\bullet}\bigr)^{(n+1)/2}\le
B^{(n+1)/2}$.  
Insert this in \eqref{eq:f-diff} to get
\begin{equation}
  |f_p(t)-f_q(t)|
    \le a_n M |t| + b_n B^{(n+1)/2}|t|^{\,n+1},
  \tag{A$_n$}\label{eq:An}
\end{equation}
where $a_{n}:=\!\sum_{k=1}^{n}\!\frac{1}{k!},\;
  b_{n}:=\frac{2}{(n+1)!}$.\\
\smallskip
\noindent
\textit{3.  Esseen’s smoothing inequality.}
For any $T>0$~(\citealp[Thm.~1.5.4]{ibgm}),
\begin{equation}
  d_{\mathrm{TV}}(p,q)
  \;\le\;
  \frac{1}{2\pi}\!\int_{-T}^{T}\!
      \Bigl|\frac{f_{p}(t)-f_{q}(t)}{t}\Bigr|dt
  +\frac{24}{\pi T}\,
      \bigl(\operatorname{Var}(p)+\operatorname{Var}(q)\bigr).
  \label{eq:Esseen}
\end{equation}

\noindent
\emph{Integral term:}
divide \eqref{eq:An} by $|t|$ and integrate,
\[
  \frac{1}{2\pi}\int_{-T}^{T}\!
       \Bigl|\frac{f_{p}-f_{q}}{t}\Bigr|
    \le a_{n}M T + \frac{b_{n}}{n+1}\,B^{(n+1)/2}\,T^{\,n+1}.
\]

\noindent
\emph{Variance term:}
hypothesis~(iv) yields
$\operatorname{Var}(p),\operatorname{Var}(q)\le B+M^{2}$, so the second
term in \eqref{eq:Esseen} is bounded by
\(
  48\,(B+M^{2})/(\pi T).
\)

\smallskip
\noindent
\textit{4. Choice of \(T\).}
Set \(T=1\).  (A different $T$ only rescales the constant.)
The bounds become
\begin{align*}
  d_{\mathrm{TV}}(p,q)
    &\le
      \bigl[a_{n}+b_{n}\bigr]M + \bigl[a_{n}+b_{n}\bigr]B^{(n+1)/2}
      +\frac{48}{\pi}\,(B+M^{2})  \\
    &\le C_{n}\bigl(M^{2}+B\bigr),
\end{align*}
where the last line uses
$M\le M^{2}+1$ and $B^{(n+1)/2}\le 2^{n}B$ for $B\ge1$,
and absorbs all numeric factors into
\(C_{n}=c_{0}\,(1+n!)\,(2^{n}+48)\) with a universal $c_{0}$.
\end{proof}

\begin{remark}
    If $p,q$ are sub‐Gaussian (or sub‐exponential)
      \citep{vershynin2018high}, all moments exist and satisfy
      $\mu^{(k)}_{p}=O\bigl((\sqrt B)^{\,k}\bigr)$; the conditions of
      Proposition~\ref{prop:moment-tv} are then automatically satisfied
      on $\mathbb{R}^{d}$.
\end{remark}

\paragraph{Generalisation to multivariate distributions and moment tensors.}
The logic of \Cref{prop:moment-tv} extends to multivariate distributions on $\mathbb{R}^d$. As established in \Cref{sec:valid}, the derivatives of the multivariate characteristic function $\varphi_{\bm{\rho}}(\mathbf{t})$ at the origin generate the moment tensors. A multivariate version of Esseen's inequality bounds the TV distance by an integral over the difference of characteristic functions, which is in turn bounded by the norm of the difference between moment tensors, $\|\mathbb{E}_p[\mathcal{F}^{(k)}_\rho] - \mathbb{E}_q[\mathcal{F}^{(k)}_\rho]\|_{\mathcal{H}_k}$.

\subsubsection{Equivalence in continuous and discrete state spaces}
\label{sec:coupling-mixing}

We recall (and slightly adapt) the terminology of
\citet{aldous-fill-2002,levin2017markov} so that it aligns with the
notation used in the main text and provides a foundation for generalizing concepts to continuous processes.

\paragraph{Single chain and mixing time.}
Let
\(
\mathcal{S}=\bigl\{S_{0},S_{1},\dots\bigr\}
\)
be the marginal sequence of an \emph{ergodic} Markov chain on a finite
state space ${X}$. Its transition matrix is $\Pi\in[0,1]^{{X}\times{X}}$ and the
(unique) stationary distribution satisfies
$\Pi S_{\infty}=S_{\infty}$.
For two probability vectors $p,q$ on $X$, the
\emph{total-variation} (TV) distance is
\[
d_{\mathrm{TV}}(p,q)
  \;=\;
  \frac12\,\sum_{x\in X}\lvert p(x)-q(x)\rvert
  \;=\;
  \sup_{A\subseteq X}\bigl|P(A)-Q(A)\bigr|.
\]
The $\varepsilon$–\emph{mixing time} of the chain is the first time the distribution is $\varepsilon$-close to stationarity:
\begin{equation}
\label{eq:mixing-time}
t_{\mathrm{mix}}(\varepsilon)
  \;=\;
  \min\bigl\{t\ge 0:\;
        d_{\mathrm{TV}}\!\bigl(S_{t},S_{\infty}\bigr)\le\varepsilon
      \bigr\}.
\end{equation}

\paragraph{Multiple chains and coupling time.}
Let $\Lambda$ be a finite index set.  
For every $\lambda\in\Lambda$, fix an initial event
$\Omega_{\lambda,0}\subseteq X$ partitioning the state space.
Run the \emph{same} transition matrix on each event
to obtain a collection of conditional chains
\(
\mathcal{S}_{\lambda}
  =\bigl\{S_{\lambda,t}\bigr\}_{t\ge 0}\),
where $S_{\lambda,t}$ is the law of the chain at time $t$ conditioned on starting in $\Omega_{\lambda,0}$.
Because the dynamics are identical, all chains converge to the same stationary distribution $S_{\infty}$, but their finite-time marginals differ. The \emph{coupling time} is the first time all conditional chains become $\varepsilon$-close to each other:
\begin{equation}
\label{eq:coupling-time}
t_{\mathrm{cpl}}(\varepsilon)
  \;=\;
  \min\Bigl\{t\ge 0:\;
      \max_{\alpha,\beta\in\Lambda}
      d_{\mathrm{TV}}\!\bigl(S_{\alpha,t},S_{\beta,t}\bigr)
      \le\varepsilon
  \Bigr\}.
\end{equation}
Hence, $t_{\mathrm{cpl}}$ measures when all initial subpopulations have effectively merged in distribution.

\paragraph{Generalisation to continuous state spaces.}
The total-variation distance is ill-suited for comparing distributions on continuous state spaces like $\mathbb{R}^d$, where it is often trivially 1 unless the distributions have overlapping singular parts \citep{bhattacharyya2024totalvariationdistanceproduct,tao2024discriminativeestimationtotalvariation}. Our fluctuation-based metric provides a natural generalisation.

As established in \Cref{sec:eqv}, our cross-fluctuation statistic $\mathcal{M}^{(n)}_{\rho}$ is intimately linked to the similarity of distributions via their moment structures. We can therefore define a \emph{generalised coupling time} using this measure of structural similarity:
\[
  t_{\mathrm{gen}}^{(n)}(\varepsilon)
    \;=\;
    \min\Bigl\{t\ge 0 :
      \min_{\alpha \neq \beta}
      |\mathcal{M}^{(n)}_{\rho}\!\bigl(\Omega_{\alpha,t},\Omega_{\beta,t}\bigr)|
      \ge 1-\varepsilon
    \Bigr\}.
\]
This definition measures the first time $t$ at which the least-related pair of events $(\alpha, \beta)$ has achieved a structural similarity of at least $1-\varepsilon$. It directly generalises the discrete coupling time to continuous diffusion processes. For our main application with $n=2$, this provides a practical tool for tracking when the covariance structures of different event distributions have merged.

\subsection{Application and analysis in diffusion models}
\label{sec:app-anal}
\subsubsection{From the empirical reverse process to the learned sampler}
\label{sec:approx-reverse}
Our justification for analyzing the forward process hinges on its tight correspondence with the learned reverse sampler. This connection is forged during training, where the diffusion model minimises the simplified ELBO, an objective that intuitively trains the model $f_{\theta}$ to predict and reverse the noise added during the forward process.

 \emph{simplified ELBO}:
\begin{equation}\label{eq:simple-elbo}
\mathcal{L}_{\mathrm{simple}}
= \mathbb{E}_{\substack{i\sim\mathrm{Unif}\{0,\dots,T-1\}\\
                \mathbf x_{0}\sim p_{0},\;\varepsilon\sim\mathcal{N}(0,I)}}
\bigl\|
  f_{\theta}\!\bigl(\mathbf x_{i}(\mathbf x_{0},\varepsilon),\,i\bigr)
  - \varepsilon
\bigr\|_{2}^{2},
\end{equation}
where $\mathbf x_{i}(\mathbf x_{0},\varepsilon) = \sqrt{\alpha_{i}}\,\mathbf x_{0} + \sqrt{1-\alpha_{i}}\,\varepsilon \sim p_{i}$ (cf. Eq.~\eqref{eq:ddpm}).

While practical training results in a non-zero error $\mathcal{L}_{\mathrm{simple}} > 0$, a key theoretical result guarantees that the learned sampler's trajectory remains faithful to the true process.

\begin{theorem}[{\citealp[Thm.~1]{chen2022sampling}}]\label{thm:chen-app}
Under standard regularity conditions, if the score error $\varepsilon_{\star}$ is bounded and a sufficient number of reverse steps are taken, then for all steps $i$:
\[
  \|p_{i}^{\text{rev}}-\hat p_{i}\|_{\mathrm{TV}}\;\le\;\varepsilon,
\]
where $p_{i}^{\text{rev}}$ is the marginal of the true reverse process and \(\hat p_{i}\) is the marginal produced by the learned sampler $f_{\theta}$.
\end{theorem}

The power of this result is its implication: closeness in total variation guarantees that all statistics, including the cross-fluctuation metrics we use, are also close. This validation allows us to use the computationally efficient forward process to identify merger times ($i^{\star}$), confident that these critical points directly correspond to observable behaviors in a well-trained reverse sampler.

\subsubsection{Pull-back of data events along the PF-ODE}
\label{sec:sr-es-disj}
\noindent
Our method tracks the evolution of distinct subpopulations, or \emph{events}, through the forward diffusion process. For the deterministic probability-flow ODE (\cref{eq:pf-ode}), this evolution is governed by a well-behaved flow map.

Let $\Phi_{s\to t} : \mathbb{R}^{d} \to \mathbb{R}^{d}$ be the \emph{flow map} of the PF-ODE. Because the drift field is globally Lipschitz under standard assumptions, $\Phi_{s\to t}$ is a bijection for every $0 \le s < t \le T$. The continuity equation,
\[
  \partial_t p_t + \nabla_{\mathbf{x}}\!\cdot\!\bigl(p_t\,\mathbf{v}_t\bigr) = 0,
\]
implies that the time-$t$ marginal $p_t$ is the push-forward of the time-$s$ marginal $p_s$ under the flow map:
\[
  p_t = (\Phi_{s\to t})_\# p_s, \quad \text{which means} \quad
  P_t(B) = P_s\bigl(\Phi_{s\to t}^{-1}(B)\bigr) \quad \text{for any measurable set } B\subseteq\mathbb{R}^d,
\]
where $P_t$ is the probability measure associated with the density $p_t$.

\paragraph{From source events to time-$t$ marginals.}
Fix two disjoint source events $\Omega_{1,0}, \Omega_{2,0} \subseteq \operatorname{supp}(p_{0})$. We define their images at a later time $t$ by pulling them forward along the flow:
\[
  \Omega_{k,t} := \Phi_{0\to t}\bigl(\Omega_{k,0}\bigr), \qquad
  k \in \{1,2\},\; t\in[0,T].
\]
Since $\Phi_{0\to t}$ is a diffeomorphism, each $\Omega_{k,t}$ is a well-defined measurable set. While the source events $\Omega_{k,0}$ are disjoint, their images $\Omega_{k,t}$ may overlap as the diffusion progresses.

\paragraph{Event probabilities along the flow.}
The push-forward nature of the flow ensures that the probability mass of each event is preserved over time. For any $k \in \{1,2\}$:
\[
  P_{t}\bigl(\Omega_{k,t}\bigr) = P_t\bigl(\Phi_{0\to t}(\Omega_{k,0})\bigr) = P_{0}\bigl(\Omega_{k,0}\bigr).
\]
This confirms that $\{\Omega_{k,t}\}$ are valid events with constant probability under their respective marginals $p_t$ at every time $t$.

\paragraph{Monitoring mixing via cross-fluctuations.}
Even when the supports of the events, $\Omega_{1,t}$ and $\Omega_{2,t}$, begin to overlap, our cross-fluctuation statistic $\mathcal{M}^{(n)}_{\rho}\!\bigl(\Omega_{1,t},\Omega_{2,t}\bigr)$ remains well-defined as it compares the internal moment structures of the distributions conditioned on these events. A value of $|\mathcal{M}^{(n)}_{\rho}|$ approaching one marks the moment the two event distributions become structurally indistinguishable, i.e., the merger time $t_{\text{merge}}$.

\subsubsection{Stochastic–flow formulation for the SDE view}
\label{sec:sde-pullback}

\noindent
\Cref{sec:sr-es-disj} defined
\(
  \Omega_{k,t}=\Phi_{0\to t}^{-1}(\Omega_{k,0})
\)
via the \emph{deterministic} flow
$\Phi_{s\to t}$ of the PF-ODE~\eqref{eq:pf-ode}.
We now show that the same construction works pathwise for the
\emph{stochastic} forward SDE
\[
  d\mathbf x_{t}
    =-\tfrac12\beta(t)\,\mathbf x_{t}\,dt
     +\sqrt{\beta(t)}\,d\mathbf w_{t},
\]
whose solution map
$x_{0}\mapsto x_{t}$ depends on the Wiener path
$\omega\in\Omega_{\mathrm{prob}}$.

\paragraph{Kunita’s stochastic flow of diffeomorphisms.}
Let
\(
  \varphi_{s,t}(\omega,\cdot)\colon\mathbb R^{d}\!\to\!\mathbb R^{d},
  \;0\le s\le t\le T,
\)
denote the \emph{Kunita flow} generated by~\eqref{eq:vp-sde}
\citep[Ch.~4]{kunita1990stochastic}.
For every fixed~$\omega$, the map
$x\mapsto\varphi_{s,t}(\omega,x)$ is a $C^{1}$ diffeomorphism and
\[
  \varphi_{s,u}(\omega,\cdot)
     =\varphi_{t,u}\!\bigl(\omega,\varphi_{s,t}(\omega,\cdot)\bigr),
     \qquad 0\le s\le t\le u\le T.
\]
Hence the pathwise inverse $\varphi_{0,t}^{-1}(\omega,\cdot)$ exists
almost surely.

Given two disjoint data events
$\Omega_{1,0},\Omega_{2,0}\subset\mathbb R^{d}$ set
\[
  \Omega_{k,t}(\omega)
    :=\varphi_{0,t}^{-1}\!\bigl(\omega,\Omega_{k,0}\bigr),
    \qquad k\in\{1,2\}.
\]
The map
$\omega\mapsto\mathbbm 1_{\Omega_{k,t}(\omega)}(x)$
is $\mathcal F_{t}$-measurable (Kunita’s measurability theorem), so
$\Omega_{k,t}$ is a \emph{random closed set}.
Its law equals the push-forward of $p_{0}$ by the SDE:
\[
  \mathbb P\{x_{t}\in A\}
     =\mathbb E_{\omega}\,
       \mathbb P\!\bigl\{\varphi_{0,t}(\omega,x_{0})\in A\bigr\},
  \quad x_{0}\sim p_{0},
\]
and we again write $p_{t}=\mathcal L(x_{t})$.

\paragraph{Annealed cross-fluctuations.}
Fix $n$. Because $\varphi_{0,t}^{-1}(\omega,\cdot)$ is $C^{1}$ and $p_{0}$ has finite $n$-th moments, the pathwise conditional expected fluctuation tensor $\mathbb{E}[\mathcal{F}^{(n)}_\rho | \Omega_{k,t}(\omega)]$ exists. To obtain deterministic statistics, we define the \emph{annealed} conditional expected fluctuation tensor, which we denote $\mathbf{E}_{k,t}^{(n)}$, by averaging its pathwise counterpart over all Wiener paths:
\[
  \mathbf{E}_{k,t}^{(n)} := \mathbb{E}_{\omega}\bigl[ \mathbb{E}[\mathcal{F}^{(n)}_\rho | \Omega_{k,t}(\omega)] \bigr].
\]
From this, the annealed normalised cross-fluctuation, $\mathcal{M}_{\text{ann}}^{(n)}$, is the cosine similarity between these deterministic annealed tensors:
\[
  \mathcal{M}_{\text{ann}}^{(n)}(\Omega_{1,t},\Omega_{2,t}) := \frac{\langle \mathbf{E}_{1,t}^{(n)}, \mathbf{E}_{2,t}^{(n)} \rangle_{\mathcal{H}_n}}{\|\mathbf{E}_{1,t}^{(n)}\|_{\mathcal{H}_n} \|\mathbf{E}_{2,t}^{(n)}\|_{\mathcal{H}_n}}.
\]
Since expectation commutes with the inner products and norms, all fluctuation identities remain valid for these annealed quantities. Thus, all merger-time results proved for the deterministic PF-ODE hold for the SDE in an expected sense, justifying the use of the same algorithms.

\subsubsection{Fluctuations offer fine-probes into the geometry of data}
\label{sec:exp-contract}
In this section, we illustrate that the analysis of fluctuations on a diffusion process provides a powerful probe into the intrinsic geometry of the data itself. Our main tool for this purpose is the following theorem,

\begin{theorem}[Exponential contraction of fluctuations and MST construction]\label{lem:geom-time}
Let \(\rho\) be a Lipschitz state operator, such that the components of the fluctuation tensor function $\mathcal{F}^{(n)}_\rho$ are square integrable with respect to the invariant measure $\mu$ of the diffusion (i.e., $\mathcal{F}^{(n)}_\rho \in L^2(\mu, \mathcal{H}_n)$). Let the evolution be governed by a diffusion semigroup \(P_t\) that is self adjoint on $L^2(\mu)$ and possesses a spectral gap $\lambda > 0$. 

Then, for any two initial event distributions $\mu_i, \mu_j$ with densities $h_i, h_j$ with respect to $\mu$, the distance between their conditional expected fluctuation tensors decays exponentially:
\[
\|\mathbb{E}_i[\mathcal{F}^{(n)}_\rho(t)] - \mathbb{E}_j[\mathcal{F}^{(n)}_\rho(t)]\|_{\mathcal{H}_n} \leq e^{-\lambda t} \, \|\mathcal{F}^{(n)}_\rho\|_{L^2(\mu, \mathcal{H}_n)} \, \|h_i - h_j\|_{L^2(\mu)}.
\]
Consequently, since the right-hand side converges to zero as $t \to \infty$, for any fixed \(\varepsilon > 0\), the graph on events with edges between pairs \((\Omega_i, \Omega_j)\) satisfying $\|\mathbb{E}_i[\mathcal{F}^{(n)}_\rho(t))] - \mathbb{E}_j[\mathcal{F}^{(n)}_\rho(t))]\|_{\mathcal{H}_n} \leq \varepsilon$ becomes a complete graph for sufficiently large $t$. A complete graph on a finite number of vertices always admits a well-defined minimal spanning tree (MST).
\end{theorem}
\begin{proof}
Before going into the details of the proof, we justify the reasonableness of the theorem's core assumptions in the context of a $d$ dimensional Variance-Preserving (VP) SDE used in diffusion models.

\begin{enumerate}
    \item \textbf{Existence of an invariant measure $\mu$}: The theorem assumes a stationary distribution $\mu$, which is clearly satisfied. This distribution for our case is $\mathcal{N}(\mathbf{0}_{d},\mathbf{I}_{d \times d})$ 

    \item \textbf{Self adjointness of the semigroup $P_t$}: The theorem critically relies on the self adjointness of the semigroup operator $P_t$ on the Hilbert space $L^2(\mu)$. The VP-SDE process is a \textbf{reversible} process with respect to its Gaussian invariant measure $\mu$. A fundamental result in the theory of Markov processes is that reversibility of a process with respect to a measure $\mu$ is equivalent to its generator being self-adjoint in $L^2(\mu)$, which in turn implies the self adjointness of the associated semigroup \citep{bakry2014analysis}.

    \item \textbf{Existence of a spectral gap $\lambda > 0$}: This is the crucial assumption providing the exponential decay rate. The VP-SDE process is a textbook example of a system with a spectral gap.

    \item \textbf{Square integrability of the fluctuation tensor $\mathcal{F}$}: The proof requires the norm $\|\mathcal{F}^{(n)}_\rho\|_{L^2(\mu, \mathcal{H}_n)}$ to be finite. The invariant measure $\mu$ is Gaussian, meaning its density decays extremely rapidly (exponentially in $\|x\|^2$). If the state operator $\rho$ is Lipschitz (a mild regularity condition), the components of the fluctuation tensor $\mathcal{F}^{(n)}_\rho$ will be polynomials in the state variables. Any polynomial function is square integrable with respect to a Gaussian measure, making this assumption easily satisfied.
\end{enumerate}
In summary, the VP-SDE process is a prime example for which all assumptions of the theorem hold, making the conclusion of exponential convergence robust and directly applicable. We now head towards the main proof.

The proof relies on the consequences of the spectral gap property of the diffusion semigroup $P_t$. Let $\mathcal{F} \equiv \mathcal{F}^{(n)}_\rho$ denote the tensor-valued function on the state space. The difference in conditional expectations at time $t$, for initial distributions $\mu_i, \mu_j$ with densities $h_i, h_j$ with respect to the invariant measure $\mu$, is a vector in the Hilbert space $\mathcal{H}_n$:
\[
D_{ij}(t) := \mathbb{E}_i[\mathcal{F}(t)] - \mathbb{E}_j[\mathcal{F}(t)] = \int (P_t \mathcal{F})(x) (h_i(x) - h_j(x)) \, d\mu(x).
\]
This is a Bochner integral of an $\mathcal{H}_n$-valued function against a scalar function. The semigroup $P_t$ is assumed to be self adjoint on the Hilbert space $L^2(\mu)$ of scalar functions. This property extends to the space of tensor-valued functions $L^2(\mu, \mathcal{H}_n)$. We can thus move the action of the semigroup from the function $\mathcal{F}$ to the density difference $(h_i - h_j)$:
\[
D_{ij}(t) = \int \mathcal{F}(x) \left(P_t(h_i - h_j)\right)(x) \, d\mu(x).
\]
We bound the norm of this integral using the Cauchy-Schwarz inequality for Bochner integrals, followed by the standard Cauchy-Schwarz inequality:
\begin{align*}
\|D_{ij}(t)\|_{\mathcal{H}_n} &\le \int \|\mathcal{F}(x)\|_{\mathcal{H}_n} |(P_t(h_i - h_j))(x)| \, d\mu(x) \\
&\le \left(\int \|\mathcal{F}(x)\|_{\mathcal{H}_n}^2 \, d\mu(x)\right)^{1/2} \left(\int |(P_t(h_i - h_j))(x)|^2 \, d\mu(x)\right)^{1/2} \\
&= \|\mathcal{F}\|_{L^2(\mu, \mathcal{H}_n)} \, \|P_t(h_i - h_j)\|_{L^2(\mu)}.
\end{align*}
The existence of a spectral gap $\lambda > 0$ is equivalent to the following contraction property on the space of mean-zero functions in $L^2(\mu)$: for any $g \in L^2(\mu)$ with $\int g \, d\mu = 0$, we have $\|P_t g\|_{L^2(\mu)} \le e^{-\lambda t} \|g\|_{L^2(\mu)}$. Since $h_i$ and $h_j$ are probability densities, their difference $g = h_i - h_j$ has zero mean. Applying the contraction property yields:
\[
\|\mathbb{E}_i[\mathcal{F}(t)] - \mathbb{E}_j[\mathcal{F}(t)]\|_{\mathcal{H}_n} \le e^{-\lambda t} \, \|\mathcal{F}\|_{L^2(\mu, \mathcal{H}_n)} \, \|h_i - h_j\|_{L^2(\mu)}.
\]
The term $\|\mathcal{F}\|_{L^2(\mu, \mathcal{H}_n)} \, \|h_i - h_j\|_{L^2(\mu)}$ is a constant for any fixed pair of events. The exponential term $e^{-\lambda t}$ guarantees that the distance converges to zero as $t \to \infty$. Therefore, for any threshold $\varepsilon > 0$, there exists a time $T$ such that for all $t > T$, the distance is less than $\varepsilon$ for all pairs $(i,j)$, making the corresponding graph complete. An MST can always be constructed on a weighted complete graph using standard algorithms such as Kruskal's \citep{Kruskal1956}.
\end{proof}

This exponential convergence justifies why we expect mergers to occur and provides a geometric interpretation of the process: initially distant event structures are pulled together until they collapse into a single point in the space of moment tensors. The above geometric perspective on the evolution of event structures has two key implications:
\begin{enumerate}
    \item \textbf{Hierarchical refinement.} We can progressively refine our analysis by choosing finer partitions of the initial state space. Starting with broad categories (e.g., animals vs. vehicles), we can track their mergers, then move to finer sub-partitions (e.g., cats vs. dogs) and track their subsequent mergers. This allows for probing the system's dynamics at increasingly fine resolutions, all on the same underlying data distribution.

    \item \textbf{Connection to manifold learning.} Tracking the evolution of the graph $G_t$ on events (where edge weights are given by the distance $\|\mathbb{E}_i[\mathcal{F}^{(n)}_\rho] - \mathbb{E}_j[\mathcal{F}^{(n)}_\rho]\|_{\mathcal{H}_n}$) is conceptually similar to algorithms that build neighborhood graphs to learn low-dimensional embeddings. Methods like t-SNE \citep{maaten2008visualizing} and UMAP \citep{mcinnes2018umap} also rely on connecting nearby points (or neighborhoods) to reveal underlying manifold structure. Our framework can be seen as applying a similar principle in the time domain, tracking how neighborhoods of events connect and merge as the diffusion process evolves. Exploring this connection in more detail is a promising direction for future work.
\end{enumerate}

\subsubsection{Unbiased monte-carlo estimators for cross fluctuations}
\label{sec:mcmc}

\begin{theorem}[One-sweep unbiasedness]\label{thm:one-sweep}
Let $\Omega_{1,0},\Omega_{2,0}\subseteq\Omega$ be disjoint events with probabilities $p_{1},p_{2}>0$. Simulate \emph{once} $N$ i.i.d.\ forward trajectories $\{\mathbf x^{(i)}_{t}\}_{t=0}^{T}$ from the VP process. Let $Z_{k}^{(i)}:=\mathbbm 1_{\Omega_{k,0}}\bigl(\mathbf x^{(i)}_{0}\bigr)$.

First, we form an unbiased estimator for the conditional expected fluctuation tensor itself:
\begin{equation}
  \widehat{\mathbf{E}}_{k,t}^{(n)} := \frac{1}{Np_k} \sum_{i=1}^N Z_k^{(i)} \mathcal{F}^{(n)}_\rho(\mathbf{x}_t^{(i)}), \qquad k \in \{1,2\},
  \label{eq:tensor_estimator}
\end{equation}
where $\mathcal{F}^{(n)}_\rho(\mathbf{x}_t^{(i)})$ is the random fluctuation tensor for sample $i$ at time $t$. This estimator is an average of tensors and is itself a tensor in $\mathcal{H}_n$.

Next, we define plug-in estimators for the unnormalised cross-fluctuation $G^{(n)}_\rho$ and the within-event fluctuation magnitude $\widehat{F}^{(2n)}_\rho$ based on these tensor estimators:
\begin{align}
  \widehat{G}^{(n)}_\rho(\Omega_{1,t},\Omega_{2,t}) &:= \langle \widehat{\mathbf{E}}_{1,t}^{(n)}, \widehat{\mathbf{E}}_{2,t}^{(n)} \rangle_{\mathcal{H}_n}, \\
  \widehat{\widehat{F}}^{(2n)}_\rho(\Omega_{k,t}) &:= \|\widehat{\mathbf{E}}_{k,t}^{(n)}\|_{\mathcal{H}_n}^2.
\end{align}
The tensor estimator $\widehat{\mathbf{E}}_{k,t}^{(n)}$ is unbiased for the true conditional expected fluctuation tensor $\mathbb{E}_k[\mathcal{F}^{(n)}_\rho(\Omega_{k,t})]$.
Consequently, the plug-in ratio estimator for the normalised cross-fluctuation,
\[
  \widehat{\mathcal M}^{(n)}_{\rho}(t)
    :=\frac{\widehat{G}^{(n)}_\rho(\Omega_{1,t},\Omega_{2,t})}
            {\sqrt{\widehat{\widehat{F}}^{(2n)}_\rho(\Omega_{1,t})\,
                   \widehat{\widehat{F}}^{(2n)}_\rho(\Omega_{2,t})}} = \frac{\langle \widehat{\mathbf{E}}_{1,t}^{(n)}, \widehat{\mathbf{E}}_{2,t}^{(n)} \rangle_{\mathcal{H}_n}}{\|\widehat{\mathbf{E}}_{1,t}^{(n)}\|_{\mathcal{H}_n} \|\widehat{\mathbf{E}}_{2,t}^{(n)}\|_{\mathcal{H}_n}},
\]
is consistent and asymptotically unbiased for the true value $\mathcal M^{(n)}_{\rho}(\Omega_{1,t},\Omega_{2,t})$:
\[
  \mathbb E \bigl[\widehat{\mathcal M}^{(n)}_{\rho}(t)\bigr]
    = \mathcal M^{(n)}_{\rho}(\Omega_{1,t},\Omega_{2,t})
      +O\!\bigl(N^{-1}\bigr).
\]
\end{theorem}

\begin{proof}
The proof proceeds in three steps, starting with the unbiasedness of the core tensor estimator.

\medskip
\noindent
\textit{Step 1: Unbiasedness of the tensor estimator $\widehat{\mathbf{E}}_{k,t}^{(n)}$.}
Fix $k\in\{1,2\}$. By the linearity of expectation, we can move the expectation inside the sum:
\begin{align*}
  \mathbb E_{\mathbb{P}}\!\bigl[\widehat{\mathbf{E}}_{k,t}^{(n)}\bigr]
   &= \frac{1}{Np_k} \sum_{i=1}^N \mathbb E_{\mathbb{P}}\!\Bigl[Z_{k}^{(i)}\mathcal{F}^{(n)}_\rho\bigl(\mathbf x^{(i)}_{t}\bigr)\Bigr].
\end{align*}
For each term in the sum, we condition on the starting point $\mathbf x_{0}^{(i)}$:
\begin{align*}
   \mathbb E_{\mathbb{P}}\!\Bigl[Z_{k}^{(i)}\mathcal{F}^{(n)}_\rho\bigl(\mathbf x^{(i)}_{t}\bigr)\Bigr]
   &= \int_{\mathbb R^d} \mathbbm{1}_{\Omega_{k,0}}(x_0) p_0(x_0) \mathbb{E}_{\mathbb{P}}[\mathcal{F}^{(n)}_\rho(x_t) | x_0] \, dx_0 \\
   &= p_k \left( \frac{1}{p_k} \int_{\Omega_{k,0}} p_0(x_0) \mathbb{E}_{\mathbb{P}}[\mathcal{F}^{(n)}_\rho(x_t) | x_0] \, dx_0 \right) \\
   &= p_k \, \mathbb{E}_k[\mathcal{F}^{(n)}_\rho(\Omega_{k,t})].
\end{align*}
Summing $N$ identical terms gives $N p_k \, \mathbb{E}_k[\mathcal{F}^{(n)}_\rho(\Omega_{k,t})]$. Dividing by the $Np_k$ prefactor proves that $\mathbb E_{\mathbb{P}}\!\bigl[\widehat{\mathbf{E}}_{k,t}^{(n)}\bigr] = \mathbb{E}_k[\mathcal{F}^{(n)}_\rho(\Omega_{k,t})]$.

\medskip
\noindent
\textit{Step 2: Consistency of plug-in estimators.}
The estimators $\widehat{G}^{(n)}_\rho$ and $\widehat{\widehat{F}}^{(2n)}_\rho$ are continuous functions (inner product and squared norm) of the tensor estimator $\widehat{\mathbf{E}}_{k,t}^{(n)}$. Since $\widehat{\mathbf{E}}_{k,t}^{(n)}$ is an average of i.i.d. random tensors, it is a consistent estimator for its mean by the Law of Large Numbers. By the continuous mapping theorem \citep{billingsley2012probability}, $\widehat{G}^{(n)}_\rho$ and $\widehat{\widehat{F}}^{(2n)}_\rho$ are therefore consistent estimators for $G^{(n)}_\rho$ and $\widehat{F}^{(2n)}_\rho$, respectively.

\medskip
\noindent
\textit{Step 3: Bias of the ratio estimator.}
The estimator $\widehat{\mathcal M}^{(n)}_{\rho}(t)$ is a smooth function of the components of the estimators $\widehat{\mathbf{E}}_{1,t}^{(n)}$ and $\widehat{\mathbf{E}}_{2,t}^{(n)}$. Each of these components is an average of $N$ i.i.d. random variables. By the multivariate Central Limit Theorem, the joint distribution of these averages converges to a normal distribution. The delta method for ratios of random variables then applies directly. A multivariate second-order Taylor expansion of the cosine similarity function around the true expected tensor values shows that the linear terms in the bias expansion vanish, and the bias is of order $O(N^{-1})$.
\end{proof}

\begin{remark}
All computations rely on the same set of $N$ forward trajectories. The primary computational step is to compute and store the random fluctuation tensors $\{\mathcal{F}^{(n)}_\rho(\mathbf{x}_t^{(i)})\}$ for each sample and time step, from which all other quantities can be derived.
\end{remark}

\subsection{Framework connections, extensions, and interpretation}
\label{sec:frm-wk}

\subsubsection{Mixing time of isotropic Gaussians under Brownian diffusion}
\label{sec:vp-mixing}

We quantify how long the forward VP–SDE~\eqref{eq:vp-sde} needs to
\emph{forget} an isotropic sub-Gaussian input and become
$\varepsilon$-close (in total variation) to its Gaussian limit.
Write
\[
  J(t)\;=\;\exp \Bigl(-\tfrac12\!\int_{0}^{t}\beta(s)\,ds\Bigr),
\]
the deterministic attenuation factor from~\eqref{eq:marginal}.

\begin{proposition}[Mixing time for sub-Gaussian data]
\label{prop:vp-mixing}
Let $\mathbf y\!\in\!\mathbb R^{d}$ have i.i.d.\ mean-zero,
unit-variance components that are $\sigma^{2}$-sub-Gaussian:
$\Pr(|y_{i}|>t)\le2e^{-t^{2}/2\sigma^{2}}$.
Evolve $\mathbf y$ with the VP–SDE \eqref{eq:vp-sde} and denote
$\mathcal L(\mathbf x_{t})=p_{t}$.
For every $\varepsilon\!\in(0,1)$ the
$\varepsilon$-mixing time
\[
  t_{\mathrm{mix}}(\varepsilon)
    :=\inf\bigl\{t\ge0:
      d_{\mathrm{TV}}\bigl(p_{t},\mathcal N(0,I_{d})\bigr)\le\varepsilon
    \bigr\}
\]
satisfies
\[
  J\!\bigl(t_{\mathrm{mix}}(\varepsilon)\bigr)
    \;\le\;
    \frac{2}{d}\,
    \Bigl(1+O\bigl(\sigma^{2}\log\tfrac1\varepsilon\bigr)\Bigr),
\quad\text{and}\quad
  J\!\bigl(t_{\mathrm{mix}}(e^{-1})\bigr)=\Theta(d^{-1}).
\]
\end{proposition}

\begin{proof}
\textit{Step\,1: tails of the input.}
Sub-Gaussianity yields
$\mathbb E\|\mathbf y\|_{2}^{2}=d$ and
$\operatorname{Var}\|\mathbf y\|_{2}^{2}\le C d$
(for $C=C(\sigma)$).  Bernstein's inequality \cite{vershynin2018high} gives
$\|\mathbf y\|_{2}^{2}=d\pm O(\sqrt d)$ w.h.p.

\textit{Step\,2: second moment under the SDE.}
Conditioned on $\mathbf y$,
\(
  \mathbb E\|\mathbf x_{t}\|_{2}^{2}
  = J(t)^{2}\|\mathbf y\|_{2}^{2}+d\bigl(1-J(t)^{2}\bigr),
\)
so averaging produces
\(
  \mathbb E\|\mathbf x_{t}\|_{2}^{2}=d+J(t)^{2}O(\sqrt d).
\)

\textit{Step\,3: bounding \(\chi^{2}\).}
Pinsker's inequality \cite{cover2006elements} gives
$d_{\mathrm{TV}}^{2}\le\frac12\chi^{2}$.
For Gaussians with equal means,
$\chi^{2}=(\det\Sigma)^{-1/2}
          \exp\bigl(\tfrac12\operatorname{tr}(I-\Sigma^{-1})\bigr)-1$.
With $\Sigma=J(t)^{2}I_{d}+(1-J(t)^{2})I_{d}$,
\[
  d_{\mathrm{TV}}^{2}(p_{t},\mathcal N)
    \le \tfrac12 d\,\frac{J(t)^{4}}{1-J(t)^{2}}\bigl(1+O(d^{-1/2})\bigr).
\]

\textit{Step\,4: solve for \(J(t)\).}
Setting the rhs to $\varepsilon^{2}$ and solving yields the claimed
bound.
\end{proof}

\paragraph{Closed form for a linear schedule.}
With $\beta(t)=\beta_{0}+(\beta_{T}-\beta_{0})t/T$,
\[
  J(t)=\exp \Bigl(
    -\tfrac12\beta_{0}t
    -\tfrac14(\beta_{T}-\beta_{0})\,\frac{t^{2}}{T}
  \Bigr).
\]
Taking $\varepsilon=e^{-1}$ in \Cref{prop:vp-mixing},
$\log J(t_{\mathrm{mix}})\simeq-\log d+\log 2$, so $t_{\mathrm{mix}}$
solves a quadratic.  For the DDPM
defaults $(\beta_{0},\beta_{T},T)=(10^{-4},0.02,1000)$:
\[
  t_{\mathrm{mix}}+0.0995\,t_{\mathrm{mix}}^{2}
    =5000\log\!\bigl(d/2\bigr).
\]

\begin{table}[H]
\centering\small
\begin{tabular}{|l|c|c|}
\hline
\rowcolor{gray!20}
\textbf{Data dimension} &
\textbf{Pred.\ $t_{\mathrm{mix}}/T$} &
\textbf{Obs.\ $i^{\star}/T$}
\\ \hline
$3\times32\times32$ (CIFAR-10) & 0.602 & 0.60 \\ \hline
$1\times28\times28$ (MNIST)    & 0.543 & 0.60 \\ \hline
$4\times32\times32$ (ImageNet latents) & 0.614 & 0.70 \\ \hline
\end{tabular}
\vspace{0.3em}
\caption{Theoretical mixing index vs.\ empirical convergence index.}
\label{tab:eq-times}
\end{table}
\Cref{tab:eq-times} shows theory versus measured convergence
indices; the $\Theta(d^{-1})$ scaling persists on real data.

Thus, it is possible to estimate the mixing time for sub-gaussian data analytically, in fact due to concentration bounds on high-dimensional sub-gaussians \cite{vershynin2018high} it could be shown that the above time manifests physically as a symmetry breaking transition, leading to an estimate consistent with \cite{raya2024spontaneousiop} for spherically symmetric distributions and in \cite{biroli2024dynamical} for gaussian mixtures. Interestingly, ImageNet latent representations align closely with these theoretical estimates. We hypothesise that this occurs because the compressed space corresponds to the latent space of a Variational Autoencoder (VAE) trained to approximate a Gaussian distribution, potentially making these latents effectively sub-Gaussian. Verification of this hypothesis and related questions remains future work.

\subsubsection{Higher-order fluctuations as a proof of concept}
\label{sec:hi-ord-fluc}
To make higher-order analysis computationally tractable, we transition from analyzing the full rank-$n$ fluctuation tensor $\mathcal{F}^{(n)}_\rho$ to analyzing the moments of a corresponding scalar-valued random variable. This simplification is equivalent to assuming the components of the state vector are i.i.d. and studying the dynamics of a single component. This bypasses tensor computations and allows us to work with scalar moments.

\begin{figure*}[htbp!]
\vskip 0.2in
\begin{center}
\begin{minipage}[b]{0.49\textwidth}
    \centering
    \includegraphics[width=\textwidth]{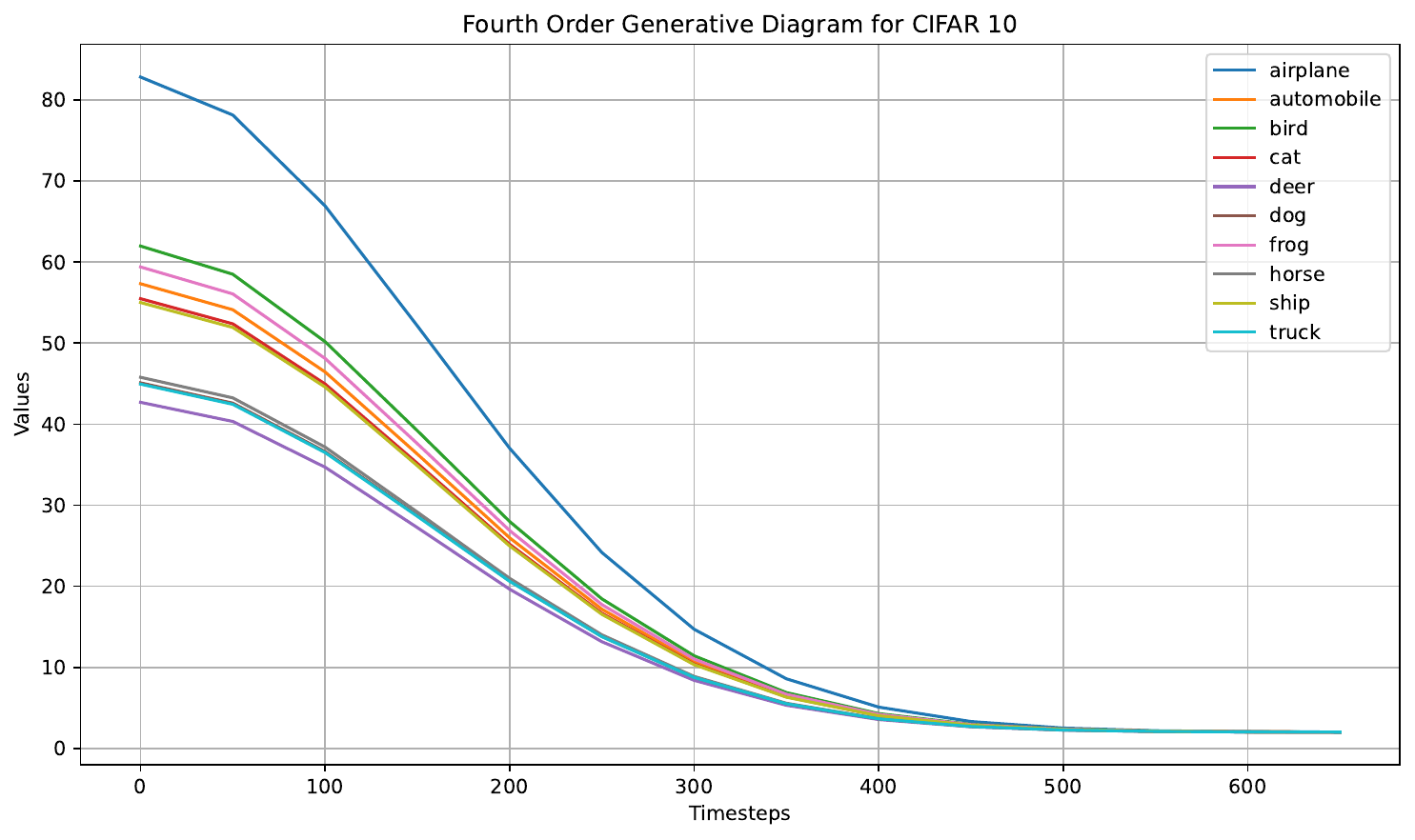}
    \caption{Fourth order generative diagram for CIFAR10.We show the emergence of classes using fourth-order correlations. }
    \label{fig:fourth-cifar}
\end{minipage}%
\hfill
\begin{minipage}[b]{0.49\textwidth}
    \centering
    \includegraphics[width=\textwidth]{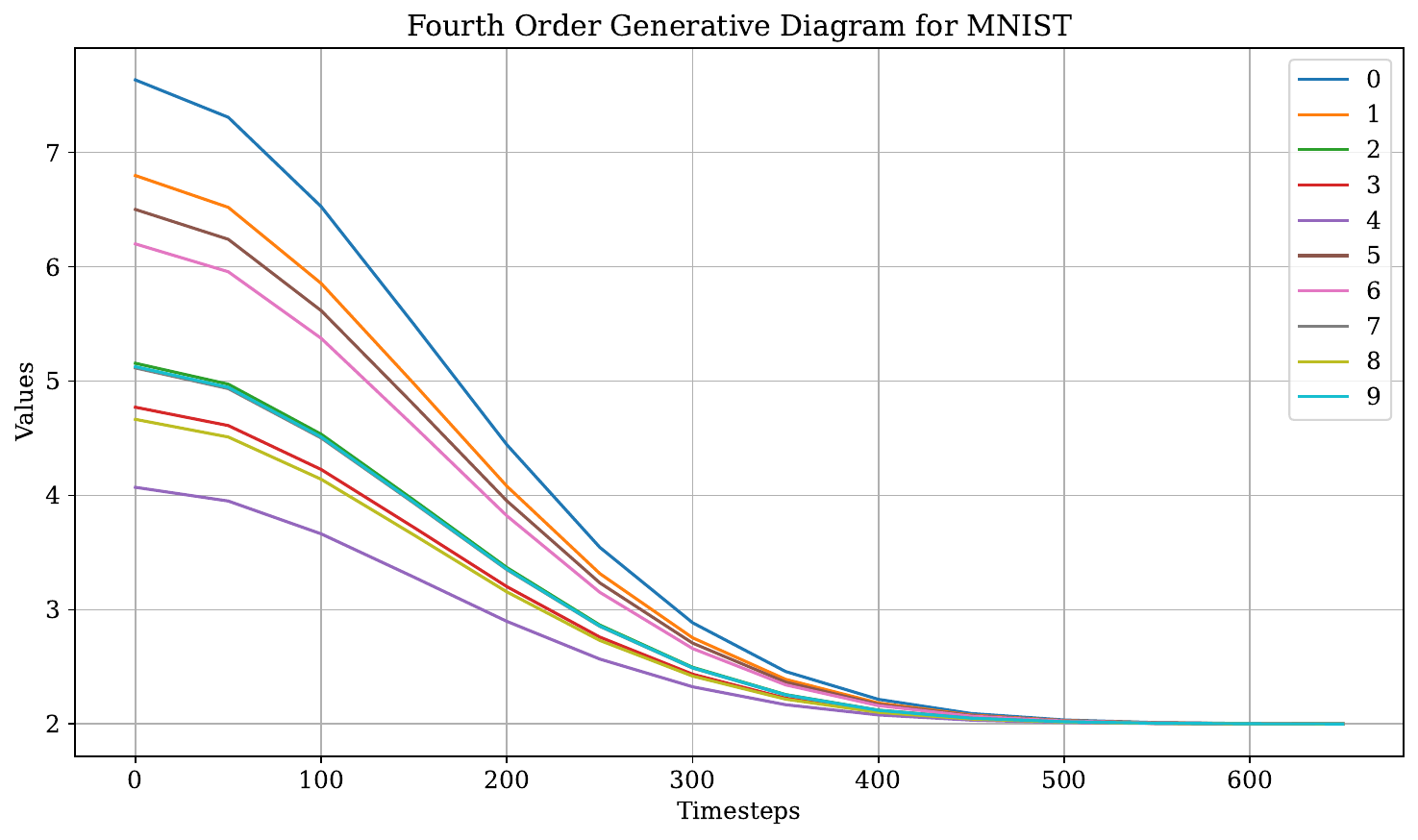}
    \caption{Fourth order generative diagram for MNIST.We show the emergence of classes using fourth-order correlations. }
    \label{fig:fourth-mnist}
\end{minipage}
\end{center}
\vskip 0.2in
\end{figure*}

Under this assumption, we can track the scalar moment identity:
\[
    \widehat{\mu}_{i}^{(n)}(t)
     = J(t)^{n}\,\widehat{\mu}_{i}^{(n)}(0)
       +\bigl(1-J(t)^{n}\bigr)\widehat{\mu}_{\mathcal N}^{(n)},
\]
where $\widehat{\mu}_{i}^{(n)}(t)$ is the $n$-th centered moment for class $i$ at time $t$. For the Gaussian limit, Isserlis'/Wick's theorem\citep{isserlis_1918_1431593,wick1950evaluation} can be applied because the components are treated as jointly Gaussian. The theorem states that higher-order moments of a Gaussian are polynomials in the variance, and odd moments are zero. This rule lets us draw the higher-order generative diagrams in Figures \ref{fig:fourth-cifar}-\ref{fig:fourth-mnist}.
Compared with the second-order diagram,
high-order curves fan out more widely at early times—evidence that
non-linear features dominate, but they collapse sooner, consistent with
the rapid \(J(t)^{n}\) decay.

\subsubsection{Centred kernel alignment (CKA)}
\label{sec:cka}

Kernel alignment measures how similarly two Gram matrices embed the
\emph{same} data.  For $K,L\in\mathbb R^{n\times n}$
the uncentred score is the cosine in $\mathbb R^{n^{2}}$
\citep{cristianini2001kernel}:
\[
  A(K,L)=\frac{\langle K,L\rangle_{F}}
               {\|K\|_{F}\,\|L\|_{F}},
  \quad
  \langle K,L\rangle_{F}:=\!\sum_{ij}K_{ij}L_{ij}.
\]
Because $A$ reacts to mean shifts,  
\citet{cortes2012algorithms} introduced \emph{centred kernel alignment}
\[
  A_{\mathrm c}(K,L):=
    A \bigl(HKH,HLH\bigr),
  \quad
  H:=I_{n}-\tfrac1n\mathbf1\mathbf1^{\!\top}.
\]
$A_{\mathrm c}$ is invariant to any feature-space translation and, for
linear kernels, to orthogonal mixing and isotropic scaling.

\paragraph{CKA for covariance kernels.}
\Cref{sec:class-cond} sets
$\tilde K=\Sigma_{i,t}$,
$\tilde L=\Sigma_{j,t}$.  For symmetric matrices
\begin{align}
  \langle\tilde K,\tilde L\rangle_{F}
    &=\sum_{m=1}^{d}\lambda_{i,t,m}\,\lambda_{j,t,m},\\
  \|\tilde K\|_{F}^{2}
    &=\sum_{m=1}^{d}\lambda_{i,t,m}^{2},\qquad
  \|\tilde L\|_{F}^{2}
    =\sum_{m=1}^{d}\lambda_{j,t,m}^{2},
\label{eq:cka-eigs}
\end{align}
where $\{\lambda_{i,t,m}\}$ are the eigenvalues of
$\Sigma_{i,t}$ and \emph{mutatis mutandis} for $\{\lambda_{i,t,m}\}$.
Under the VP flow they contract as
\(
  \lambda_{i,t,m}
   = \lambda_{i,0,m}J(t)^{2}+\bigl(1-J(t)^{2}\bigr)
\)
\citep{biroli2024dynamical}.  The largest eigenvalue dominates once
$J(t)^{2}\ll\lambda_{i,0,2}/\lambda_{i,0,1}$, so
\[
  A_{\mathrm c}\bigl(\Sigma_{i,t},
                     \Sigma_{j,t})\bigr)\!\uparrow 1
  \iff
  \lambda_{i,t,1}\simeq\lambda_{j,t,1}.
\]
Therefore tracking the top eigenvalues
$\lambda_{i,t,1},\lambda_{j,t,1}$ gives an efficient proxy for merger of
$\Omega_{i,t}$ and $\Omega_{j,t}$ in \Cref{sec:class-cond}.

\subsubsection{Structural regularity bounds fluctuations}
\label{sec:reg-fluc}

This section provides a more rigorous justification for the Fluctuation adaptation lemma from \Cref{sec:zero-shot-style}, which posits that structurally similar distributions exhibit similar fluctuation dynamics. The premise is the Fourier regularity condition from \cref{eq:fourier-regularity}, stating that the $L^2$ distance between the characteristic functions (CFs) of the source and target-style distributions is small: $\|\widehat{p}_0 - \widehat{p}^\star\|_{L^2} \le \delta$. The challenge is to show that this integral bound on the CFs implies a bound on the fluctuation tensors, which are determined by pointwise derivatives of the CFs at the origin.

The bridge between these two concepts is provided by the \emph{Sobolev Embedding Theorem} \citep{adams2003sobolev}. This theorem connects a function's \emph{average} smoothness, measured by its Sobolev norm $\|f\|_{H^s}$, to its \emph{pointwise} smoothness, such as the boundedness of its derivatives. However, the Sobolev norm, $\|g\|_{H^s}^2 = \int (1 + |\boldsymbol{\xi}|^2)^s |g(\boldsymbol{\xi})|^2 \, d\boldsymbol{\xi}$, is stronger than the standard $L^2$ norm due to the frequency-weighting factor $(1 + |\boldsymbol{\xi}|^2)^s$. Therefore, a small $L^2$ norm does not automatically guarantee a small Sobolev norm. To bridge this gap, we must introduce a more explicit regularity assumption on the style transfer process itself:

\emph{Assumption:} The style transfer transformation $\mathcal{T}_{\text{style}}$ is assumed to be sufficiently regular such that the difference in characteristic functions, $g(\boldsymbol{\xi}) = \widehat{p}_0(\boldsymbol{\xi}) - \widehat{p}^\star(\boldsymbol{\xi})$, not only has a small $L^2$ norm but also has rapidly decaying high-frequency content. Formally, this means there exists a constant $K$ such that $\|g\|_{H^s} \le K \|g\|_{L^2}$ for the required Sobolev order $s$. This assumption holds if, for instance, the style transfer primarily modifies low-to-mid frequency components of the distribution, a common case for artistic stylisation that preserves core structures.

With this assumption, the argument is complete. The Sobolev Embedding Theorem can now be directly applied:
\[
|D^\alpha g(0)| \le C \|g\|_{H^s} \le CK \|g\|_{L^2} \le CK\delta.
\]
This step guarantees that the difference between each individual \emph{scalar component} of the corresponding moment tensors is bounded by $O(\delta)$. This component-wise bound is then extended to a norm on the entire fluctuation tensor. The conditional expected fluctuation tensor, $\mathbb{E}_{0,\Omega}[\mathcal{F}^{(n)}_\rho]$, is an element of the Hilbert space $\mathcal{H}_n$ of rank-$n$ tensors. Let $\Delta_{\mathcal{F}} = \mathbb{E}_{0,\Omega}[\mathcal{F}^{(n)}_\rho] - \mathbb{E}_{\star,\Omega^\star}[\mathcal{F}^{(n)}_\rho]$ be the difference tensor. The squared Frobenius norm of this tensor is the sum of the squared magnitudes of its components: $\|\Delta_{\mathcal{F}}\|_{\mathcal{H}_n}^2 = \sum_{\mathbf{j}} |(\Delta_{\mathcal{F}})_{\mathbf{j}}|^2$. Since we established that each component is bounded, $|(\Delta_{\mathcal{F}})_{\mathbf{j}}| \le C_{\mathbf{j}}\delta$, the norm of the full tensor is also bounded:
\[
  \left\| \mathbb{E}_{0,\Omega}[\mathcal{F}^{(n)}_\rho] - \mathbb{E}_{\star,\Omega^\star}[\mathcal{F}^{(n)}_\rho] \right\|_{\mathcal{H}_n} \le C'_{n}\delta.
\]
Finally, since the normalised cross-fluctuation $\mathcal{M}^{(n)}_\rho$ is defined as the cosine of the angle between these tensors in the Hilbert space $\mathcal{H}_n$, it is a continuous function of its tensor arguments \citep{conway1990course}. Therefore, a small perturbation in the input tensors, bounded by $O(\delta)$, leads to a correspondingly small change in the value of $\mathcal{M}^{(n)}_\rho$. This ensures that the merger times are stable under the stylistic transformation, rigorously justifying the use of a schedule computed on the source distribution for the zero-shot style transfer task.

\subsubsection{Extending the framework to certain non Markovian samplers}
\label{sec:nonmarkov-tail}

\noindent
For a \emph{non Markovian} latent chain the marginal \(p_{i}\) depends on
the entire future tail 
\(\{p_{i+1},p_{i+2},\dots ,p_{n}\}\).
Hence the pull–back
\(\Omega_{k,0}\mapsto\Omega_{k,i}\) is well defined only if every
conditional kernel beyond step \(i\) is known.  
This hurdle disappears when the latent family belongs to a
\emph{natural exponential family} (NEF).

\paragraph{Tail statistic Markovisation.}
An NEF on \(\mathbb R^{d}\) has densities $  p_{\boldsymbol\theta}(x)
     = h(x)\exp \bigl(\langle \boldsymbol\theta,T(x)\rangle
                       -A(\boldsymbol\theta)\bigr),$ with sufficient statistic \(T\) and log-partition function \(A\).
For an \emph{independent} sequence \(\{X_{i}\}_{i=1}^{n}\) drawn from an
NEF, define the \emph{tail statistic}
\[
  G_{i}:=\sum_{t=i}^{n}T(X_{t}),
  \qquad i=1,\dots ,n.
\]
The Pitman–Koopman–Darmois theorem gives
\begin{theorem}[Tail statistic Markov property
  {\citealp{pitman1936sufficient}}]\label{thm:pkd}
  \leavevmode
  \begin{enumerate}[label=(\roman*), leftmargin=*]
  \item If \(\{X_{i}\}\) are i.i.d.\ from an NEF, the conditional sequence
        \(\{\mathcal L(X_{i}\mid G_{i})\}_{i=1}^{n}\) is
        first-order Markov.
  \item Conversely, if a statistic sequence \(\{G_{i}\}\) makes
        \(\{\mathcal L(X_{i}\mid G_{i})\}\) Markov for \emph{all} \(n\),
        then the marginals must form an NEF.
  \end{enumerate}
\end{theorem}

Thus the random vector \(G_{i}\) captures \emph{all} future information
relevant at step~\(i\).

\paragraph{Injecting the tail statistic into fluctuations.}
Fix disjoint initial events \(\Omega_{1,0},\Omega_{2,0}\).
Condition on \(G_{i}=g\) and apply the deterministic PF-ODE of \Cref{sec:sr-es-disj} \emph{inside the fibre}
\(\{X_{i}\mid G_{i}=g\}\):
\[
  \Omega_{k,i}(g)
    :=\bigl\{x\in\mathbb R^{d} :
              \varphi^{-1}_{0\to i}(g,x)\in\Omega_{k,0}\bigr\},
  \qquad k\in\{1,2\}.
\]
Because the conditioned kernels are Markov
(\Cref{thm:pkd}), this construction mirrors the purely Markovian
case. The pathwise cross-fluctuation $\mathcal M^{(n)}_{\rho}\!\bigl(\Omega_{1,i}(g),\Omega_{2,i}(g)\bigr)$ is the cosine similarity between the conditional expected fluctuation tensors for a given $g$. Averaging over the law of \(G_{i}\) yields the \emph{annealed} cross-fluctuation:
\[
  \overline{\mathcal M}^{(n)}_{\rho}(\Omega_{1,i},\Omega_{2,i})
    :=\int
        \mathcal M^{(n)}_{\rho}\!\bigl(
          \Omega_{1,i}(g),\Omega_{2,i}(g)\bigr)\,
      d\mathbb P_{G_{i}}(g).
\]
Every algebraic identity from the main framework now carries over to this tail-averaged counterpart. A merger is detected when $|\overline{\mathcal M}^{(n)}_{\rho}| \to 1$. Star-DDPM \cite{okhotin2023star} is a recent work that uses a similar formulation to obtain non Markovian diffusion generative models. 

Thus, whenever a non Markovian diffusion admits a finite-dimensional
\emph{tail statistic}—a property guaranteed for exponential-family
latents—conditioning on that statistic restores the Markov property and lets the fluctuation framework operate unchanged.  Identifying broader classes of tail markovizable samplers is a promising direction for
future work.

\subsubsection{Phases of diffusion model dynamics}
\label{sec:phase-transit}

We analyze diffusion model dynamics by distinguishing between two types of phase transitions. Our approach is inspired by discretisation based analyses in physics \citep{kogut1983lattice}, but we treat the system's discrete nature, its time steps and finite precision—as a fundamental property, not a computational artifact. This leads us to differentiate between:

\begin{enumerate}[label=(\arabic*), leftmargin=*]
\item \textbf{Thermodynamic.}
A discontinuity in the $n$-th classical derivative of a system-wide property, $(\Phi^{(n)})$, at some time $t_0$ (assuming $(\Phi\in C^{n})$). Such transitions, studied in the continuum limit, are typically \emph{exclusive}, meaning at most one can occur for the whole system at a given time \citep{biroli2024dynamical}.

\item \textbf{Lattice.}
For a given step size $\tau > 0$, if there exists a discontinuity threshold $\varrho_{\text{disc}} > 0$ such that for the $n$-th finite differences from the left ($\mathrm{LD}^{(n)}_{\tau}\Phi$) and right ($\mathrm{RD}^{(n)}_{\tau}\Phi$), the below inequality holds, then the observable $\Phi$ undergoes a \emph{lattice} transition at step $t_0$ of \emph{order $n$}.
\[
    \|\mathrm{LD}^{(n)}_{\tau}\Phi(t_0) - \mathrm{RD}^{(n)}_{\tau}\Phi(t_0)\| \geq \varrho_{\text{disc}}.
\]
Lattice transitions include thermodynamic ones in the limit $\tau \to 0$ but are generally \emph{non-exclusive}.
\end{enumerate}

Our core observable for detecting these transitions is the absolute normalised cross-fluctuation, $|\mathcal{M}^{(n)}_{\rho}|$. This quantity measures the cosine similarity of the conditional expected fluctuation tensors of two evolving events. A merger signifies that these tensors have aligned, i.e., $|\mathcal{M}^{(n)}_{\rho}| \to 1$. Our merger time $i^\star$ is precisely a lattice transition point for this observable, defined by our chosen \emph{merger distance threshold} $\varepsilon$ (as in \cref{eq:time-dependent-M}).

\begin{theorem}
The merger event detected by $\tilde{\mathcal{M}}^{(n)}_{\rho}$ as defined by our \emph{merger distance threshold} $\varepsilon$ in \cref{eq:time-dependent-M} constitutes a lattice transition at the merger time $i^\star$.
\end{theorem}
\begin{proof}
Consider two initially disjoint events $\Omega_{1,0},\Omega_{2,0} \subset \Omega_0$. Let $i^\star$ denote the merger time given by \cref{eq:time-dependent-M}. From the topological equivalence proved in \Cref{lem:cross-straight}, there exists a constant $\vartheta>0$ such that at the merger,
\[
    \big||\mathcal{M}^{(n)}_{\rho}(\Omega_{1,i^\star}, \Omega_{2,i^\star})| - 1\big| \leq \vartheta.
\]
By selecting a suitable \emph{merger distance threshold} $\varepsilon > 0$ in the metric $d_n(\cdot,\cdot)$ used in \cref{eq:time-dependent-M}, we can ensure $\vartheta < 1/2$. Let our observable be $\Phi(i) = |\mathcal{M}^{(n)}_{\rho}(\Omega_{1,i}, \Omega_{2,i})|$. It follows that $\Phi(i^\star) \geq 1 - \vartheta$. Due to the hypercontractivity of the underlying Brownian motion, the structural similarity between events is monotonically increasing, hence $\Phi(i)$ is monotonically increasing in $i$. This monotonicity implies that for a discrete step size $\tau$, the rate of approach to the merger is bounded. For the $n$-th order finite differences, this can be shown to satisfy:
\[
    \mathrm{RD}^{(n)}_{\tau} \bigl(\Phi(i^\star)\bigr) \leq \frac{\vartheta}{\tau^n},
    \quad \text{and} \quad
    \mathrm{LD}^{(n)}_{\tau} \bigl(\Phi(i^\star)\bigr) \geq \frac{1 - \vartheta - \Phi(i^\star-n\tau)}{\tau^n}.
\]
The key insight is that the jump into the merged state is a finite, discrete event. Applying the reverse triangle inequality to the difference in the finite derivatives gives:
\[
    \big\|\mathrm{LD}^{(n)}_{\tau}(\Phi(i^\star)) - \mathrm{RD}^{(n)}_{\tau}(\Phi(i^\star))\big\| > 0,
\]
which establishes the existence of a lattice transition at the merger time, by demonstrating a non-zero jump in the finite differences, which can be thresholded by a chosen $\varepsilon_{\text{disc}}$. Notice that as this relies on finite $\tau$ and $\vartheta$, this transition is not necessarily thermodynamic.
\end{proof}

\paragraph{Thermodynamic phases from prior works.}
For class-conditioned VP diffusion,
\citet{biroli2024dynamical} proved two
thermodynamic boundaries
\(t_{\mathrm{u}\to\mathrm{s}}\) (unbiased \(\to\) speciation) and
\(t_{\mathrm{s}\to\mathrm{c}}\) (speciation \(\to\) condensation):
\[
  \text{unbiased}\;[0,t_{\mathrm{u}\to\mathrm{s}})
  \subset
  \text{speciation}\;
    (t_{\mathrm{u}\to\mathrm{s}},t_{\mathrm{s}\to\mathrm{c}})
  \subset
  \text{condensation}\;(t_{\mathrm{s}\to\mathrm{c}},T].
\]

\paragraph{Relation of class conditional lattice mergers to thermodynamic phases.} For two classes \(k\neq\ell\) define the centred cross-fluctuation \(\mathcal{M}_{k\ell}(t)\) 
(\eqref{eq:mk-l} in \Cref{sec:class-cond}).  
Its \(\varepsilon\)-merger time is
\[
  t^{\mathrm{lat}}_{k\ell}(\varepsilon)
     :=\inf\{t\ge0:\mathcal{M}_{k\ell}(t)\ge1-\varepsilon\},
     \qquad \varepsilon\in(0,1).
\]

\begin{lemma}[Merger times lie inside the speciation phase]
\label{lem:speciation-only}
For all \(k\neq\ell\) and \(\varepsilon\in(0,1)\),
\[
  t_{\mathrm{u}\to\mathrm{s}}
     < t^{\mathrm{lat}}_{k\ell}(\varepsilon)
     \le t_{\mathrm{s}\to\mathrm{c}} .
\]
\end{lemma}

\begin{proof}
\textit{Unbiased phase.}
If \(t<t_{\mathrm{u}\to\mathrm{s}}\) then \(p_{k,t} = p_{l,t}\), so
\(\mathcal{M}_{k\ell}(t)=1\); no upward crossing can occur.

\smallskip
\noindent\textit{Condensation phase.}
For \(t>t_{\mathrm{s}\to\mathrm{c}}\) each covariance \eqref{eq:sigma-t} satisfies  
\(\lambda_{\max}(\Sigma_{k,t})
     \le e^{-c(t-t_{\mathrm{s}\to\mathrm{c}})}\)
\citep[Prop.~4]{biroli2024dynamical}.
Using \eqref{eq:cka-eigs},
\(1-\mathcal{M}_{k\ell}(t)=O(e^{-c(t-t_{\mathrm{s}\to\mathrm{c}})})\);
hence \(\mathcal{M}_{k\ell}(t)\ge1-\varepsilon\) for \emph{all} large \(t\).
No new crossing can start after \(t_{\mathrm{s}\to\mathrm{c}}\).

\smallskip
\noindent\textit{Speciation phase.}
Because a crossing cannot start before \(t_{\mathrm{u}\to\mathrm{s}}\)
or after \(t_{\mathrm{s}\to\mathrm{c}}\), any merger time must lie in
\((t_{\mathrm{u}\to\mathrm{s}},t_{\mathrm{s}\to\mathrm{c}}]\).
\end{proof}

\begin{figure*}[t!]
    \centering
    \captionsetup{font=small, labelfont=bf}
    \begin{subfigure}[b]{0.49\textwidth}
        \centering
        \includegraphics[width=\textwidth]{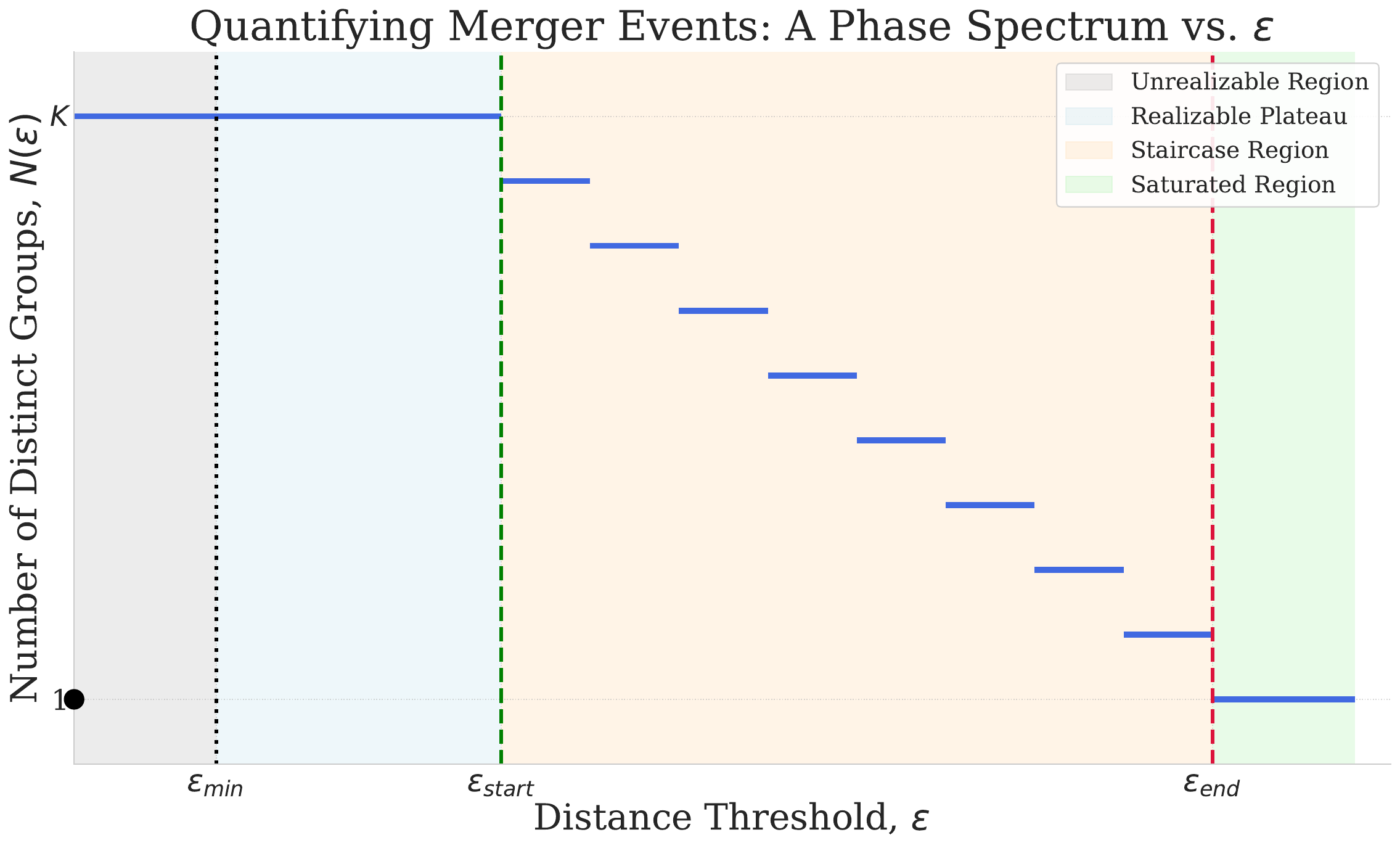}
        \caption{The phase spectrum of merger events}
        \label{fig:phase_diagram_app}
    \end{subfigure}
    \hfill
    \begin{subfigure}[b]{0.49\textwidth}
        \centering
        \includegraphics[width=\textwidth]{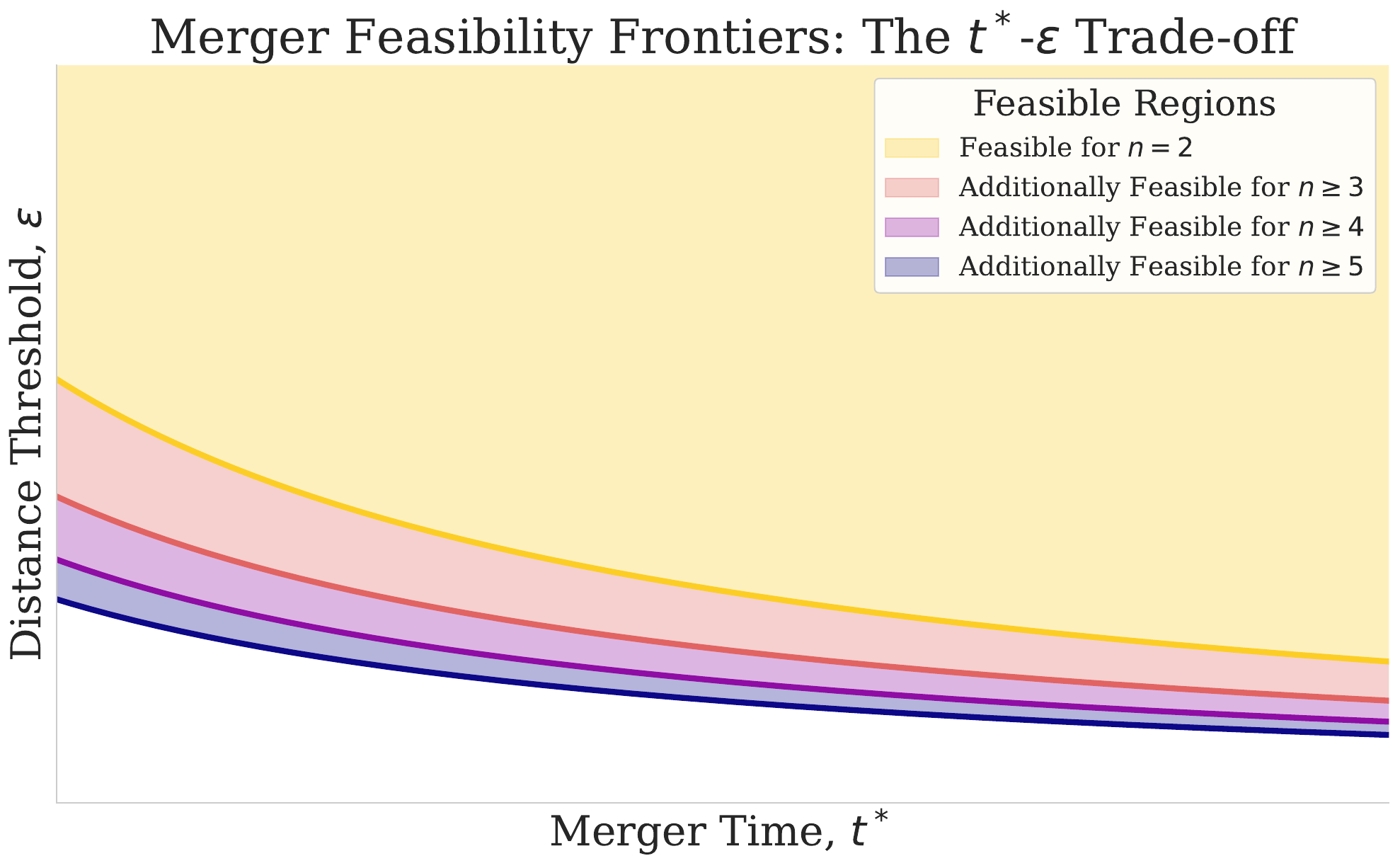}
        \caption{Merger feasibility frontiers}
        \label{fig:merger_curves_app}
    \end{subfigure}
    \caption{\textbf{Conceptual visualisation of lattice transition dynamics.} 
    \textbf{(a)}~The phase diagram illustrates how the number of merger events changes with the threshold $\epsilon$. Adopting a lattice perspective reveals a spectrum of distinct phases: an unrealizable  thermodynamic phase at $\epsilon =0$ which admits a single speciation transition,  an unrealizable region below the system's precision limit, a realizable plateau of maximum potential mergers, a staircase of sequential transitions where the system "splits" into fine-grained merger states, and a final saturated state.
    \textbf{(b)}~The feasibility frontiers visualise the trade-off between merger time $t^*$ and threshold $\epsilon$. The relationship is analogous to an uncertainty principle, where higher-order fluctuations merge faster, shifting their feasibility frontiers to the left.}
    \label{fig:conceptual_plots_app}
\end{figure*}

\paragraph{The phase spectrum.}
\Cref{fig:conceptual_plots_app}(a) details the spectrum of phases revealed by our lattice approach. For any choice of $\epsilon > 0$, the system can be characterised by the number of merger events required to connect all events in a partition. The plot visualises a sequence of distinct states:
\begin{enumerate}[leftmargin=*,label=\roman*.]
    \item \textbf{Unrealizable thermodynamic phase $\epsilon = 0$:} An unattainable limit which admits only a single possible merger at $t \to  \infty$. 
    \item \textbf{Unrealizable phase ($0 < \epsilon < \epsilon_{min}$):} Below the computational precision limit, a maximal number of $K$ potential mergers exist, but they are not observable.
    \item \textbf{Realizable plateau ($\epsilon_{min} \le \epsilon < \epsilon_{start}$):} The system is observable and the threshold allows for maximum number of mergers possible.
    \item \textbf{Staircase phase ($\epsilon_{start} \le \epsilon < \epsilon_{end}$):} As $\epsilon$ increases, it becomes less sensitive to the structural differences between events. The single macro-state of "maximum potential mergers" splits into a spectrum of fine-grained states, visualised as a downward staircase. Each step is a lattice transition where the number of distinct merger events required to connect the class graph decreases.
    \item \textbf{Saturated phase ($\epsilon \ge \epsilon_{end}$):} The threshold is so large that a single overarching rule merges all events. The system becomes a singleton for discrimination purposes, losing all distinguishing power.
\end{enumerate}

\paragraph{Feasibility and uncertainty.}
\Cref{fig:conceptual_plots_app}(b) visualises a fundamental trade-off in our framework. To precisely identify a merger (small $\epsilon$), one must observe the system for a long time (large $t^*$). Conversely, to detect a merger quickly (small $t^*$), one must accept a lenient threshold (large $\epsilon$). This relationship, which holds for any fluctuation tensor, is analogous to an uncertainty principle. For the specific, practical case where the state operator is the identity, $\rho(x)=x$, the plot shows an additional phenomenon: higher-order fluctuations ($n$) merge faster. This is because when analyzing the data vectors directly, high-order statistics capture fine-grained details that are more fragile and are erased more quickly by the diffusion process. The layered shading illustrates how the "feasible region" of detectable mergers expands as one moves to higher orders at any given time for this specific choice of $\rho$.

Thus, lattice transitions give a fine-grained view inside the most relevant operational regimes of standard diffusion models, which are invisible to purely classical criteria.

\clearpage
\section{Experimental details and further results}
\label{sec:experimental}
\subsection{Compute and reproducibility}
We conducted all experiments using a single Nvidia A100 GPU and provided sample code for reproducibility. Our implementation builds on open-source code from Hugging Face (diffusers library) and publicly available code from \cite{kynkäänniemi2024applyingguidancelimitedinterval,li2023diffusionmodelsecretlyzeroshot,Peebles2022DiT}. Our method is plug-and-play, requiring simple hyperparameter adjustments for these techniques without any major code modifications.
For zero-shot style transfer (\Cref{sec:zero-shot-style}), we used the Img2Img transfer pipeline in diffusers (\cite{meng2021sdedit,rombach2022highresolutionimagesynthesislatent}), fixing the VP/DDPM schedule and adjusting the strength parameter.

Baseline experiments using grid search were computationally intensive, typically requiring 24–48 hours of GPU time.
Fluctuation computations (\Cref{sec:methodology}) were primarily constrained by covariance matrix calculations and eigendecompositions. These can be efficiently optimised using multi-processing, though we used a single-process approach in this work. Our method is directly compatible with any standard implementation of the baselines. Preliminary small-scale experiments verified our theoretical framework but were excluded from the final paper.

\subsection{Convergence of data}
\label{sec:conv}
We present images visualizing the normality tests as stated in \Cref{sec:warmup} 
\begin{figure*}[htbp]
\centering
    \begin{minipage}{0.7\textwidth}
        \centering
        \includegraphics[width=\textwidth]{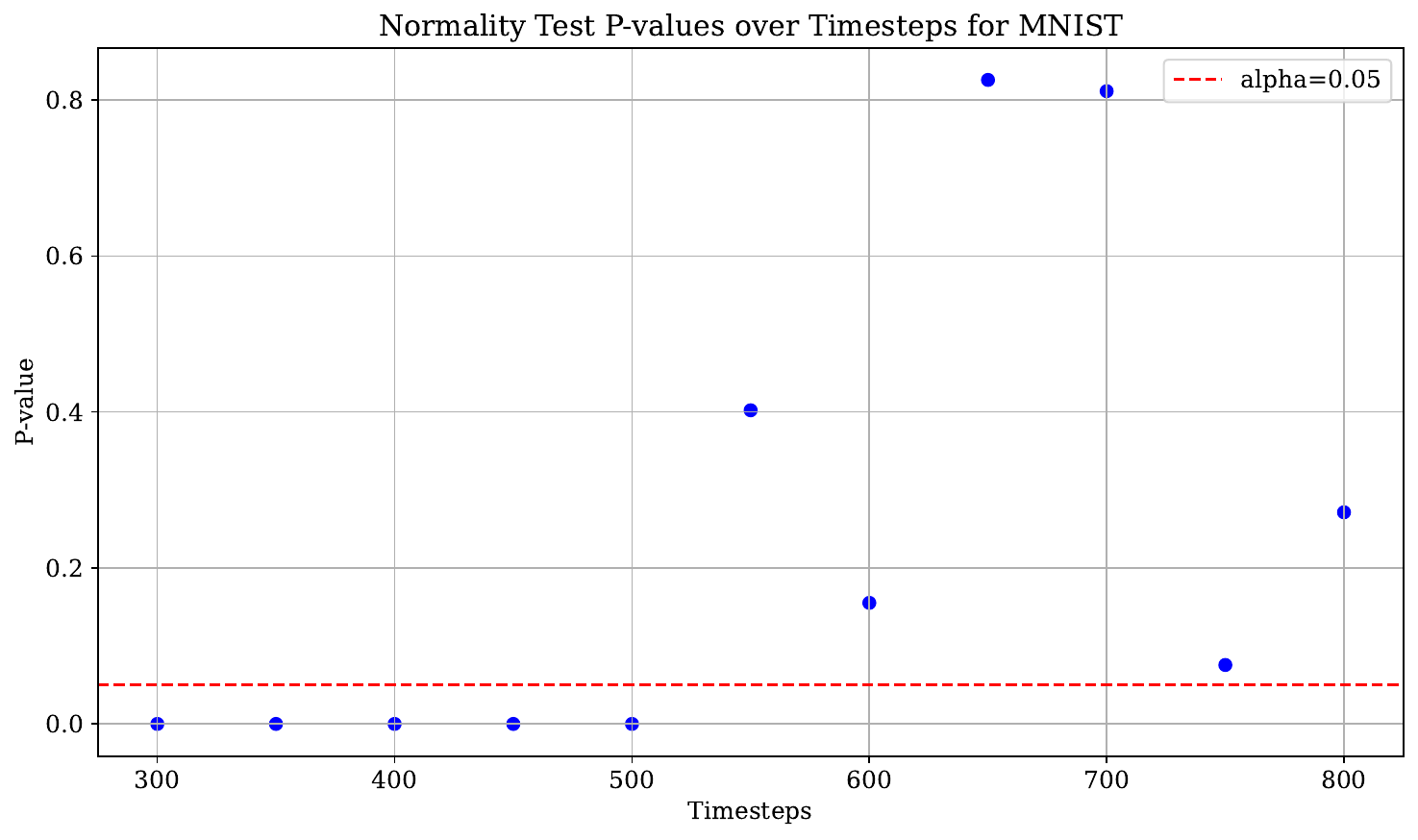}
        \subcaption{Normality test p-values over time for MNIST}\label{fig:mnist-norm}
    \end{minipage}

    \vskip 0.25in 

    \begin{minipage}{0.7\textwidth}
        \centering
        \includegraphics[width=\textwidth]{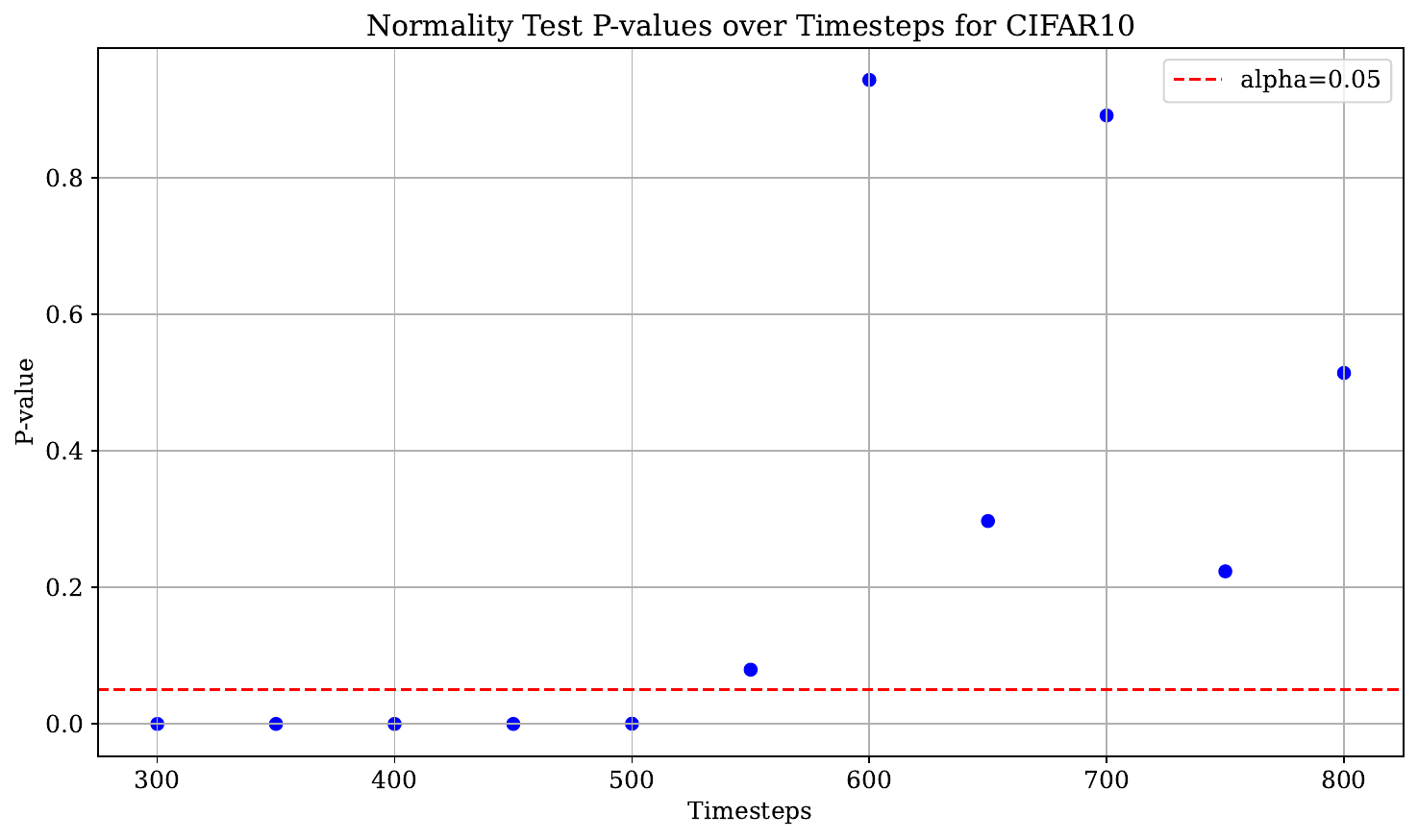}
        \subcaption{Normality test p-values over time for CIFAR10}\label{fig:cifar10-norm}
    \end{minipage}



\end{figure*}

\clearpage
\begin{figure*}[htbp]
\centering
\begin{minipage}{0.7\textwidth}
        \centering
        \includegraphics[width=\textwidth]{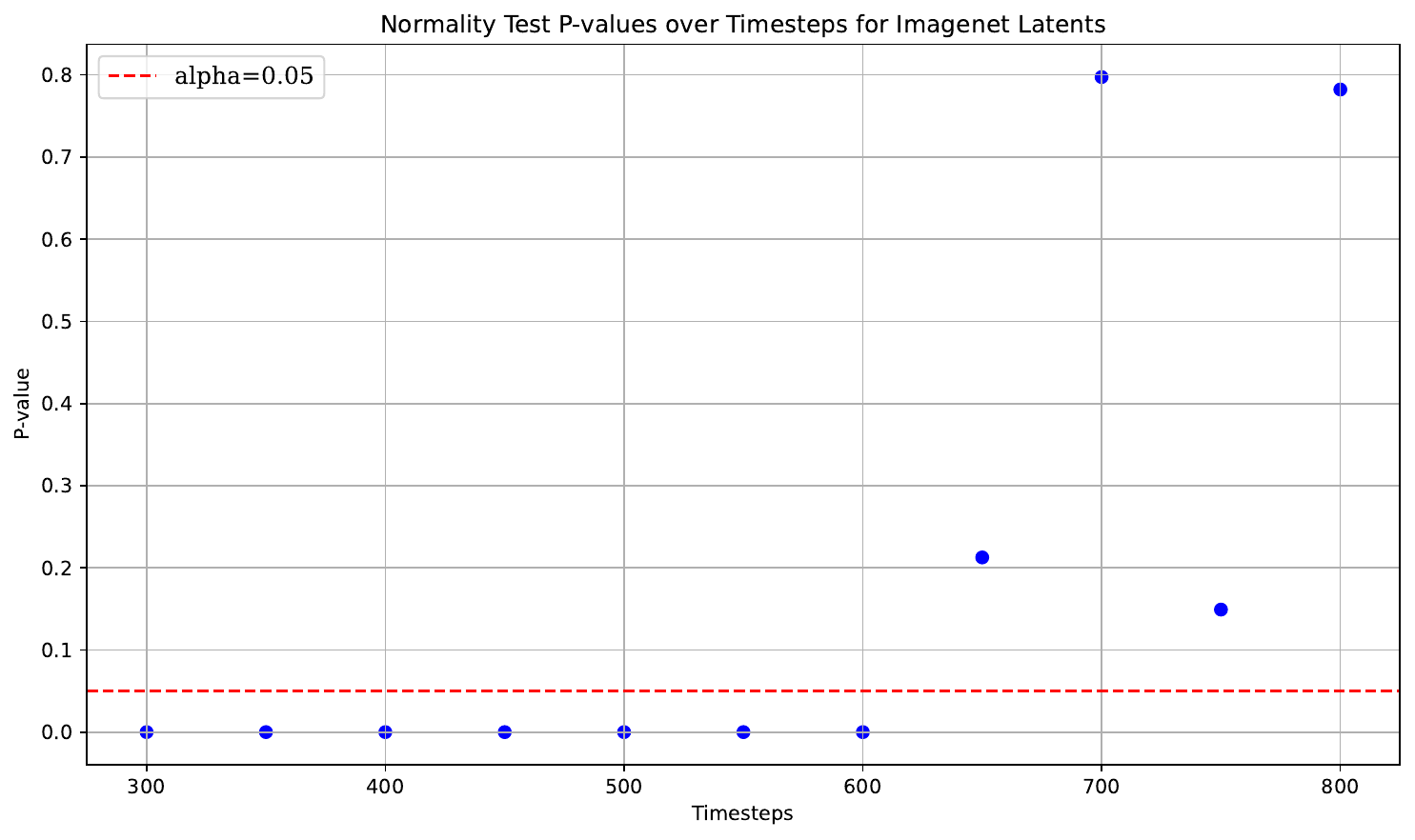}
        \subcaption{Normality test p-values over time for Imagenet}\label{figimgnet-norm}
    \end{minipage}
\caption{Normality test p-values for the DDPM schedule}\label{fig:ddpm-norm}
\vskip 0.1in

\end{figure*}
\subsection{Class-conditional generation}
\label{sec:class-cond-exp}
We compare two guidance schedules:

\begin{enumerate}
\item a \emph{grid-search baseline} that follows
      Interval Guidance (IG)~\citep{kynkäänniemi2024applyingguidancelimitedinterval}
      with a \textit{single} dataset-level interval found by brute force\footnote{In \cite{kynkäänniemi2024applyingguidancelimitedinterval}, it is argued that a class-level or even a sample-level search is preferable. However, both of these settings require at least $10^{3}-10^{6} \times$ the baseline compute, making them infeasible for us. Our primary objective is to demonstrate that leveraging finer hierarchical levels can compensate for compute limitations. Specifically,  we reason that for class-level brute-force search, it is more practical to consider the transitions of a fine-grained hierarchy derived from the representations of a semi/self-supervised learning model \cite{chen2020simple, grill2020bootstrap, he2021masked}. A theoretical backing for the same is found in \Cref{lem:geom-time}. We leave such an extension to future work. We note that asymptotically, our method converges to the same output as a per-sample-level grid search, as ultimately each data sample can be treated as a distinct singleton event $\Omega_{k,0}$.};
\item our \emph{merger-aware schedule}, in which each class $k$
      receives its own window  
      $\bigl(t_{\mathrm{start},k},\,t_{\mathrm{end},k}\bigr)$
      derived from fluctuation theory
      (\Cref{sec:warmup,sec:class-cond}).
\end{enumerate}

\paragraph{Interval guidance baseline.}
Let $w\!>\!0$ be the classifier-free guidance  \cite{ho2022classifierfreediffusionguidance} (CFG) weight,  
and let  
$T$ be the full diffusion horizon.  
During reverse sampling we switch
CFG on only for $t\in(t_{\mathrm{end},c},\,t_{\mathrm{start},c})$:

\begin{algorithm}[H]
\caption{Interval Guidance (class $c$)}
\label{alg:opt-cfg}
\begin{algorithmic}[1]\small
\Require latent $x_{T}\!\sim\!\mathcal{N}(0,I)$, CFG weight $w$
\For{$t=T-1,\dots,0$}
   \If{$t_{\mathrm{end},c}<t<t_{\mathrm{start},c}$}
      \State $x_{t}\!\gets\!\operatorname{CFG}_{w}(x_{t+1},c)$
   \Else
      \State $x_{t}\!\gets\!\operatorname{CFG}_{0}(x_{t+1},c)$
   \EndIf
\EndFor
\State\Return $x_{0}$
\end{algorithmic}
\end{algorithm}

\noindent
\textbf{Hyper-parameters and Grid-search baseline.}
Following \citet{Peebles2022DiT} we set $w=1.5$ for DiT-XL/2.
Stable Diffusion requires stronger guidance; we use a fixed
$w\in[3.5,4.5]$ per dataset.
For every dataset we sweep
\[
t_{\mathrm{end},c}\in\{0.1T,0.2T,\dots,0.8T\},
\quad
t_{\mathrm{start},c}\in\{0.2T,\dots,T\},
\]
under the constraint $t_{\mathrm{start},c}>t_{\mathrm{end},c}$,
yielding $44$ admissible pairs.  
On ImageNet the best pair is $(0.8T,0.2T)$
(lowest FID);  
MNIST and CIFAR-10 select $(0.6T,0.1T)$ and $(0.7T,0.1T)$, respectively.
For Stable Diffusion we replace FID by CLIP similarity and obtain
$(0.8T,0.1T)$ for both ImageNet and Oxford-IIIT Pet.
One exhaustive sweep on DiT-XL/2 costs
$4100\;\text{GFLOPs}\times 50\,000$ samples\,$\times 44$ configs
$\approx9.0$\,PFLOPs; five repeats per pair multiply the cost five-fold. Results for Imagenet, MNIST and CIFAR are in \Cref{tab:class-cond} in the main paper while that for Imagenet and Oxford-IIITPets using Stable Diffusion is in \Cref{tab:cond-sd}. Note that for the case of Imagenet, identical intervals are used for both settings as the empirical forward process trajectory is independent of the model choice.  

\paragraph{Merger-aware schedule (ours).}
For each class $k$, 
$
t_{\mathrm{start},k}=i^{\star},
~
t_{\mathrm{end},k}=t_{\text{merge},k},
$
where $i^{\star}$ is the global convergence index
(\Cref{sec:warmup}) and $t_{\text{merge},k}$ is the first
$t$ at which
$M^{(2)}_{\rho}\!\bigl(\Omega_{k,t},\Omega_{\ell,t}\bigr)=1$
for some $\ell\neq k$
(\Cref{sec:class-cond}).  
No search is required; windows differ automatically across classes.
\vskip 0.1in

\begin{table}[H]
  \centering
  \small
  \resizebox{\textwidth}{!}{%
    \begin{tabular}{|l|c|c|c|c|c|}
      \hline
      \rowcolor[gray]{0.8}
      \textbf{Model/Dataset} & \textbf{CLIP Similarity} ($\uparrow$) & \textbf{Precision} ($\uparrow$) &
      \textbf{Recall} ($\uparrow$) & \textbf{Density} ($\uparrow$) & \textbf{Coverage} ($\uparrow$) \\ \hline
      SD (Imagenet, IG baseline) & $0.26 \pm 0.03$ & $0.75 \pm 0.04$ & $0.18 \pm 0.02$ & $0.80 \pm 0.05$ & $0.30 \pm 0.03$ \\ \hline
      SD (Imagenet, IG Ours) & $\mathbf{0.31 \pm 0.02}$ & $\mathbf{0.78 \pm 0.02}$ & $\mathbf{0.23 \pm 0.01}$ & $\mathbf{0.88 \pm 0.03}$ & $\mathbf{0.34 \pm 0.02}$ \\ \hline
      SD (OxfordIIITPet, IG baseline) & $0.28 \pm 0.02$ & $0.79 \pm 0.03$ & $0.21 \pm 0.03$ & $0.84 \pm 0.06$ & $0.33 \pm 0.05$ \\ \hline
      SD (OxfordIIITPet, IG Ours) & $\mathbf{0.34 \pm 0.03}$ & $\mathbf{0.81 \pm 0.01}$ & $\mathbf{0.26 \pm 0.04}$ & $\mathbf{0.89 \pm 0.01}$ & $\mathbf{0.36 \pm 0.02}$ \\ \hline
    \end{tabular}
  }
  \caption{Class conditional generation using Stable Diffusion}
  \label{tab:cond-sd}
\end{table}

\begin{figure*}[htbp]
  \captionsetup[subfigure]{justification=centering,labelformat=simple}
  \centering
  \begin{subfigure}[t]{0.48\textwidth}
    \includegraphics[width=\linewidth]{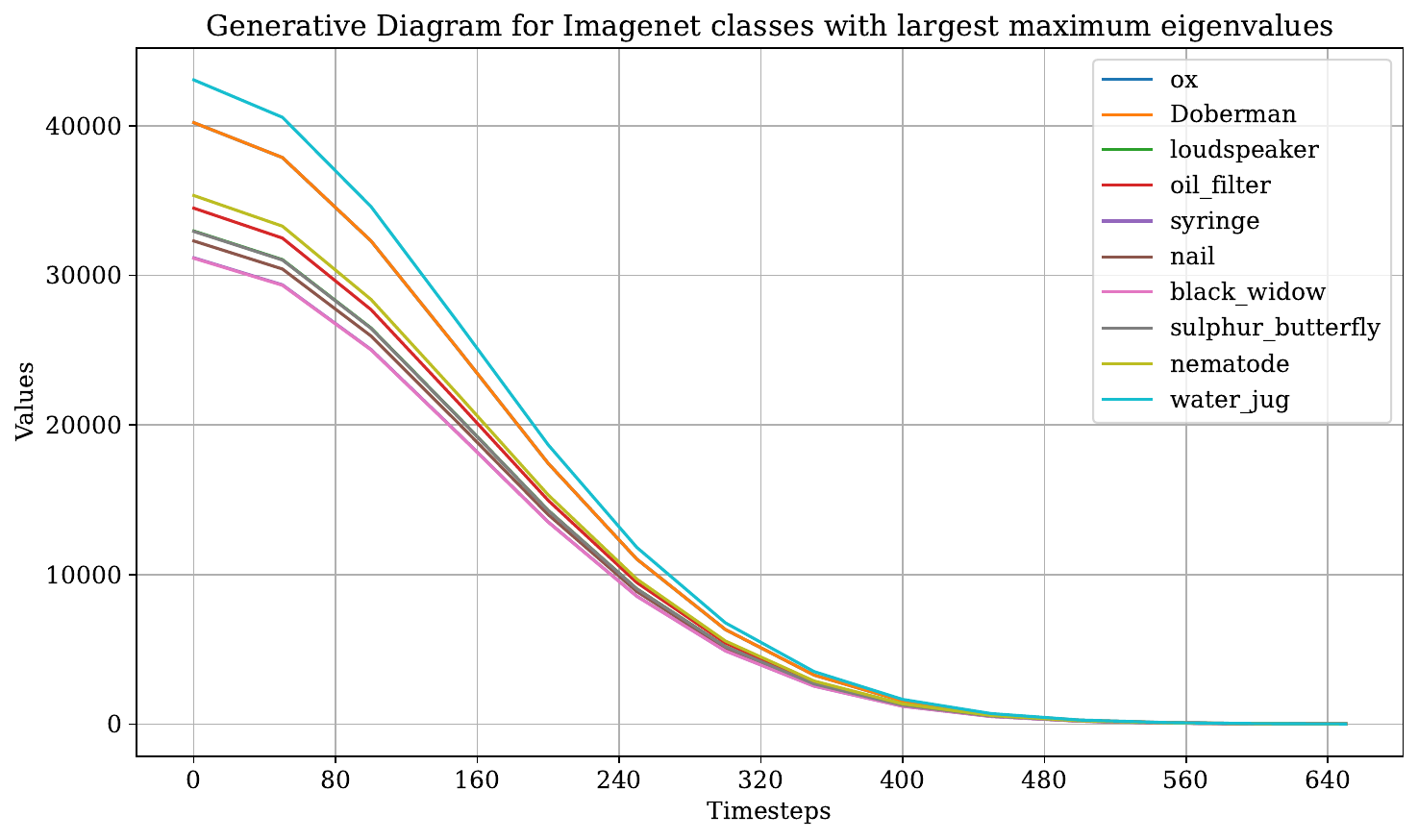}
    \caption{ImageNet — top 10 classes}\label{fig:imagenet-top10}
  \end{subfigure}\hfill
  \begin{subfigure}[t]{0.48\textwidth}
    \includegraphics[width=\linewidth]{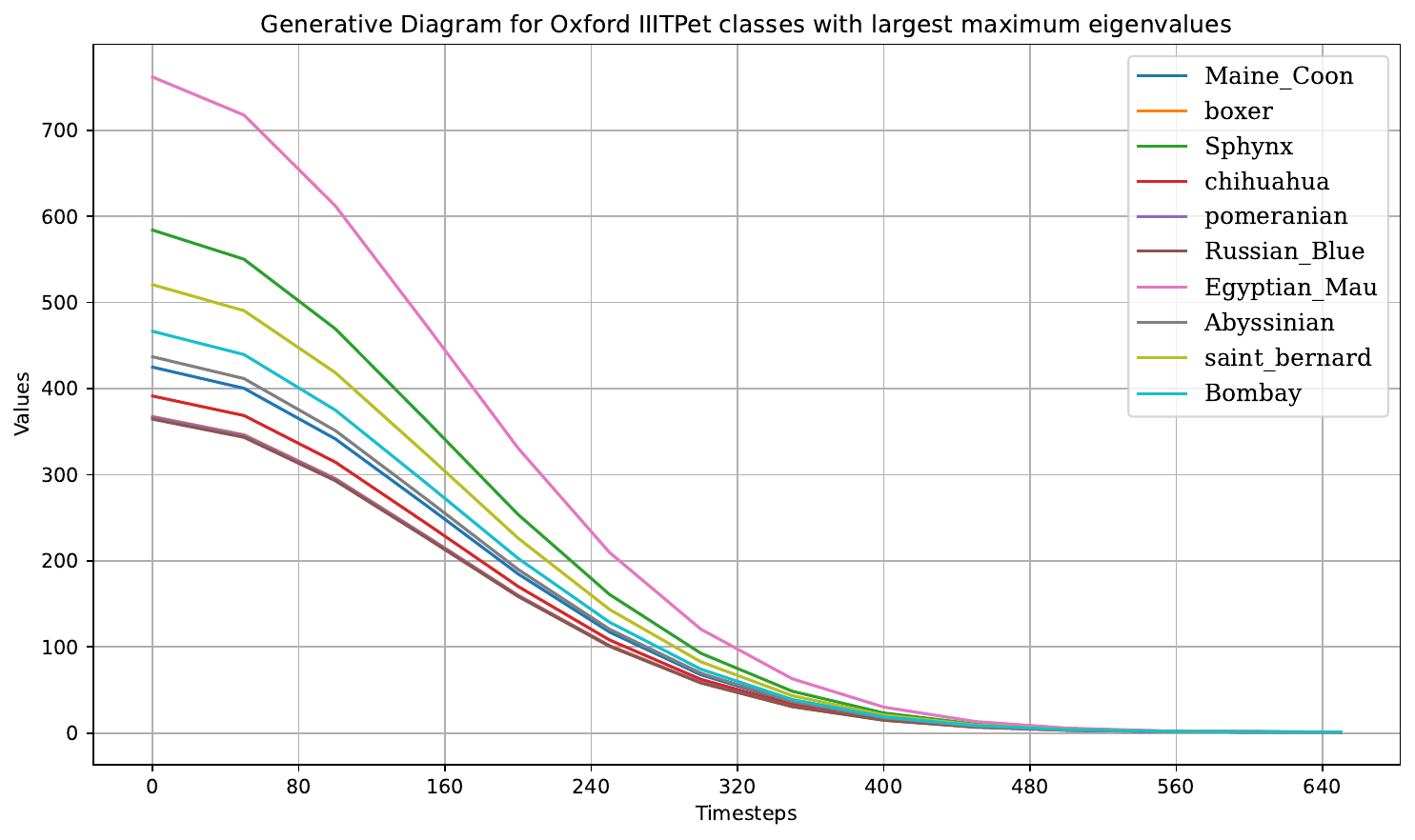}
    \caption{Oxford-IIIT-Pet — top 10 classes}\label{fig:oxpet-top10}
  \end{subfigure}
  \caption{Merger transitions are upper-bounded by those of the ten classes with the largest principal-eigenvalue magnitudes in their covariance matrices.}
  \label{fig:top10-bounds-img}
\end{figure*}

\begin{figure*}[htbp]
  \captionsetup[subfigure]{justification=centering,labelformat=simple}
  \centering

  \begin{minipage}[t]{0.48\textwidth}
    \centering
    \begin{subfigure}[t]{\linewidth}
      \includegraphics[width=\linewidth]{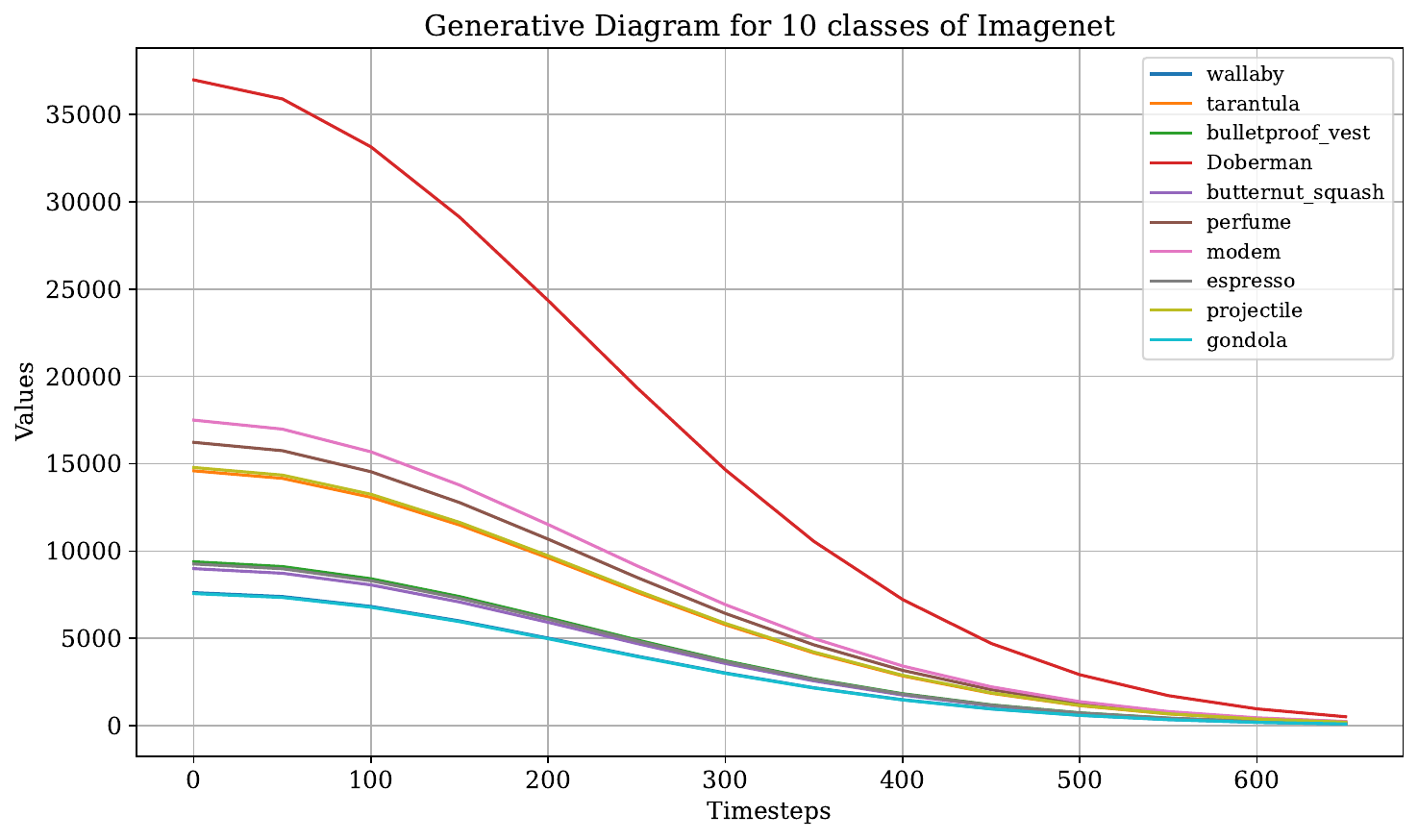}
      \caption{}\label{fig:g1}
    \end{subfigure}\par\vspace{0.6em}

    \begin{subfigure}[t]{\linewidth}
      \includegraphics[width=\linewidth]{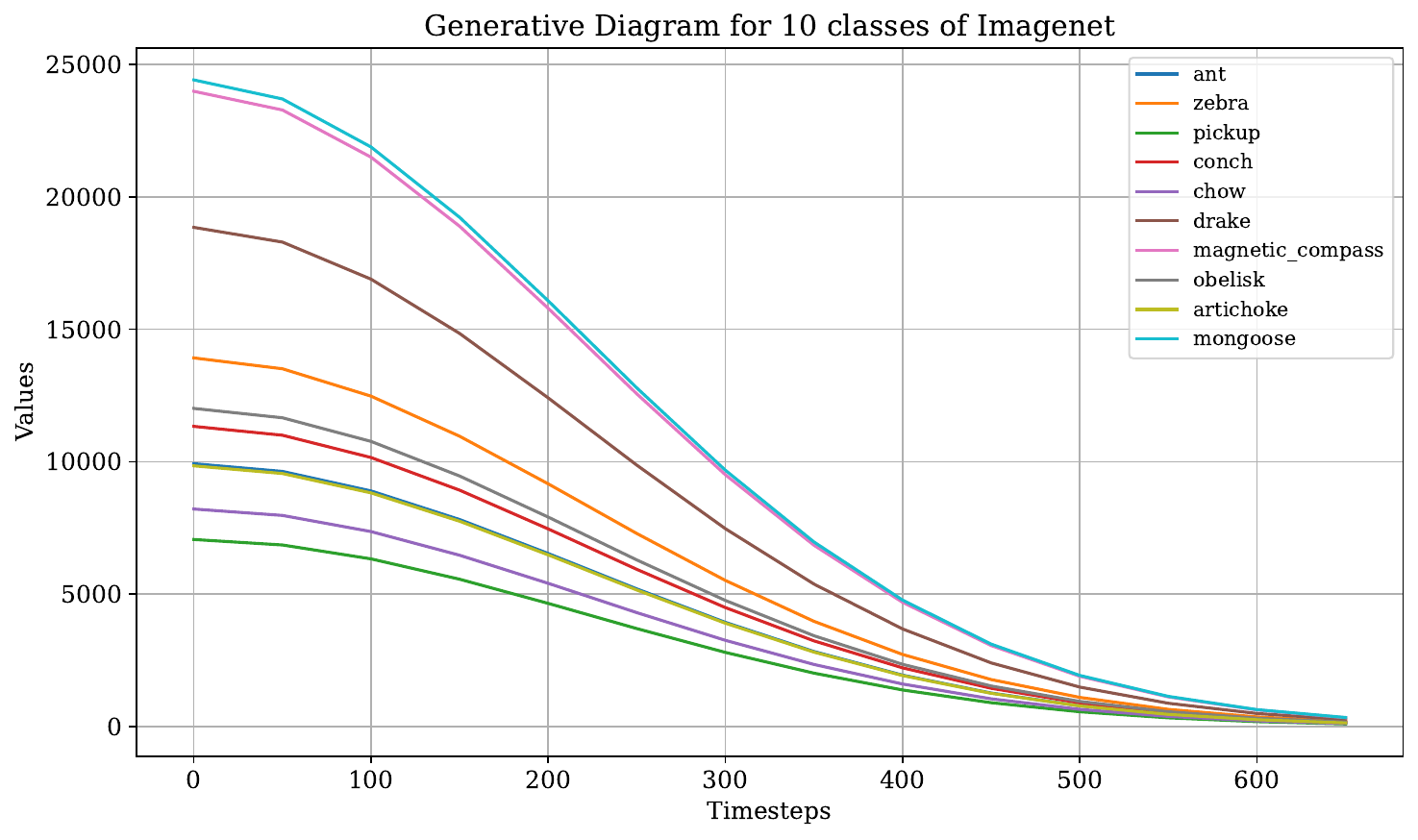}
      \caption{}\label{fig:g2}
    \end{subfigure}\par\vspace{0.6em}

    \begin{subfigure}[t]{\linewidth}
      \includegraphics[width=\linewidth]{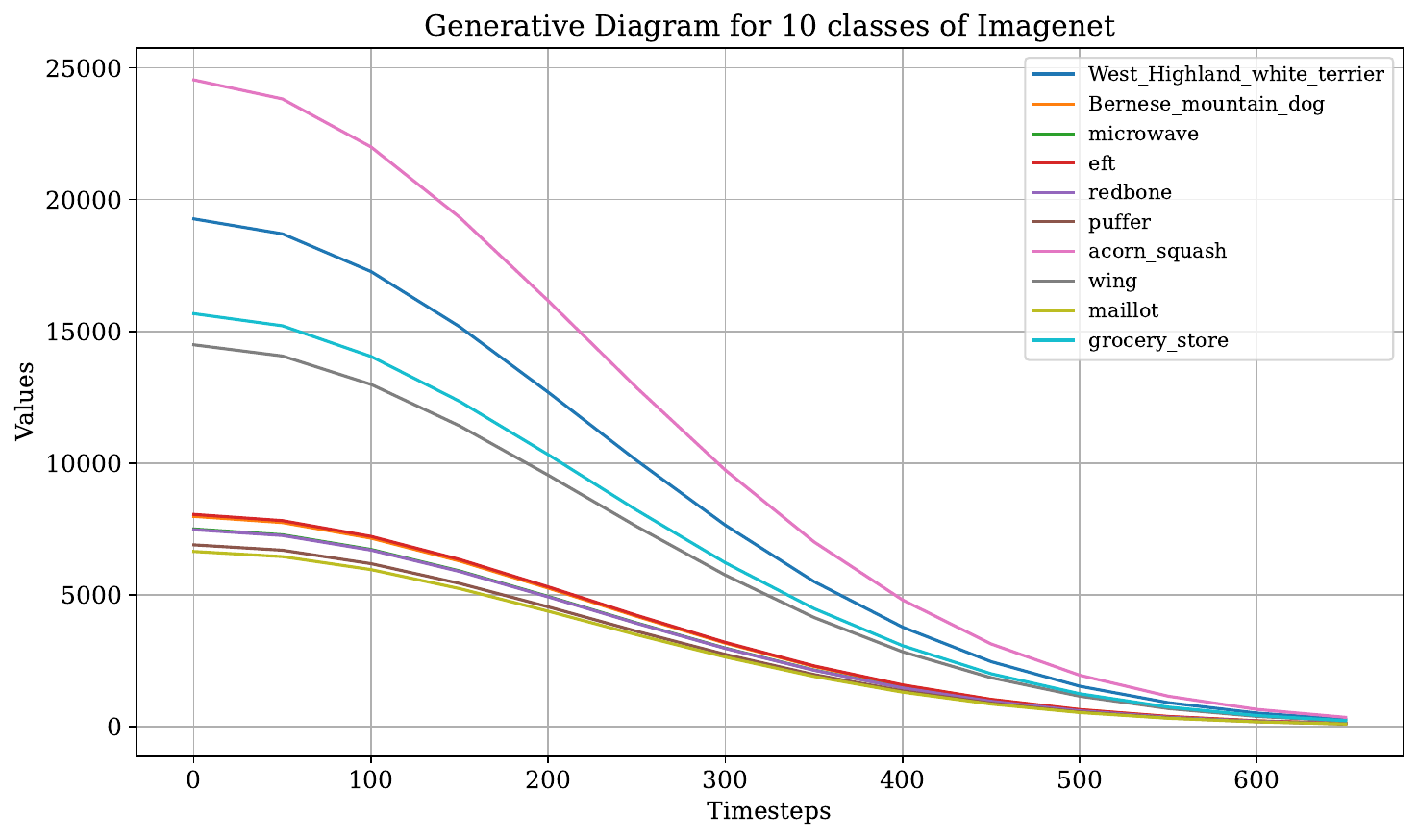}
      \caption{}\label{fig:g3}
    \end{subfigure}\par\vspace{0.6em}

    \begin{subfigure}[t]{\linewidth}
      \includegraphics[width=\linewidth]{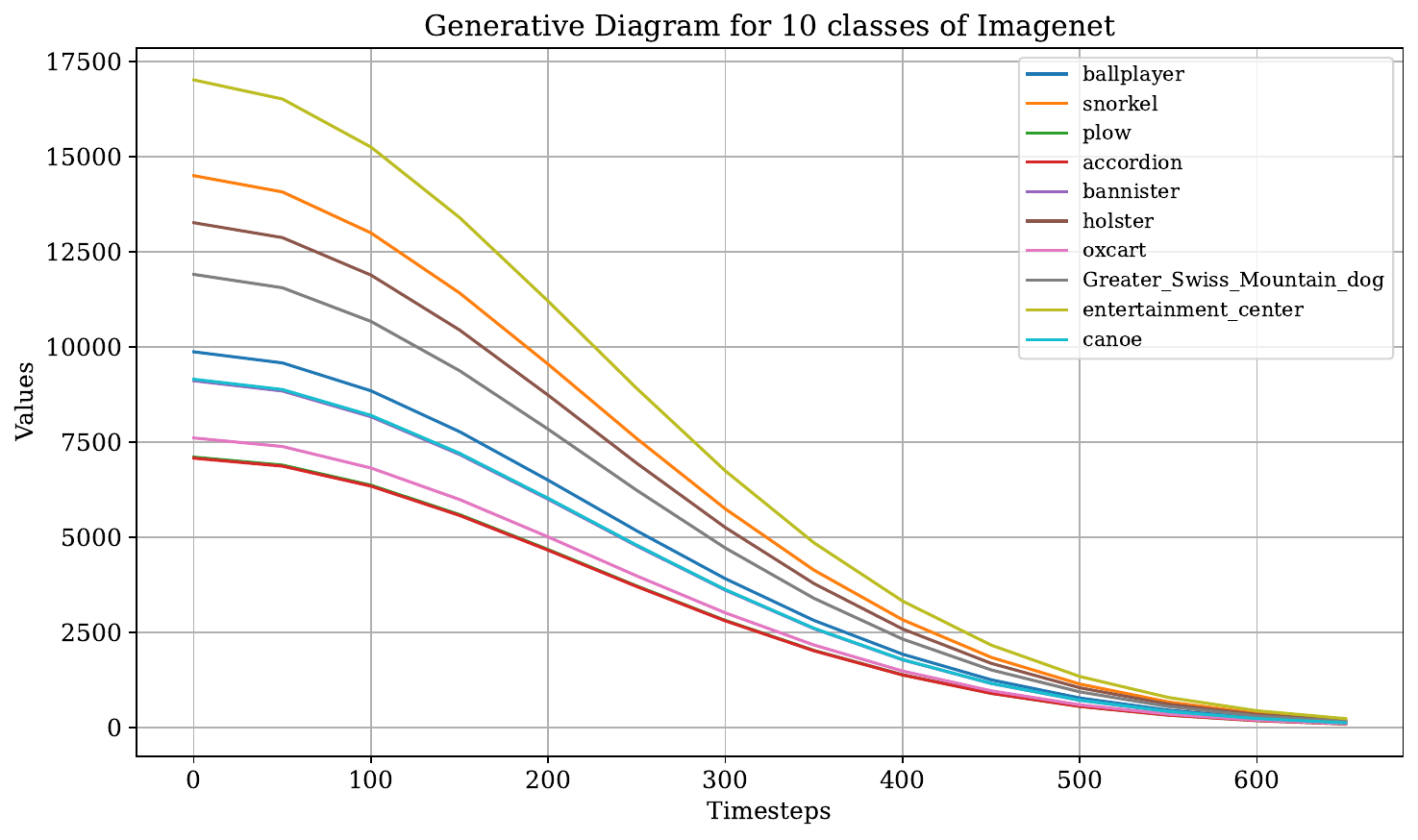}
      \caption{}\label{fig:g4}
    \end{subfigure}\par\vspace{0.6em}

    \begin{subfigure}[t]{\linewidth}
      \includegraphics[width=\linewidth]{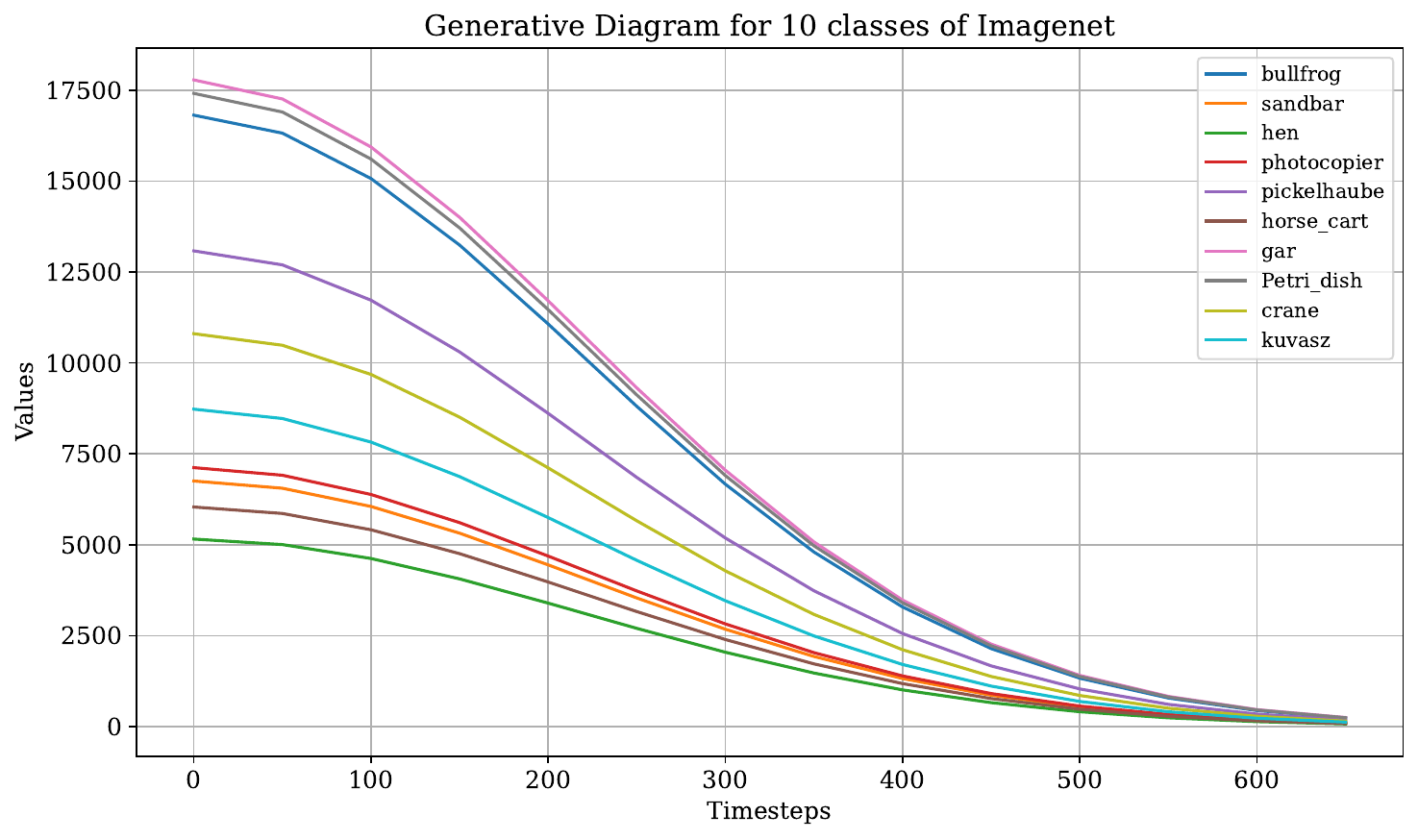}
      \caption{}\label{fig:g5}
    \end{subfigure}
  \end{minipage}
  \hfill
  \begin{minipage}[t]{0.48\textwidth}
    \centering
    \begin{subfigure}[t]{\linewidth}
      \includegraphics[width=\linewidth]{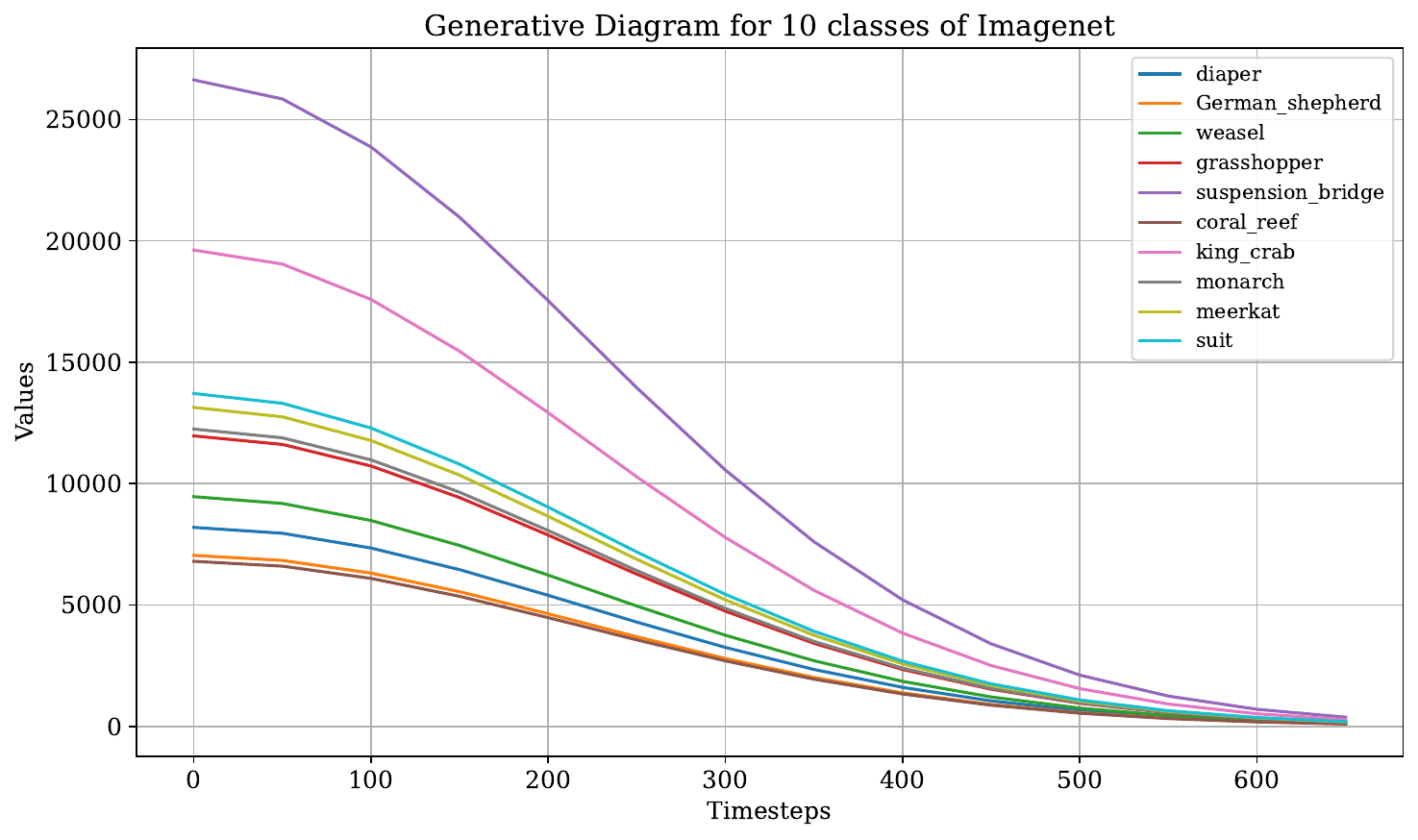}
      \caption{}\label{fig:g6}
    \end{subfigure}\par\vspace{0.6em}

    \begin{subfigure}[t]{\linewidth}
      \includegraphics[width=\linewidth]{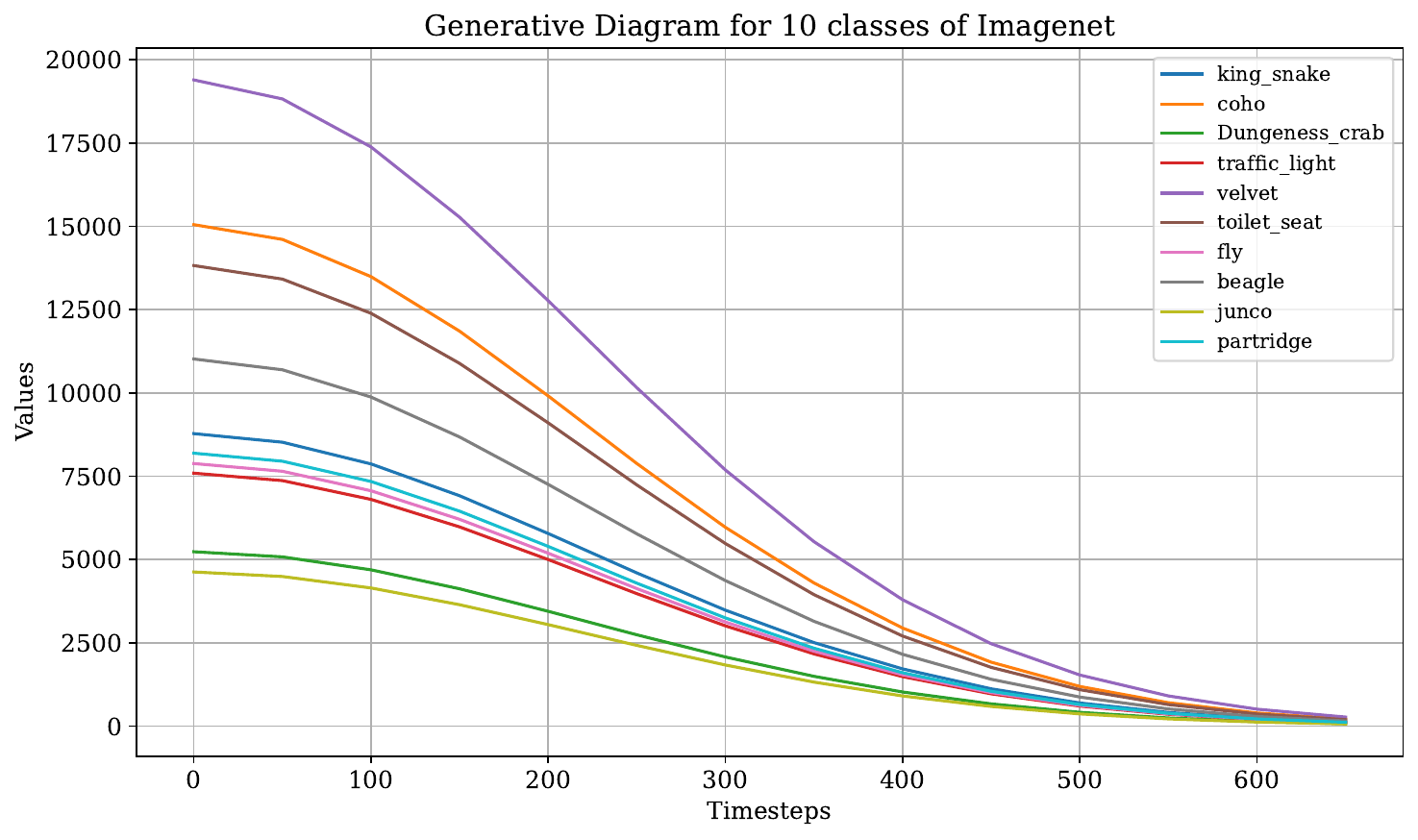}
      \caption{}\label{fig:g7}
    \end{subfigure}\par\vspace{0.6em}

    \begin{subfigure}[t]{\linewidth}
      \includegraphics[width=\linewidth]{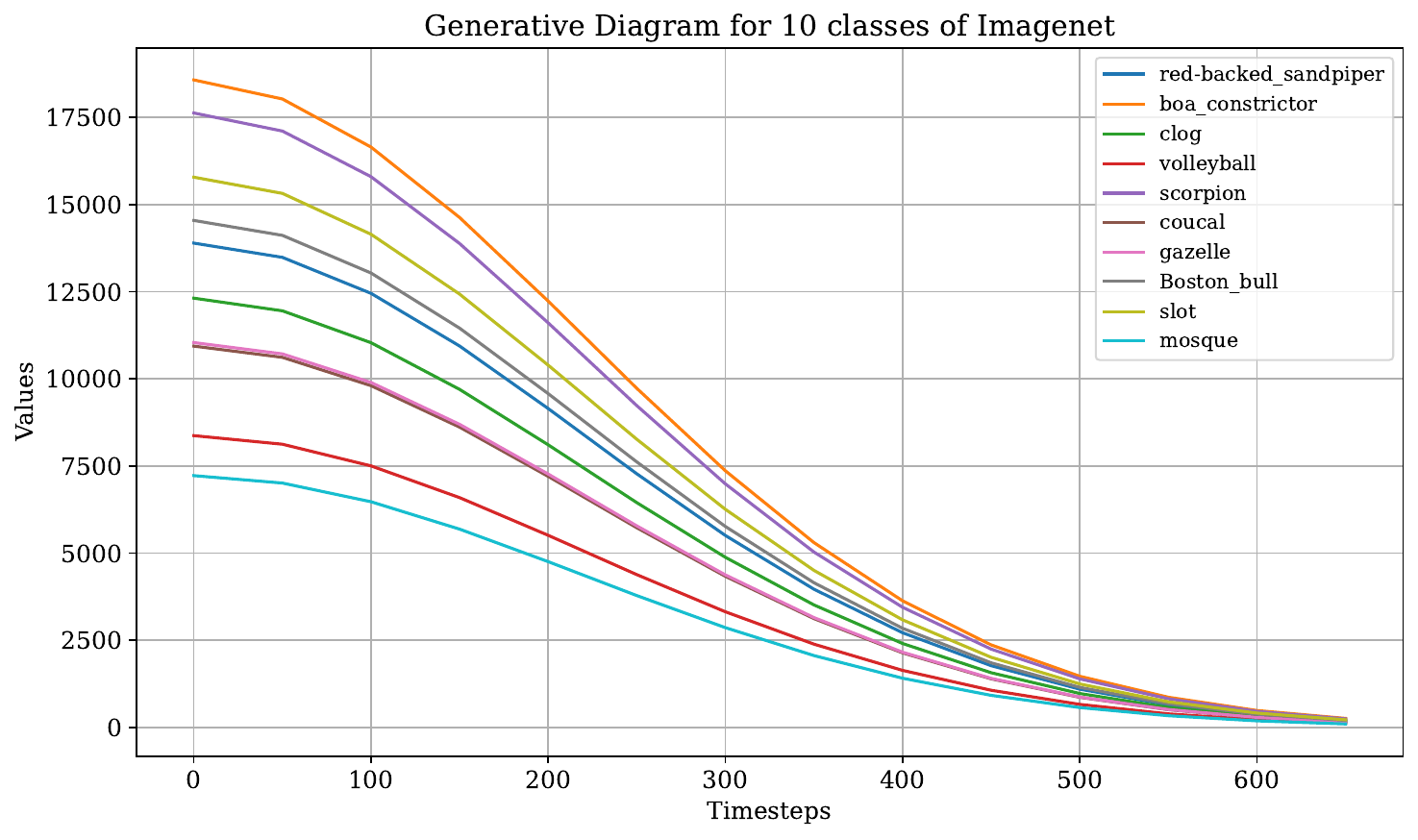}
      \caption{}\label{fig:g8}
    \end{subfigure}\par\vspace{0.6em}

    \begin{subfigure}[t]{\linewidth}
      \includegraphics[width=\linewidth]{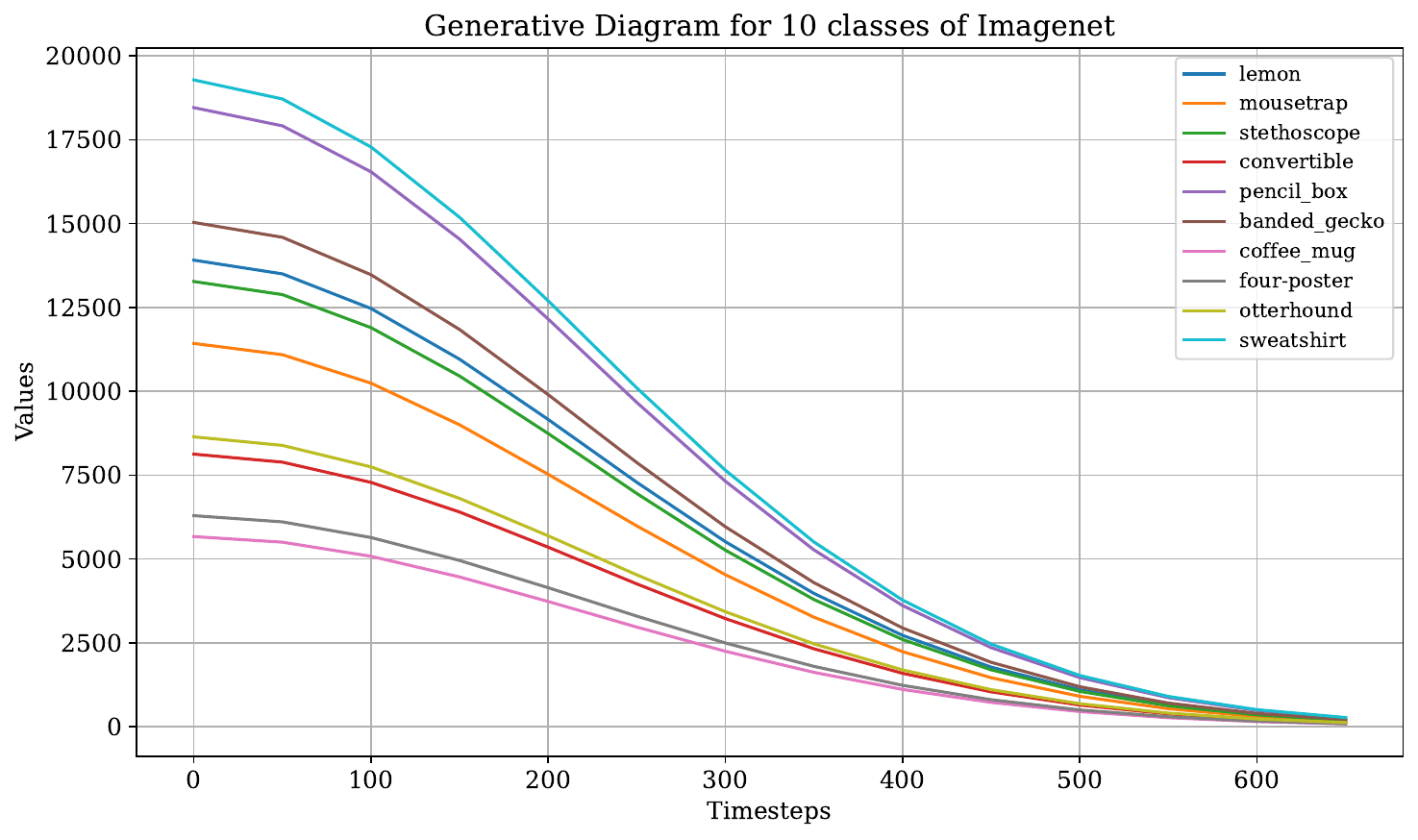}
      \caption{}\label{fig:g9}
    \end{subfigure}\par\vspace{0.6em}

    \begin{subfigure}[t]{\linewidth}
      \includegraphics[width=\linewidth]{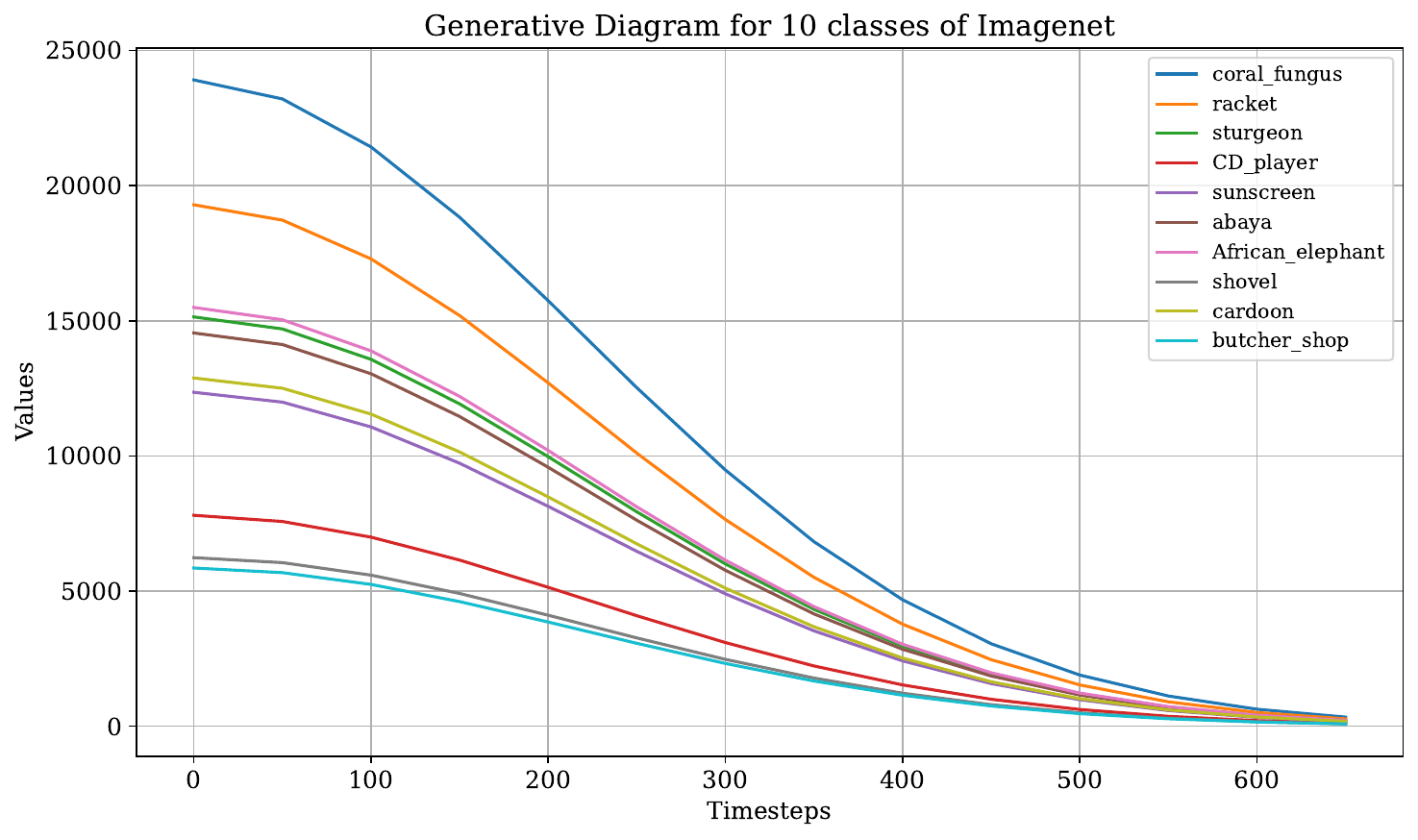}
      \caption{}\label{fig:g10}
    \end{subfigure}
  \end{minipage}
  \caption{Merger-transition subplots for ImageNet (ten classes shown in two parallel columns).}
  \label{fig:imagenet_parallel}
\end{figure*}

\paragraph{Generative diagrams.}
\Cref{fig:gen-diags} plots the leading eigenvalues
$\lambda_{\max}\bigl(\Sigma_{k,t}\bigr)$
of the class covariances $\Sigma_{k,t}$
for ImageNet, CIFAR-10, and MNIST,
highlighting merger points (proofs in \Cref{sec:cka}).  
Additional zoom-ins for ImageNet appear in
\Cref{fig:imagenet_parallel}.
For long-tail datasets used in \Cref{sec:rare-class}, analogous diagrams are given in
\Cref{fig:inat-gen,fig:cub-gen}. We also plot the subplots for the $10$ classes for Imagenet and OxfordIIITPet, having the greatest magnitude of principal eigenvalues in \Cref{fig:top10-bounds-img}. 
\vspace{0.1in}

\begin{figure*}[htbp]
  \captionsetup[subfigure]{justification=centering,labelformat=simple}
  \centering
  \begin{minipage}[t]{0.48\textwidth}
    \centering
    \begin{subfigure}[t]{\linewidth}
      \includegraphics[width=\linewidth]{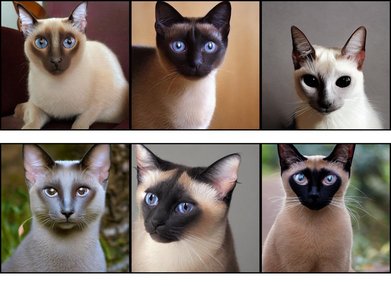}
      \caption{SD output for prompt “Siamese Cat”}\label{fig:sd-siamese}
    \end{subfigure}\par\vspace{0.8em}
    \begin{subfigure}[t]{\linewidth}
      \includegraphics[width=\linewidth]{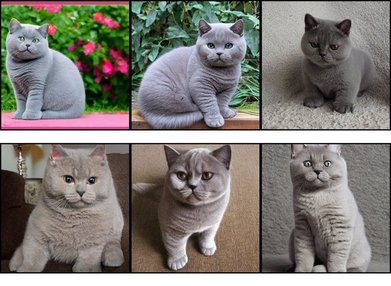}
      \caption{SD output for prompt “British Shorthair”}\label{fig:sd-british}
    \end{subfigure}
  \end{minipage}
  \hfill
  \begin{minipage}[t]{0.48\textwidth}
    \centering
    \begin{subfigure}[t]{\linewidth}
      \includegraphics[width=\linewidth]{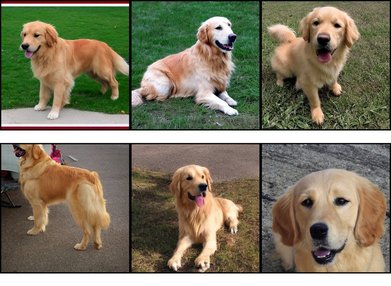}
      \caption{SD output for prompt “Golden Retriever”}\label{fig:sd-golden}
    \end{subfigure}\par\vspace{0.8em}
    \begin{subfigure}[t]{\linewidth}
      \includegraphics[width=\linewidth]{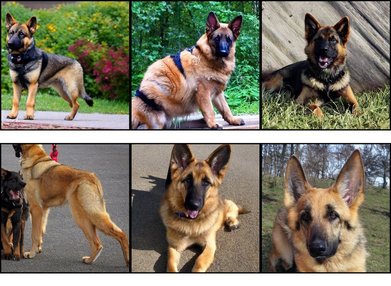}
      \caption{SD output for prompt “German Shepherd”}\label{fig:sd-shepherd}
    \end{subfigure}
  \end{minipage}

  \caption{Stable-Diffusion samples for four Oxford-IIIT-Pet classes, generated with naïve interval guidance (top of each column) versus our method (bottom).}
  \label{fig:oxiit-vis}
\end{figure*}

\begin{figure*}[htbp]
\centering
\captionsetup{width=\textwidth}

\begin{minipage}{0.48\textwidth}
\subfloat[Naive interval guidance for Imagenet]{
  \includegraphics[width=\linewidth]{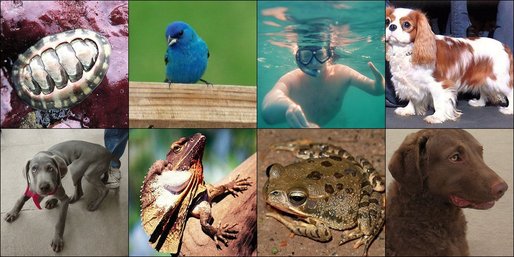}
}
\end{minipage}
\hspace{0.02\textwidth}
\begin{minipage}{0.48\textwidth}
\subfloat[Our optimised interval guidance for Imagenet]{
  \includegraphics[width=\linewidth]{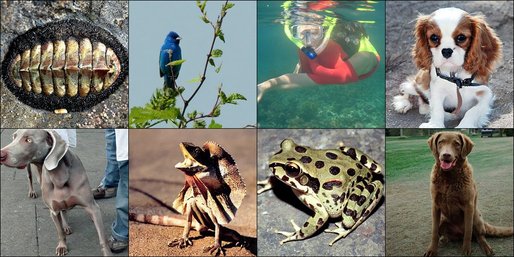}
}
\end{minipage}

\vspace{2mm}

\begin{minipage}{0.48\textwidth}
\subfloat[Naive interval guidance for Imagenet]{
  \includegraphics[width=\linewidth]{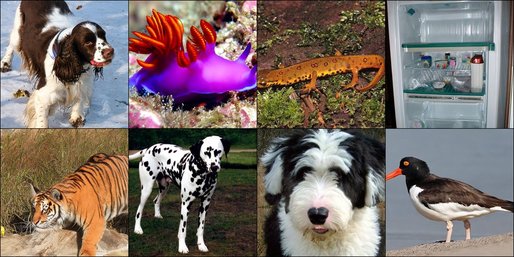}
}
\end{minipage}
\hspace{0.02\textwidth}
\begin{minipage}{0.48\textwidth}
\subfloat[Our optimised interval guidance for Imagenet]{
  \includegraphics[width=\linewidth]{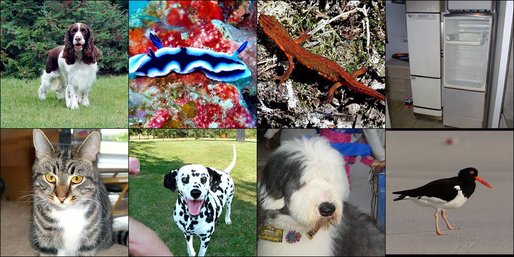}
}
\end{minipage}

\vspace{2mm}

\begin{minipage}{0.48\textwidth}
\subfloat[Naive interval guidance for Imagenet]{
  \includegraphics[width=\linewidth]{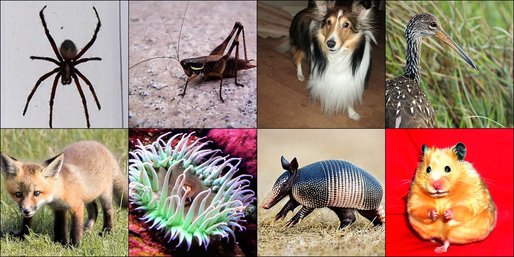}
}
\end{minipage}
\hspace{0.02\textwidth}
\begin{minipage}{0.48\textwidth}
\subfloat[Our optimised interval guidance for Imagenet]{
  \includegraphics[width=\linewidth]{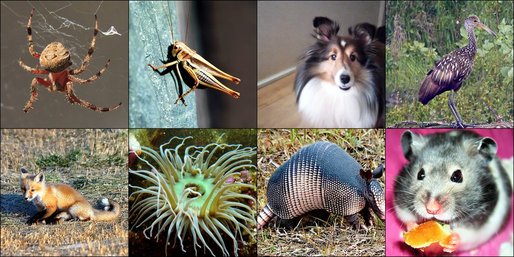}
}
\end{minipage}

\vspace{2mm}

\begin{minipage}{0.48\textwidth}
\subfloat[Naive interval guidance for Imagenet]{
  \includegraphics[width=\linewidth]{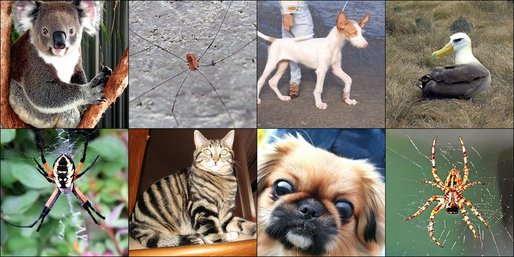}
}
\end{minipage}
\hspace{0.02\textwidth}
\begin{minipage}{0.48\textwidth}
\subfloat[Our optimised interval guidance for Imagenet]{
  \includegraphics[width=\linewidth]{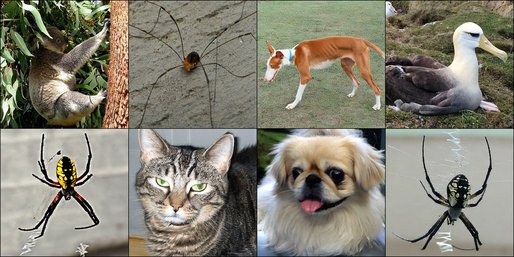}
}
\end{minipage}

\vspace{2mm}

\begin{minipage}{0.48\textwidth}
\subfloat[Naive interval guidance for Imagenet]{
  \includegraphics[width=\linewidth]{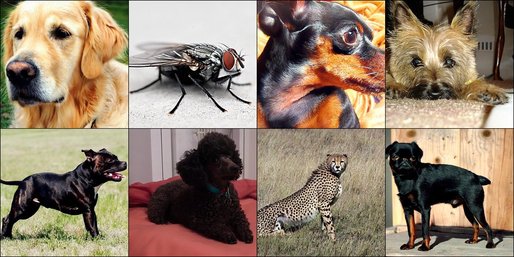}
}
\end{minipage}
\hspace{0.02\textwidth}
\begin{minipage}{0.48\textwidth}
\subfloat[Our optimised interval guidance for Imagenet]{
  \includegraphics[width=\linewidth]{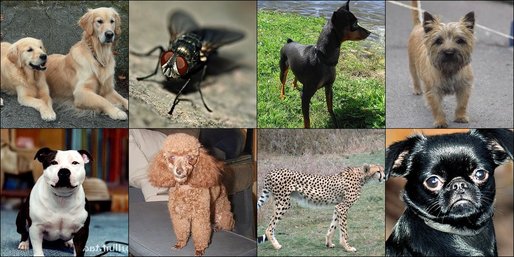}
}
\end{minipage}

\caption{Visual Comparison of guidance for the Imagenet dataset ~\citep{deng2009imagenet,imagenet_int8}. All samples were originally generated in $512\times512$ resolution.}
\label{fig:imgnt-all}
\end{figure*}

\Cref{fig:imgnt-all} compares ImageNet samples from our merger-aware IG with the grid-search baseline, using a guidance weight of $4.5$ for visual clarity;  
\Cref{fig:oxiit-vis} does the same for
Oxford-IIIT Pet under Stable Diffusion.  
Our method yields crisper details and fewer artefacts—despite
eliminating $\approx 9$ PFLOPs of search for Imagenet. Thus, class-wise guidance windows obtained from cross-fluctuation mergers can
match or exceed the quality of an
\emph{exhaustive} dataset-level search,  
while slashing computational cost by orders of magnitude.

\vspace{0.1in}
\begin{table}[H]
\centering\small
\begin{tabular}{|c|c|c|}
\hline
\rowcolor{gray!20}
$\varepsilon$ & Merge prob.\ $(t=0)$ & FID $(\downarrow)$ \\ \hline
200 & 0.027 & $3.06\!\pm\!0.19$ \\ \hline
100 & 0.013 & $\mathbf{2.86\!\pm\!0.15}$ \\ \hline
 80 & 0.010 & $2.92\!\pm\!0.17$ \\ \hline
 67 & 0.009 & $2.90\!\pm\!0.11$ \\ \hline
 20 & 0.002 & $2.88\!\pm\!0.14$ \\ \hline
\end{tabular}
\vspace{0.3em}
\caption{Effect of the MAE threshold $\varepsilon$ (ImageNet).}
\label{tab:merge-prob}
\end{table}

\begin{figure*}[htbp]
  \captionsetup[subfigure]{justification=centering,labelformat=simple}
  \centering
  \begin{minipage}[t]{0.48\textwidth}
    \centering
    \begin{subfigure}[t]{\linewidth}
      \includegraphics[width=\linewidth]{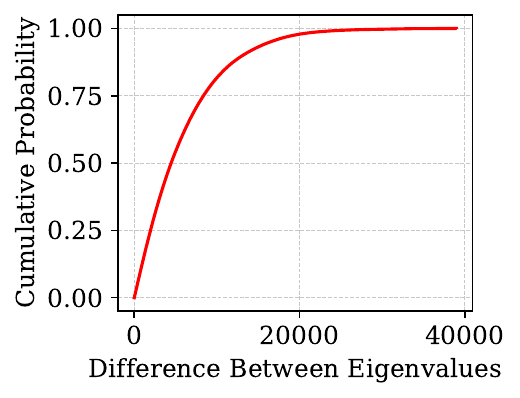}
      \caption{CDF of principal-eigenvalue differences, ImageNet}\label{fig:imgnet-cdf}
    \end{subfigure}\par\vspace{0.8em}
    \begin{subfigure}[t]{\linewidth}
      \includegraphics[width=\linewidth]{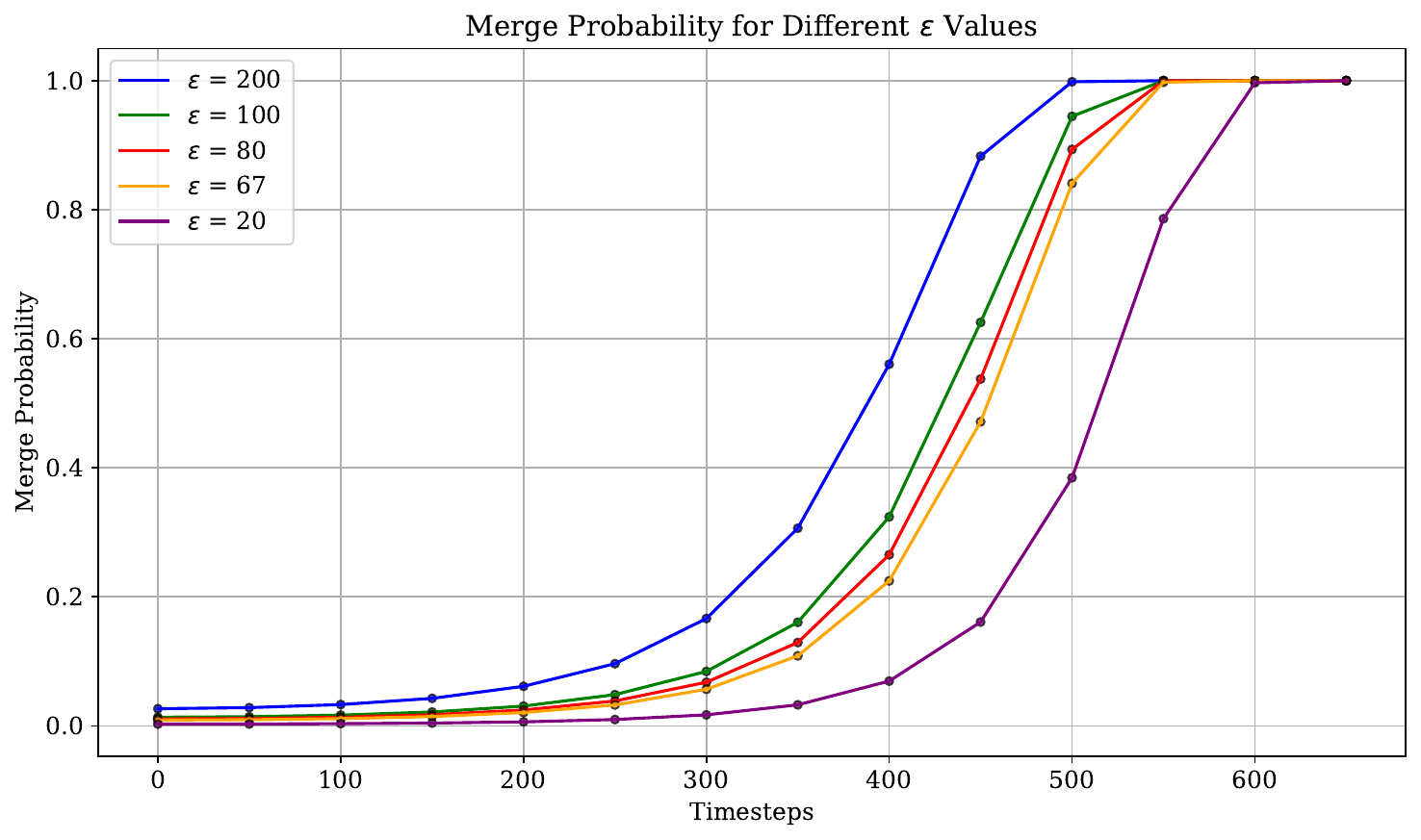}
      \caption{Merge-probability vs.~threshold, ImageNet}\label{fig:imgnet-mix}
    \end{subfigure}
  \end{minipage}
  \hfill
  \begin{minipage}[t]{0.48\textwidth}
    \centering
    \begin{subfigure}[t]{\linewidth}
      \includegraphics[width=\linewidth]{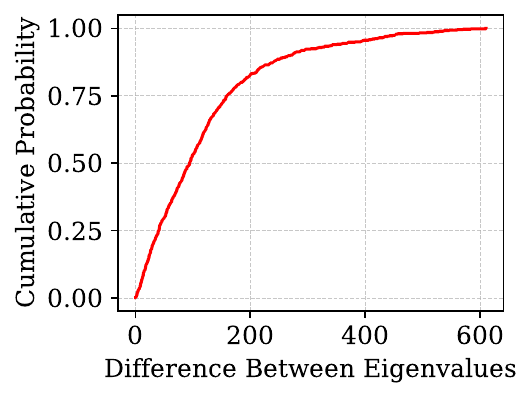}
      \caption{CDF of principal-eigenvalue differences, Oxford-IIIT-Pet}\label{fig:oxford-cdf}
    \end{subfigure}\par\vspace{0.8em}
    \begin{subfigure}[t]{\linewidth}
      \includegraphics[width=\linewidth]{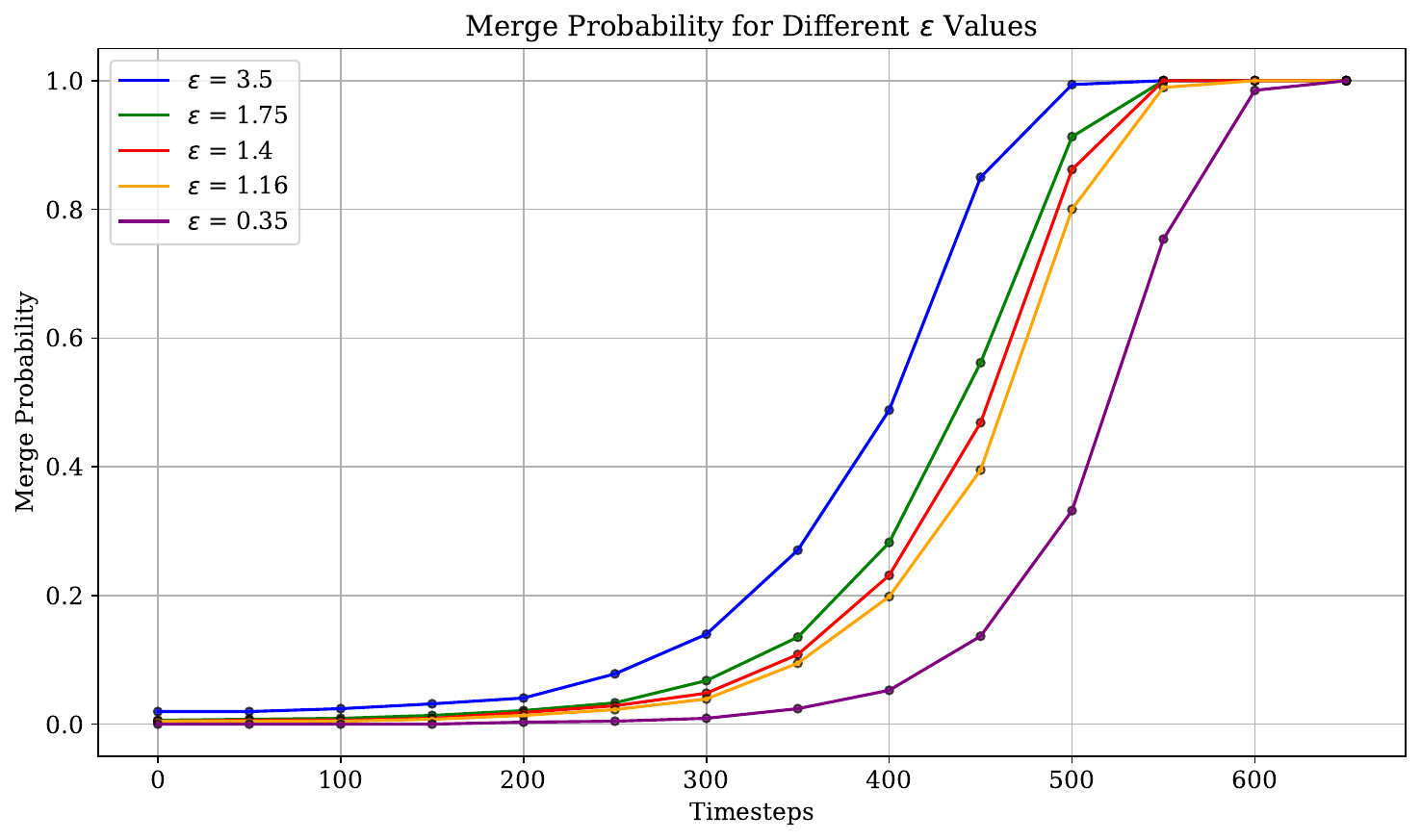}
      \caption{Merge-probability vs.~threshold, Oxford-IIIT-Pet}\label{fig:oxford-mix}
    \end{subfigure}
  \end{minipage}

  \caption{Eigenvalue-difference distributions and their associated merge-probability curves for ImageNet and Oxford-IIIT-Pet.}
  \label{fig:cdf-mix-comparison}
\end{figure*}

To understand the sensitivity of the parameter $\varepsilon$, we plot the distribution of the difference in eigenvalues (CDF) and evolution of merge probabilities for different choices of $\varepsilon$ (\Cref{fig:cdf-mix-comparison}). Our default choice $\varepsilon = 100$ has a merge probability $\approx 0.01$ at $t=0$. We show corresponding FIDs in \Cref{tab:merge-prob}.   

\vspace{0.4in}

\clearpage
\newpage

    

\newpage

\clearpage
\newpage

\begingroup
  \setlength{\abovecaptionskip}{4pt}
  \setlength{\belowcaptionskip}{4pt}
  \setlength{\textfloatsep}{6pt}
  \setlength{\floatsep}{6pt}
  \setlength{\intextsep}{6pt}

  \begin{figure*}[!p]
    \centering
    \begin{minipage}[t]{0.48\textwidth}
      \centering
      \begin{subfigure}[t]{\linewidth}
        \includegraphics[width=\linewidth]{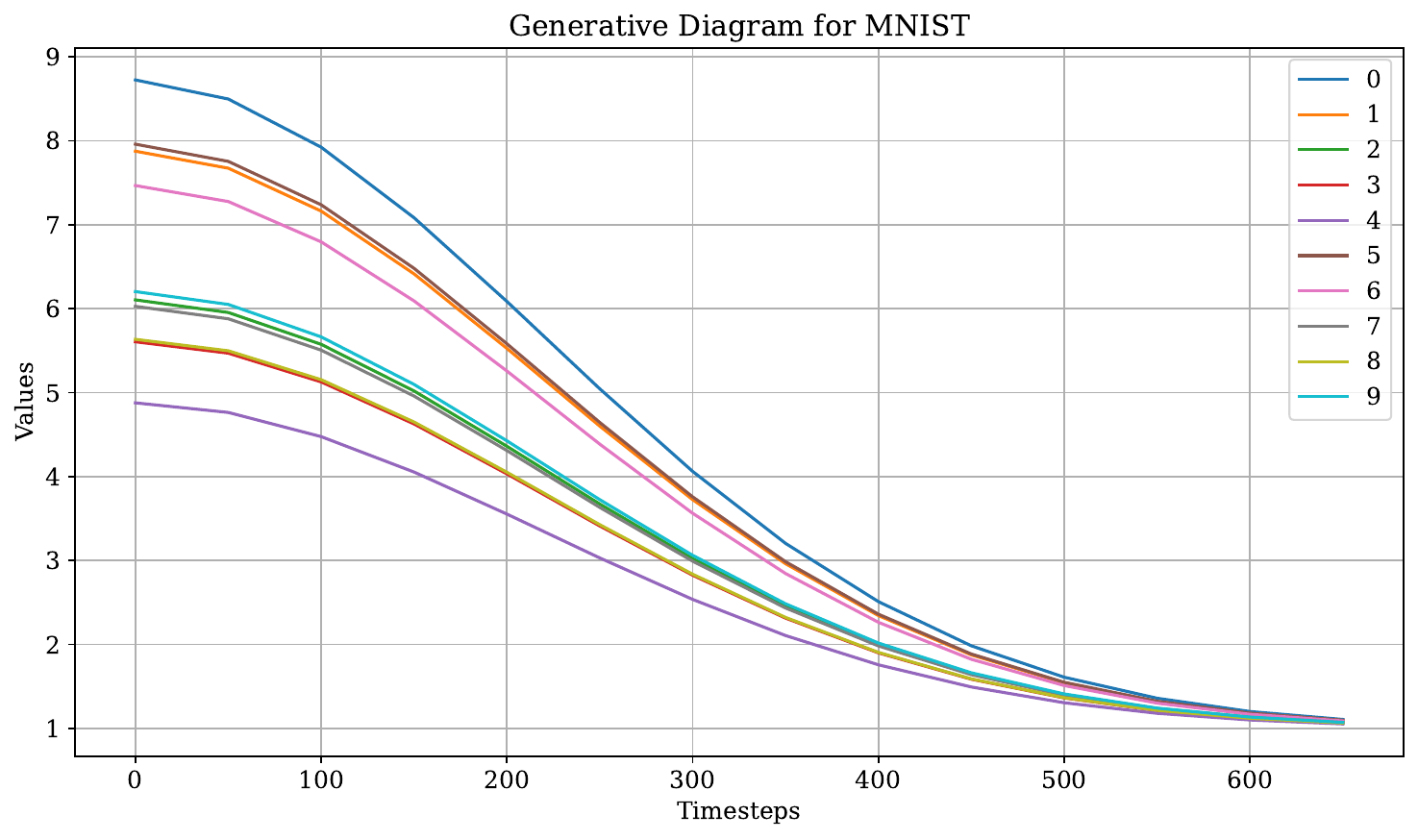}
        \caption{Generative diagram of MNIST}\label{fig:mnist-gen}
      \end{subfigure}\par\vspace{0.3em}
      \begin{subfigure}[t]{\linewidth}
        \includegraphics[width=\linewidth]{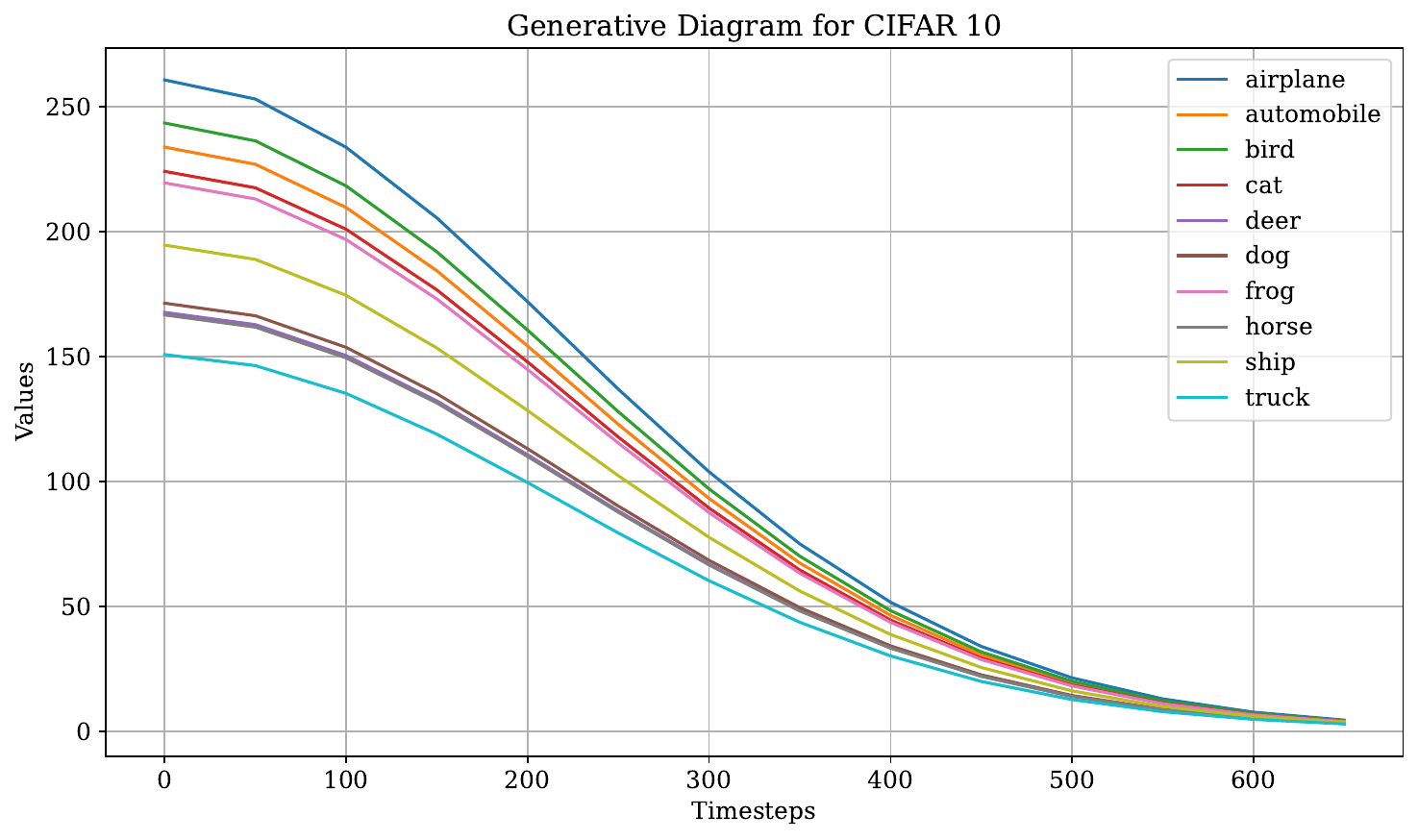}
        \caption{Generative diagram of CIFAR-10}\label{fig:cifar10-gen}
      \end{subfigure}
    \end{minipage}
    \hfill
    \begin{minipage}[t]{0.48\textwidth}
      \centering
      \begin{subfigure}[t]{\linewidth}
        \includegraphics[width=\linewidth]{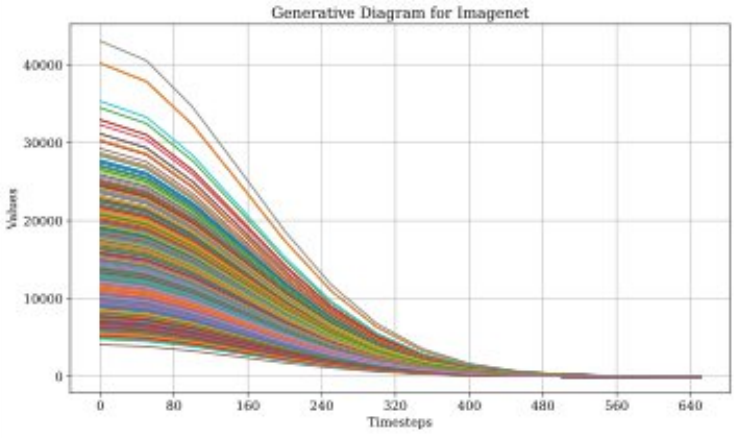}
        \caption{Generative diagram of ImageNet}\label{fig:imagenet-gen}
      \end{subfigure}\par\vspace{0.3em}
      \begin{subfigure}[t]{\linewidth}
        \includegraphics[width=\linewidth]{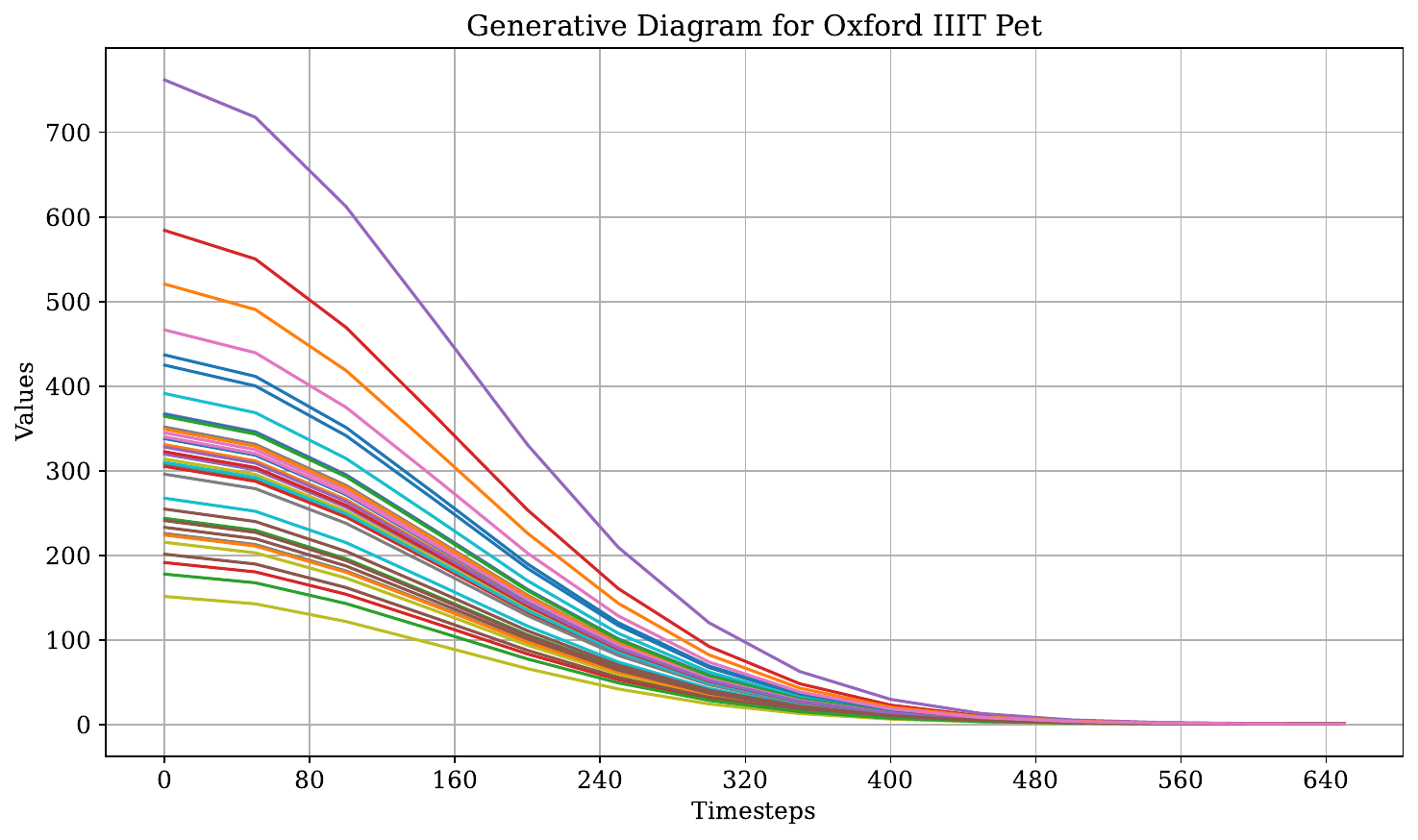}
        \caption{Generative diagram of Oxford–IIIT Pet}\label{fig:pet-gen}
      \end{subfigure}
    \end{minipage}
    \caption{Merger–transition measure obtained from the intersection time of the principal eigenvalues of the class-covariance matrices.}
    \label{fig:gen-diags}
  \end{figure*}
\endgroup


\begingroup
  \setlength{\abovecaptionskip}{4pt}
  \setlength{\belowcaptionskip}{4pt}
  \setlength{\textfloatsep}{6pt}
  \setlength{\floatsep}{6pt}
  \setlength{\intextsep}{6pt}

  \begin{figure*}[!p]
    \centering
    \begin{minipage}[t]{0.48\textwidth}
      \centering
      \begin{subfigure}[t]{\linewidth}
        \includegraphics[width=\linewidth]{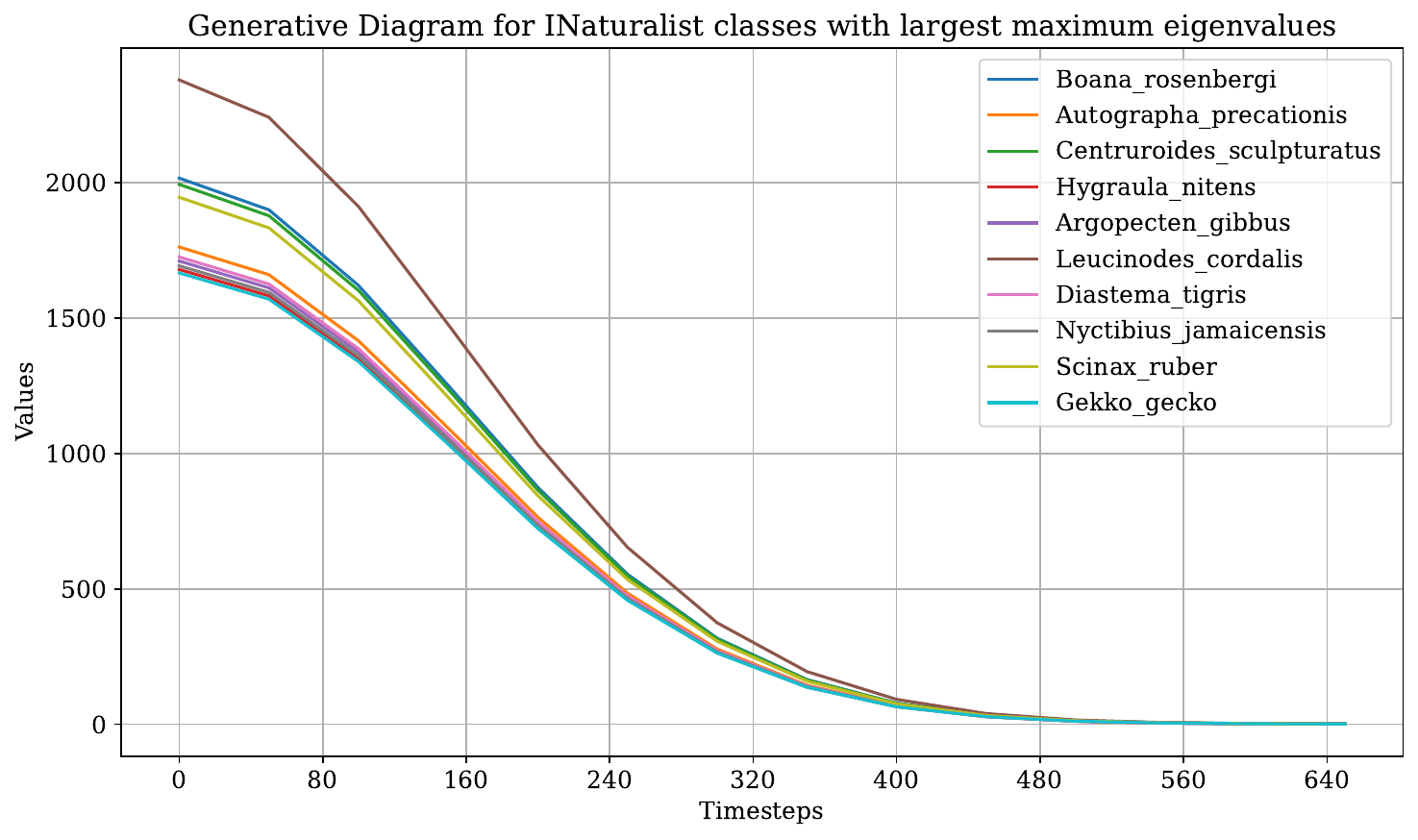}
        \caption{iNaturalist — all classes}\label{fig:inat-gen}
      \end{subfigure}\par\vspace{0.3em}
      \begin{subfigure}[t]{\linewidth}
        \includegraphics[width=\linewidth]{icml/images/inat_gen_diag_ub_eps_500_edm.pdf}
        \caption{iNaturalist — top 10 classes}\label{fig:inat-top10}
      \end{subfigure}
    \end{minipage}
    \hfill
    \begin{minipage}[t]{0.48\textwidth}
      \centering
      \begin{subfigure}[t]{\linewidth}
        \includegraphics[width=\linewidth]{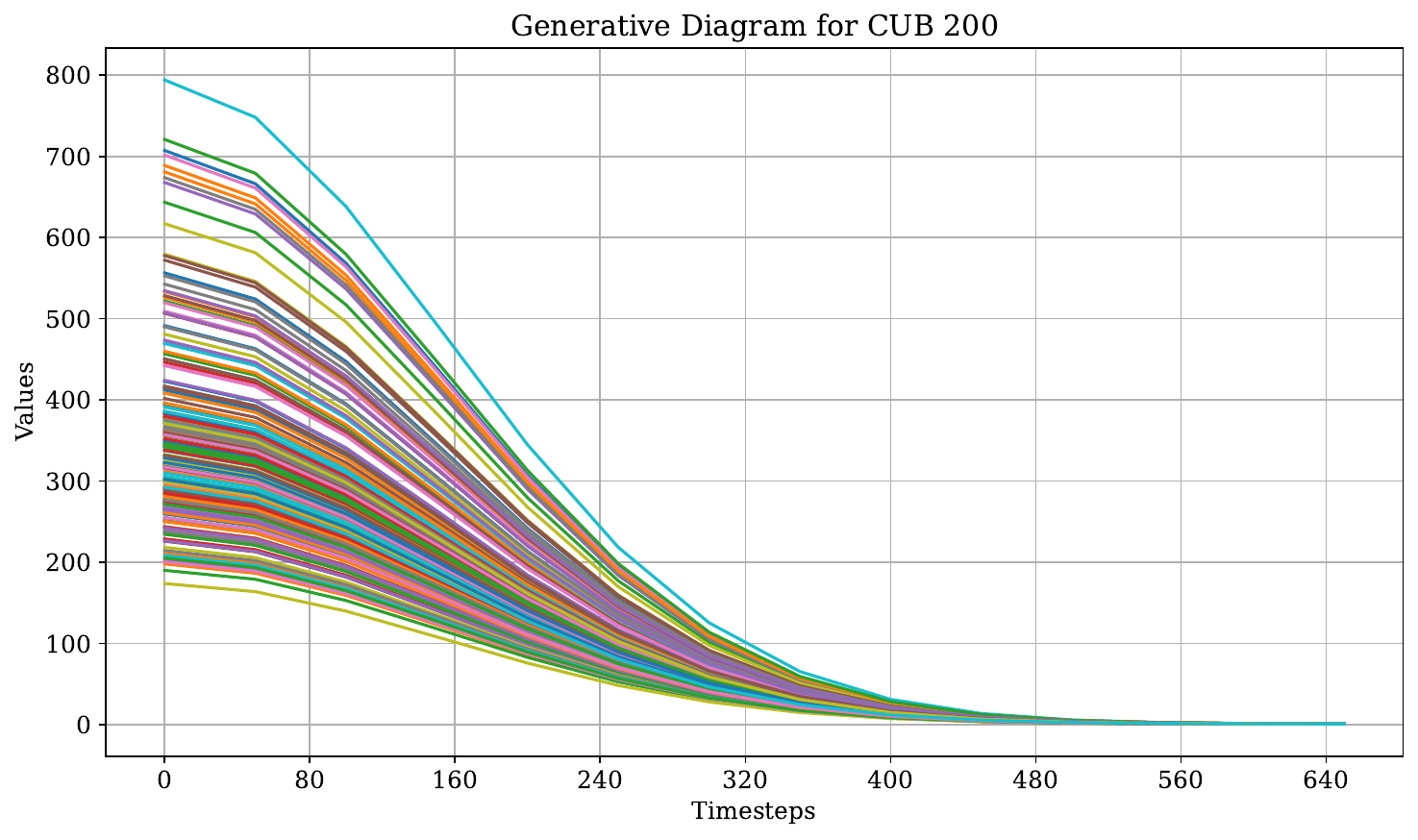}
        \caption{CUB-200 — all classes}\label{fig:cub-gen}
      \end{subfigure}\par\vspace{0.3em}
      \begin{subfigure}[t]{\linewidth}
        \includegraphics[width=\linewidth]{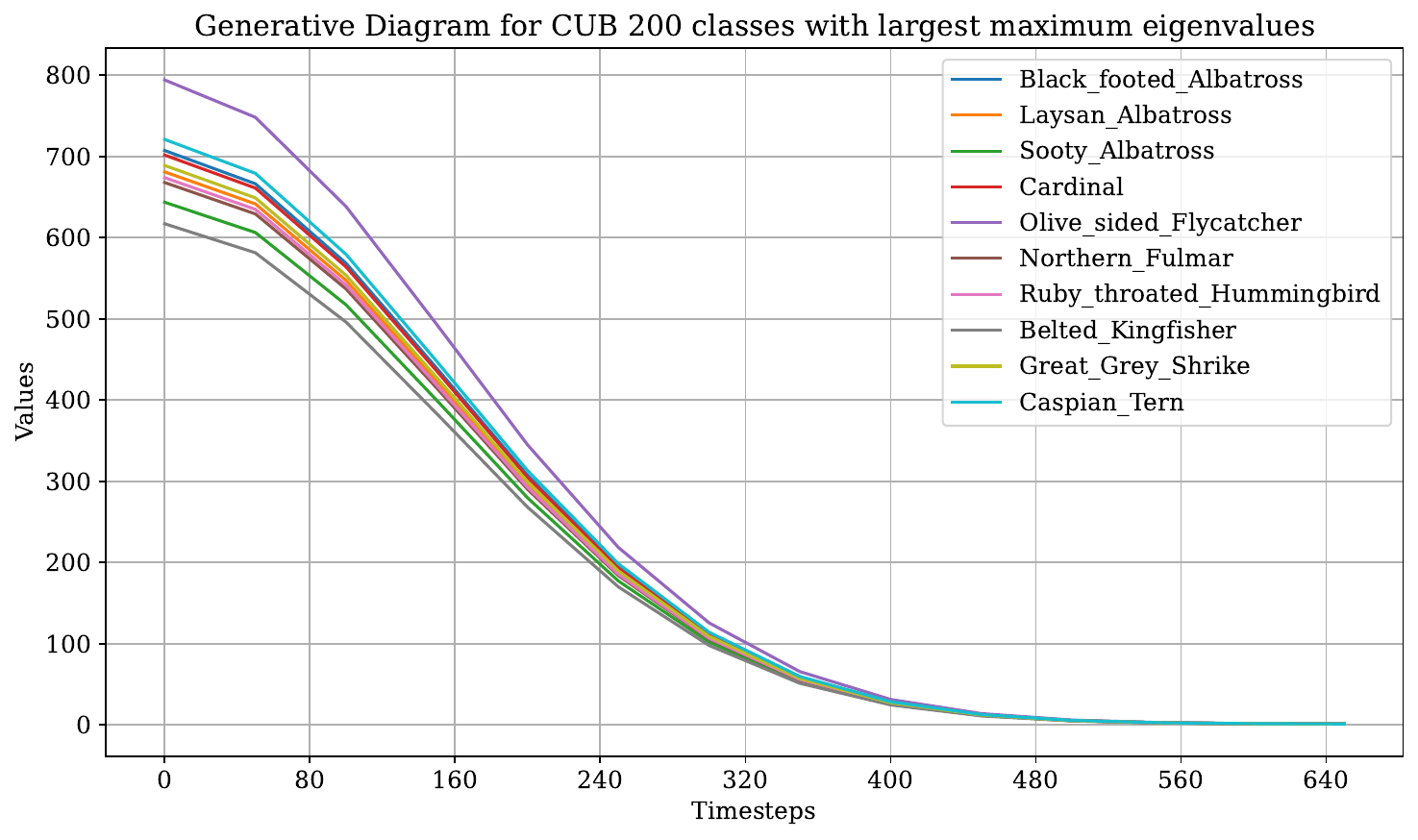}
        \caption{CUB-200 — top 10 classes}\label{fig:cub-top10}
      \end{subfigure}
    \end{minipage}
    \caption{Merger–transition measures obtained from the intersection times of the principal eigenvalues for iNaturalist and CUB-200. The bottom row shows that the classes with the ten largest eigenvalue magnitudes upper-bound all other transitions.}
    \label{fig:nat-cub-combined}
  \end{figure*}
\endgroup

\clearpage
\newpage

\subsection{Rare-class generation}
\label{sec:rare-class-app}

\paragraph{Interpolation-based interval guidance.}
In \Cref{sec:rare-class} we observed that augmenting
\Cref{alg:opt-cfg} with an \emph{interpolation
correction}—akin to ILVR~\citep{choi2021ilvrconditioningmethoddenoising}—gives
the best fidelity for CUB-200 and iNaturalist-2019.
The full procedure is listed in
\Cref{alg:opt-intp}; it differs from standard
Interval Guidance only in lines~5–7.

\vspace{0.1in}
\begin{algorithm}[hbtp]
\caption{Interpolation-based interval guidance (class $c$)}
\label{alg:opt-intp}
\begin{algorithmic}[1]\small
\Require
    \begin{itemize}
    \item $x_{T}\!\sim\!\mathcal{N}(\mathbf{0},\mathbf{I})$ \hfill\emph{latent to be denoised}
    \item $\hat{x}_{0}\!\sim\!p_{0}$ with $\operatorname{shape}(x_{T})=\operatorname{shape}(\hat{x}_{0})$
          \hfill\emph{class-$c$ exemplar}
    \item guidance weights $\text{CFG}_{w}$, $\text{CFG}_{0}$
    \item interpolation schedule $\eta=\{\eta_{t}\}_{t=0}^{T-1}$,
          where $0\le\eta_{t}\le1$
    \end{itemize}
\For{$t=T-1,\dots,0$}
  \If{$t_{\text{start},c} < t < t_{\text{end},c}$} \Comment{guidance window}
      \State $x_{t}\;\gets\;\text{CFG}_{w}(x_{t+1},c)$
      \State $\hat{x}_{t}\;\gets\;\text{FWD}(\hat{x}_{0},t)$
      \State $x_{t}\;\gets\;\eta_{t}\,x_{t} \;+\;(1-\eta_{t})\,\hat{x}_{t}$
  \Else
      \State $x_{t}\;\gets\;\text{CFG}_{0}(x_{t+1},c)$
  \EndIf
\EndFor
\State\Return $x_{0}$
\end{algorithmic}
\end{algorithm}

Note that in \Cref{alg:opt-intp}, \texttt{CFG\(_{w}\)} applies classifier-free guidance with strength~$w$;
      \texttt{CFG\(_{0}\)} disables conditioning. \texttt{FWD}($\hat{x}_{0},t$) generates the \emph{forward-noised}
      version of the exemplar at step~$t$. The convex update in line $5$ nudges the current latent
      towards the exemplar’s trajectory, counteracting class drift.

\paragraph{Choosing the interpolation schedule $\eta$.}
Because guidance corrections are most valuable late in the reverse chain,
we derive $\eta_{t}$ from the noise schedule $\beta_{t}$:
$\eta_{t}
\;=\;s(\beta_{t}/\max_{u<T}\beta_{u}),~0<s\le1$. The scale factor $s$ is found by a binary search over
$[10^{-4},10^{-2}]$:
larger values degrade sharpness, while smaller ones have negligible
impact.

\vspace{0.05 in}
\begin{figure*}[htbp]
  \centering
  \scriptsize
  \captionsetup{skip=0pt,belowskip=2pt,aboveskip=2pt}
  \setlength{\tabcolsep}{4pt}  

  \begin{tabular}{cc}
    \subcaptionbox{CDF of principal‐eigenvalue differences (iNaturalist)\label{fig:inat-cdf}}{%
      \includegraphics[width=0.45\textwidth,trim=5 5 5 5,clip]{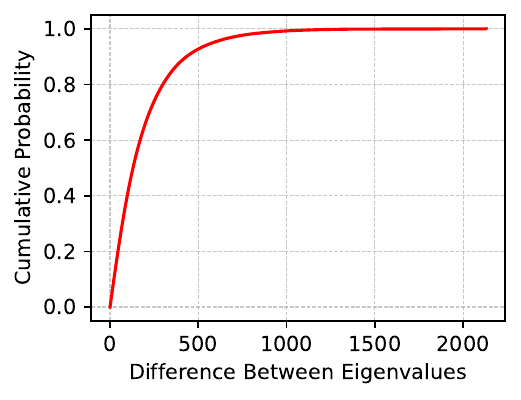}%
    } &
    \subcaptionbox{CDF of principal‐eigenvalue differences (CUB200)\label{fig:cub-cdf}}{%
      \includegraphics[width=0.45\textwidth,trim=5 5 5 5,clip]{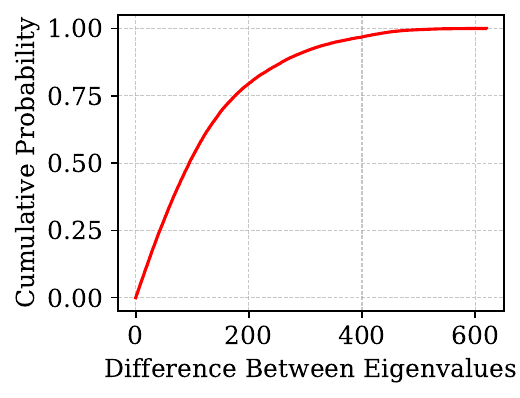}%
    } \\[-2pt]
    \subcaptionbox{Merge‐probability vs.~threshold (iNaturalist)\label{fig:inat-mix}}{%
      \includegraphics[width=0.45\textwidth,trim=5 5 5 5,clip]{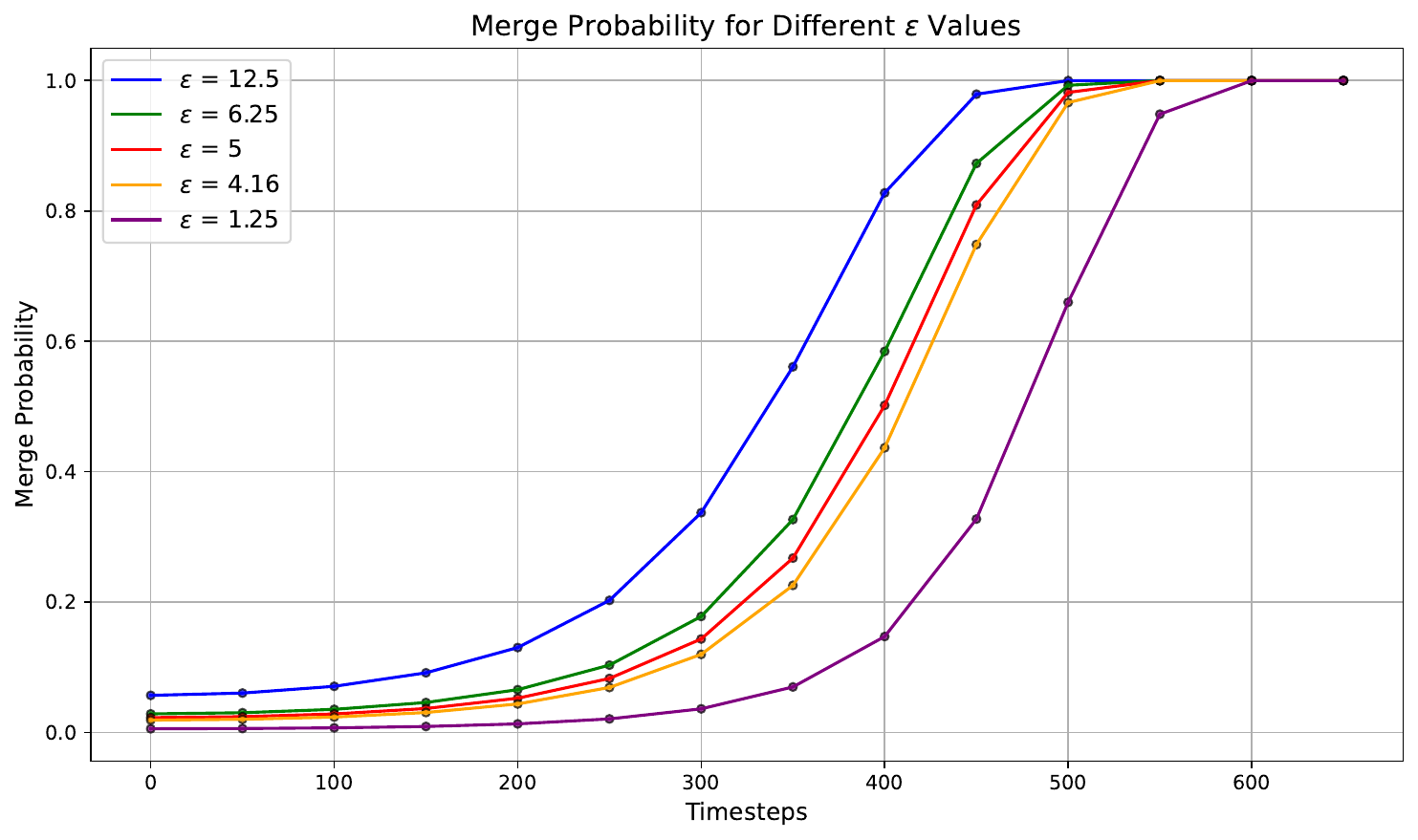}%
    } &
    \subcaptionbox{Merge‐probability vs.~threshold (CUB200)\label{fig:cub-mix}}{%
      \includegraphics[width=0.45\textwidth,trim=5 5 5 5,clip]{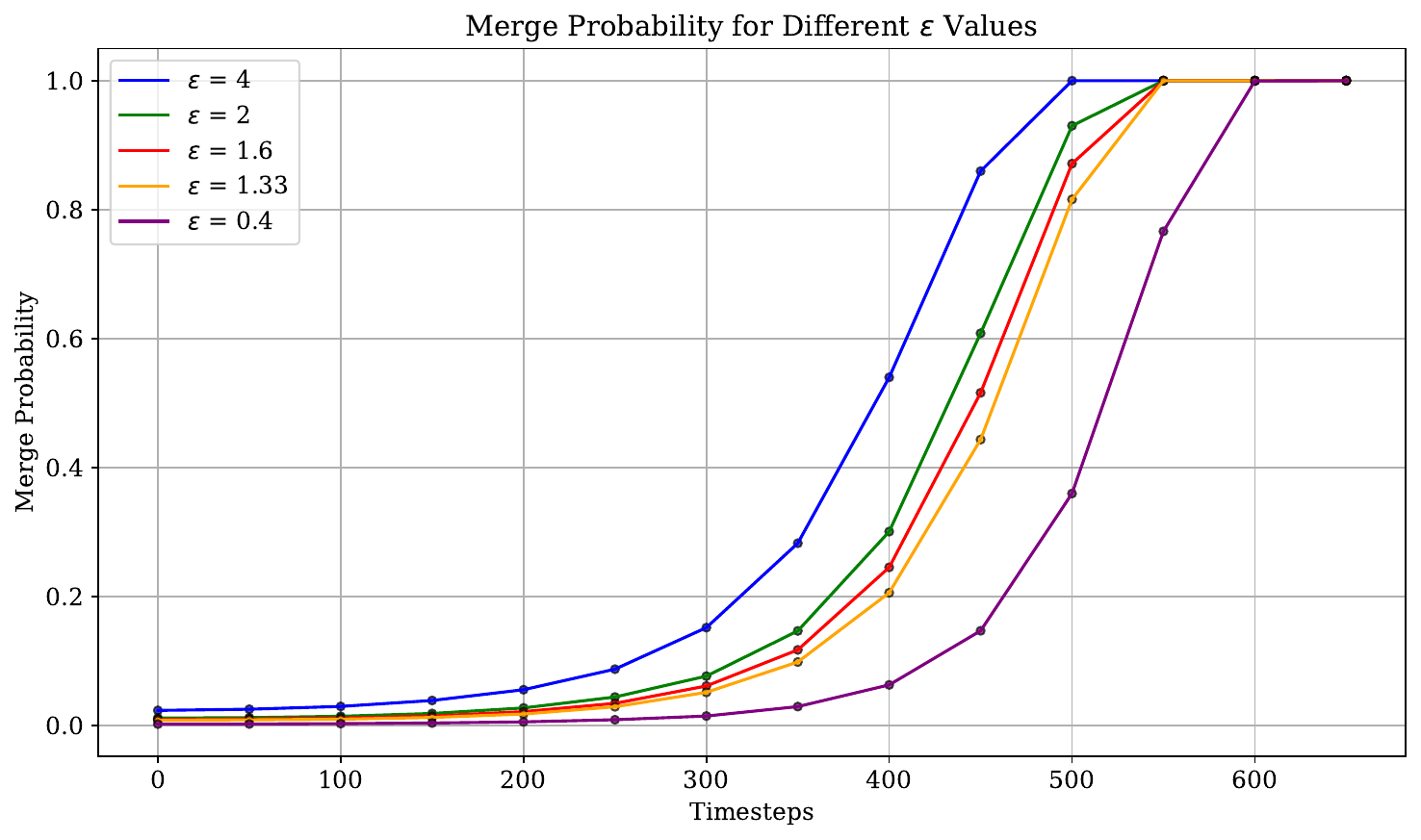}%
    }
  \end{tabular}

  \caption{Comparison of eigenvalue‐difference distributions and merge‐probability curves for iNaturalist vs.\ CUB200.}
  \label{fig:cdf-mix-comparison-2}
\end{figure*}

\begin{figure*}[hbtp!]
  \centering
  \resizebox{0.8\textwidth}{!}{%
    \begin{minipage}{\textwidth}
    \centering
      \subfloat[Original SD Image]{%
        \includegraphics[width=35mm,height=35mm,keepaspectratio]{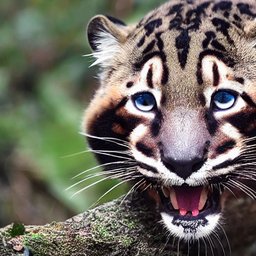}%
      }
      \subfloat[Original SD Image]{%
        \includegraphics[width=35mm,height=35mm,keepaspectratio]{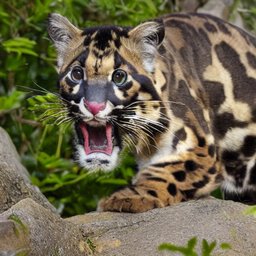}%
      }
      \subfloat[Original SD Image]{%
        \includegraphics[width=35mm,height=35mm,keepaspectratio]{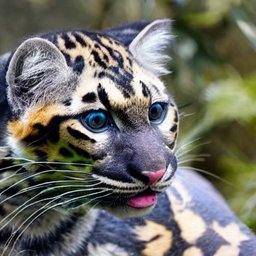}%
      }
      \hspace{0mm}
      \subfloat[Conditioning Image 1]{%
        \includegraphics[width=35mm,height=35mm,keepaspectratio]{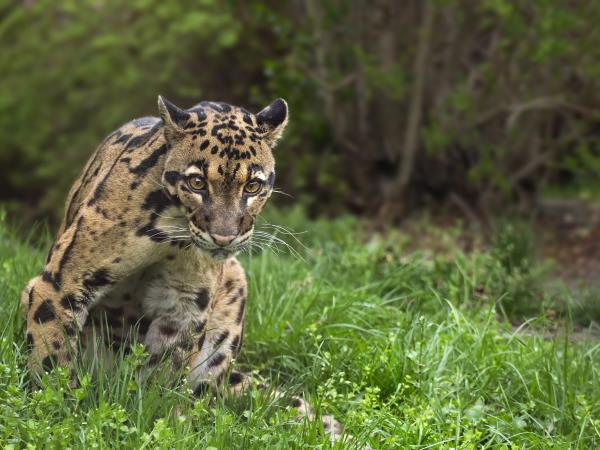}%
      }
      \subfloat[Conditioning Image 2]{%
        \includegraphics[width=35mm,height=35mm,keepaspectratio]{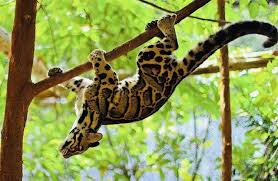}%
      }
      \subfloat[Conditioning Image 3]{%
        \includegraphics[width=35mm,height=35mm,keepaspectratio]{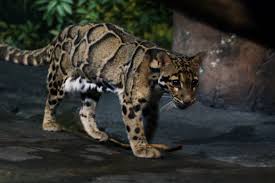}%
      }
      \hspace{0mm}
      \subfloat[New SD Image 1]{%
        \includegraphics[width=35mm,height=35mm,keepaspectratio]{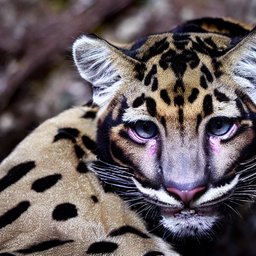}%
      }
      \subfloat[New SD Image 2]{%
        \includegraphics[width=35mm,height=35mm,keepaspectratio]{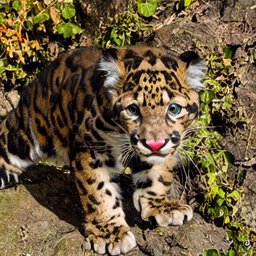}%
      }
      \subfloat[New SD Image 3]{%
        \includegraphics[width=35mm,height=35mm,keepaspectratio]{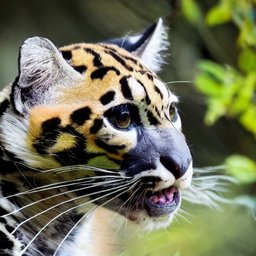}%
      }
    \end{minipage}%
  }
  \caption{Visual comparison of generation algorithms for stable diffusion for the iNaturalist class prompt “clouded leopard walking”.}
  \label{fig:comp_2}
\end{figure*}

\begin{figure*}[htbp!]
  \centering
  \makebox[\textwidth][c]{%
    \resizebox{0.8\textwidth}{!}{%
      \begin{minipage}{\textwidth}
      \centering
        \subfloat[Original SD Image]{
          \includegraphics[width=35mm,height=35mm,keepaspectratio]{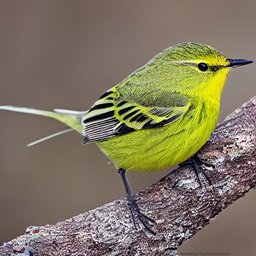}%
        }
        \subfloat[Original SD Image]{
          \includegraphics[width=35mm,height=35mm,keepaspectratio]{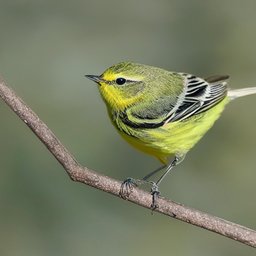}%
        }
        \subfloat[Original SD Image]{
          \includegraphics[width=35mm,height=35mm,keepaspectratio]{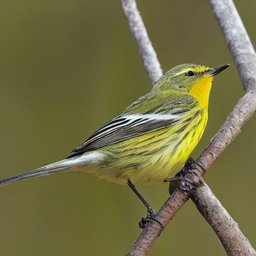}%
        }
        \hspace{0mm}
        \subfloat[Conditioning Image 1]{
          \includegraphics[width=35mm,height=35mm,keepaspectratio]{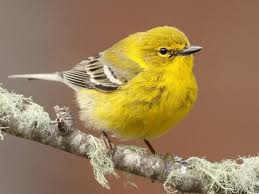}%
        }
        \subfloat[Conditioning Image 2]{
          \includegraphics[width=35mm,height=35mm,keepaspectratio]{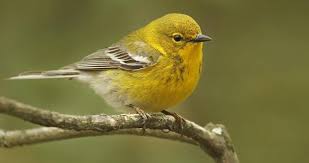}%
        }
        \subfloat[Conditioning Image 3]{
          \includegraphics[width=35mm,height=35mm,keepaspectratio]{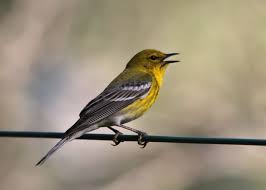}%
        }
        \hspace{0mm}
        \subfloat[New SD Image 1]{
          \includegraphics[width=35mm,height=35mm,keepaspectratio]{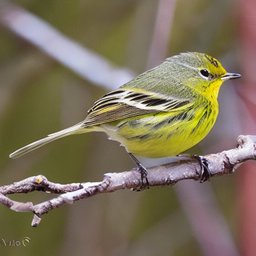}%
        }
        \subfloat[New SD Image 2]{
          \includegraphics[width=35mm,height=35mm,keepaspectratio]{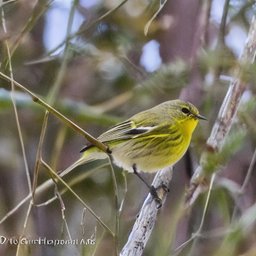}%
        }
        \subfloat[New SD Image 3]{
          \includegraphics[width=35mm,height=35mm,keepaspectratio]{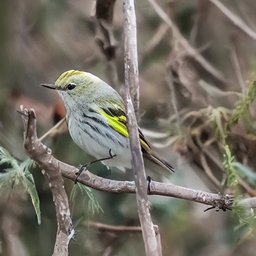}%
        }
      \end{minipage}%
    }%
  }
  \caption{Visual comparison of generation algorithms for stable diffusion for the CUB-200 prompt “pine warbler.”}
  \label{fig:comp_s}
\end{figure*}

\clearpage
\newpage
\paragraph{Experimental settings.} For all runs, we fix
$\epsilon=5$ on iNaturalist and $\epsilon=2$ on CUB-200 when computing merger statistics.
\Cref{fig:inat-gen,fig:cub-gen} visualise per-class merger probabilities,
with complementary CDF and eigenvalue trajectories in
\Cref{fig:inat-cdf,fig:inat-mix,fig:cub-cdf,fig:cub-mix}.
Top-10 eigenvalue plots appear in
\Cref{fig:inat-top10,fig:cub-top10}.
Qualitative comparisons between \Cref{alg:opt-intp} and naïve
Stable Diffusion, using identical random seeds, is given in
\Cref{fig:comp_2,fig:comp_s}.
\vspace{0.1in}

\vspace{0.1 in}

\clearpage
\newpage

\subsection{Zero-shot classification}
\label{sec:zsc}

In \citet{li2023diffusionmodelsecretlyzeroshot}, a pretrained
\emph{class-conditional} diffusion network $f_{\theta}$ is repurposed as a
classifier by averaging softmax logits over forward-diffused replicas of the query image.  
We keep that spirit but (i)~restrict the average to the
\emph{class-specific guidance window}
$[\,t_{\mathrm{start},\lambda},\,t_{\mathrm{stop},\lambda}]$
identified in \Cref{sec:class-cond}, and  
(ii)~attach an importance weight $w(t)$ to every timestep~$t$.
Setting $w(t)=1/T$ and  the bounds $t_{\mathrm{start},\lambda} =0, t_{\mathrm{stop},\lambda}=T$ recovers the estimator
of \citet{li2023diffusionmodelsecretlyzeroshot}. 

Throughout this section, $\Lambda=\{1,\dots,K\}$ is the label set;
$\varepsilon_{n}\stackrel{\mathrm{iid}}{\sim}\mathcal{N}(0,I_{d})$ are noise seeds ($n=1,\dots,N$);  
 $\operatorname{FWD}_{t}(x,\varepsilon)$ denotes the \emph{forward} map that produces the noisy latent $z_{t}$ at step~$t$; $w(t)\!\ge\!0$ is a probability mass on the integer interval
      $[t_{\mathrm{start},\lambda},t_{\mathrm{stop},\lambda}]$.
\vspace{0.1in}

\begin{algorithm}[h]
\caption{Zero-shot class probability $p_{\lambda}(z)$}
\label{alg:class}
\begin{algorithmic}[1]\small
\Require
  query $z\!\sim\!p_{0}$, class label $\lambda\in\Lambda$,
  time window $t_{\mathrm{start},\lambda}\le t\le t_{\mathrm{stop},\lambda}$,
  weights $\{w(t)\}$ summing to~$1$,
  noise seeds $\{\varepsilon_{n}\}_{n=1}^{N}$
\State $p_{\lambda}\gets 0$
\For{$t=t_{\mathrm{start},\lambda},\dots,t_{\mathrm{stop},\lambda}$}
  \For{$n=1,\dots,N$}
    \State $z_{t}\gets\operatorname{FWD}_{t}(z,\varepsilon_{n})$
    \State $s_{\lambda}\gets
           -\bigl\lVert f_{\theta}(z_{t},\varepsilon_{n},\lambda)
                       -\varepsilon_{n}\bigr\rVert^{2}$
    \State $\displaystyle
           p_{\lambda}\gets
           p_{\lambda}\;+\;
           \frac{w(t)}{N}\;
           \frac{\exp(s_{\lambda})}
                {\sum_{\mu\in\Lambda}\exp\!\bigl(
                   -\lVert f_{\theta}(z_{t},\varepsilon_{n},\mu)
                         -\varepsilon_{n}\rVert^{2}\bigr)}$
  \EndFor
\EndFor
\State\Return $p_{\lambda}$
\end{algorithmic}
\label{alg:zs-cl}
\end{algorithm}
\vspace{0.05in}
Let $\alpha_{t}$ be the signal coefficient of the VP process
\eqref{eq:marginal}.  The signal-to-noise ratio is $\operatorname{SNR}(t)=\alpha_{t}^{2}/(1-\alpha_{t}^{2})$. We consider three discrete weight laws: \emph{Uniform:} $w(t)=1/(t_{\mathrm{stop},\lambda}-t_{\mathrm{start},\lambda}+1)$.
\emph{Inverse-SNR:} $w(t)\propto\operatorname{SNR}(t)^{-1}$ with
      $t_{\mathrm{start},\lambda}=0$. \emph{Truncated inverse-SNR:} same as previous but restricted to
      $t\ge 20$ (empirically, very early steps degrade performance as the score model is not exact at these time scales
      \citep{chen2022sampling,karras2022edm}). We fix $N=250$ for all of our experiments following \cite{li2023diffusionmodelsecretlyzeroshot}.

\subsubsection{Binary classification via linear probes}
\label{sec:bin-clf}

To understand why non-uniform weights help, we study binary accuracy along the
forward chain with \emph{no} diffusion model involved. Pick two ImageNet classes $\{\lambda,\mu\}$ at random and sample $\min\{\text{card}(\lambda),\text{card}(\mu),10000\}$ number of images for each class.  At each step $t$, we extract the noisy
embedding $z_{t}$ (VP schedule) and train a linear MLP on $80\%$ of
the embeddings; the rest forms the test set.  We repeat the experiment
$20$ times and report mean\,$\pm$\,s.d.

\begin{figure*}[htbp]
  \centering
  \begin{minipage}[t]{0.48\textwidth}
    \centering
    \begin{subfigure}[t]{\linewidth}
      \includegraphics[width=\linewidth]{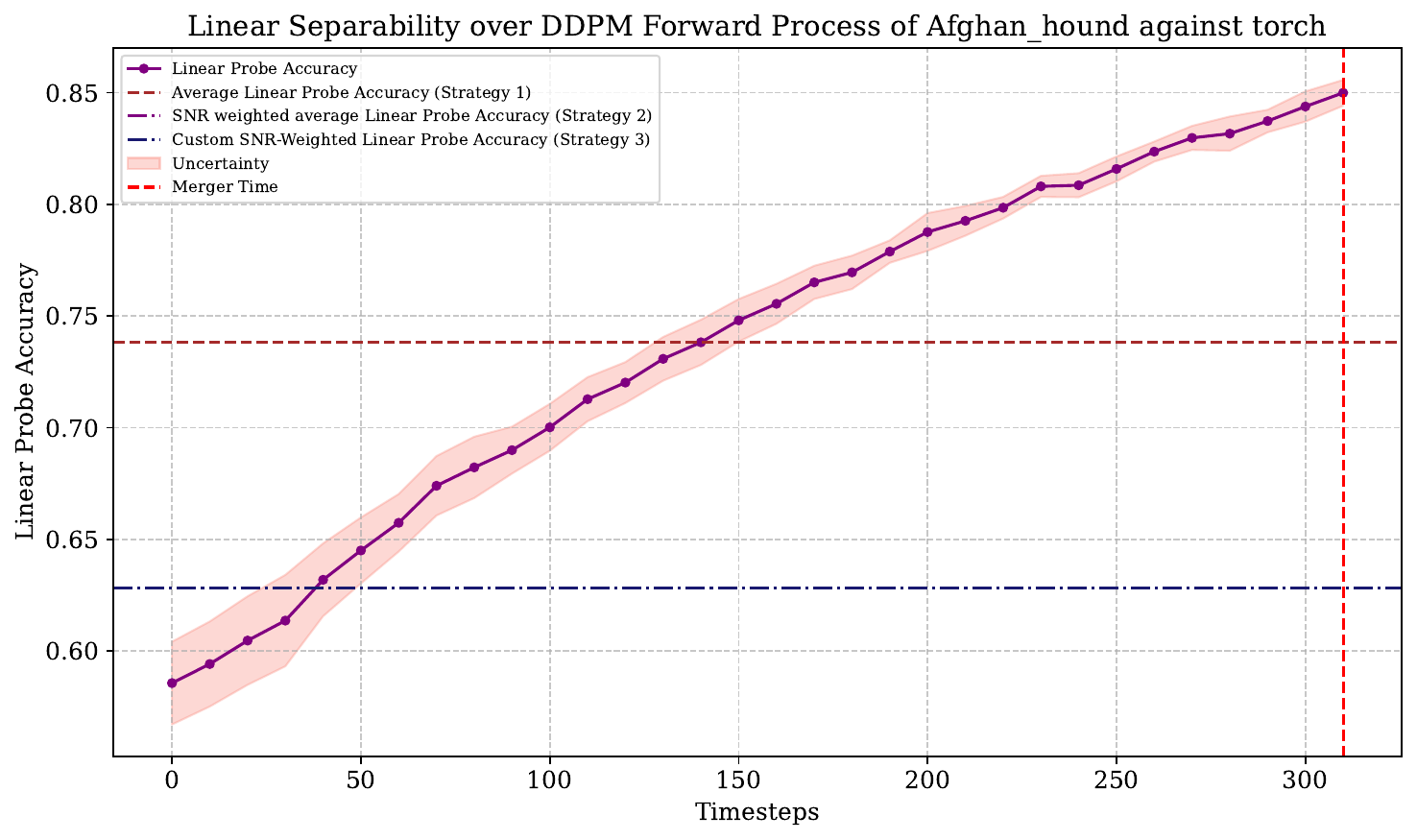}
      \caption{}\label{fig:lp0}
    \end{subfigure}\par\vspace{0.6em}

    \begin{subfigure}[t]{\linewidth}
      \includegraphics[width=\linewidth]{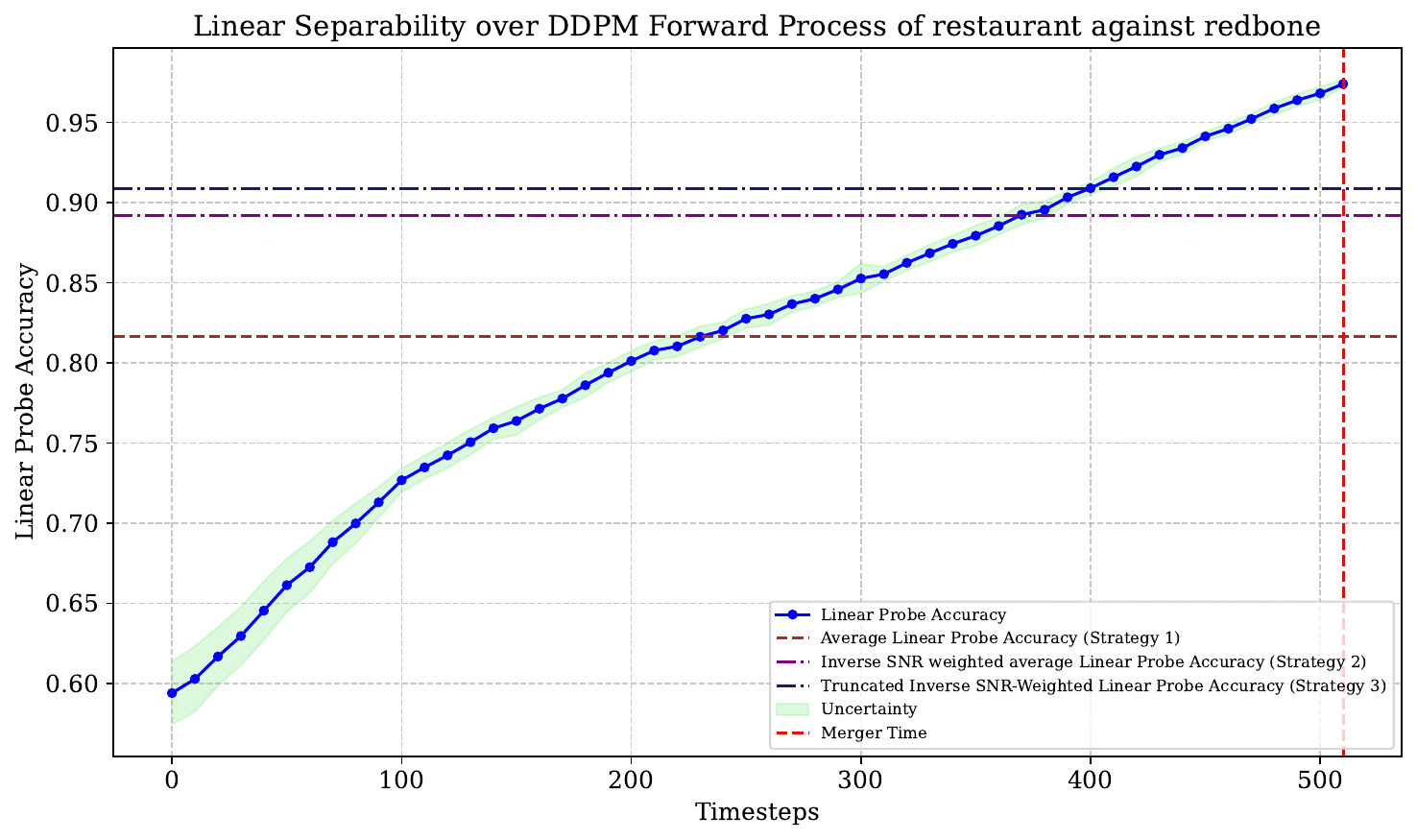}
      \caption{}\label{fig:lp1}
    \end{subfigure}\par\vspace{0.6em}

    \begin{subfigure}[t]{\linewidth}
      \includegraphics[width=\linewidth]{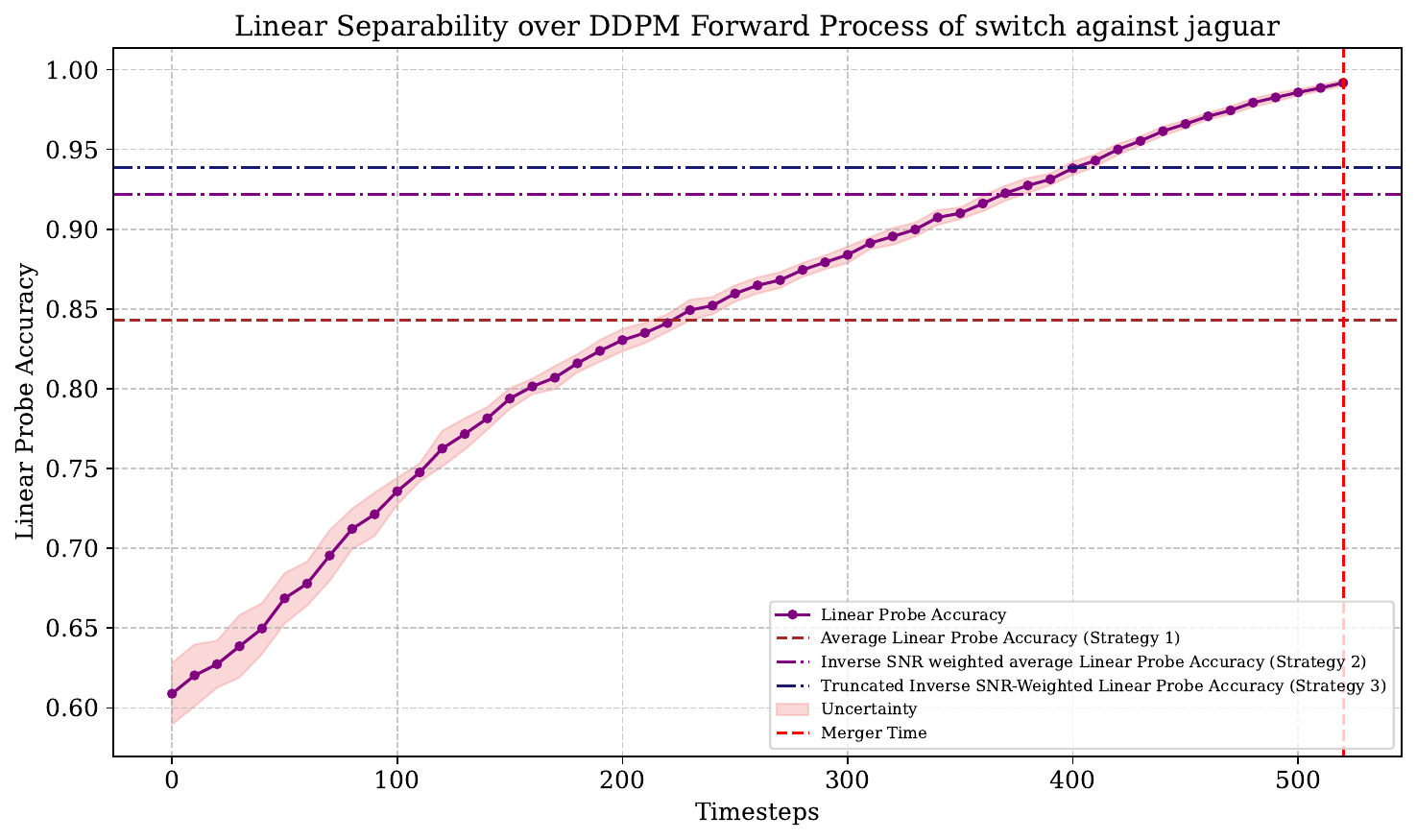}
      \caption{}\label{fig:lp2}
    \end{subfigure}\par\vspace{0.6em}

    \begin{subfigure}[t]{\linewidth}
      \includegraphics[width=\linewidth]{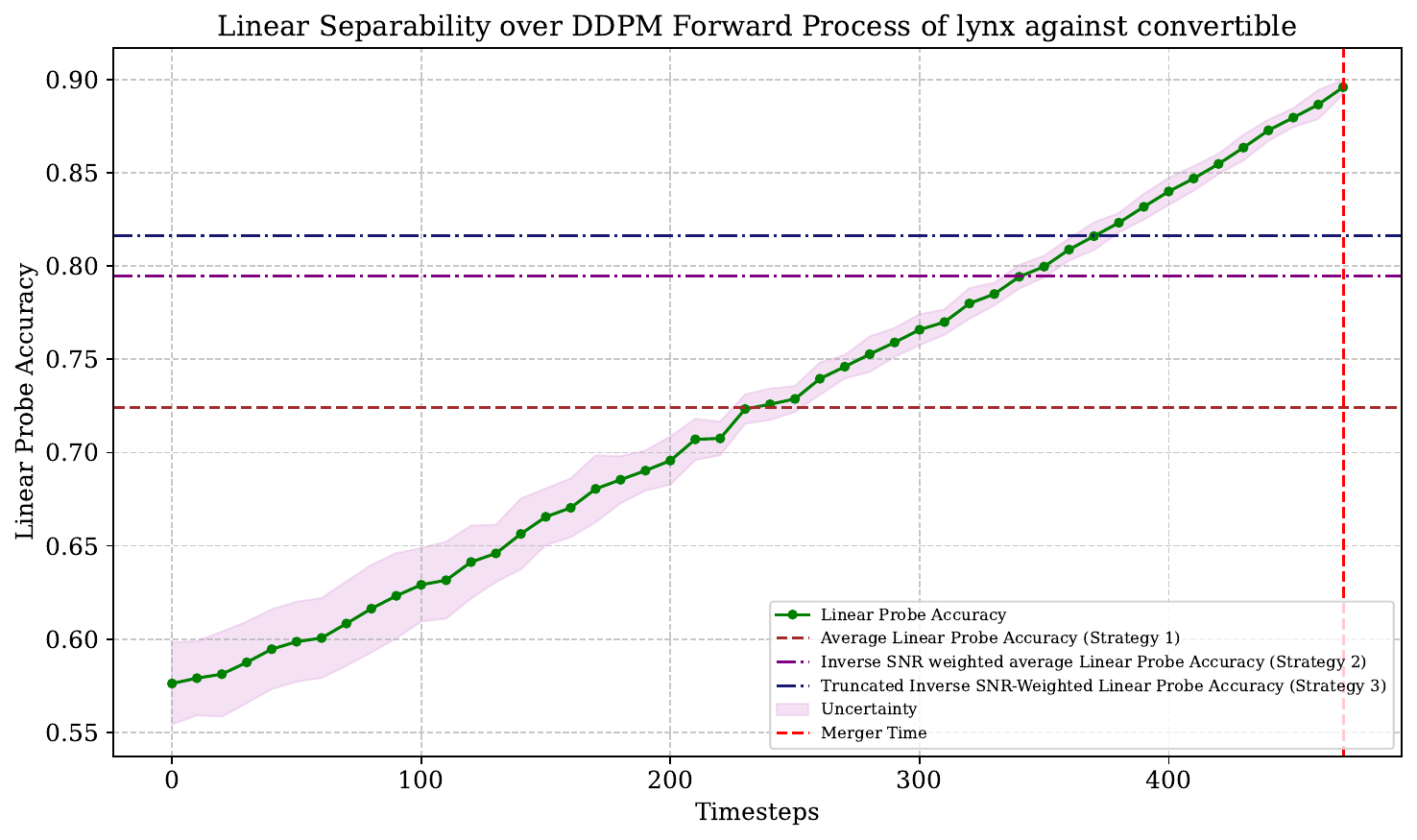}
      \caption{}\label{fig:lp3}
    \end{subfigure}\par\vspace{0.6em}

    \begin{subfigure}[t]{\linewidth}
      \includegraphics[width=\linewidth]{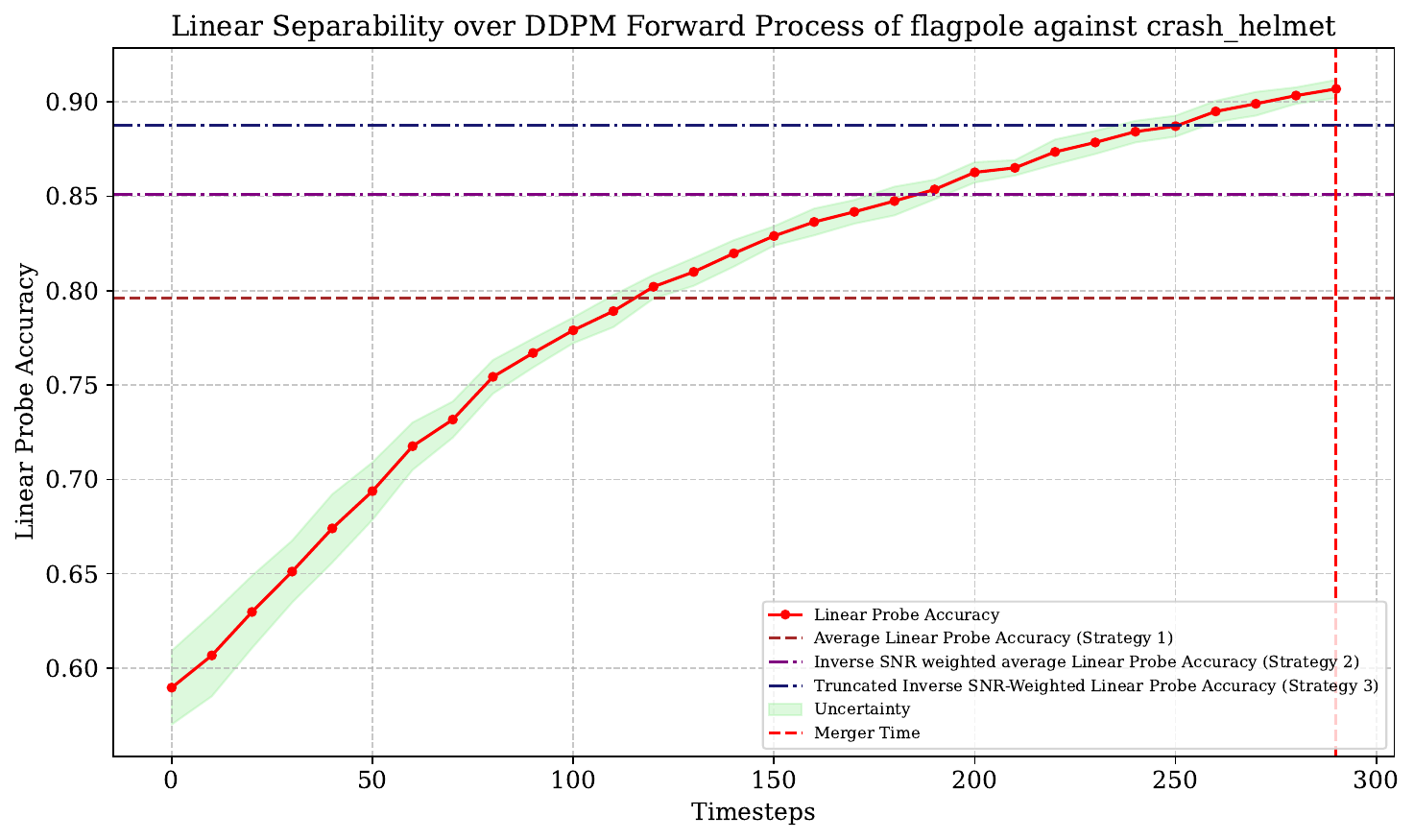}
      \caption{}\label{fig:lp4}
    \end{subfigure}
  \end{minipage}
  \hfill
  \begin{minipage}[t]{0.48\textwidth}
    \centering
    \begin{subfigure}[t]{\linewidth}
      \includegraphics[width=\linewidth]{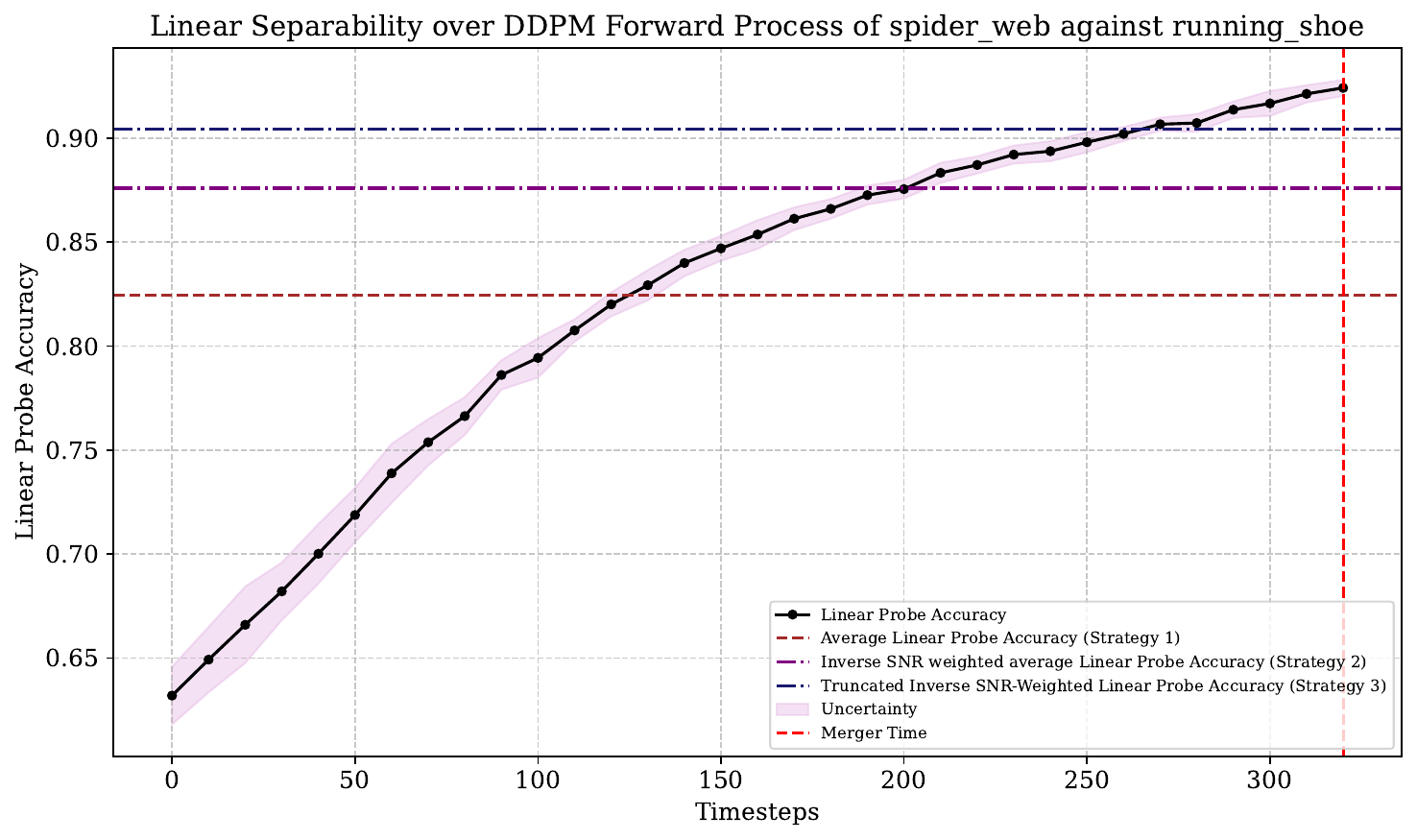}
      \caption{}\label{fig:lp5}
    \end{subfigure}\par\vspace{0.6em}

    \begin{subfigure}[t]{\linewidth}
      \includegraphics[width=\linewidth]{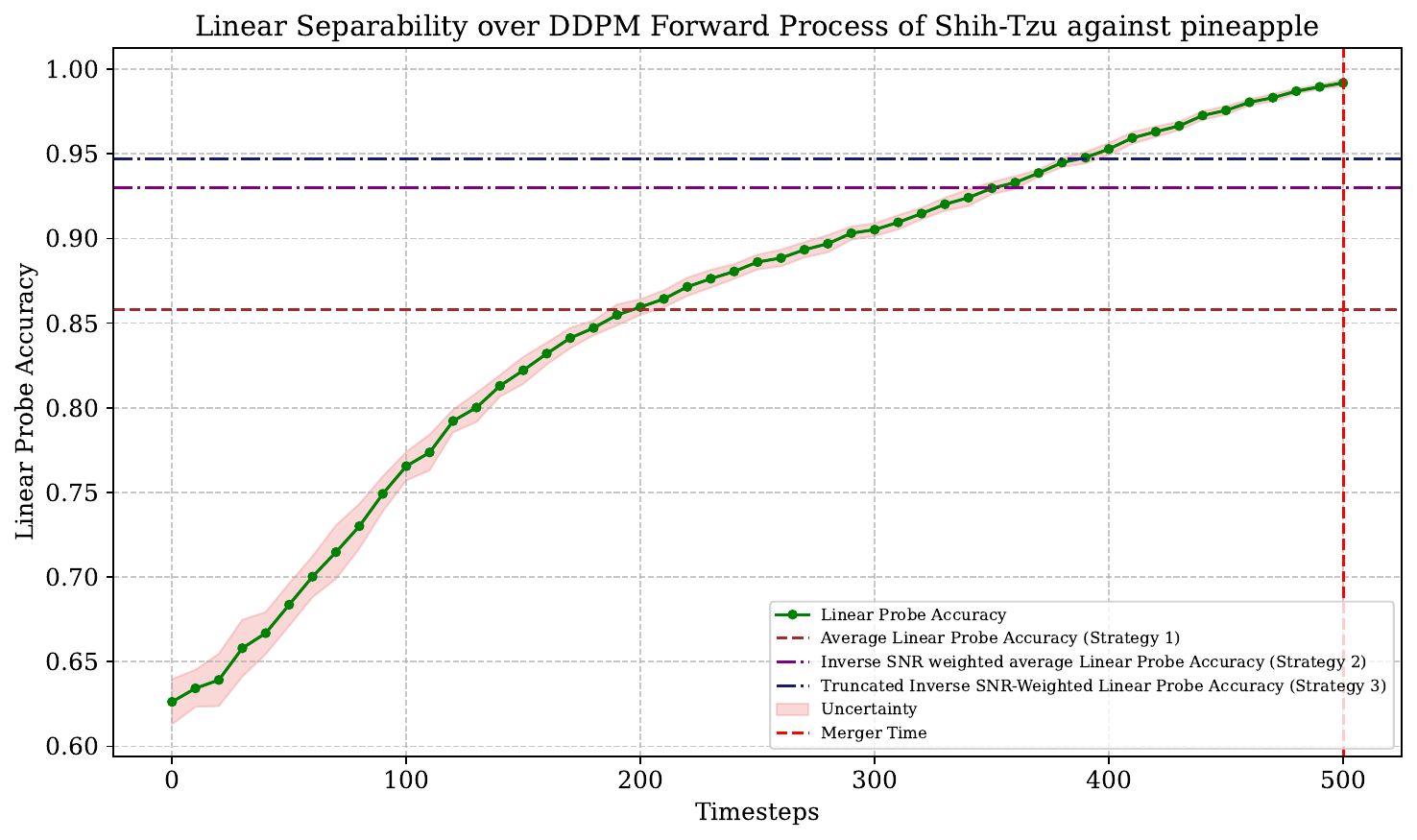}
      \caption{}\label{fig:lp6}
    \end{subfigure}\par\vspace{0.6em}

    \begin{subfigure}[t]{\linewidth}
      \includegraphics[width=\linewidth]{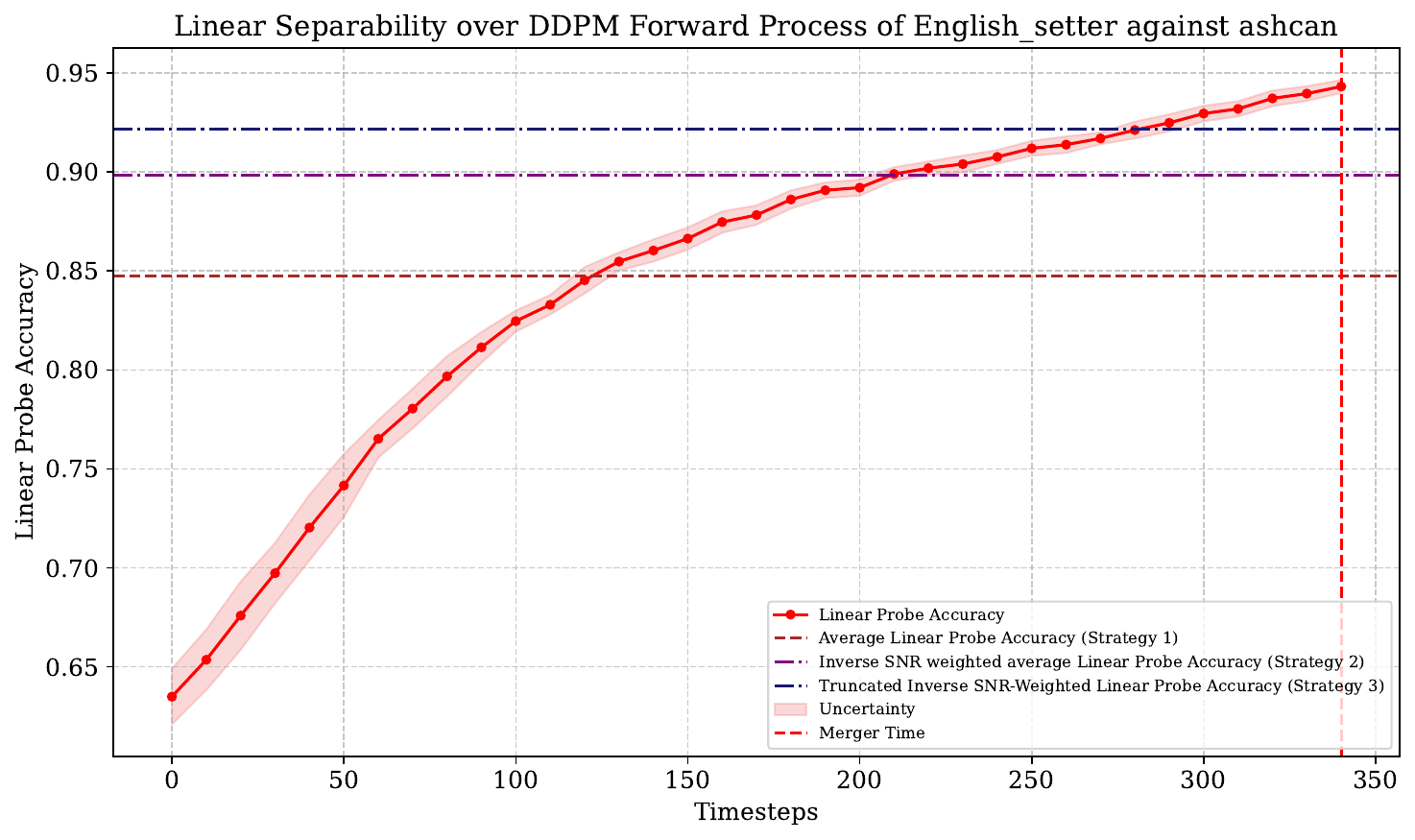}
      \caption{}\label{fig:lp7}
    \end{subfigure}\par\vspace{0.6em}

    \begin{subfigure}[t]{\linewidth}
      \includegraphics[width=\linewidth]{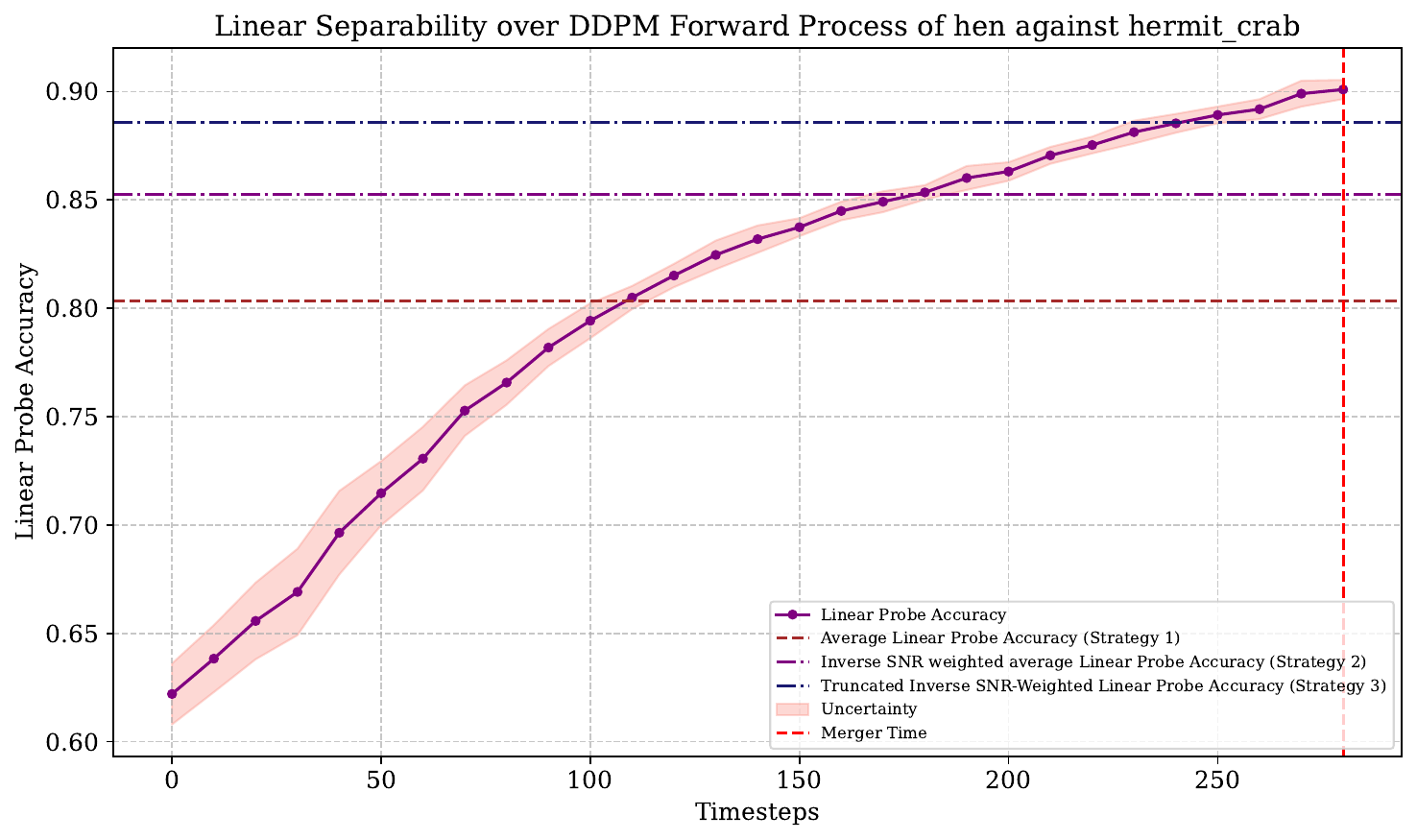}
      \caption{}\label{fig:lp8}
    \end{subfigure}\par\vspace{0.6em}

    \begin{subfigure}[t]{\linewidth}
      \includegraphics[width=\linewidth]{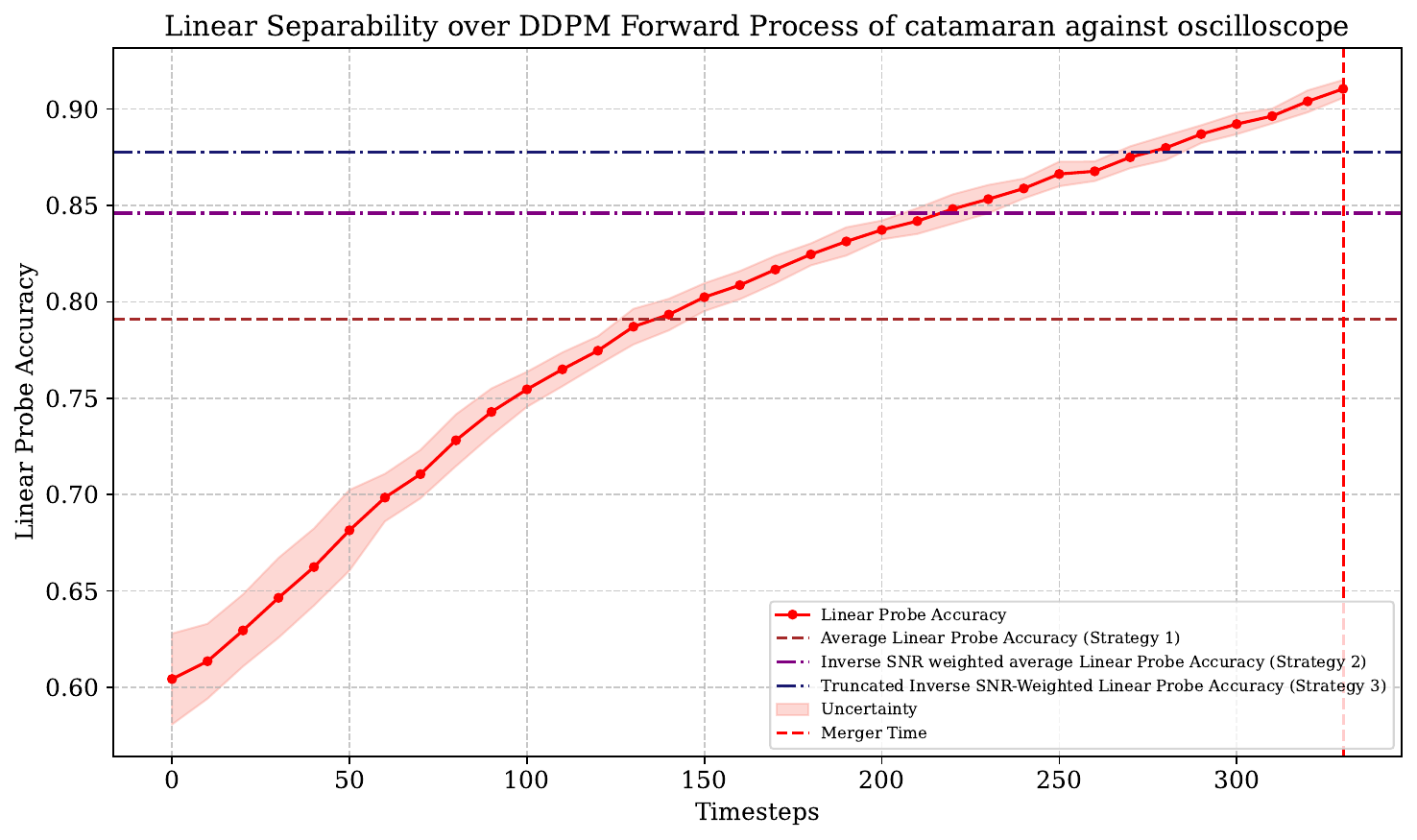}
      \caption{}\label{fig:lp9}
    \end{subfigure}
  \end{minipage}

  \caption{Linear-probe accuracy through the forward diffusion process.  
           Later timesteps hold greater discriminative information between the two classes.}
  \label{fig:zsc-column}
\end{figure*}

\Cref{fig:zsc-column} shows that test accuracy rises sharply and
peaks near the \emph{merger time} of $\lambda$ and $\mu$, after which it is not defined since the embeddings corresponding to $\lambda,\mu$ become practically indistinguishable.
Averaging the accuracies with the three weight laws yields the means in
~\Cref{tab:avg-table}; using the single best $t$ (the merger time)
achieves the highest score but is undefined for the multi-class case, however, the general trend for the other strategies that are valid for the multi-class setting remains the same. 
\vspace{0.05in}
\begin{table}[H]
\centering
\small
\begin{tabular}{|l|c|}
\hline
\rowcolor{gray!20}
Weighting strategy & Avg. binary accuracy $\uparrow$ \\ \hline
Uniform              & $0.72\pm0.03$ \\ \hline
Inverse-SNR          & $0.79\pm0.06$ \\ \hline
Trunc.\ inv.\ SNR    & $0.82\pm0.04$ \\ \hline
Merge-time probe     & $\mathbf{0.90\pm0.02}$ \\ \hline
\end{tabular}
\vspace{0.2em}
\caption{Binary accuracy for random ImageNet pairs.}
\label{tab:avg-table}
\end{table}
\vspace{-0.1in}
Each forward step convolves the data with a Gaussian kernel, progressively
smoothing non-linear features.  
Just before the classes merge, the representation is \emph{simpler} yet still
separates the two manifolds, making linear decision boundaries easiest to
learn.  
This explains why inverse-SNR weighting, which emphasises mid-to-late
timesteps—beats uniform weighting, and provides an additional reason why further truncation gives a small
extra gain. Thus, merger-aware weighting focuses the zero-shot estimator on the most discriminative region of the diffusion trajectory, narrowing the
accuracy gap to dedicated representation learners such as CLIP.

\clearpage
\newpage

\clearpage
\newpage

\subsection{Zero-shot style transfer}
\label{sec:zero-shot-app}

In ~\citep{meng2021sdedit} the parameter $t_{\mathrm{stop},z}$ is estimated by a grid search over the interval ${0.1T,0.2T,\dots,T}$ for an entire dataset for each style which we follow for our baseline implementation using the PSNR metric. $t_{\mathrm{stop},z}$ estimated in this way for both datasets were observed to lie in the range ${0.3T,0.4T,0.5T}$ across styles. For our implementation $t_{stop,z}$ is estimated class-wise as the smallest merger time of the class(es) to which $z$ belongs.
\vspace{0.2in}
\begin{algorithm}[H]
\caption{Zero-shot Style Transfer}
\label{alg:zs-st}
\begin{algorithmic}[1]\small
\Require
  query $z\!\sim\!p_{0}$, noise seed $\epsilon_{n}$, maximum time $t_{\mathrm{stop},z}$

    \State $z_{t_{\mathrm{stop},z}}\gets\operatorname{FWD}_{t_{\mathrm{stop},z}}(z,\varepsilon_{n})$
    \State\Return $z^{\ast} \gets \operatorname{BWD}_{\theta,t_{\mathrm{stop},z}}(z_{t_{\mathrm{stop},z}})$
\end{algorithmic}
\end{algorithm}
\vspace{0.2in}
We use open-source fine-tuned stable diffusion models available on Hugging Face for all of our experiments. These models are trained on the artistic styles of Studio Ghibli \cite{miyazaki_ghibli_2014}, Van Gogh \cite{vanrooijen_vangogh_2019} and the styles of the animation game Elden Ring \cite{eldenring_2022} and series Arcane \cite{arcane_2021}\footnote{Disclaimer: All referenced trademarks, copyrighted characters, and original artworks remain the property of their respective owners. \Cref{alg:zs-st} was utilised for research purposes only and is not intended for commercial use or to infringe upon the intellectual property rights of the original creators.}. Note that some samples from the OxfordIIITPet Dataset may be randomly censored by the automatic filter present in these models. We observed that this happens mostly for dog images, where some breeds have samples with their mouth open. This can be mitigated to some extent by center cropping and realigning; however, if preprocessing fails, we discard such images.\Cref{tab:stf} in \Cref{sec:zero-shot-style} of the main paper has results for the Studio Ghibli and Van Gogh styles while \Cref{tab:stf-2} has results for the Elden Ring and Arcane styles.  
\vspace{0.2in}
\begin{table}[H]
\centering
\small
\begin{tabular}{|l|cc|cc|}
\hline
\rowcolor[gray]{0.8}
Style & \multicolumn{2}{l|}{Elden Ring} & \multicolumn{2}{l|}{Arcane} \\ \hline
\rowcolor[gray]{0.9}
Models/Metrics & \multicolumn{1}{l|}{PSNR ($\uparrow$)} & MSE ($\downarrow$) & \multicolumn{1}{l|}{PSNR ($\uparrow$)} & MSE ($\downarrow$) \\ \hline
SD Edit (OxfordIIITPets) & \multicolumn{1}{l|}{$24.19 \pm 0.72$} & $0.09 \pm 0.006$ & \multicolumn{1}{l|}{$25.17 \pm 0.42$} & $0.09 \pm 0.005$ \\ \hline
\textbf{Ours} (OxfordIIITPets) & \multicolumn{1}{l|}{$\mathbf{28.14 \pm 0.69}$} & $\mathbf{0.03 \pm 0.002}$ & \multicolumn{1}{l|}{$\mathbf{28.35 \pm 0.72}$} & $\mathbf{0.03 \pm 0.006}$ \\ \hline
SD Edit (AFHQv2) & \multicolumn{1}{l|}{$26.08 \pm 0.37$} & $0.06 \pm 0.003$ & \multicolumn{1}{l|}{$26.49 \pm 0.34$} & $0.05 \pm 0.004$ \\ \hline
\textbf{Ours} (AFHQ v2) & \multicolumn{1}{l|}{$\mathbf{27.56 \pm 0.37}$} & $\mathbf{0.04 \pm 0.009}$ & \multicolumn{1}{l|}{$\mathbf{28.23 \pm 0.18}$} & $\mathbf{0.03 \pm 0.001}$ \\ \hline
\end{tabular}
\vskip 0.2in
\caption{Style Transfer results for Elden Ring and Arcane styles}
\label{tab:stf-2}
\end{table}

\begin{figure}[hbtp!]
    \centering
    \includegraphics[width=\textwidth,height=0.75\textheight]{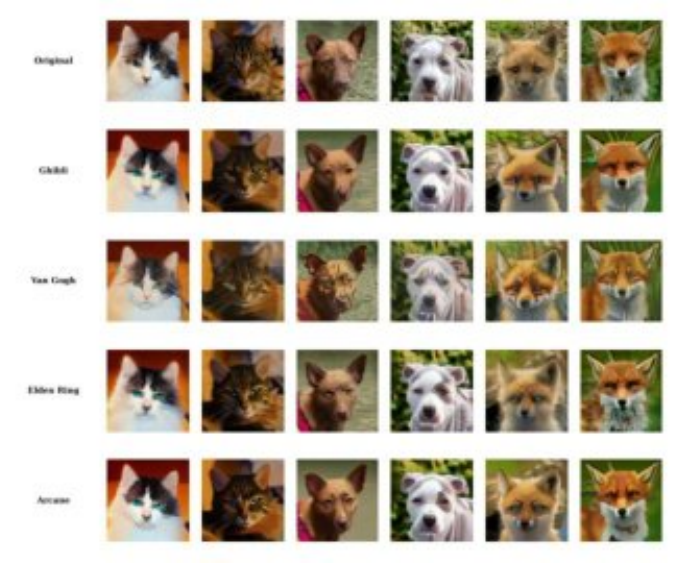}
         \caption{Zero Shot Transfer results on AFHQ v2}\label{fig:afhqv2-style}
  \end{figure}

\begin{figure}[hbtp!]
    \centering
    \includegraphics[width=\textwidth,height=0.8\textheight]{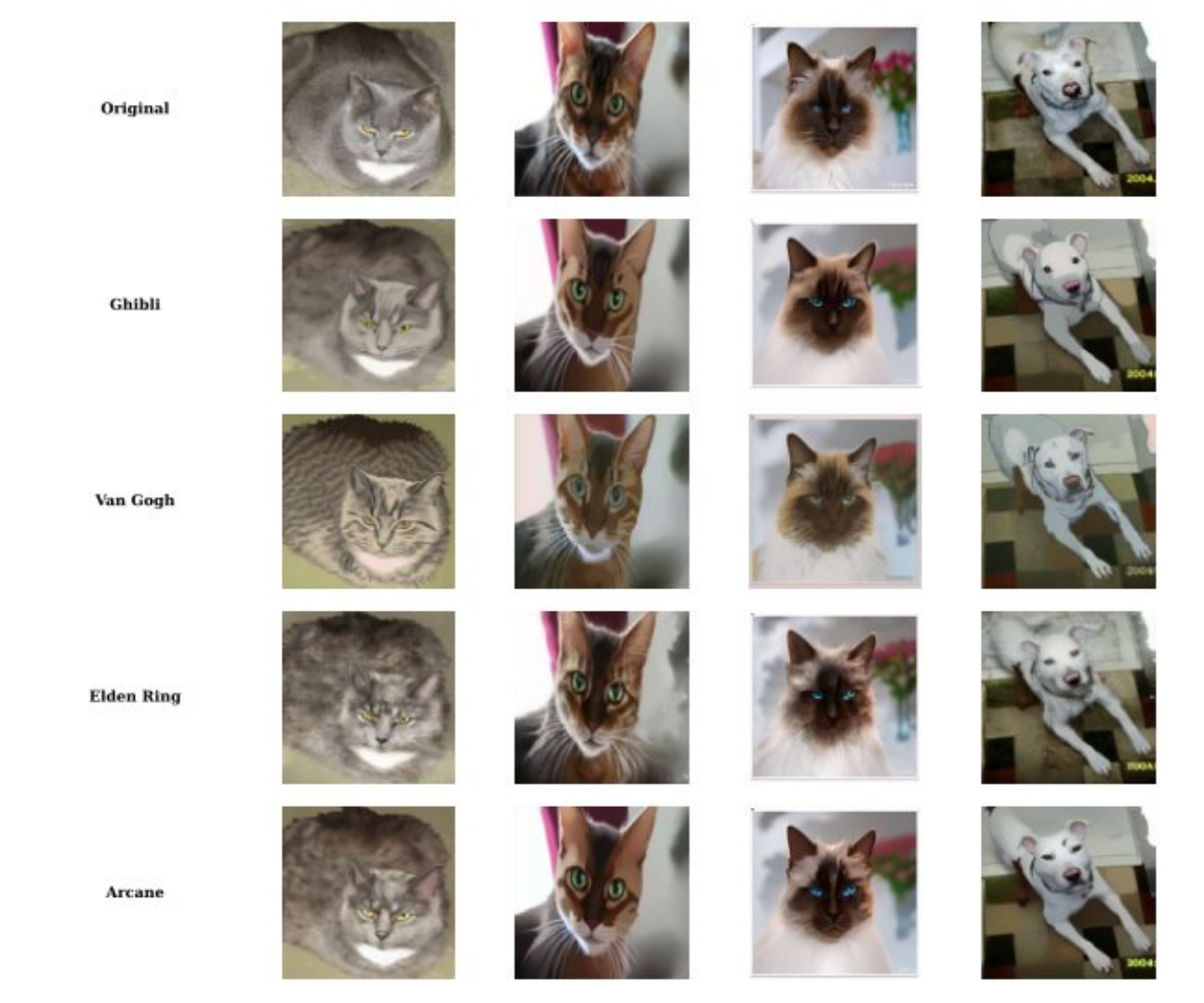}
         \caption{Zero Shot Transfer results on OxfordIIITPet }\label{fig:oxf-style}
  \end{figure}

\begin{figure}[hbtp!]
    \centering
    \includegraphics[width=\textwidth,height=0.95\textheight]{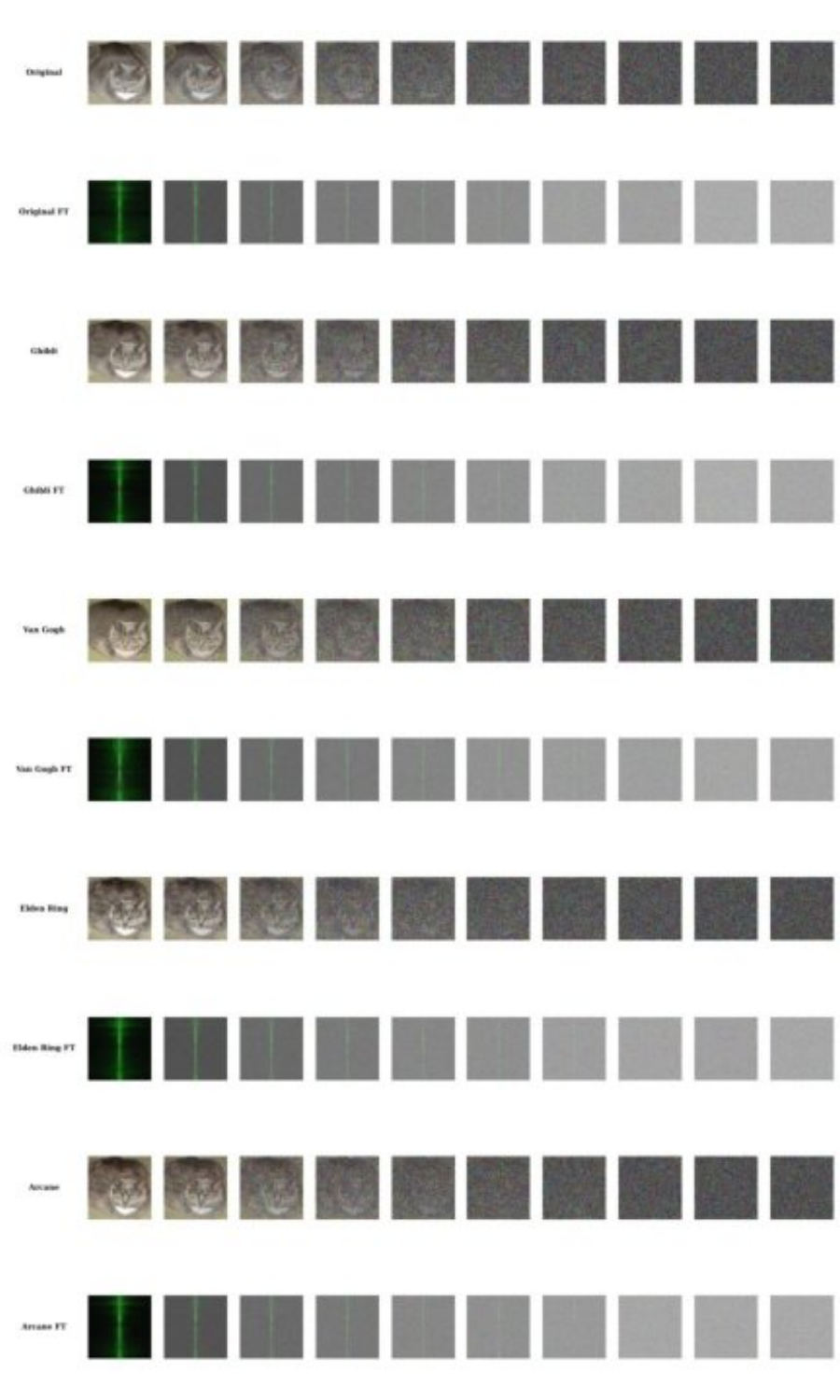}
         \caption{Fourier transforms and decay in Fourier spectra are close for structurally similar data }\label{fig:oxf-ft-1}
  \end{figure}

Visual results are in \Cref{fig:afhqv2-style,fig:oxf-style} for the AFHQv2 and OxfordIIITPet datasets, respectively. We also show visual proof of our core assumption from \Cref{sec:zero-shot-style} elaborated in \Cref{sec:reg-fluc} through visual plots of the evolution of the fourier transforms through the forward process of images mainly differing only in style, in \Cref{fig:oxf-ft-1}. Here we set a gap of $0.1T$ between the two images. These plots empirically show that for images differing only in style but with the same essential structure, their fourier transforms are \emph{close} and this fact extends to their noised versions as well due to the nonlinear decay in Brownian Motion (\Cref{sec:cka}).

\clearpage

\section{Limitations and future work}
\label{sec:limitations}

\paragraph{Modelling assumptions.}
Our analysis is built on three structural premises:
(i) a variance–preserving (VP) noise schedule,
(ii) isotropic Brownian perturbations, and
(iii) the Fourier-regularity bound of
\eqref{eq:fourier-regularity}.
These hold for the canonical SDE/ODE pairs
\citep{song2021scorebased,ho2020denoisingdiffusionprobabilisticmodels},
but have not yet been extended to non-VP schedules
(e.g.\ EDM~\citep{karras2022edm}) or to anisotropic diffusions.
Relaxing any of the three assumptions remains open.

\paragraph{Moment truncation.}
Empirically, the first few centered moments suffice to expose the phase
boundaries, yet difficult data (heavy-tailed, multimodal) may demand
higher orders.  Reliable estimation of tensor moments beyond order four is computationally expensive and lacks sharp concentration inequalities.

\paragraph{Memory overhead.}
Computing covariances $\Sigma_{k,t}\in\mathbb{R}^{d\times d}$ for many
classes incurs \(O(Kd^{2})\) memory.
Random projections, sketching, or block-diagonal
approximations could mitigate this without degrading detection power.

\paragraph{Modalities beyond 2-D images.}
All experiments target natural images.  How merger-based guidance
interacts with 3-D diffusion, video or audio, or with text 
diffusion is unexplored.

\paragraph{Training-time usage.}
We use cross-fluctuations \emph{post hoc}.
Injecting merger times into the loss—
as a curriculum on noise levels or as a regulariser enforcing class
separation—may accelerate or stabilise training; we leave this to
future work.

\section{Related work}
\label{sec:related}

\paragraph{Statistical physics of diffusion models} Thermodynamic transitions in diffusion models have been studied in prior work, such as \cite{raya2024spontaneousiop}, which identifies a mixing time transition akin to ours in \Cref{sec:warmup}, but for a narrow class of initial distributions. This is similar to the Curie-Weiss transition in ferromagnetic systems \cite{kivelson2024statistical}, with an analytical estimate. We provide an alternative derivation of this dependency for a broader class of sub-Gaussian data in \Cref{sec:vp-mixing}, but our approach in \Cref{sec:warmup} is more general, relying solely on the assumption that the supports of the data and the isotropic Gaussian are \emph{essentially disjoint}. \cite{biroli2023generative,biroli2024dynamical} extend \cite{raya2024spontaneousiop} by modeling data as a Gaussian mixture and show three distinct \emph{phases} in a class-conditional setup similar to \Cref{sec:class-cond}. Our framework further builds on these ideas in \Cref{sec:phase-transit}, where we showcase a general framework relying on discrete \emph{lattice transitions} more common in quantum systems.

\paragraph{Fast sampling for diffusion models.}
DDIM~\citep{song2020ddim}, IDDPM~\citep{nichol2021improved},
DPM-Solver~\citep{lu2022dpm}, and EDM~\citep{karras2022edm}
reduce the reverse step count; early-exit criteria such as ours are
orthogonal and in principle could be combined with any of them. We leave such extensions to future work. As discussed earlier, the occurrence of this criterion has also been demonstrated by ~\citep{raya2024spontaneousiop,biroli2023generative,biroli2024dynamical} through a different analysis based on symmetry breaking of a potential function; we show that our framework extends such considerations in \Cref{sec:phase-transit}.

\paragraph{Theoretical lenses on diffusion.}
Hyper-contractivity~\citep{saloff1994precise,chen2022sampling},
mixing-time bounds~\citep{levin2017markov}, and classical coupling
\citep{aldous-fill-2002} traces back to Markov-chain theory.
We recast these ideas as \emph{cross-fluctuation mergers},
bridging discrete and continuous settings in \Cref{sec:coupling-mixing}.

\paragraph{Conditional guidance.}
Score distillation~\citep{poole2022dreamfusion},
ILVR~\citep{choi2021ilvrconditioningmethoddenoising},
classifier-free guidance \citep{ho2022classifierfree},
and Interval Guidance (IG) \citep{kynkäänniemi2024applyingguidancelimitedinterval}
dominate conditional generation.  Our merger-aware IG trims the
interval search from per-sample to per-class with no extra
hyper-parameters with details in \Cref{sec:class-cond} and \Cref{sec:class-cond-exp}

\paragraph{Zero-shot classification using diffusion networks.}
\citet{li2023diffusionmodelsecretlyzeroshot} showed diffusion backbones
encode class information.  Importance weighting by intermediate times with a cutoff at merger times
boosts their zero-shot accuracy at equal compute (\Cref{sec:zero-shot}). We also demonstrate that near the merger times, the diffusion process demonstrates near-perfect discriminability \Cref{sec:bin-clf}.  

\paragraph{Rare-class synthesis.}
Long-tail generation typically relies on fine-tuning
\citep{bansal2023mining,samuel2024generating}.
Our fluctuation-based guidance is tuning-free and alleviates the mode
dropping at inference time by utilizing optimised IG based guidance (\Cref{sec:rare-class}, \Cref{sec:rare-class-app})
\paragraph{Zero-shot style transfer using diffusion models} The intriguing property of style transfer by learning a trained VP-SDE only on the target style after utilizing the forward process to corrupt inputs from the source style was initially shown in \cite{meng2021sdedit} and has become standard practice across setups \cite{rombach2022highresolutionimagesynthesislatent}. To our knowledge, we are the first to provide a theoretical foundation for this approach in terms of Fourier regularity that manifests in observed transitions (\Cref{sec:zero-shot-style}, \Cref{sec:reg-fluc}) while showing that setting the noising parameter based on merger times can boost fidelity (\Cref{sec:zero-shot-style},\Cref{sec:zero-shot-app})

\end{document}